\documentclass{report}
\usepackage[final,nonatbib]{nips_2018}
\usepackage{natbib}
\usepackage[utf8]{inputenc}
\usepackage{paralist}
\usepackage[dvipsnames]{xcolor}
\definecolor{linkColor}{rgb}{0.18,0.39,0.62}
\usepackage[colorlinks=true,linkcolor=linkColor,citecolor=linkColor,filecolor=linkColor,urlcolor=linkColor]{hyperref}       

\usepackage{url}
\usepackage{graphicx}
\graphicspath{{fig/}}
\usepackage{wrapfig,graphicx,lipsum}

\usepackage{tikz}
\usepackage[export]{adjustbox}

\title{Multimodal Foundation Models: \\ From Specialists to General-Purpose Assistants}

\author{
{\bf Chunyuan Li$^{*\spadesuit}$, Zhe Gan$^{*}$, Zhengyuan Yang$^{*}$, Jianwei Yang$^{*}$, Linjie Li$^{*}$, } \\
\textbf{
Lijuan Wang, Jianfeng Gao
}
\\
Microsoft Corporation\\
{\tt\small \{chunyl,zhgan,zhengyang,jianwyan,linjli,lijuanw,jfgao\}@microsoft.com}
\and
\footnotesize{
$^*$~Core Contribution \;
$^{\spadesuit}$~Project Lead \;
}
}

\usepackage{mwe}
\usepackage{amsmath}
\usepackage{amssymb}
\usepackage{wasysym}
\usepackage{float}
\usepackage{bbm}
\usepackage{enumitem}
\usepackage{multirow}
\usepackage{arydshln}
\usepackage{amssymb}
\usepackage{pifont}
\usepackage{booktabs}
\usepackage{subcaption}
\usepackage[edges]{forest}
\usepackage{changepage}
\newsavebox{\measurebox}
\definecolor{paired-light-blue}{RGB}{198, 219, 239}
\definecolor{paired-dark-blue}{RGB}{49, 130, 188}
\definecolor{paired-light-orange}{RGB}{251, 208, 162}
\definecolor{paired-dark-orange}{RGB}{230, 85, 12}
\definecolor{paired-light-green}{RGB}{199, 233, 193}
\definecolor{paired-dark-green}{RGB}{49, 163, 83}
\definecolor{paired-light-purple}{RGB}{218, 218, 235}
\definecolor{paired-dark-purple}{RGB}{117, 107, 176}
\definecolor{paired-light-gray}{RGB}{217, 217, 217}
\definecolor{paired-dark-gray}{RGB}{99, 99, 99}
\definecolor{paired-light-pink}{RGB}{222, 158, 214}
\definecolor{paired-dark-pink}{RGB}{123, 65, 115}
\definecolor{paired-light-red}{RGB}{231, 150, 156}
\definecolor{paired-dark-red}{RGB}{131, 60, 56}
\definecolor{paired-light-yellow}{RGB}{231, 204, 149}
\definecolor{paired-dark-yellow}{RGB}{141, 109, 49}
\tikzset{%
    parent/.style =          {align=center,text width=1.5cm,rounded corners=3pt, line width=0.3mm, fill=gray!10,draw=gray!80},
    child/.style =           {align=center,text width=2.3cm,rounded corners=3pt, fill=blue!10,draw=blue!80,line width=0.3mm},
    grandchild/.style =      {align=center,text width=2cm,rounded corners=3pt},
    greatgrandchild/.style = {align=center,text width=1.5cm,rounded corners=3pt},
    greatgrandchild2/.style = {align=center,text width=1.5cm,rounded corners=3pt},    
    referenceblock/.style =  {align=center,text width=1.5cm,rounded corners=2pt},
    top_class/.style =           {align=center,text width=2cm,rounded corners=3pt, fill=paired-light-gray!50,draw=paired-dark-gray!65,line width=0.3mm},
    generation/.style =           {align=center,text width=2cm,rounded corners=3pt, fill= paired-light-green!50,draw=paired-dark-green!75,line width=0.3mm}, 
    generation_wide/.style =           {align=center,text width=2.5cm,rounded corners=3pt, fill= paired-light-green!50,draw=paired-dark-green!75,line width=0.3mm}, 
    generation_more/.style =           {align=center,text width=4cm,rounded corners=3pt, fill= paired-light-green!50,draw=paired-dark-green!75,line width=0.3mm},   
    generation_work/.style =           {align=center,text width=4.0cm,rounded corners=3pt, fill= paired-light-green!50,draw= cyan!0,line width=0.3mm},
    encoder/.style =           {align=center,text width=2cm,rounded corners=3pt, fill=paired-light-orange!50,draw=paired-dark-orange!65,line width=0.3mm},  
    encoder_more/.style =           {align=center,text width=4cm,rounded corners=3pt, fill=paired-light-orange!50,draw=paired-dark-orange!65,line width=0.3mm}, 
    encoder_work/.style =           {align=center,text width=4.0cm,rounded corners=3pt, fill=paired-light-orange!50,draw=red!0,line width=0.3mm},    
    gpa/.style =           {align=center,text width=2cm,rounded corners=3pt, fill=paired-light-blue!50,draw=paired-dark-blue!65,line width=0.3mm},
    gpa_wide/.style =           {align=center,text width=4cm,rounded corners=3pt, fill=paired-light-blue!50,draw=paired-dark-blue!65,line width=0.3mm},   
    gpa_work/.style =           {align=center, text width=4.0cm,rounded corners=3pt, fill=paired-light-blue!50,draw=blue!0,line width=0.3mm},
    data/.style =           {align=center,text width=2cm,rounded corners=3pt, fill=paired-light-blue!50,draw=paired-dark-blue!65,line width=0.3mm},
    data_wide/.style =           {align=center,text width=3cm,rounded corners=3pt, fill=paired-light-blue!50,draw=paired-dark-blue!65,line width=0.3mm},   
    data_work/.style =           {align=center, text width=4.5cm,rounded corners=3pt, fill=paired-light-blue!50,draw=blue!0,line width=0.3mm},  
    model/.style =           {align=center,text width=2cm,rounded corners=3pt, fill=paired-light-orange!50,draw=paired-dark-orange!65,line width=0.3mm},  
    model_more/.style =           {align=center,text width=4cm,rounded corners=3pt, fill=paired-light-orange!50,draw=paired-dark-orange!65,line width=0.3mm}, 
    model_work/.style =           {align=center,text width=4.5cm,rounded corners=3pt, fill=paired-light-orange!50,draw=red!0,line width=0.3mm},    
    pretraining/.style =           {align=center,text width=2cm,rounded corners=3pt, fill= paired-light-green!50,draw=paired-dark-green!75,line width=0.3mm}, 
    pretraining_wide/.style =           {align=center,text width=2.5cm,rounded corners=3pt, fill= paired-light-green!50,draw=paired-dark-green!75,line width=0.3mm}, 
    pretraining_more/.style =           {align=center,text width=4cm,rounded corners=3pt, fill= paired-light-green!50,draw=paired-dark-green!75,line width=0.3mm},   
    pretraining_work/.style =           {align=center,text width=4.5cm,rounded corners=3pt, fill= paired-light-green!50,draw= cyan!0,line width=0.3mm},      
    finetuning/.style =           {align=center,text width=2cm,rounded corners=3pt, fill= paired-light-purple!50,draw=paired-dark-purple!75,line width=0.3mm},   
    finetuning_work/.style =           {align=center,text width=4.5cm,rounded corners=3pt, fill= paired-light-purple!50,draw= orange!0,line width=0.3mm},        
    inference/.style =           {align=center,text width=2cm,rounded corners=3pt, fill= paired-light-red!35,draw=paired-light-red!90,line width=0.3mm},           
    inference_more/.style =           {align=center,text width=4cm,rounded corners=3pt, fill= paired-light-red!35,draw=paired-light-red!90,line width=0.3mm},
    inference_work/.style =           {align=center,text width=4.5cm,rounded corners=3pt, fill= paired-light-red!35,draw= magenta!0,line width=0.3mm},         
}

\newcommand\blfootnote[1]{%
  \begingroup
  \renewcommand\thefootnote{}\footnote{#1}%
  \addtocounter{footnote}{-1}%
  \endgroup
}

\definecolor{chp5orange}{RGB}{238, 135, 114}
\definecolor{chp5green}{RGB}{169, 209, 142}
\definecolor{chp5blue}{RGB}{115, 149, 211}

\begin{document}

\maketitle

 \begin{figure*}[h!]
  \centering
    \includegraphics[width=0.9\linewidth]{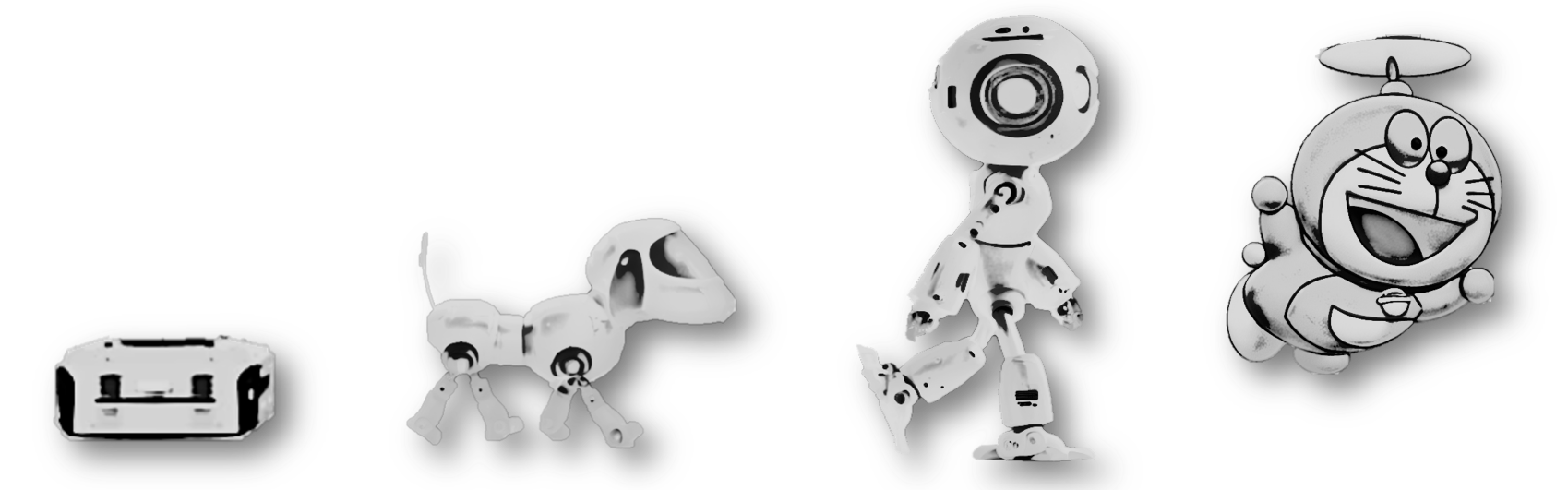}
  \label{fig:front_image}
\end{figure*}

\vspace{0cm}
\begin{abstract}
This paper presents a comprehensive survey of the taxonomy and evolution of multimodal foundation models that demonstrate vision and vision-language capabilities, focusing on the transition from specialist models to general-purpose assistants. The research landscape encompasses five core topics, categorized into two classes. $(i)$ We start with a survey of well-established research areas: multimodal foundation models pre-trained for specific purposes, including two topics -- methods of learning vision backbones for visual understanding and text-to-image generation. $(ii)$ 
Then, we present recent advances in exploratory, open research areas:  multimodal foundation models that aim to play the role of general-purpose assistants, including three topics -- unified vision models inspired by large language models (LLMs), end-to-end training of multimodal LLMs, and chaining multimodal tools with LLMs. 
The target audiences of the paper are researchers, graduate students, and professionals in computer vision and vision-language multimodal communities who are eager to learn the basics and recent advances in multimodal foundation models.
\end{abstract}

\blfootnote{$^{1}$~Chunyuan Li initiated the project, and took lead in the writing of Chapter~\ref{chp:intro}, \ref{chp:training_with_llm} and \ref{chp:conclusion}. Zhe Gan, Zhengyuan Yang, Jianwei Yang, Linjie Li took lead in the writing of Chapter~\ref{chp:understanding}, ~\ref{chp:generation}, ~\ref{chp:generalist} and \ref{chp:chaining_with_llm}, respectively. Lijuan Wang and Jianfeng Gao provided comprehensive suggestions and edits of the entire paper. All the authors provided project advice, and contributed to paper review, editing and proofreading.
}
\blfootnote{$^{2}$~Zhe Gan is currently with Apple AI/ML.
}


\newcommand{\todo}[1]{{\color{red}{[{\bf TODO}: #1]}}}
\newcommand{\todoJG}[1]{{\color{blue}{[{\bf JG}: #1]}}}
\newcommand{\todoMG}[1]{{\color{purple}{[{\bf MG}: #1]}}}
\newcommand{\todoLL}[1]{{\color{cyan}{[{\bf LL}: #1]}}}

\newcommand{\figref}[1]{Fig.~\ref{#1}}
\newcommand{\eqnref}[1]{Eqn.~\ref{#1}}
\newcommand{\chref}[1]{Chapter~\ref{#1}}
\newcommand{\secref}[1]{Sec.~\ref{#1}}
\newcommand{\tabref}[1]{Table~\ref{#1}}
\newcommand{\ie}{{\it  i.e.~}}
\newcommand{\eg}{{\it e.g.~}}
\newcommand{\etc}{{etc.}}
\newcommand{\cf}{{\it cf.}}

\newcommand{\dnfont}[1]{{\texttt{#1}}}  
\newcommand{\dafont}[1]{{\texttt{#1}}}  
\newcommand{\slotfont}[1]{\texttt{#1}}  
\newcommand{\valuefont}[1]{\textcolor{gray}{\texttt{#1}}}
\newcommand{\vecb}[1]{\mathbf{#1}}

\newcommand{\exbox}[1]{ 
{\begin{center}\fbox{%
    \begin{minipage}{.95\textwidth} 
      \centering #1
    \end{minipage}%
  }\end{center}} 
}

\newcommand{\defeq}{:=}
\newcommand{\E}{\mathbb{E}}
\newcommand{\Rset}{\mathbb{R}}
\newcommand{\mt}{{\operatorname{T}}}  			
\newcommand{\mi}{{-1}}  											
\newcommand{\argmin}{\operatorname{argmin}}
\newcommand{\argmax}{\operatorname{argmax}}

\newcommand{\Sset}{\mathcal{S}}
\newcommand{\Aset}{\mathcal{A}}
\newcommand{\Dset}{\mathcal{D}}

\newcommand{\mMSE}{\operatorname{MSE}}
\newcommand{\mFone}{\operatorname{F1}}
\newcommand{\mREC}{\operatorname{RECALL}}
\newcommand{\mACC}{\operatorname{ACCURACY}}
\newcommand{\mPRE}{\operatorname{PRECISION}}
\newcommand{\mAUC}{\operatorname{AUC}}
\newcommand{\1}{\mathbf{1}}

\newcommand{\av}{{\boldsymbol{a}}}
\newcommand{\bv}[0]{{\boldsymbol{b}}}
\newcommand{\cv}[0]{{\boldsymbol{c}}}
\newcommand{\dv}{\boldsymbol{d}}
\newcommand{\ev}[0]{{\boldsymbol{e}}\xspace}
\newcommand{\fv}[0]{{\boldsymbol{f}}}
\newcommand{\gv}[0]{{\boldsymbol{g}}}
\newcommand{\hv}[0]{{\boldsymbol{h}}}
\newcommand{\iv}[0]{{\boldsymbol{i}}\xspace}
\newcommand{\jv}[0]{{\boldsymbol{j}}\xspace}
\newcommand{\kv}[0]{{\boldsymbol{k}}\xspace}
\newcommand{\lv}[0]{{\boldsymbol{l}}}
\newcommand{\mv}[0]{{\boldsymbol{m}}}
\newcommand{\nv}[0]{{\boldsymbol{n}}\xspace}
\newcommand{\ov}[0]{{\boldsymbol{o}}\xspace}
\newcommand{\pv}[0]{{\boldsymbol{p}}}
\newcommand{\qv}[0]{{\boldsymbol{q}}}
\newcommand{\rv}{\boldsymbol{r}}
\newcommand{\sv}{{\boldsymbol{s}}}
\newcommand{\tv}[0]{{\boldsymbol{t}}}
\newcommand{\uv}{\boldsymbol{u}}
\newcommand{\vv}{\boldsymbol{v}}
\newcommand{\wv}{\boldsymbol{w}}
\newcommand{\xv}{\boldsymbol{x}}
\newcommand{\yv}{\boldsymbol{y}}
\newcommand{\zv}{\boldsymbol{z}}
\newcommand{\cdotv}{\boldsymbol{\cdot}}

\newcommand{\Amat}[0]{{{\bf A}}}
\newcommand{\Bmat}{{\bf B}}
\newcommand{\Cmat}{{\bf C}}
\newcommand{\Dmat}{{\bf D}}
\newcommand{\Emat}[0]{{{\bf E}}}
\newcommand{\Fmat}[0]{{{\bf F}}\xspace}
\newcommand{\Gmat}{{\bf G}}
\newcommand{\Hmat}{{\bf H}}
\newcommand{\Imat}{{\bf I}}
\newcommand{\Jmat}[0]{{{\bf J}}\xspace}
\newcommand{\Kmat}[0]{{{\bf K}}\xspace}
\newcommand{\Lmat}[0]{{{\bf L}}}
\newcommand{\Mmat}{{\bf M}}
\newcommand{\Nmat}[0]{{{\bf N}}\xspace}
\newcommand{\Omat}[0]{{{\bf O}}}
\newcommand{\Pmat}{{\bf P}}
\newcommand{\Qmat}[0]{{{\bf Q}}\xspace}
\newcommand{\Rmat}[0]{{{\bf R}}}
\newcommand{\Smat}[0]{{{\bf S}}}
\newcommand{\Tmat}[0]{{{\bf T}}}
\newcommand{\Umat}[0]{{{\bf U}}}
\newcommand{\Vmat}[0]{{{\bf V}}}
\newcommand{\Wmat}[0]{{{\bf W}}}
\newcommand{\Xmat}[0]{{{\bf X}}}
\newcommand{\Ymat}{{\bf Y}}
\newcommand{\Zmat}{{\bf Z}}

\newcommand{\Xcal}{\mathcal{X}}
\newcommand{\Ycal}{\mathcal{Y}}
\newcommand{\Ncal}{\mathcal{N}}
\newcommand{\Acal}{\mathcal{A}}
\newcommand{\Bcal}{\mathcal{B}}
\newcommand{\Dcal}{\mathcal{D}}
\newcommand{\Fcal}{\mathcal{F}}
\newcommand{\Tcal}{\mathcal{T}}
\newcommand{\Mcal}{\mathcal{M}}
\newcommand{\Lcal}{\mathcal{L}}
\newcommand{\Ocal}{\mathcal{O}}
\newcommand{\Pcal}{\mathcal{P}}
\newcommand{\Ical}{\mathcal{I}}
\newcommand{\Kcal}{\mathcal{K}}
\newcommand{\Gcal}{\mathcal{G}}
\newcommand{\Qcal}{\mathcal{Q}}
\newcommand{\Rcal}{\mathcal{R}}
\newcommand{\Scal}{\mathcal{S}}
\newcommand{\Hcal}{\mathcal{H}}
\newcommand{\Vcal}{\mathcal{V}}
\newcommand{\Zcal}{\mathcal{Z}}
\newcommand{\Ucal}{\mathcal{U}}
\newcommand{\Jcal}{\mathcal{J}}
\newcommand{\scal}{\mathcal{s}}

\newcommand{\alphav}{\boldsymbol{\alpha}}
\newcommand{\chiv}{\boldsymbol{\chi}}
\newcommand{\betav}[0]{{\boldsymbol{\beta}}}
\newcommand{\gammav}[0]{{\boldsymbol{\gamma}}\xspace}
\newcommand{\deltav}[0]{{\boldsymbol{\delta}}\xspace}
\newcommand{\epsilonv}{\boldsymbol{\epsilon}}
\newcommand{\zetav}{\boldsymbol{\zeta}}
\newcommand{\etav}{\boldsymbol{\eta}}
\newcommand{\ellv}[0]{{\boldsymbol{\ell}}}
\newcommand{\thetav}{\boldsymbol{\theta}}
\newcommand{\iotav}[0]{{\boldsymbol{\iota}}}
\newcommand{\kappav}[0]{{\boldsymbol{\kappa}}\xspace}
\newcommand{\lambdav}[0]{{\boldsymbol{\lambda}}}
\newcommand{\muv}[0]{{\boldsymbol{\mu}}}
\newcommand{\nuv}[0]{{\boldsymbol{\nu}}}
\newcommand{\xiv}[0]{{\boldsymbol{\xi}}}
\newcommand{\omicronv}[0]{{\boldsymbol{\omicron}}\xspace}
\newcommand{\piv}{\boldsymbol{\pi}}
\newcommand{\rhov}[0]{{\boldsymbol{\rho}}\xspace}
\newcommand{\sigmav}[0]{{\boldsymbol{\sigma}}}
\newcommand{\tauv}[0]{{\boldsymbol{\tau}}}
\newcommand{\upsilonv}[0]{{\boldsymbol{\upsilon}}\xspace}
\newcommand{\phiv}{\boldsymbol{\phi}}
\newcommand{\psiv}{\boldsymbol{\psi}}
\newcommand{\varthetav}{\boldsymbol{\vartheta}}
\newcommand{\omegav}[0]{{\boldsymbol{\omega}}}
\newcommand{\R}{\mathbb{R}}
\newcommand{\Z}{\mathbb{Z}}
\newcommand{\specialcell}[2][c]{%
  \begin{tabular}[#1]{@{}c@{}}#2\end{tabular}}
\newcommand{\specialcelll}[2][l]{%
  \begin{tabular}[#1]{@{}l@{}}#2\end{tabular}}

\tableofcontents
\newpage

\chapter{Introduction}
\label{chp:intro}

Vision is one of the primary channels for humans and many living creatures to perceive and interact with the world.  One of the core aspirations in artificial intelligence (AI) is to develop AI agents to mimic such an ability to effectively perceive and generate visual signals, and thus reason over and interact with the visual world. Examples include recognition of the objects and actions in the scenes, and creation of sketches and pictures for communication.
Building foundational models with visual capabilities is a prevalent research field striving to accomplish this objective.

\begin{figure*}[h!]
  \centering
    \includegraphics[width=0.99\linewidth]{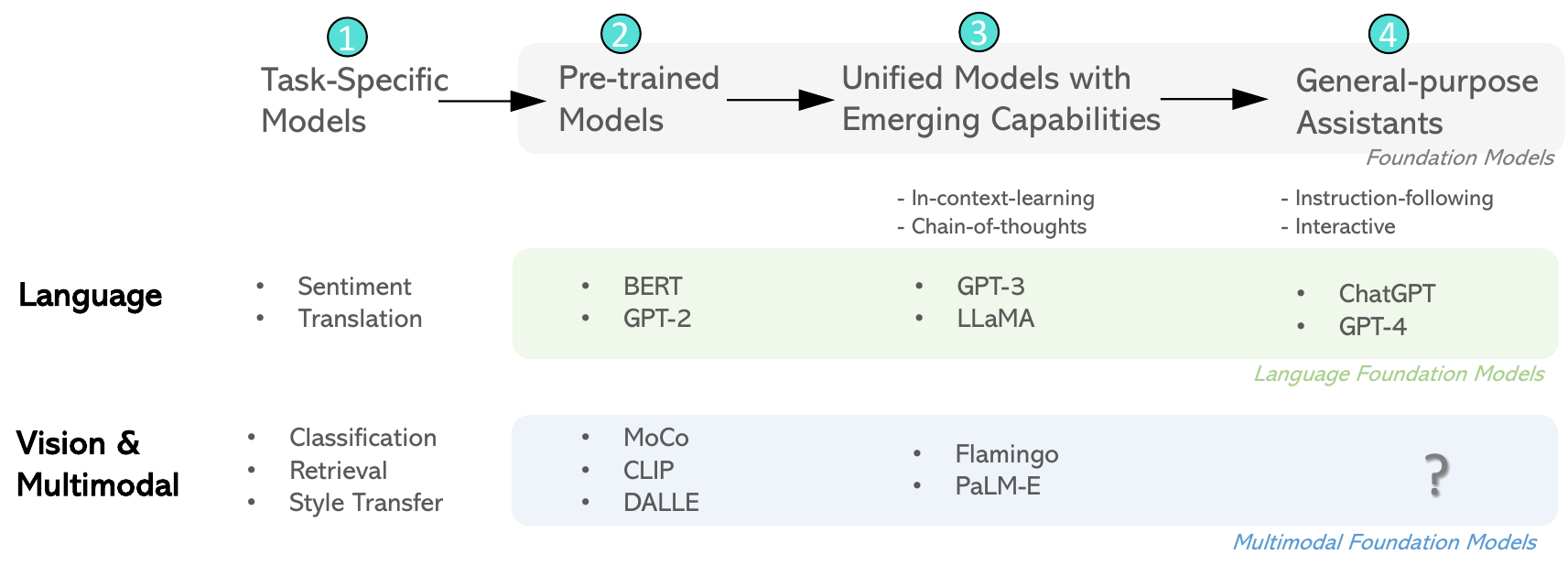}
  \caption{Illustration of foundation model development trajectory for language and vision/multi-modality. Among the four categories, the first category is the task-specific model, and the last three categories belong to foundation models, where these foundation models for language and vision are grouped in green and blue blocks, respectively. Some prominent properties of models in each category are highlighted. By comparing the  models between language and vision, we are foreseeing that the transition of multimodal foundation models follows a similar trend: from the pre-trained model for specific purpose, to unified models and general-purpose assistants.  However, research exploration is needed to figure out the best recipe, which is indicated as the question mark in the figure, as multimodal GPT-4 and Gemini stay private. 
  }
  \label{fig:chp1_stages_nlp_cv}
\end{figure*}

Over the last decade, the field of AI has experienced a fruitful trajectory in the development of models. We divide them into four categories, as illustrated in Figure~\ref{fig:chp1_stages_nlp_cv}. The categorization can be shared among different fields in AI, including language, vision and multimodality. We first use language models in NLP to illustrate the evolution process. $(i)$ At the early years, task-specific models are developed for individual datasets and tasks, typically being trained from scratch. 
$(ii)$ With large-scale pre-training, language models achieve state-of-the-art performance on many established language understanding and generation tasks, such as BERT~\citep{devlin2018bert}, RoBERTa~\citep{liu2019roberta}, T5~\citep{raffel2020exploring}, DeBERTa~\citep{he2020deberta} and GPT-2~\citep{radford2019language}). These pre-trained models serve the basis for downstream task adaptation.
$(iii)$ Exemplified by GPT-3~\citep{brown2020language}, large language models (LLMs) unify various language understanding and generation tasks into one model. With web-scale training and unification, some emerging capabilities appear, such as in-context-learning and chain-of-thoughts.
$(iv)$
With recent advances in human-AI alignment, LLMs start to play the role of general-purpose assistants to follow human intents to complete a wide range of language tasks in the wild, such as ChatGPT~\citep{chatgpt} and GPT-4~\citep{gpt4}. These assistants exhibit interesting capabilities, such as interaction and tool use, and lay a foundation for developing general-purpose AI agents.
It is important to note that the latest iterations of foundation models build upon the noteworthy features of their earlier counterparts while also providing additional capabilities.

Inspired by the great successes of LLMs in NLP, it is natural for researchers in the computer vision and vision-language community to ask the question: what is the counterpart of ChatGPT/GPT-4 for vision, vision-language and multi-modal models?
There is no doubt that vision pre-training and vision-language pre-training (VLP) have attracted a growing attention since the birth of BERT, and has become the mainstream learning paradigm for vision, with the promise to learn universal transferable visual and vision-language representations, or to generate highly plausible images. Arguably, they can be considered as the early generation of multimodal foundation models, just as BERT/GPT-2 to the language field. While the road-map to build general-purpose assistants for language such as ChatGPT is clear, it is becoming increasingly crucial for the research community to explore feasible solutions to building its counterpart for computer vision: the general-purpose visual assistants. Overall, building general-purpose agents has been a long-standing goal for AI. 
LLMs with emerging properties have significantly reduced the cost of building such agents for language tasks. Similarly, we foresee emerging capabilities from vision models, such as following the instructions composed by various visual prompts like user-uploaded images, human-drawn clicks, sketches and mask, in addition to text prompt.
Such strong zero-shot visual task composition capabilities can significantly reduce the cost of building AI agents.

In this paper, we limit the scope of multimodal foundation models to the vision and vision-language domains. Recent survey papers on related topics include $(i)$ {\it image understanding models} such as self-supervised learning~\citep{jaiswal2020survey,jing2020self,ozbulak2023know}, segment anything (SAM)~\citep{zhang2023comprehensive,zhang2023survey}, 
$(ii)$ {\it  image generation models}~\citep{zhang2023text,zhou2023vision}, and 
$(iii)$  {\it vision-language pre-training (VLP)}.
Existing VLP survey papers cover VLP methods for task-specific VL problems before the era of pre-training, image-text tasks, core vision tasks, and/or video-text tasks~\citep{zhang2020multimodal,du2022survey,li2022vision,ruan2022survey,chen2022vlp,gan2022vision,zhang2023vision}. Two recent survey papers cover the integration of vision models with LLM~\citep{awais2023foundational,yin2022survey}.

Among them, \cite{gan2022vision} is a survey on VLP that covers the CVPR tutorial series on {\it Recent Advances in Vision-and-Language Research} in 2022 and before. This paper summarizes the CVPR tutorial on {\it Recent Advances in Vision Foundation Models} in 2023.
Different from the aforementioned survey papers that focus on literature review of a given research topic, this paper presents our perspectives on the role transition of multimodal foundation models from specialists to general-purpose visual assistants, in the era of large language models.
The contributions of this survey paper are summarized as follows.
\begin{itemize}[leftmargin=*]
    \item We provide a comprehensive and timely survey on modern multimodal foundation models, not only covering well-established models for visual representation learning and image generation, but also summarizing emerging topics for the past 6 months inspired by LLMs, including unified vision models, training and chaining with LLMs.
    \item The paper is positioned to provide the audiences with the perspective to advocate a transition in developing multimodal foundation models. On top of great modeling successes for specific vision problems, we are moving towards building general-purpose assistants that can follow human intents to complete a wide range of computer vision tasks in the wild. We provide in-depth discussions on these advanced topics, demonstrating the potential of developing general-purpose visual assistants.
\end{itemize}

\section{What are Multimodal Foundation Models?}

As elucidated in the Stanford foundation model paper~\citep{bommasani2021opportunities}, AI has been undergoing a paradigm shift with the rise of models (\emph{e.g.}, BERT, GPT family, CLIP~\citep{radford2021learning} and DALL-E~\citep{ramesh2021dalle}) trained on broad data that can be adapted to a wide range of downstream tasks. They call these models {\it  foundation models} to underscore their critically central yet incomplete character: homogenization of the methodologies across research communities and emergence of new capabilities. 
From a technical perspective, it is \emph{transfer learning} that makes foundation models possible, and it is \emph{scale} that makes them powerful.
The emergence of foundation models has been predominantly observed in the NLP domain, with examples ranging from BERT to ChatGPT. This trend has gained traction in recent years, extending to computer vision and other fields. In NLP, the introduction of BERT in late 2018 is considered as the inception of the foundation model era. The remarkable success of BERT rapidly stimulates interest in self-supervised learning in the computer vision community, giving rise to models such as SimCLR~\citep{chen2020simple}, MoCo~\citep{he2020momentum}, BEiT~\citep{bao2021beit}, and MAE~\citep{he2022masked}. During the same time period, the success of pre-training also significantly promotes the vision-and-language multimodal field to an unprecedented level of attention.

\begin{figure*}[t!]
  \centering
    \includegraphics[width=0.9\linewidth]{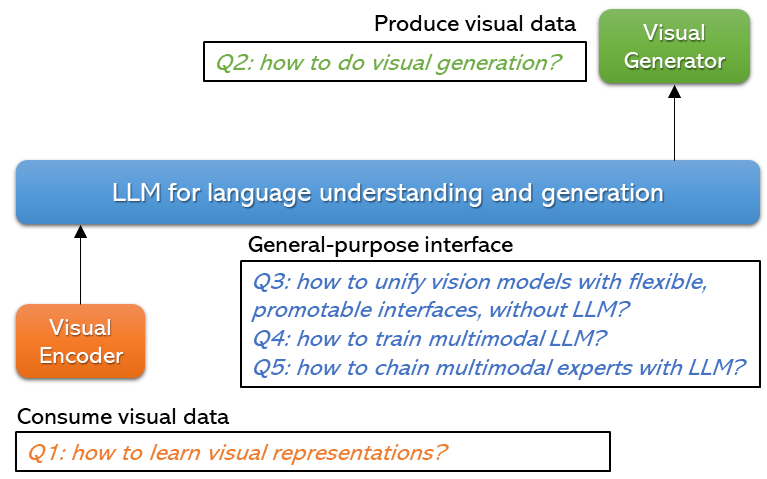}
  \caption{Illustration of three representative problems that multimodal foundation models aim to solve in this paper:  \colorbox{paired-light-orange!120}{visual understanding tasks}, \colorbox{paired-light-green!120}{visual generation tasks}, and \colorbox{paired-light-blue!120}{general-purpose interface}  with language understanding and generation.}
  \label{fig:chp1_tasks}
\end{figure*}

In this paper, {\it we focus on multimodal foundation models, which inherit all properties of foundation models discussed in the Stanford paper~\citep{bommasani2021opportunities}, but with an emphasis on models with the capability to deal with vision and vision-language modalities.}
Among the ever-growing literature, we categorize multimodal foundation models in Figure~\ref{fig:chp1_tasks}, based on their functionality and generality. For each category, we present exemplary models that demonstrate the primary capabilities inherent to these multimodal foundation models.

\begin{itemize}[leftmargin=*]
    \item \textbf{Visual Understanding Models.} (Highlighted with orange in Figure~\ref{fig:chp1_tasks}) Learning general visual representations is essential to build vision foundation models, as pre-training a strong vision backbone is foundamental to all types of computer vision downstream tasks, ranging from image-level (\emph{e.g.}, image classification, retrieval, and captioning), region-level (\emph{e.g.}, detection and grounding) to pixel-level tasks (\emph{e.g.}, segmentation). We group the methods into three categories, depending on the types of supervision signals used to train the models. 
    \begin{itemize}[leftmargin=*]
        \item \textbf{Label supervision.} Datasets like ImageNet~\citep{krizhevsky2012imagenet} and ImageNet21K~\citep{ridnik2021imagenet} have been popular for supervised learning, and larger-scale proprietary datasets are also used in industrial labs~\citep{sun2017revisiting,singh2022revisiting,zhai2022scaling}. 
        \item \textbf{Language supervision.} Language is a richer form of supervision.
        Models like CLIP~\citep{radford2021learning} and ALIGN~\citep{jia2021scaling}
        are pre-trained using a contrastive loss over millions or even billions of noisy image-text pairs mined from the Web. These models enable zero-shot image classification, and make traditional computer vision (CV) models to perform open-vocabulary CV tasks. We advocate the concept of \emph{computer vision in the wild},\footnote{\href{https:https://github.com/Computer-Vision-in-the-Wild/CVinW_Readings/blob/main/README.md}{Computer-Vision-in-the-Wild Readings.}} and encourage the development and evaluation of future foundation models for this.
        \item \textbf{Image-only self-supervision.} This line of work aims to learn image representations from supervision signals mined from the images themselves, ranging from contrastive learning~\citep{chen2020simple,he2020momentum}, non-contrastive learning~\citep{grill2020bootstrap,chen2021exploring,caron2021emerging}, to masked image modeling~\citep{bao2021beit,he2022masked}. 
        
        \item \textbf{Multimodal fusion, region-level and pixel-level pre-training.} Besides the methods of pre-training image backbones, we will also discuss pre-training methods that allow multimodal fusion (\emph{e.g.}, CoCa~\citep{yu2022coca}, Flamingo~\citep{alayrac2022flamingo}), region-level and pixel-level image understanding, such as open-set object detection (\emph{e.g.}, GLIP~\citep{li2021grounded}) and promptable semgentation (\emph{e.g.}, SAM~\citep{kirillov2023segment}). These methods typically rely on a pre-trained image encoder or a pre-trained image-text encoder pair.

    \end{itemize}
    
    \item \textbf{Visual Generation Models.}  
     (Highlighted with green in Figure~\ref{fig:chp1_tasks}) Recently, foundation image generation models have been built, due to the emergence of large-scale image-text data. The techniques that make it possible include the vector-quantized VAE methods~\citep{razavi2019generating}, diffusion-based models~\citep{dhariwal2021diffusion} and auto-regressive models. 
        \begin{itemize}[leftmargin=*]
            

            \item \textbf{Text-conditioned visual generation.} This research area focuses on generating faithful visual content, including images, videos, and more, conditioned on open-ended text descriptions/prompts. Text-to-image generation develops generative models that synthesize images of high fidelity to follow the text prompt. Prominent examples include DALL-E~\citep{ramesh2021dalle}, DALL-E 2~\citep{ramesh2022hierarchical}, Stable Diffusion~\citep{rombach2021highresolution,stable_diffusion}, Imagen~\citep{saharia2022photorealistic}, and Parti~\citep{yu2022scaling}. Building on the success of text-to-image generation models, text-to-video generation models generate videos based on text prompts, such as Imagen Video~\citep{ho2022imagen} and Make-A-Video~\citep{singer2022make}.

            \item \textbf{Human-aligned visual generator.} This research area focuses on improving the pre-trained visual generator to better follow human intentions. Efforts have been made to address various challenges inherent to base visual generators. These include improving spatial controllability~\citep{zhang2023adding,yang2023reco}, ensuring better adherence to text prompts~\citep{black2023training}, supporting flexible text-based editing~\citep{brooks2023instructpix2pix}, and facilitating visual concept customization~\citep{ruiz2023dreambooth}.
        \end{itemize}

    \item \textbf{General-purpose Interface.} 
    (Highlighted with blue in Figure~\ref{fig:chp1_tasks}) The aforementioned multimodal foundation models are designed for specific purposes -- tackling a specific set of CV problems/tasks. Recently, we see an emergence of general-purpose models that lay the basis of AI agents. Existing efforts focus on three research topics. The first topic aims to unify models for visual understanding and generation. These models are inspired by the unification spirit of LLMs in NLP, but do not explicitly leverage pre-trained LLM in modeling. In contrast, the other two topics embrace and involve LLMs in modeling, including training and chaining with LLMs, respectively.
        \begin{itemize}[leftmargin=*]
            \item \textbf{Unified vision models for understanding and generation.}  In computer vision, several attempts have been made to build a general-purpose foundation model by combining the functionalities of specific-purpose multimodal models. To this end, a unified model architecture is adopted for various downstream computer vision and vision-language (VL) tasks. There are different levels of unification. First, a prevalent effort is to bridge vision and language by converting all closed-set vision tasks to open-set ones, such as CLIP~\citep{radford2021learning}, GLIP~\citep{li2022grounded}, OpenSeg~\citep{ghiasi2021open}, \textit{etc.} Second, the unification of different VL understanding tasks across different granularity levels is also actively explored, such as I/O unification methods like UniTAB~\citep{yang2021crossing}, Unified-IO~\citep{lu2022unified}), Pix2Seq-v2~\citep{chen2022unified} and functional unification methods like GPV~\citep{gupta2022towards}, GLIP-v2~\citep{zhang2022glipv2}) and X-Decoder~\citep{zou2023generalized}. In the end, it is also necessitated to make the models more interactive and promptable like ChatGPT, and this has been recently studied in SAM~\citep{kirillov2023segment} and SEEM~\citep{zou2023segment}.             
            \item \textbf{Training with LLMs.} Similar to the behavior of LLMs, which can address a language task by following the instruction and processing examples of the task in their text prompt, it is desirable to develop a visual and text interface to steer the model towards solving a multimodal task. By extending the capability of LLMs to multimodal settings and training the model end-to-end, multimodal LLMs or large multimodal models are developed, including Flamingo~\citep{alayrac2022flamingo} and Multimodal GPT-4~\citep{gpt4}.
            \item \textbf{Chaining tools with LLM.} Exploiting the tool use capabilities of LLMs, an increasing number of studies integrate LLMs such as ChatGPT with various multimodal foundation models to facilitate image understanding and generation through a conversation interface. This interdisciplinary approach combines the strengths of NLP and computer vision, enabling researchers to develop more robust and versatile AI systems that are capable of processing visual information and generating human-like responses via human-computer conversations. Representative works include Visual ChatGPT~\citep{wu2023visual} and \textsc{MM-ReAct}~\citep{yang2023mmreact}.
        \end{itemize} 
\end{itemize}



\section{Definition and Transition from Specialists to General-Purpose Assistants}

Based on the model development history and taxonomy in NLP, we group multimodal foundation models in Figure~\ref{fig:chp1_tasks} into two categories. 
\begin{itemize}[leftmargin=*]
    \item \textbf{Specific-Purpose Pre-trained Vision Models} cover most existing multimodal foundation models, including visual understanding models (\emph{e.g.}, CLIP~\citep{radford2021learning}, SimCLR~\citep{chen2020simple}, BEiT~\citep{bao2021beit}, SAM~\citep{kirillov2023segment}) and visual generation models (\emph{e.g.}, Stable Diffusion~\citep{rombach2021highresolution,stable_diffusion}), as they present powerful transferable ability for specific vision problems. 

    \item \textbf{General-Purpose Assistants} refer to AI agents that can follow human intents to complete various computer vision tasks in the wild. The meanings of general-purpose assistants are two-fold: $(i)$ generalists with unified architectures that could complete tasks across different problem types, and $(ii)$ easy to follow human instruction, rather than replacing humans. To this end, several research topics have been actively explored, including unified vision modeling~\citep{lu2022unified,zhang2022glipv2,zou2023generalized}, training and chaining with LLMs~\citep{liu2023visual,zhu2023minigpt4,wu2023visual,yang2023mmreact}.
\end{itemize}

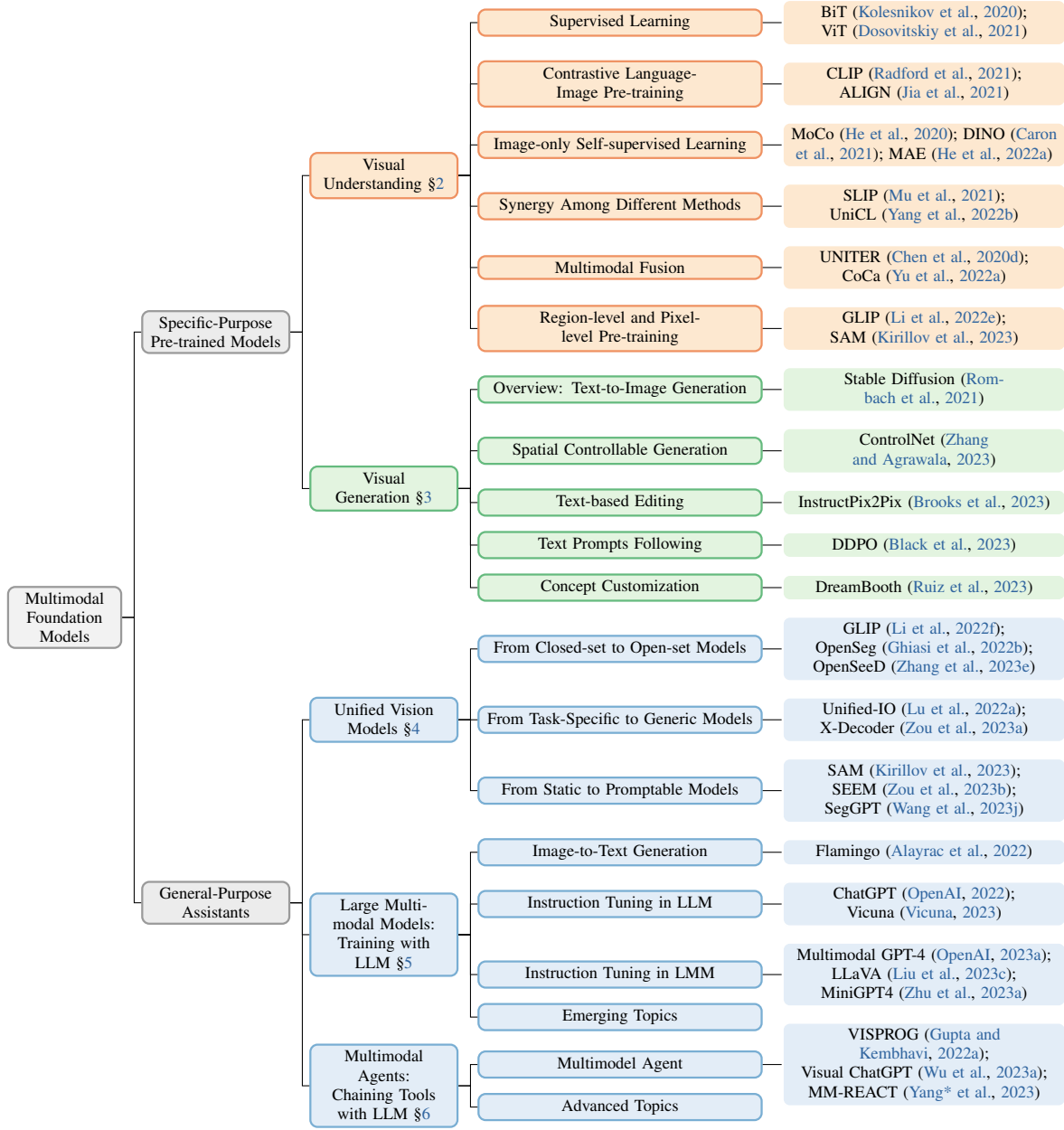
\begin{figure*}
\scriptsize
\hspace*{-30pt}
    \begin{forest}
        for tree={
            forked edges,
            grow'=0,
            draw,
            rounded corners,
            node options={align=center,},
            text width=2.7cm,
            s sep=6pt,
            calign=edge midpoint,
        },
        [Multimodal \\Foundation~\\ Models, fill=gray!45, parent
            [Specific-Purpose Pre-trained Models, for tree={ top_class}
                [Visual \\ Understanding \S\ref{chp:understanding}, for tree={fill=red!45,encoder}
                    [Supervised Learning,  encoder_more
                        [BiT~\citep{kolesnikov2020big}; ViT~\citep{dosovitskiy2020image}, encoder_work]
                    ]
                    [Contrastive Language-Image Pre-training, encoder_more 
                        [CLIP~\citep{radford2021learning};
                        ALIGN~\citep{jia2021scaling}, encoder_work]
                    ]
                    [Image-only Self-supervised Learning,  encoder_more
                        [MoCo~\citep{he2020momentum}; 
                        DINO~\citep{caron2021emerging};
                        MAE~\citep{he2022masked}, encoder_work]
                    ]
                    [Synergy Among Different Methods,  encoder_more
                        [SLIP~\citep{mu2021slip}; UniCL~\citep{yang2022unicl}, encoder_work]
                    ]
                    [Multimodal Fusion,  encoder_more
                        [UNITER~\citep{chen2020uniter};
                        CoCa~\citep{yu2022coca}, encoder_work]
                    ]
                    [Region-level and Pixel-level Pre-training,  encoder_more
                        [GLIP~\citep{li2021grounded}; SAM~\citep{kirillov2023segment}, encoder_work]
                    ]                    
                ]
                [Visual \\ Generation \S\ref{chp:generation}, for tree={fill=green!45,generation}
                    [Overview: Text-to-Image Generation, generation_more
                        [Stable Diffusion~\citep{rombach2021highresolution}, generation_work]
                    ]
                    [Spatial Controllable Generation, generation_more
                        [ControlNet~\citep{zhang2023adding}, generation_work]
                    ]  
                    [Text-based Editing, generation_more
                        [InstructPix2Pix~\citep{brooks2023instructpix2pix}, generation_work]
                    ]                 
                    [Text Prompts Following, generation_more
                        [DDPO~\citep{black2023training}, generation_work]                    ]      
                    [Concept Customization, generation_more
                        [DreamBooth~\citep{ruiz2023dreambooth}, generation_work]
                    ]      
                ]
            ]
            [General-Purpose Assistants, for tree={ top_class}
                [Unified Vision \\ Models \S\ref{chp:generalist}, for tree={fill=green!45,gpa}
                    [From Closed-set to Open-set Models, gpa_wide
                        [GLIP~\citep{li2022grounded};
                        OpenSeg~\citep{ghiasi2022scaling}; OpenSeeD~\citep{zhang2023simple}, gpa_work]
                    ]
                    [From Task-Specific to Generic Models, gpa_wide
                        [Unified-IO~\citep{lu2022unified}; 
                        X-Decoder~\citep{zou2023generalized}, gpa_work]
                    ]        
                    [From Static to Promptable Models, gpa_wide
                        [SAM~\citep{kirillov2023segment}; SEEM~\citep{zou2023segment}; SegGPT~\citep{wang2023seggpt}, gpa_work]
                    ]         
                ] 
                [Large Multimodal Models:\\ Training with LLM \S\ref{chp:training_with_llm}, for tree={fill=green!45,gpa}
                    [Image-to-Text Generation, gpa_wide
                        [Flamingo~\citep{alayrac2022flamingo}, gpa_work]
                    ]  
                    [Instruction Tuning in LLM, gpa_wide
                        [ChatGPT~\citep{chatgpt};\\Vicuna~\citep{vicuna}, gpa_work]
                    ]                 
                    [Instruction Tuning in LMM, gpa_wide
                        [Multimodal GPT-4~\citep{gpt4};\\
                        LLaVA~\citep{liu2023visual}; \\MiniGPT4~\citep{zhu2023minigpt4}, gpa_work]
                    ]   
                    [Emerging Topics, gpa_wide] 
                ]
                [Multimodal Agents: \\ Chaining Tools with LLM \S\ref{chp:chaining_with_llm}, for tree={fill=green!45,gpa}
                    [Multimodel Agent, gpa_wide
                        [VISPROG~\citep{gupta2022visual};
                         \\Visual ChatGPT~\citep{wu2023visual};
                         \\MM-REACT~\citep{yang2023mmreact}, gpa_work]
                    ]  
                    [Advanced Topics, gpa_wide]         
                ]
            ]
        ]
    \end{forest}
    \caption{An overview of the paper's structure, detailing Chapters \ref{chp:understanding}-\ref{chp:chaining_with_llm}.}
    \label{fig:paper_structure}
\end{figure*}
    
\section{Who Should Read this Paper?}
This paper is based on our CVPR 2023 tutorial,\footnote{\url{https://vlp-tutorial.github.io/2023/index.html}} with researchers in the computer vision and vision-language multimodal communities as our primary target audience.  It reviews the literature and explains topics to those who seek to learn the basics and recent advances in multimodal foundation models. The target audiences are graduate students, researchers and professionals who are not experts of multimodal foundation models but are eager to develop perspectives and learn the trends in the field. The structure of this paper is illustrated in Figure~\ref{fig:paper_structure}. It consists of 7 chapters. 

\begin{itemize}[leftmargin=*]
    \item Chapter~\ref{chp:intro} introduces the landscape of multimodal foundation model research, and presents a historical view on the transition of research from specialists to general-purpose assistants.
    \item Chapter~\ref{chp:understanding} introduces different ways to consume visual data, with a focus on how to learn a strong image backbone.
    
    \item Chapter~\ref{chp:generation} describes how to produce visual data that aligns with human intents.

    \item Chapter~\ref{chp:generalist} describes how to design unified vision models, with an interface that is interactive and promptable, especially when LLMs are not employed.
    
    \item Chapter~\ref{chp:training_with_llm} describes how to train an LLM in an end-to-end manner to consume visual input for understanding and reasoning.
    
    \item Chapter~\ref{chp:chaining_with_llm} describes how to chain multimodal tools with an LLM to enable new capabilities.
    
    \item Chapter~\ref{chp:conclusion} concludes the paper and discusses research trends.
\end{itemize}

\paragraph{Relations among Chapters 2-6.} Chapter~\ref{chp:understanding}-\ref{chp:chaining_with_llm} are the core chapters of this survey paper. An overview of the structure for these chapters are provided in Figure~\ref{fig:chp1_tasks}. We start with a discussion of two typical multimodal foundation models for specific tasks, including visual understanding in Chapter~\ref{chp:understanding} and visual generation in Chapter~\ref{chp:generation}. As the notion of multimodal foundation models are originally based on visual backbone/representation learning for understanding tasks, we first present a comprehensive review to the transition of image backbone learning methods, evolving from early supervised methods to the recent language-image contrastive methods, and extend the discussion on image representations from image-level to region-level and pixel-level (Chapter~\ref{chp:understanding}). Recently, generative AI is becoming increasingly popular, where vision generative foundation models have been developed. In Chapter~\ref{chp:generation}, we discuss large pre-trained text-to-image models, and various ways that the community leverage the generative foundation models to develop new techniques to make them better aligned with human intents. Inspired by the recent advances in NLP that LLMs serve as general-purpose assistants for a wide range of language tasks in daily life, the computer vision community has been anticipating and attempting to build general-purpose visual assistants. We discuss three different ways to build general-purpose assistants. Inspired by the spirit of LLMs, Chapter~\ref{chp:generalist} focuses on unifying different vision models of understanding and generation without explicitly incorporating LLMs in modeling. In contrast, Chapter~\ref{chp:training_with_llm} and Chapter~\ref{chp:chaining_with_llm} focus on embracing LLMs to build general-purpose visual assistants, by explicitly augmenting LLMs in modeling. Specifically,  Chapter~\ref{chp:training_with_llm} describes end-to-end training methods, and  Chapter~\ref{chp:chaining_with_llm} focuses on training-free approaches that chain various vision models to LLMs.

\paragraph{How to read the paper.} Different readers have different backgrounds, and may have different purposes of reading this paper. Here, we provide a few guidance. 
\begin{itemize}[leftmargin=*]
    \item Each chapter is mostly self-contained. If you have a clear goal and a clear research direction that you want to focus on, then just jump to the corresponding chapter. For example, if you are interested in building a mini prototype using OpenAI's multimodal GPT-4, then you can directly jump to Chapter~\ref{chp:training_with_llm}.
    
    \item If you are a beginner of multimodal foundation models, and are interested in getting a glimpse of the cutting-edge research, we highly recommend that you read the whole paper chapter by chapter in order, as the early chapters serve as the building blocks of later chapters, and each chapter provides the description of the key concepts to help you understand the basic ideas, and a comprehensive literature review that to help you grasp the landscape and state of the art.

    \item If you already have rich experience in multimodal foundation models and are familiar with the literature, feel free to jump to specific chapters you want to read. In particular, we include in most chapters a section to discuss advanced topics and sometimes provide our own perspectives, based on the up-to-date literature. For example, in Chapter~\ref{chp:chaining_with_llm}, we discuss several important aspects of multimodal agents in tool use, including tool creation and its connection to retrieval-augmented methods.
\end{itemize}



\section{Related Materials: Slide Decks and Pre-recorded Talks}

This survey paper extends what we present in the CVPR 2023 tutorial by covering the most recent advances in the field. Below, we provide a list of slide decks and pre-recorded talks, which are related to the topics in each chapter, for references.

\begin{itemize}[leftmargin=*]
    \item \textbf{Chapter 2}: 
    \href{https://datarelease.blob.core.windows.net/tutorial/vision_foundation_models_2023/slides/Zhe_CVPR2023_Tutorial.pdf}{ Visual and Vision-Language Pre-training}~(\href{https://youtu.be/hE135guhTQo}{Youtube}, \href{https://www.bilibili.com/video/BV1Hk4y1M77T/}{Bilibili})
        
    \item \textbf{Chapter 3}: 
    \href{https://datarelease.blob.core.windows.net/tutorial/vision_foundation_models_2023/slides/Zhengyuan_Tutorial_T2I2023.pdf}{Alignments in Text-to-Image Generation}~(\href{https://youtu.be/iixMLxeuOqU}{Youtube}, \href{https://www.bilibili.com/video/BV14P411v7Un/}{Bilibili})
        
    \item \textbf{Chapter 4}:
     \href{https://datarelease.blob.core.windows.net/tutorial/vision_foundation_models_2023/slides/Jianwei_CVPR2023_Tutorial.pdf}{From Representation to Interface: The Evolution of Foundation for Vision Understanding}~(\href{https://youtu.be/wIcTyutOlDs}{Youtube}, \href{https://www.bilibili.com/video/BV1ds4y1k7pj/}{Bilibili})
        
    \item \textbf{Chapter 5}: 
     \href{https://datarelease.blob.core.windows.net/tutorial/vision_foundation_models_2023/slides/Chunyuan_cvpr2023_tutorial_lmm.pdf}{Large Multimodal Models}~(\href{https://youtu.be/mkI7EPD1vp8}{Youtube}, \href{https://www.bilibili.com/video/BV1Ng4y1T7v3/}{Bilibili})
        
    \item \textbf{Chapter 6}: 
     \href{https://datarelease.blob.core.windows.net/tutorial/vision_foundation_models_2023/slides/Linjie_Multimodal%20Agents.pptx}{Multimodal Agents: Chaining Multimodal Experts with LLMs}~(\href{https://youtu.be/Wb5ZkZUNYc4}{Youtube}, \href{https://www.bilibili.com/video/BV1Dg4y1K7wJ/}{Bilibili})

\end{itemize}

\chapter{Visual Understanding}
\label{chp:understanding}
\begin{wrapfigure}{r}{4cm}
  \centering
  \vspace{-6cm}
  \includegraphics[width=1.0\linewidth]{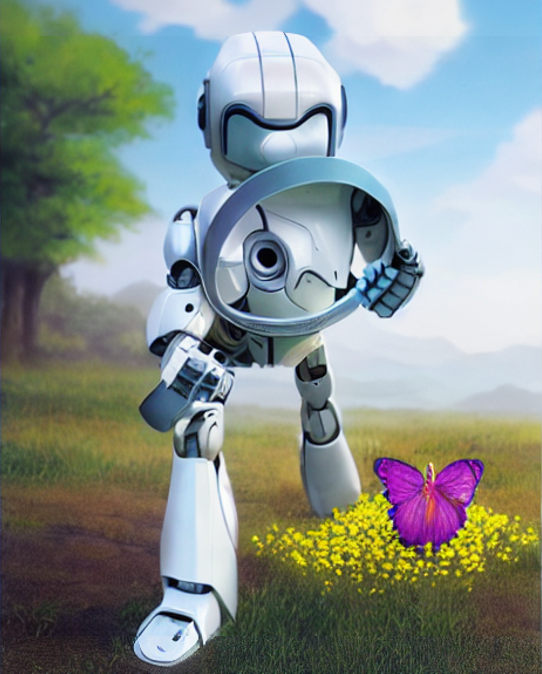}
\end{wrapfigure}

Over the past decade, the research community has devoted
significant efforts to study the acquisition of high-quality, general-purpose image representations. This is essential to build vision foundation models, as pre-training a strong vision backbone to learn image representations is fundamental to all types of computer vision downstream tasks, ranging from image-level (\emph{e.g.}, image classification~\citep{krizhevsky2012imagenet}, image-text retrieval~\citep{frome2013devise}, image captioning~\citep{chen2015microsoftcoco}), region-level (\emph{e.g.}, object detection~\citep{girshick2015fast}, phrase grounding~\citep{plummer2015flickr30k}), to pixel-level (\emph{e.g.}, semantic/instance/panoptic segmentation~\citep{long2015fully,hafiz2020survey,kirillov2019panoptic}) tasks. 

In this chapter, we present how image representations can be learned, either using supervision signals mined inside the images, or through using language supervision of image-text datasets mined from the Web. Specifically, Section~\ref{sec:overview} presents an overview of different learning paradigms, including supervised pre-training, contrastive language-image pre-training (CLIP), and image-only self-supervised learning. Section~\ref{sec:sup_pt} discusses supervised pre-training. Section~\ref{sec:clip} focuses on CLIP. Section~\ref{sec:ssl} discusses image-only self-supervised learning, including contrastive learning, non-contrastive learning, and masked image modeling. Given the various learning approaches to training vision foundation models, Section~\ref{sec:synergy} reviews how they can be incorporated for better performance. Lastly, Section~\ref{sec:visual_understanding_advanced_topics} discusses how vision foundation models can be used for finer-grained visual understanding tasks, such as fusion-encoder-based pre-training for image captioning and visual question answering that require multimodal fusion, region-level pre-training for grounding, and pixel-level pre-training for segmentation. 

\begin{figure*}[t!]
  \centering
    \includegraphics[width=0.98\linewidth]{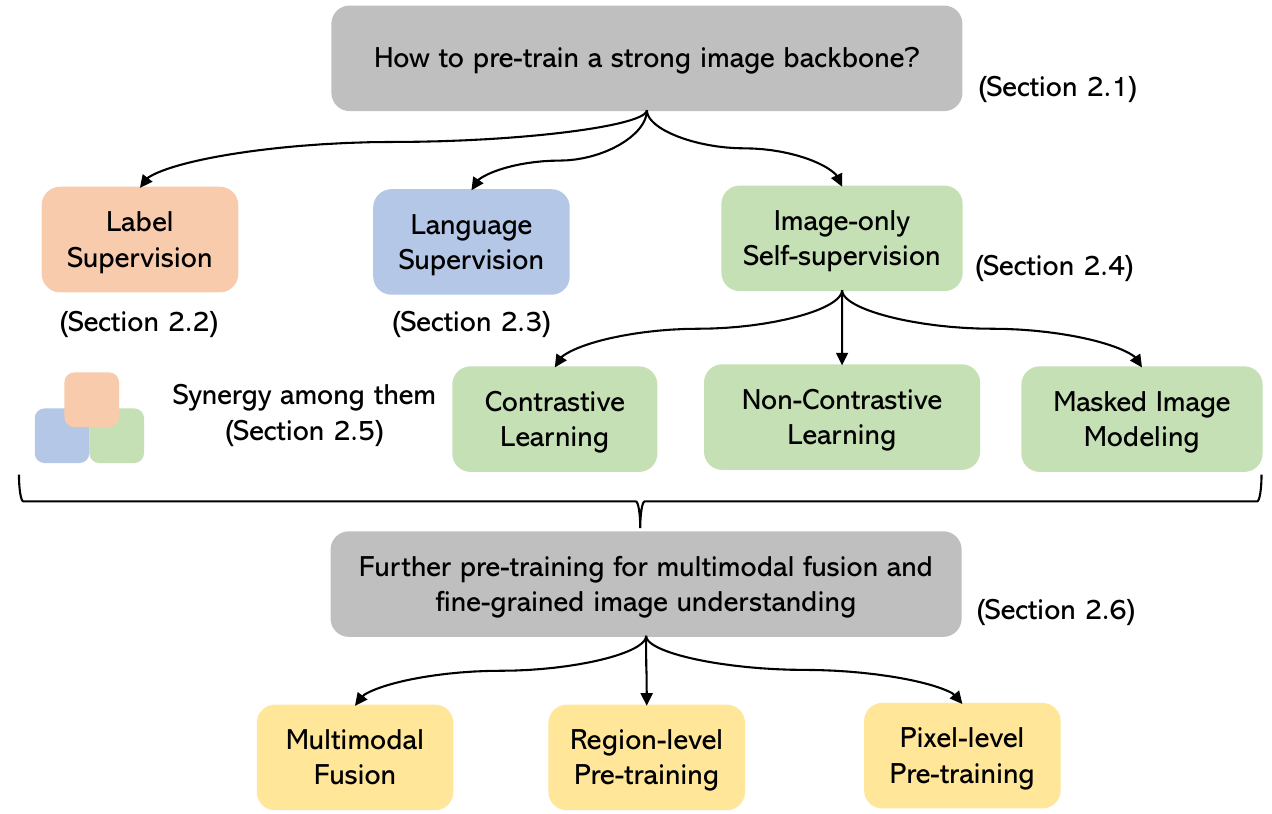}
  \caption{An overview of the structure of Chapter \ref{chp:understanding}.}
  \label{fig:chp2_flow}
\end{figure*}

\section{Overview}\label{sec:overview}
There is a vast amount of literature on various methods of learning general-purpose vision backbones. As illustrated in Figure~\ref{fig:chp2_flow}, we group these methods into three categories, depending on the types of supervision signals used to train the models, including:  
\begin{itemize}[leftmargin=*]
    \item \textbf{Label supervision}: Arguably, the most well-studied image representation learning methods are based on label supervisions (typically in the form of image classification)~\citep{sun2017revisiting}, where datasets like ImageNet~\citep{krizhevsky2012imagenet} and ImageNet21K~\citep{ridnik2021imagenet} have been popular, and larger-scale proprietary datasets are also used in industrial labs~\citep{sun2017revisiting,singh2022revisiting,zhai2022scaling,wu2023mofi}. 
    \item \textbf{Language supervision}: Another popular approach to learning image representations leverages weakly supervised signals from text, which is easy to acquire in large scale. For instance, CLIP~\citep{radford2021learning} and ALIGN~\citep{jia2021scaling} are pre-trained using a contrastive loss and billions of image-text pairs mined from the internet. The resultant models achieve strong zero-shot performance on image classification and image-text retrieval, and the learned image and text encoders have been widely used for various downstream tasks and allow traditional computer vision models to perform open-vocabulary CV tasks~\citep{gu2021open,ghiasi2021open,qian2022multimodal,ding2022open,liang2023open,zhang2023simple,zou2023generalized,minderer2022simple}.
    \item \textbf{Image-only self-supervision}: There is also a vast amount of literature on exploring image-only self-supervised learning methods to learn image representations. As the name indicates, the supervision signals are mined from the images themselves, and popular methods range from contrastive learning~\citep{chen2020simple,he2020momentum}, non-contrastive learning~\citep{grill2020bootstrap,chen2021exploring,caron2021emerging}, to masked image modeling~\citep{bao2021beit,he2022masked}. 
\end{itemize}

\begin{figure*}[t!]
  \centering
    \includegraphics[width=0.98\linewidth]{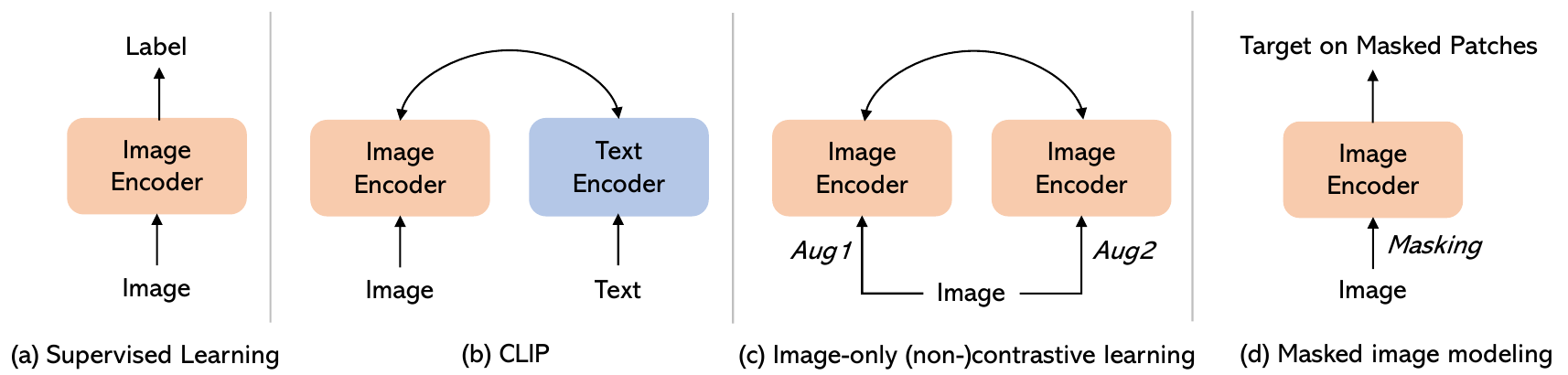}
  \caption{A high-level overview of different approaches to learn general image representations, including supervised learning~\citep{krizhevsky2012imagenet}, contrastive language-image pre-training~\citep{radford2021learning,jia2021scaling}, and image-only self-supervised learning, including contrastive learning~\citep{chen2020simple,he2020momentum}, non-contrastive learning~\citep{grill2020bootstrap,chen2021exploring}, and masked image modeling~\citep{bao2021beit,he2022masked}.}
  \label{fig:chp2_overview}
\end{figure*}

\begin{figure*}[t!]
  \centering
    \includegraphics[width=0.98\linewidth]{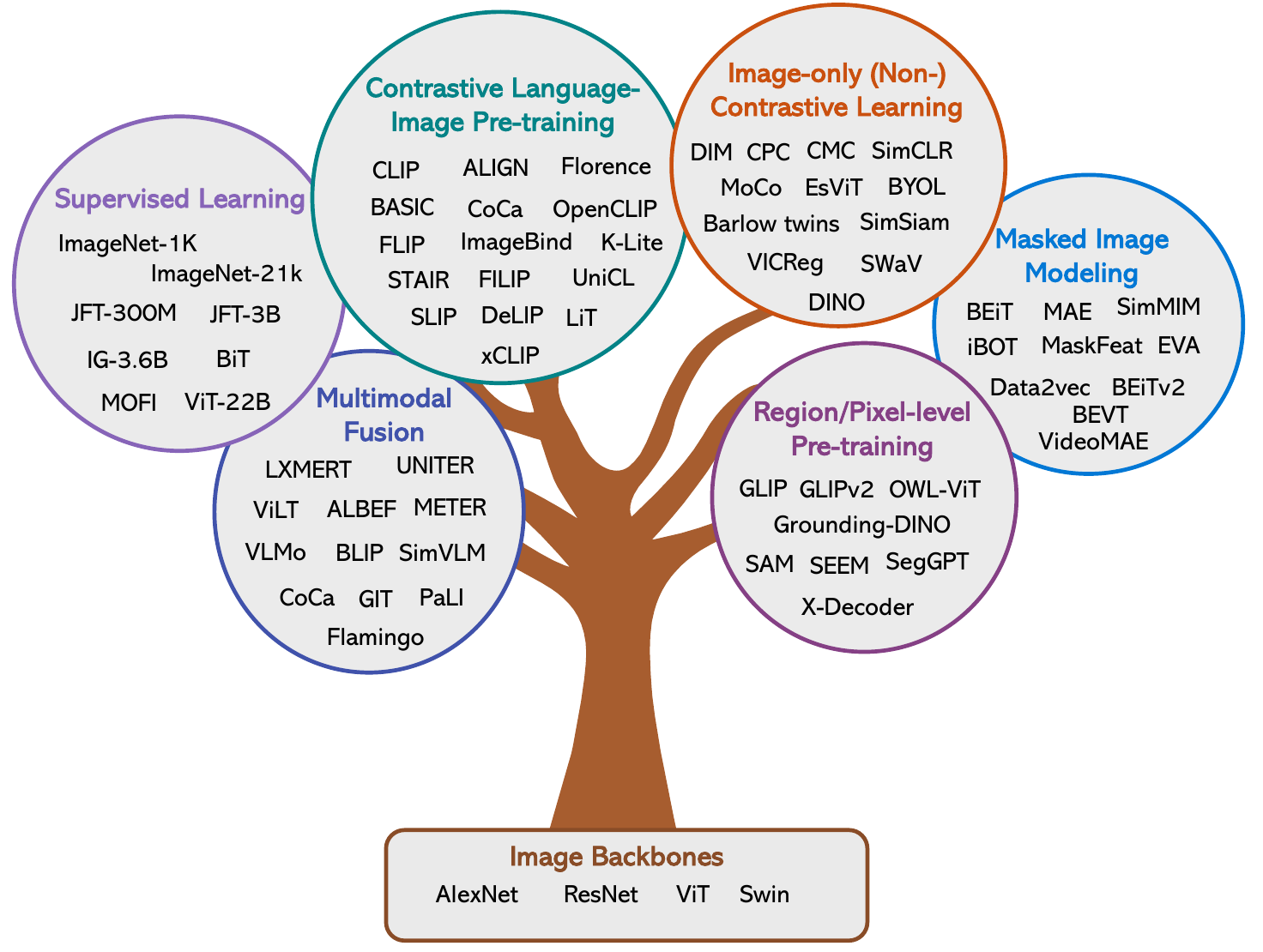}
  \caption{An overview of the topics covered in this chapter and representative works in each topic. We start from supervised learning and CLIP, and then move on to image-only self-supervised learning, including contrastive learning, non-contrastive learning, and masked image modeling. Lastly, we discuss pre-training methods that empower multimodal fusion, region-level and pixel-level image understanding.}
  \label{fig:chp2_tree}
\end{figure*}

An illustration of these learning methods is shown in Figure~\ref{fig:chp2_overview}.
Besides the methods of pre-training image backbones, we will also discuss pre-training methods that allow multimodal fusion (\emph{e.g.}, CoCa~\citep{yu2022coca}, Flamingo~\citep{alayrac2022flamingo}), region-level and pixel-level image understanding (\emph{e.g.}, GLIP~\citep{li2021grounded} and SAM~\citep{kirillov2023segment}). These methods typically rely on a pre-trained image encoder or a pre-trained image-text encoder pair. Figure~\ref{fig:chp2_tree} shows an overview of the topics covered in this chapter and some representative works in each topic.


\section{Supervised Pre-training}\label{sec:sup_pt}

Supervised pre-training on large-scale human-labeled datasets, such as ImageNet~\citep{krizhevsky2012imagenet} and ImageNet21K~\citep{ridnik2021imagenet}, has emerged as a widely adopted approach to acquiring transferable visual representations. It aims to map an image to a discrete label, which is associated with a visual concept. This approach has greatly expedited progress in designing various vision backbone architectures (\emph{e.g.}, AlexNet~\citep{krizhevsky2012imagenet}, ResNet~\citep{he2016deep}, vision transformer~\citep{dosovitskiy2020image}, and Swin transformer~\citep{liu2021swin}), and is the testbed for all the modern vision backbones. It also powered computer vision tasks across the whole spectrum, ranging from image classification, object detection/segmentation, visual question answering, image captioning, to video action recognition. However, the effectiveness of learned representations is often limited by the scale and diversity of supervisions in pre-training datasets, as human annotation is expensive. 

\paragraph{Large-scale datasets.}
For larger-scale pre-training, noisy labels can be derived in large quantities from image-text pairs crawled from the Web. Using noisy labels, many industrial labs have successfully constructed comprehensive classification datasets using semi-automatic pipelines, such as JFT~\citep{sun2017revisiting,zhai2022scaling} and I2E~\citep{wu2023mofi}, or by leveraging proprietary data like Instagram hashtags~\citep{singh2022revisiting}. The statistics of existing large-scale image classification datasets are shown in Table~\ref{tab:chp2_img_class_datasets}. The labels are typically in the form of fine-grained image entities with a long-tailed distribution. 
Though classical, this approach has been very powerful for learning universal image representations. For example, JFT-300M~\citep{sun2017revisiting}  has been used for training the BiT (``Big Transfer'') models~\citep{kolesnikov2020big}, and JFT-3B~\citep{zhai2022scaling} has been used to scale up the training of a plain vision transformer~\citep{dosovitskiy2020image} to 22B in model size. 
LiT~\citep{zhai2021lit} proposes to first learn the image backbone on JFT-3B~\citep{zhai2022scaling}, and keep it frozen and learn another text tower to align the image and text embedding space to make the model open-vocabulary and is capable of performing zero-shot image classification. 

\paragraph{Model training.} There are many loss functions that can be used to promote embedding properties (\emph{e.g.}, separability)~\citep{MetricLearningRC}. For example, the large margin loss~\citep{Wang2018CosFaceLM} is used for MOFI training~\citep{wu2023mofi}.
Furthermore, if the datasets have an immense number of labels (can potentially be over 2 million as in MOFI~\citep{wu2023mofi}), predicting all the labels in each batch becomes computationally costly. In this case, a fixed number of labels is typically used for each batch, similar to sampled softmax~\citep{gutmann2010noise}.

\begin{table}[t!]
    \small
    \centering
    \begin{tabular}{l|rr}
        \toprule
        Dataset & \# Images & \# Classes \\
        \midrule
        ImageNet-1K \citep{russakovsky2015imagenet} & 1.2M & 1K \\ 
        ImageNet-21K \citep{ridnik2021imagenet} & 14M & 21K \\
        JFT-300M \citep{sun2017revisiting} & 300M & ~18K \\
        JFT-3B \citep{zhai2022scaling} & 3B & ~30K \\
        IG-3.6B \citep{singh2022revisiting} & 3.6B & ~27K \\
        I2E \citep{wu2023mofi} & 1.1B & 2M \\
        \bottomrule
    \end{tabular}
    \vspace{1mm}
    \caption{Statistics of existing large-scale image classification datasets.}
    \label{tab:chp2_img_class_datasets}
\end{table}

\begin{figure}[t!]
\begin{center}
\includegraphics[width=\textwidth]{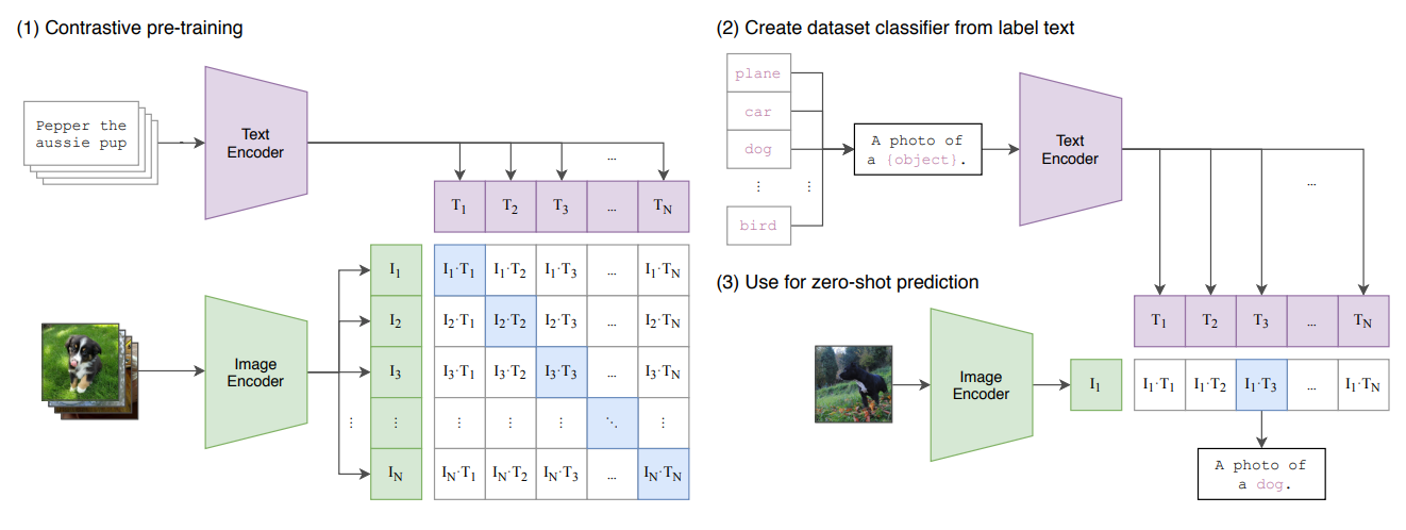}
\end{center}
\caption{Illustration of contrastive language-image pre-training, and how the learned model can be used for zero-shot image classification. Image credit: \cite{radford2021learning}.}
\label{fig:chp2_clip}
\end{figure}

\section{Contrastive Language-Image Pre-training}\label{sec:clip}

\subsection{Basics of CLIP Training}
Language is a richer form of supervision than classical closed-set labels.
Rather than deriving noisy label supervision from web-crawled image-text datasets, the alt-text can be directly used for learning transferable image representations, which is the spirit of contrastive language-image pre-training (CLIP)~\citep{radford2021learning}.
In particular, models trained in this way, such as ALIGN~\citep{jia2021scaling}, Florence~\citep{yuan2021florence}, BASIC~\citep{pham2021combined}, and OpenCLIP~\citep{ilharco_gabriel_2021_5143773}, have showcased impressive zero-shot image classification and image-text retrieval capabilities by mapping images and text into a shared embedding space. Below, we discuss how the CLIP model is pre-trained and used for zero-shot prediction.
\begin{itemize}[leftmargin=*]
    \item \textbf{Training}: As shown in Figure~\ref{fig:chp2_clip}(1), CLIP is trained via simple contrastive learning. CLIP is an outstanding example of ``\emph{simple algorithms that scale well}''~\citep{li2023scaling}. To achieve satisfactory performance, model training needs to be scaled along three dimensions: batch size, data size, and model size~\citep{pham2021combined}. Specifically, the typical batch size used for CLIP training can be 16k or 32k. The number of image-text pairs in the pre-training datasets is frequently measured in billions rather than millions. A vision transformer trained in this fashion can typically vary from 300M (Large) to 1B (giant) in model size. 
    \item \textbf{Zero-shot prediction}: As shown in Figure~\ref{fig:chp2_clip}~(2) and (3), CLIP empowers zero-shot image classification via reformatting it as a retrieval task and considering the semantics behind labels. It can also be used for zero-shot image-text retrieval by its design. Besides this, the aligned image-text embedding space makes it possible to make all the traditional vision models open vocabulary and has inspired a rich line of work on open-vocabulary object detection and segmentation~\citep{li2021grounded,zhang2022glipv2,zou2023generalized,zhang2023simple}.
\end{itemize}

\subsection{CLIP Variants}
Since the birth of CLIP, there have been tons of follow-up works to improve CLIP models, as to be discussed below. 
We do not aim to provide a comprehensive literature review of all the methods, but focus on a selected set of topics.  

\paragraph{Data scaling up.} Data is the fuel for CLIP training. For example, OpenAI's CLIP was trained on 400M image-text pairs mined from the web, while ALIGN used a proprietary dataset consisting of 1.8B image-text pairs. In BASIC~\citep{pham2021combined}, the authors have carefully studied the scaling among three dimensions: batch size, data size, and model size. However, most of these large-scale datasets are not publicly available, and training such models requires massive computing resources.

In academic settings, researchers~\citep{li2022elevater} have advocated the use of a few millions of image-text pairs for model pre-training, such as CC3M~\citep{sharma2018conceptual}, CC12M~\citep{changpinyo2021conceptual},  YFCC~\citep{thomee2016yfcc100m}. 
Relatively small-scale image-text datasets that are publicly available include SBU~\citep{ordonez2011im2text}, RedCaps~\citep{desai2021redcaps}, and WIT~\citep{srinivasan2021wit}. 
Large-scale public available image-text datasets include Shutterstock~\citep{nguyen2022quality}, LAION-400M~\citep{schuhmann2021laion}, COYO-700M~\citep{coyo700m}, and LAION-2B~\citep{schuhmann2022laion}, to name a few. For example, LAION-2B~\citep{schuhmann2022laion} has been used by researchers to study the reproducible scaling laws for CLIP training~\citep{cherti2023reproducible}.  

Interestingly, in search of the next-generation image-text datasets, in DataComp~\citep{gadre2023datacomp}, instead of fixing the dataset and designing different algorithms, the authors propose to select and rank datasets using the fixed CLIP training method. Besides paired image-text data mined from the Web for CLIP training, inspired by the interleaved image-text dataset M3W introduced in Flamingo~\citep{alayrac2022flamingo}, there have been recent efforts of collecting interleaved image-text datasets, such as MMC4~\citep{zhu2023multimodal} and OBELISC~\citep{laurenccon2023obelisc}.


\begin{figure}[t!]
\begin{center}
\includegraphics[width=\textwidth]{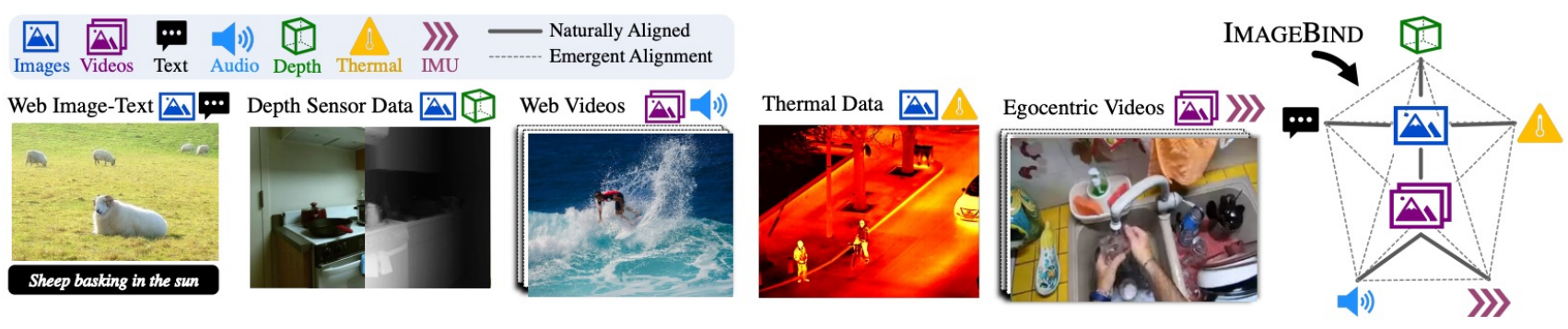}
\end{center}
\vspace{-0.2cm}
\caption{ImageBind~\citep{girdhar2023imagebind} proposes to link a total of six modalities into a common embedding space via leveraging pre-trained CLIP models, enabling new emergent alignments and capabilities. Image credit: \cite{girdhar2023imagebind}.}
\label{fig:chp2_imagebind}
\end{figure}

\paragraph{Model design and training methods.} CLIP training has been significantly improved. Below, we review some representative works.
\begin{itemize}[leftmargin=*]
    \item \textbf{Image tower}: On the image encoder side, FLIP~\citep{li2023scaling} proposes to scale CLIP training via masking. By randomly masking out image patches with a high masking ratio, and only encoding the visible patches as in MAE~\citep{he2022masked}, the authors demonstrate that masking can improve training efficiency without hurting the performance. The method can be adopted for all CLIP training. \cite{cao2023less} found that filtering out samples that contain text regions in the image improves CLIP training efficiency and robustness.
    \item \textbf{Language tower}: On the language encoder side, K-Lite~\citep{shen2022k} proposes to use external knowledge in the form of Wiki definition of entities together with the original alt-text for contrastive pre-training. Empirically, the use of enriched text descriptions improves the CLIP performance. LaCLIP~\citep{fan2023improving} shows that CLIP can be improved via rewriting the noisy and short alt-text using large language models such as ChatGPT. 
    \item \textbf{Interpretability}: The image representation is typically a dense feature vector. In order to improve the interpretability of the shared image-text embedding space, 
    STAIR~\citep{chen2023stair} proposes to map images and text to a high-dimensional, sparse, embedding space, where each dimension in the sparse embedding is a (sub-)word in a large dictionary in which the predicted non-negative scalar corresponds to the weight associated with the token. The authors show that STAIR achieves better performance than the vanilla CLIP with improved interpretability. 
    \item \textbf{More modalities}: The idea of contrastive learning is general, and can go beyond just image and text modalities. For example, as shown in Figure~\ref{fig:chp2_imagebind}, ImageBind~\citep{girdhar2023imagebind} proposes to encode six modalities into a common embedding space, including images, text, audio, depth, thermal, and IMU modalities. In practice, a pre-trained CLIP model is used and kept frozen during training, which indicates that other modality encoders are learned to align to the CLIP embedding space, so that the trained model can be applied to new applications such as audio-to-image generation and multimodal LLMs (\emph{e.g.}, PandaGPT~\citep{su2023pandagpt}).
\end{itemize}

\begin{figure}[t!]
\begin{center}
\includegraphics[width=\textwidth]{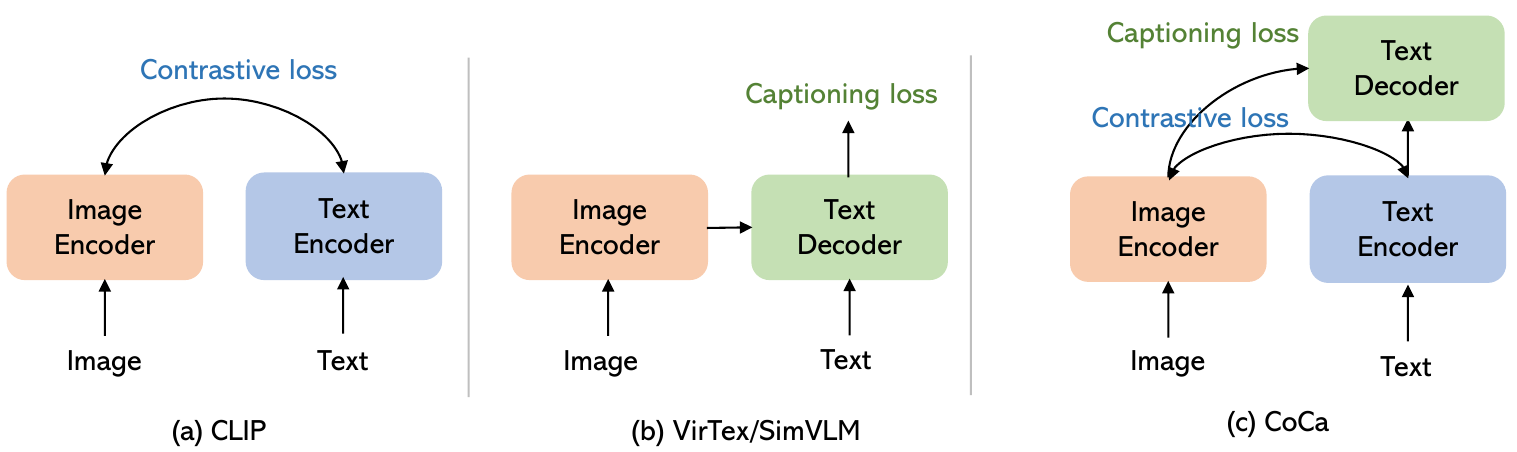}
\end{center}
\vspace{-0.2cm}
\caption{A high-level comparison of contrastive loss and captioning loss for image encoder pre-training. (a) CLIP~\citep{radford2021learning} uses contrastive loss alone for pre-training, which enables zero-shot image classification and has demonstrated strong scaling behavior. (b) VirTex~\citep{desai2021virtex} uses captioning loss alone for pre-training. SimVLM~\citep{wang2021simvlm} uses prefix language modeling for pre-training in a much larger scale.  The model architecture is similar to multimodal language models (\emph{e.g.}, GIT~\citep{wang2022git} and Flamingo~\citep{alayrac2022flamingo}), but VirTex and SimVLM aim to pre-train the image encoder from scratch. (c) CoCa~\citep{yu2022coca} uses both contrastive and captioning losses for pre-training. The model architecture is similar to ALBEF~\citep{li2021align}, but CoCa aims to pre-train the image encoder from scratch, instead of using a pre-trained one.}
\label{fig:clip_virtex_coca}
\end{figure}

\paragraph{Objective function.} The use of contrastive loss alone is powerful, especially when the model is scaled up. However, other objective functions can also be applied.
\begin{itemize}[leftmargin=*]
    \item \textbf{Fine-grained supervision}: Instead of using a simple dot-product to calculate the similarity of an image-text pair, the supervision can be made more fine-grained via learning word-patch alignment. In FILIP~\citep{yao2021filip}, the authors propose to first compute the loss by calculating the token-wise similarity, and then aggregating the matrix by max-pooling for word-patch alignment. 
    \item \textbf{Contrastive captioner}: Besides the contrastive learning branch, CoCa~\citep{yu2022coca} (shown in Figure~\ref{fig:clip_virtex_coca}(c)) adds a generative loss to improve performance and allow new capabilities that require multimodal fusion (\emph{e.g.}, image captioning and VQA). This is similar to many fusion-encoder-based vision-language models such as ALBEF~\citep{li2021align}, but with the key difference in that CoCa aims to learn a better image encoder from scratch. A detailed discussion on multimodal fusion is in Section~\ref{sec:multimodal_fusion}.
    \item \textbf{Captioning loss alone}: How about using the captioning loss alone to pre-train an image encoder? Actually, before CLIP was invented, VirTex~\citep{desai2021virtex} (shown in Figure~\ref{fig:clip_virtex_coca}(b)) and ICMLM~\citep{sariyildiz2020learning} learn encoders using a single image captioning loss, but the scale is very small (restricted to COCO images) and the performance is poor. CLIP also shows that contrastive pre-training is a much better choice. In SimVLM~\citep{wang2021simvlm}, the authors found that the learned image encoder was not as competitive as CLIP. However, in the recent work Cap/CapPa~\citep{tschannen2023image}, the authors argue that image captioners are scalable vision learners, too. Captioning can exhibit the same or even better scaling behaviors. 
    \item \textbf{Sigmoid loss for language-image pre-training}: Unlike standard contrastive learning with softmax normalization, \cite{zhai2023sigmoid} uses a simple pairwise sigmoid loss for image-text pre-training, which operates on image-text pairs and does not require a global view of the pairwise similarities for normalization. The authors show that the use of simple sigmoid loss can also achieve strong performance on zero-shot image classification.
\end{itemize}

\begin{figure}[t!]
\begin{center}
\includegraphics[width=\textwidth]{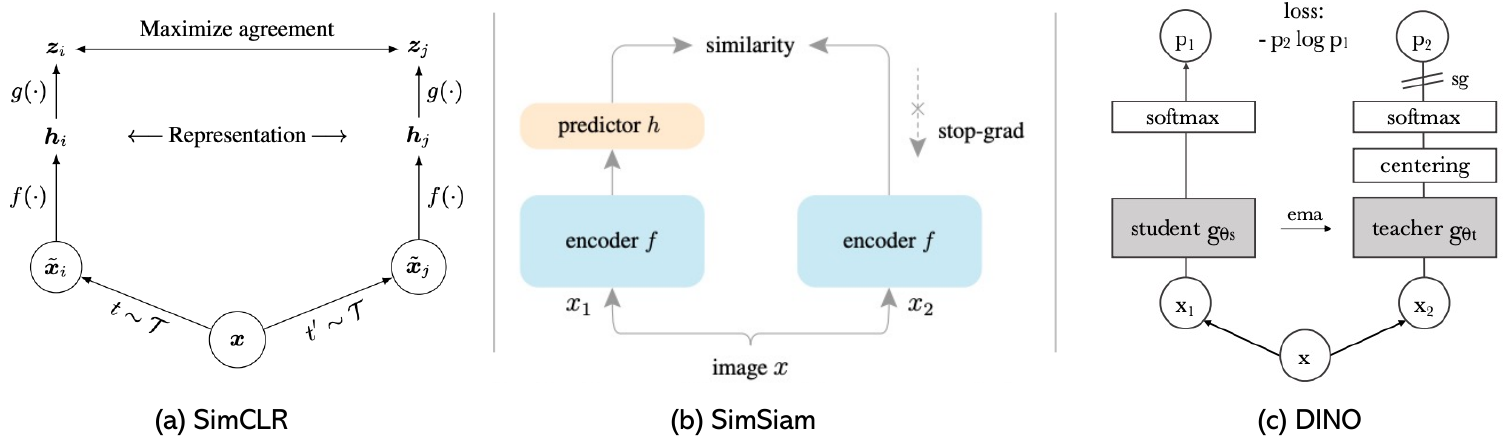}
\end{center}
\caption{Overview of SimCLR~\citep{chen2020simple}, SimSiam~\citep{chen2021exploring}, and DINO~\citep{caron2021emerging} for self-supervised image representation learning. SimCLR uses contrastive learning for model training, while SimSiam and DINO explores non-contrastive learning methods. Image credit: \cite{chen2020simple}, \cite{chen2021exploring}, \cite{caron2021emerging}.}
\label{fig:chp2_ssl}
\end{figure}

\section{Image-Only Self-Supervised Learning}\label{sec:ssl}
Now, we shift our focus to image-only self-supervised learning, and divide the discussion into three parts: ($i$) contrastive learning, ($ii$) non-contrastive learning, and ($iii$) masked image modeling.

\subsection{Contrastive and Non-contrastive Learning}

\paragraph{Contrastive learning.} The core idea of contrastive learning~\citep{gutmann2010noise,arora2019theoretical} is to promote the positive sample pairs and repulse the negative sample pairs. Besides being used in CLIP, contrastive learning has also been a popular concept in self-supervised image representation learning~\citep{wu2018unsupervised,ye2019unsupervised,tian2020contrastive,chen2020simple,he2020momentum,misra2020self,chen2020improved}. It has been shown that the contrastive objective, known as the InfoNCE loss~\citep{oord2018representation}, can be interpreted as maximizing the
lower bound of mutual information between different views of the data~\citep{hjelm2018learning,bachman2019learning,henaff2020data}.

In a nutshell, all the image-only contrastive learning methods (\emph{e.g.}, SimCLR~\citep{chen2020simple}, see Figure~\ref{fig:chp2_ssl}(a), MoCo~\citep{he2020momentum}, SimCLR-v2~\citep{chen2020big}, MoCo-v2~\citep{chen2020improved}) share the same high-level framework, detailed below.
\begin{itemize}[leftmargin=*]
    \item Given one image, two separate data augmentations are applied;
    \item A base encoder is followed by a project head, which is trained to maximize agreement using a contrastive loss (\emph{i.e.}, they are from the same image or not);
    \item The project head is thrown away for downstream tasks.
\end{itemize}
However, a caveat of contrastive learning is the requirement of a large number of negative samples. These samples can be maintained in a memory bank~\citep{wu2018unsupervised}, or directly from the current batch~\citep{chen2020simple}, which suggests the requirement of a large batch size. MoCo~\citep{he2020momentum} maintains a queue of negative samples and turns one branch into a momentum encoder to improve the consistency of the queue. Initially, contrastive learning was primarily studied for pre-training convolutional networks. However, with the rising popularity of vision transformers (ViT), researchers have also explored its application in the context of ViT.~\citep{chen2021empirical,li2021efficient,xie2021self}.

\paragraph{Non-contrastive learning.}
Recent self-supervised learning methods do not depend on negative samples. The use of negatives is replaced by asymmetric architectures (\emph{e.g.}, BYOL~\citep{grill2020bootstrap}, SimSiam~\citep{chen2021exploring}), dimension de-correlation (\emph{e.g.}, Barlow twins~\citep{zbontar2021barlow}, VICReg~\citep{bardes2021vicreg}, Whitening~\citep{ermolov2021whitening}), and clustering (\emph{e.g.}, SWaV~\citep{caron2020unsupervised}, DINO~\citep{caron2021emerging}, \cite{caron2018deep,amrani2022self,assran2022masked,wang2023self}), \emph{etc}. 

For example, as illustrated in Figure~\ref{fig:chp2_ssl}(b), in SimSiam~\citep{chen2021exploring}, two augmented views of a single image are processed by an identical encoder network. Subsequently, a prediction MLP is applied to one view, while a stop-gradient operation is employed on the other. The primary objective of this model is to maximize the similarity between the two views. It is noteworthy that SimSiam relies on neither negative pairs nor a momentum encoder. 

Another noteworthy method, known as DINO~\citep{caron2021emerging} and illustrated in Figure~\ref{fig:chp2_ssl}(c), takes a distinct approach. 
DINO involves feeding two distinct random transformations of an input image into both the student and teacher networks. Both networks share the same architecture but have different parameters. The output of the teacher network is centered by computing the mean over the batch. Each network outputs a feature vector that is normalized with a temperature softmax applied to the feature dimension. The similarity between these features is quantified using a cross-entropy loss. Additionally, a stop-gradient operator is applied to the teacher network to ensure that gradients propagate exclusively through the student network. Moreover, DINO updates the teacher's parameters using an exponential moving average of the student's parameters.

\begin{figure}[t!]
\begin{center}
\includegraphics[width=0.9\textwidth]{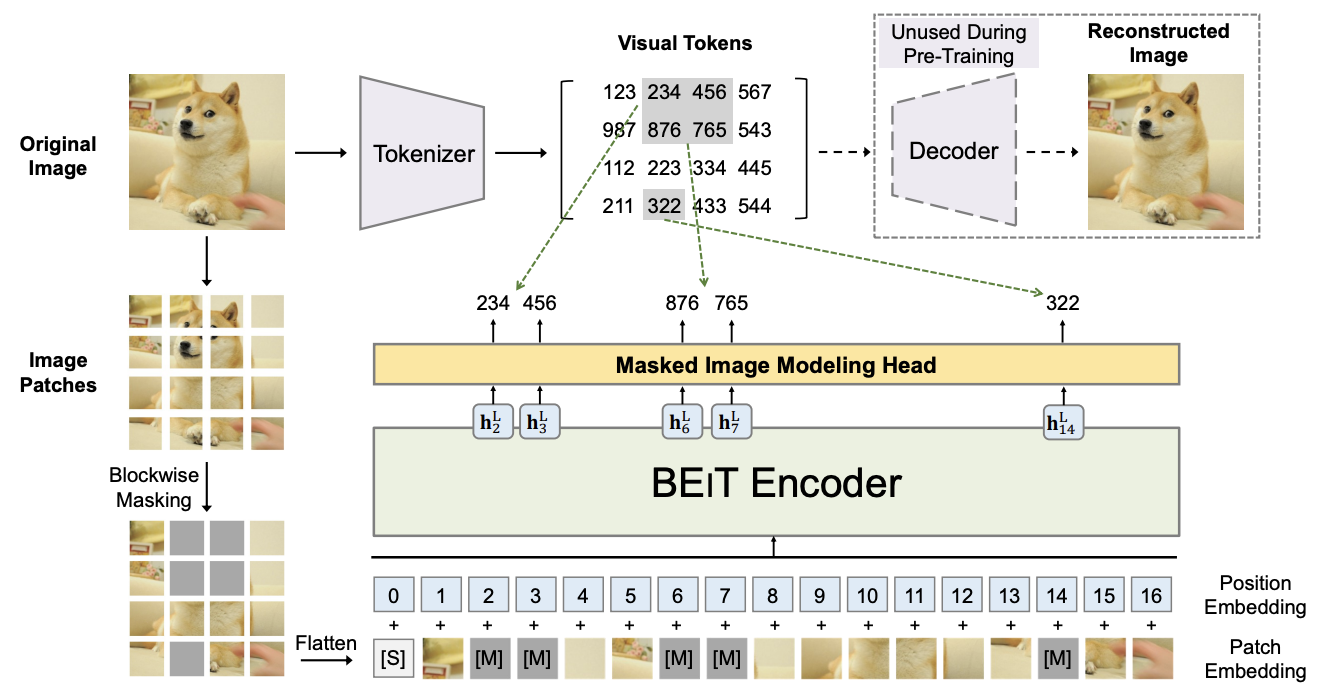}
\end{center}
\caption{Overview of BEiT pre-training for image transformers. Image credit: \cite{bao2021beit}.}
\label{fig:chp2_beit}
\end{figure}

\subsection{Masked Image Modeling}

Masked language modeling~\citep{devlin2018bert} is a powerful pre-training task that has revolutionized the NLP research. To mimic the success of BERT pre-training for NLP, the pioneering work BEiT~\citep{bao2021beit}, as illustrated in Figure~\ref{fig:chp2_beit}, proposes to perform masked image modeling (MIM) to pre-train image transformers. Specifically, 
\begin{itemize}[leftmargin=*]
    \item \textbf{Image tokenizer}: In order to perform masked token prediction, an image tokenizer is required to tokenize an image into discrete visual tokens, so that these tokens can be treated just like an additional set of language tokens. Some well-known learning methods for image tokenziers include VQ-VAE~\citep{van2017neural}, VQ-VAE-2~\citep{razavi2019generating}, VQ-GAN~\citep{esser2021taming}, ViT-VQGAN~\citep{yu2021vector}, \emph{etc}. These image tokenizers have also been widely used for autoregressive image generation, such as DALLE~\citep{ramesh2021dalle}, Make-A-Scene~\citep{gafni2022make}, Parti~\citep{yu2022scaling}, to name a few. 
    \item \textbf{Mask-then-predict}: The idea of MIM is conceptually simple: models accept the corrupted input image (\emph{e.g.}, via random masking of image patches), and then predict the target of the masked content (\emph{e.g.}, discrete visual tokens in BEiT). As discussed in iBOT~\citep{zhou2021ibot}, this training procedure can be understood as knowledge distillation between the image tokenizer (which serves as the teacher) and the BEiT encoder (which serves as the student), while the student only sees partial of the image. 
\end{itemize}

\begin{figure}[t!]
\begin{center}
\includegraphics[width=\textwidth]{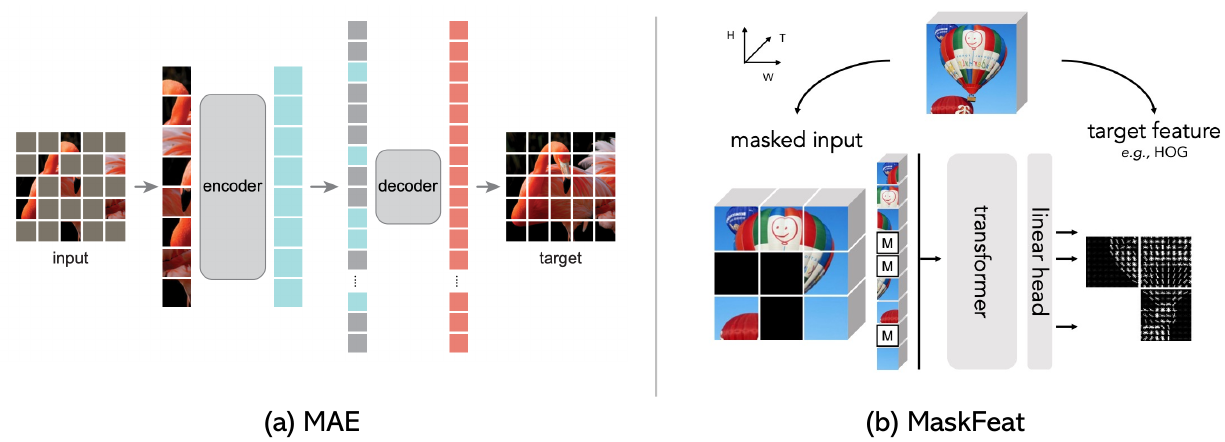}
\end{center}
\vspace{-2mm}
\caption{Illustration of Masked Autoencoder (MAE)~\citep{he2022masked} that uses raw pixel values for MIM training, and MaskFeat~\citep{wei2021masked-feat} that uses different features as the targets. HOG, a hand-crafted feature descriptor, was found to work particularly well in terms of both performance and efficiency. Image credit: \cite{he2022masked} and \cite{wei2021masked-feat}.}
\label{fig:chp2_mae_maskfeat}
\end{figure}

\paragraph{Targets.} In \cite{peng2022unified}, the authors have provided a unified view of MIM: a teacher model, a normalization layer, a student model, an MIM head, and a proper loss function. The most significant difference among all these models lies in the reconstruction targets, which can be pixels, discrete image tokens, features from pre-trained models, and outputs from the momentum updated teacher. Specifically, the targets can be roughly grouped into two categories.
\begin{itemize}[leftmargin=*]
    \item \textbf{Low-level pixels/features as targets}:  MAE~\citep{he2022masked}, SimMIM~\citep{xie2022simmim}, ConvMAE~\citep{gao2022convmae}, HiViT~\citep{zhang2022hivit}, and GreenMIM~\citep{huang2022green} leverage either original or normalized pixel values as the target for MIM. These methods have typically explored the use of a plain Vision Transformer~\citep{dosovitskiy2020image} or the Swin Transformer~\citep{liu2021swin} as the backbone architecture.
    MaskFeat~\citep{wei2021masked-feat} introduced the Histogram of Oriented Gradients (HOG) feature descriptor as the target for MIM (see Figure~\ref{fig:chp2_mae_maskfeat}(b)). Meanwhile, Ge$^2$-AE~\citep{liu2023devil} employed both pixel values and frequency information obtained from the 2D discrete Fourier transform as the target. Taking MAE~\citep{he2022masked} as an example (Figure~\ref{fig:chp2_mae_maskfeat}(a)), the authors show that using pixel values as targets works particularly well. Specifically, a large random subset of images (\emph{e.g.}, 75\%) is masked out; then, the image encoder is only applied to visible patches, while mask tokens are introduced after the encoder. It was shown that such pre-training is especially effective for object detection and segmentation tasks, which require finer-grained image understanding.
    \item \textbf{High-level features as targets}: 
    BEiT~\citep{bao2021beit}, CAE~\citep{chen2022context}, SplitMask~\citep{el2021large}, and PeCo~\citep{dong2023peco} involve the prediction of discrete tokens using learned image tokenizers.  MaskFeat~\citep{wei2021masked-feat} takes a different approach by proposing direct regression of high-level features extracted from models like DINO~\citep{caron2021emerging} and DeiT~\citep{touvron2021training}.
    Expanding this idea, MVP~\citep{wei2022mvp} and EVA~\citep{fang2023eva} make feature prediction using image features from CLIP as target features.
    Additionally, other methods such as data2vec~\citep{baevski2022data2vec}, MSN~\citep{assran2022masked}, ConMIM~\citep{yi2022masked}, SIM~\citep{tao2023siamese}, and BootMAE~\citep{dong2022bootstrapped} propose to construct regression feature targets by leveraging momentum-updated teacher models to enhance online learning. The choice of loss functions depends on the nature of the targets: cross-entropy loss is typically used when the targets are discrete tokens, while $\ell_1$, $\ell_2$, or cosine similarity losses are common choices for pixel values or continuous-valued features.
\end{itemize}

\paragraph{MIM for video pre-training.} 
Naturally, there are recent works on extending MIM to video pre-training. Prominent examples include BEVT~\citep{wang2022bevt}, MAE as spatiotemporal learner~\citep{feichtenhofer2022masked}, VideoMAE~\citep{tong2022videomae}, and VideoMAEv2~\citep{wang2023videomae}. Taking~\cite{feichtenhofer2022masked} as an example. This paper studies a conceptually simple extension of MAE to video pre-training via randomly masking out space-time patches in videos and learns an autoencoder to reconstruct them in pixels. Interestingly, the authors found that MAE learns strong video representations with almost no inductive bias on space-time, and spacetime-agnostic random masking performs the best, with an optimal masking ratio as high as 90\%.


\paragraph{Lack of learning global image representations.}
MIM is an effective pre-training method that provides a good parameter initialization for further model finetuning. However, the vanilla MIM pre-trained model does not learn a global image representation. In iBOT~\citep{zhou2021ibot}, the authors propose to enhance BEiT~\citep{bao2021beit} with a DINO-like self-distillation loss~\citep{caron2021emerging} to force the \texttt{[CLS]} token to learn global image representations. The same idea has been extended to DINOv2~\citep{oquab2023dinov2}.

\begin{figure}[t!]
\begin{center}
\includegraphics[width=\textwidth]{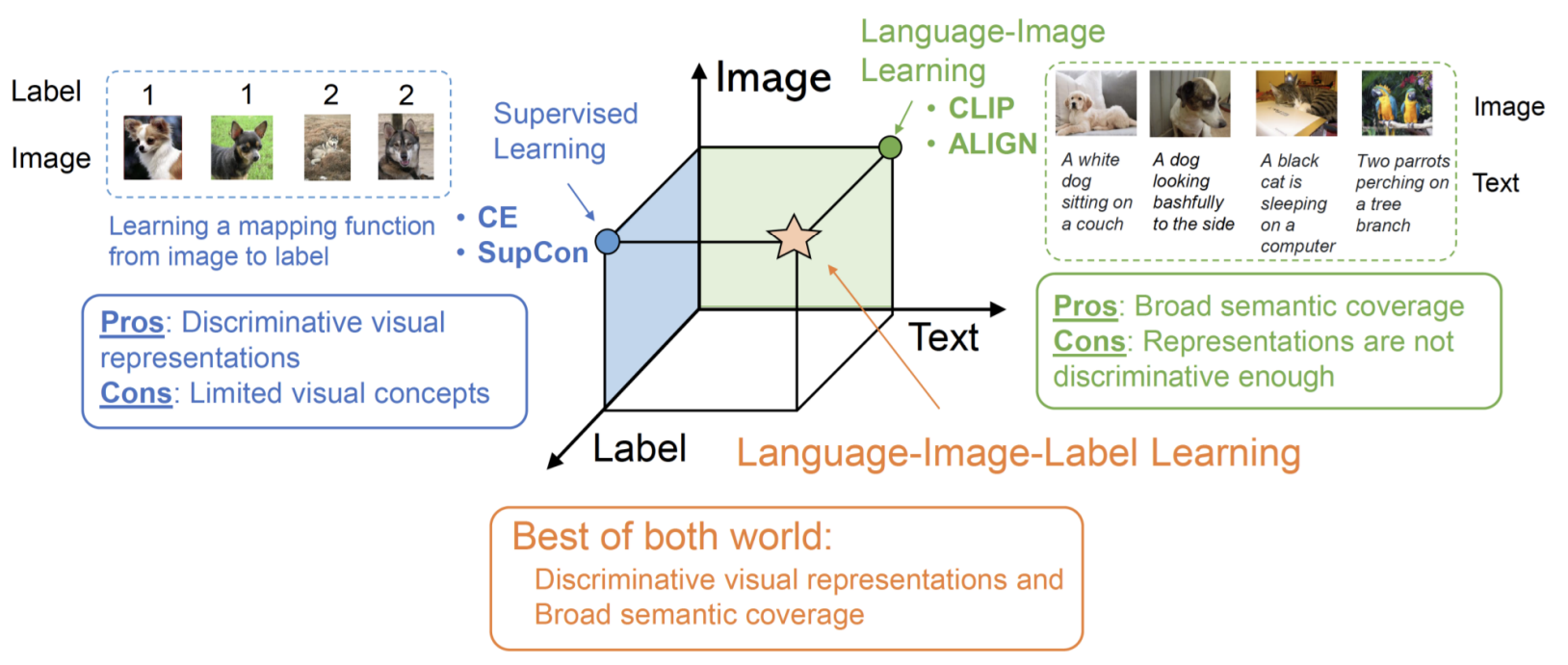}
\end{center}
\vspace{-2mm}
\caption{Overview of UniCL~\citep{yang2022unified} that performs unified contrastive pre-training on image-text and image-label data. Image credit: \cite{yang2022unified}.}
\label{fig:chp2_unicl}
\end{figure}

\paragraph{Scaling properties of MIM.} MIM is scalable in terms of model size. For example, we can perform MIM pre-training of a vision transformer with billions of parameters. However, the scaling property with regard to data size is less clear. There are some recent works that aim to understand the data scaling of MIM~\citep{xie2023data,lu2023delving}; however, the data scale is limited to millions of images, rather than billions, except \cite{singh2023effectiveness} that studies the effectiveness of MAE as a so-called ``pre-pretraining'' method for billion-scale data. Generally, MIM can be considered an effective regularization method that helps initialize a billion-scale vision transformer for downstream tasks; however, whether or not scaling the MIM pre-training to billion-scale image-only data requires further exploration. 

\section{Synergy Among Different Learning Approaches}\label{sec:synergy}
Till now, we have reviewed different approaches to pre-training image backbones, especially for vision transformers. Below, we use CLIP as the anchor point, and discuss how CLIP can be combined with other learning methods.

\paragraph{Combining CLIP with label supervision.} Noisy labels and text supervision can be jointly used for image backbone pre-training. Some representative works are discussed below. 
\begin{itemize}[leftmargin=*]
    \item UniCL~\citep{yang2022unified} proposes a principled way to use image-label and image-text data together in a joint image-text-label space for unified contrastive learning, and Florence~\citep{yuan2021florence} is a scaled-up version of UniCL. See Figure~\ref{fig:chp2_unicl} for an illustration of the framework.
    \item LiT~\citep{zhai2021lit} uses a pre-trained ViT-g/14 image encoder learned from supervised pre-training on the JFT-3B dataset, and then makes the image encoder open-vocabulary by learning an additional text tower via contrastive pre-training on image-text data. Essentially, LiT teaches a text model to read out good representations from a pre-trained image model for new tasks.
    \item MOFI~\citep{wu2023mofi} proposes to learn image representations from 1 billion noisy entity-annotated images, and uses both image classification and contrastive losses for model training. For image classification, entities associated with each image are considered as labels, and supervised pre-training on a large number of entities is conducted; for constrastive pre-training,  entity names are treated as free-form text, and are further enriched with entity descriptions.
\end{itemize}

\paragraph{Combining CLIP with image-only (non-)contrastive learning.} CLIP can also be enhanced with image-only self-supervision. Specifically,
\begin{itemize}[leftmargin=*]
    \item SLIP~\citep{mu2021slip} proposes a conceptually simple idea to combine SimCLR~\citep{chen2020simple} and CLIP for model training, and shows that SLIP outperforms CLIP on both zero-shot transfer and linear probe settings. DeCLIP~\citep{li2021supervision} mines self-supervised learning signals on each modality to make CLIP training data-efficient. In terms of image supervision, the SimSam framework~\citep{chen2021exploring} is used. 
    \item xCLIP~\citep{zhou2023non} makes CLIP non-contrastive via introducing additional sharpness and smoothness regularization terms borrowed from the image-only non-contrastive learning literature. However, the authors show that only non-contrastive pre-training (nCLIP) is not sufficient to achieve strong performance on zero-shot image classification, and it needs to be combined with the original CLIP for enhanced performance. 
\end{itemize}

\begin{figure}[t!]
\begin{center}
\includegraphics[width=\textwidth]{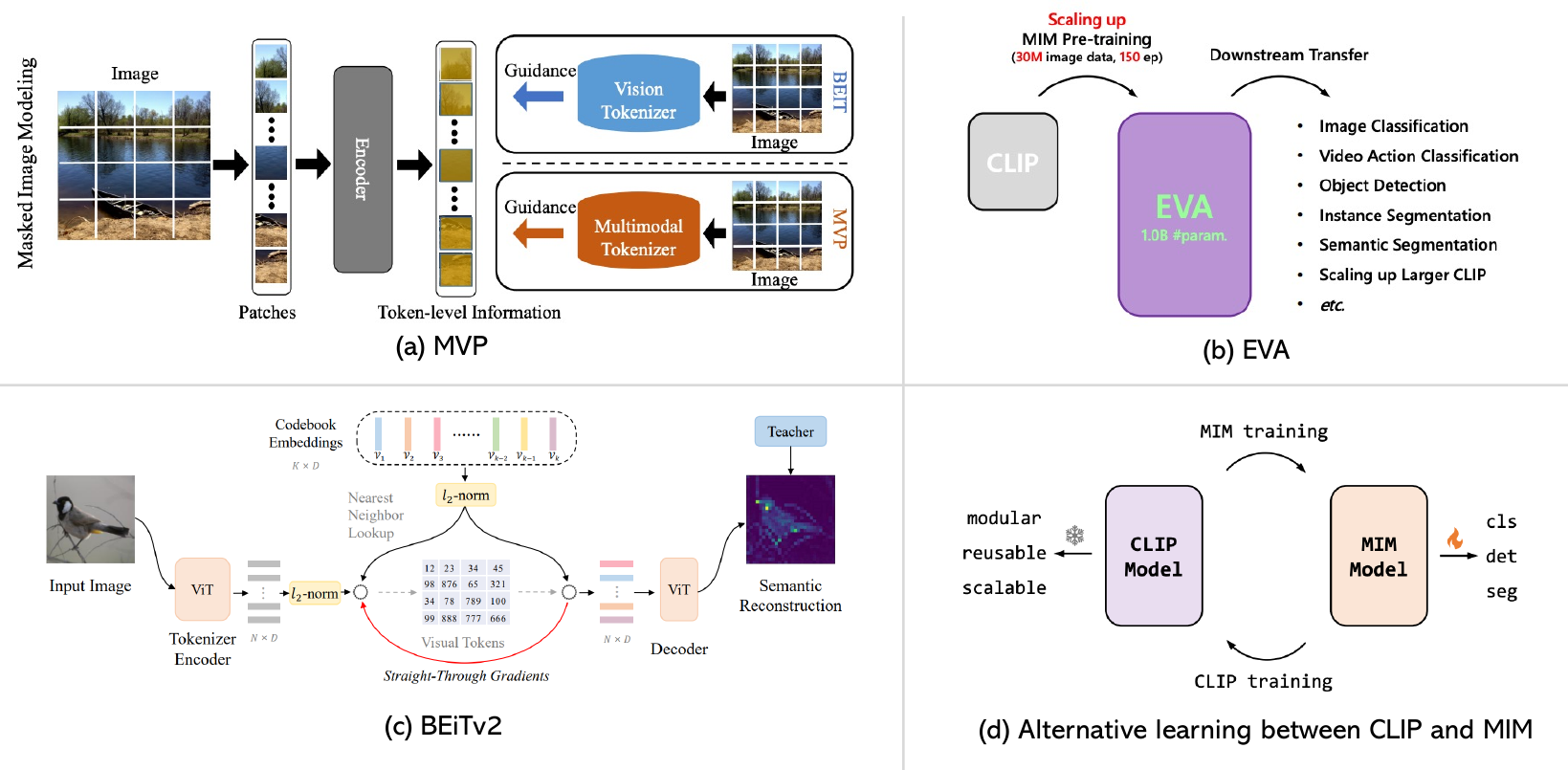}
\end{center}
\vspace{-2mm}
\caption{Illustration of MVP~\citep{wei2022mvp}, EVA~\citep{fang2023eva} and BEiTv2~\citep{peng2022beit}. (a) \& (b) MVP and EVA directly regress CLIP features for MIM pre-training. (c) BEiTv2 compresses the information inside CLIP features into discrete visual tokens, and then performing regular BEiT training. (d) Alternative learning between CLIP and MIM. Image credit: \cite{wei2022mvp}, \cite{fang2023eva}, \cite{peng2022beit}, \cite{fang2023eva}.}
\label{fig:chp2_mvp_beitv2}
\end{figure}

\begin{figure}[t!]
\begin{center}
\includegraphics[width=0.8\textwidth]{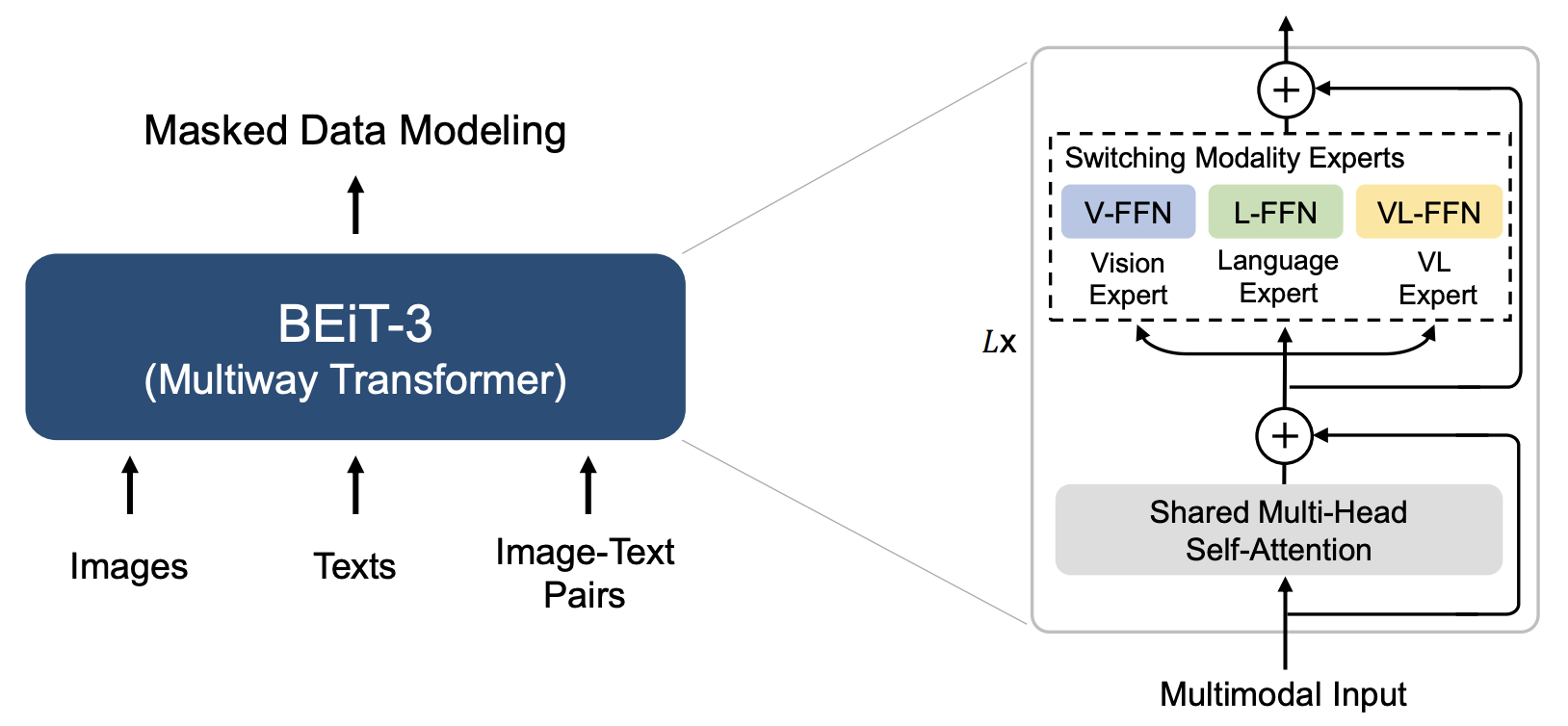}
\end{center}
\vspace{-2mm}
\caption{Overview of BEiT-3 that performs masked data modeling on both image/text and joint image-text data via a multiway transformer. Image credit: \cite{wang2022image}.}
\label{fig:chp2_beit3}
\end{figure}

\paragraph{Combining CLIP with MIM.} There are recent works that aim to combine CLIP and MIM for model training. We group them into two categories.
\begin{itemize}[leftmargin=*]
    \item \textbf{Shallow interaction.} It turns out that image features extracted from CLIP are a good target for MIM training, as the CLIP image features potentially capture the semantics that are missing in MIM training. Along this line of work, as shown in Figure~\ref{fig:chp2_mvp_beitv2}, MVP~\citep{wei2022mvp} proposes to regress CLIP features directly, while BEiTv2~\citep{peng2022beit} first compresses the information inside CLIP features into discrete visual tokens, and then performs regular BEiT training. Similar use of CLIP features as MIM training target has also been investigated in EVA~\citep{fang2023eva},  CAEv2~\citep{zhang2022cae}, and MaskDistill~\citep{peng2022unified}. In EVA-02~\citep{fang2023eva}, the authors advocate alternative learning of MIM and CLIP representations. Specifically, an off-the-shelf CLIP model is used to provide a feature target for MIM training; while the MIM pre-trained image backbone is used to initialize CLIP training. The MIM representations are used to finetune various downstream tasks while the learned frozen CLIP embedding enables zero-shot image classification and other applications.   
    \item \textbf{Deeper integration.} However, instead of using CLIP as targets for MIM training, if one aims to combine CLIP and MIM for joint model training, MIM does not seem to improve a CLIP model at scale \citep{weers2023masked,li2023scaling}.
    \item Although the combination of CLIP and MIM does not lead to a promising result at the current stage, the combination of BERT and BEiT is very promising, as evidenced in BEiT-3~\citep{wang2022image} (see Figure~\ref{fig:chp2_beit3}), where the authors show that masked data modeling can be performed on both image/text and joint image-text data via the design of a multiway transformer, and state-of-the-art performance can be achieved on a wide range of vision and vision-language tasks. 
\end{itemize}

\section{Multimodal Fusion, Region-Level and Pixel-Level Pre-training}
\label{sec:visual_understanding_advanced_topics}
Till now, we have focused on the methods of pre-training image backbones from scratch, but not on pre-training methods that power multimodal fusion, region-level and pixel-level image understanding. These methods typically use a pre-trained image encoder at the first hand to perform a second-stage pre-training. Below, we briefly discuss these topics. 

\begin{figure}[t!]
\begin{center}
\includegraphics[width=0.8\textwidth]{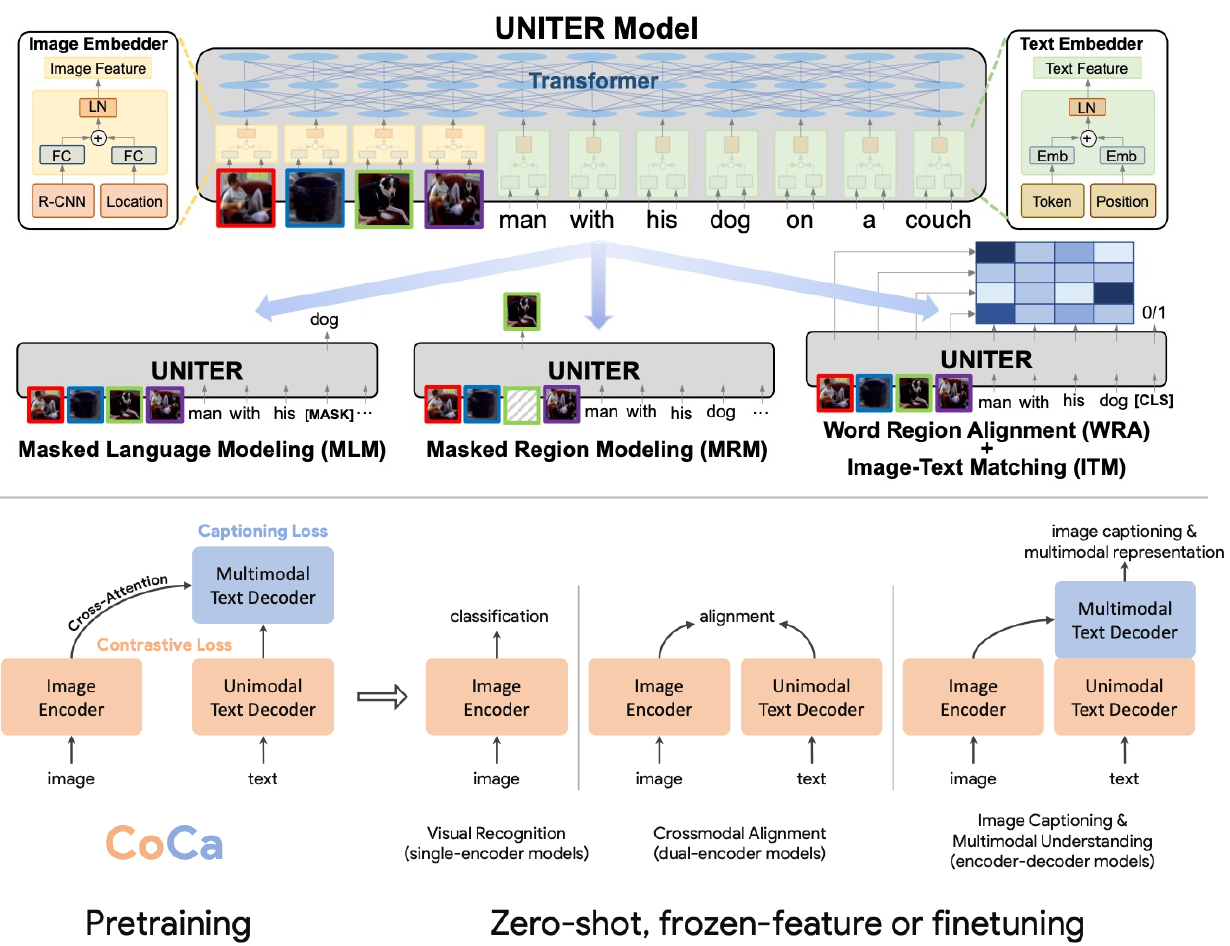}
\end{center}
\vspace{-2mm}
\caption{Illustration of UNITER~\citep{chen2020uniter} and CoCa~\citep{yu2022coca}, which serve as a classical and a modern model that performs pre-training on multimodal fusion. CoCa also pre-trains the image backbone from scratch. Specifically, UNITER extracts image features via an off-the-shelf object detector and treat image features as soft prompts of the text input to be sent into a multimodal transformer. The model is pre-trained over a few millions of image-text pairs. For CoCa, an image encoder and a text encoder is used, with a multimodal transformer stacked on top. Both contrastive loss and captioning loss are used for model training, and the model is trained over billions of image-text pairs and JFT data.  Image credit: \cite{chen2020uniter}, \cite{yu2022coca}.}
\label{fig:chp2_uniter_coca}
\end{figure}

\subsection{From Multimodal Fusion to Multimodal LLM}
\label{sec:multimodal_fusion}
For dual encoders such as CLIP~\citep{radford2021learning}, image and text are encoded separately, and modality interaction is only handled via a simple dot product of image and text feature vectors. This can be very effective for zero-shot image classification and image-text retrieval. However, due to the lack of deep multimodal fusion, CLIP alone performs poorly on the image captioning~\citep{vinyals2015show} and visual question answering~\citep{antol2015vqa} tasks. This requires the pre-training of a fusion encoder, where additional transformer layers are typically employed to model the deep interaction between image and text representations. Below, we review how these fusion-encoder pre-training methods are developed over time.

\paragraph{OD-based models.} Most early methods use pre-trained object detectors (ODs) to extract visual features. Among them, ViLBERT~\citep{lu2019vilbert} and LXMERT~\citep{tan-bansal-2019-lxmert} use co-attention for multimodal fusion, while methods like VisualBERT~\citep{li2019visualbert}, Unicoder-VL~\citep{li2020unicoder}, VL-BERT~\citep{su2019vl}, UNITER~\citep{chen2020uniter}, OSCAR~\citep{li2020oscar}, VILLA~\citep{gan2020large} and VinVL~\citep{zhang2021vinvl} treat image features as soft prompts of the text input to be sent into a multimodal transformer.

\paragraph{End-to-end models.} Now, end-to-end pre-training methods become the mainstream. Some early methods use CNNs to extract image features, such as PixelBERT~\citep{huang2020pixel}, SOHO~\citep{huang2021seeing}, and CLIP-ViL~\citep{shen2021much}, while ViLT~\citep{kim2021vilt} and ViTCAP~\citep{fang2022injecting} directly feed image patch features and text token embeddings into a multimodal transformer. Due to the popularity of vision transformer (ViT), now most methods simply use ViT as the image encoder (\emph{e.g.}, plain ViT~\citep{dosovitskiy2020image} and Swin transformer~\citep{liu2021swin}). Prominent examples include ALBEF~\citep{li2021align}, METER~\citep{dou2021empirical}, VLMo~\citep{wang2021vlmo}, X-VLM~\citep{zeng2021multi}, BLIP~\citep{li2022blip}, SimVLM~\citep{wang2021simvlm}, FLAVA~\citep{singh2022flava} and CoCa~\citep{yu2022coca}. 

An illustration of UNITER~\citep{chen2020uniter} and CoCa~\citep{yu2022coca} is shown in Figure~\ref{fig:chp2_uniter_coca}. They serve as two examples of  a classical model and a modern model, respectively, which performs pre-training on multimodal fusion. CoCa also performs image backbone pre-training directly, as all the model components are trained from scratch. Please refer to Chapter 3 of \cite{gan2022vision} for a comprehensive literature review.

\paragraph{Trend to multimodal LLM.} Instead of using masked language modeling, image-text matching and image-text contrastive learning, SimVLM~\citep{wang2021simvlm} uses a simple PrefixLM loss for pre-training. Since then, multimodal language models have become popular. Early models focus on large-scale pre-training, such as Flamingo~\citep{alayrac2022flamingo}, GIT~\citep{wang2022git}, PaLI~\citep{chen2022pali}, PaLI-X~\citep{chen2023pali}, while recent works focus on using pre-trained LLMs for instruction tuning, such as LLaVA~\citep{liu2023visual} and MiniGPT-4~\citep{zhu2023minigpt4}. A detailed discussion on this topic is provided in Chapter~\ref{chp:training_with_llm}.

\subsection{Region-Level Pre-training}

CLIP learns global image representations via contrastive pre-training. However, for tasks that require fine-grained image understanding such as object detection, CLIP is not enough. Object detection contains two sub-tasks: localization and recognition. ($i$) Localization aims to locate the presence of objects in an image and indicate the position with a bounding box, while ($ii$) recognition determines what object categories are present in the bounding box. By following the reformulation that converts image classification to image retrieval used in CLIP, generic open-set object detection can be achieved. 

\begin{figure}[t!]
\begin{center}
\includegraphics[width=\textwidth]{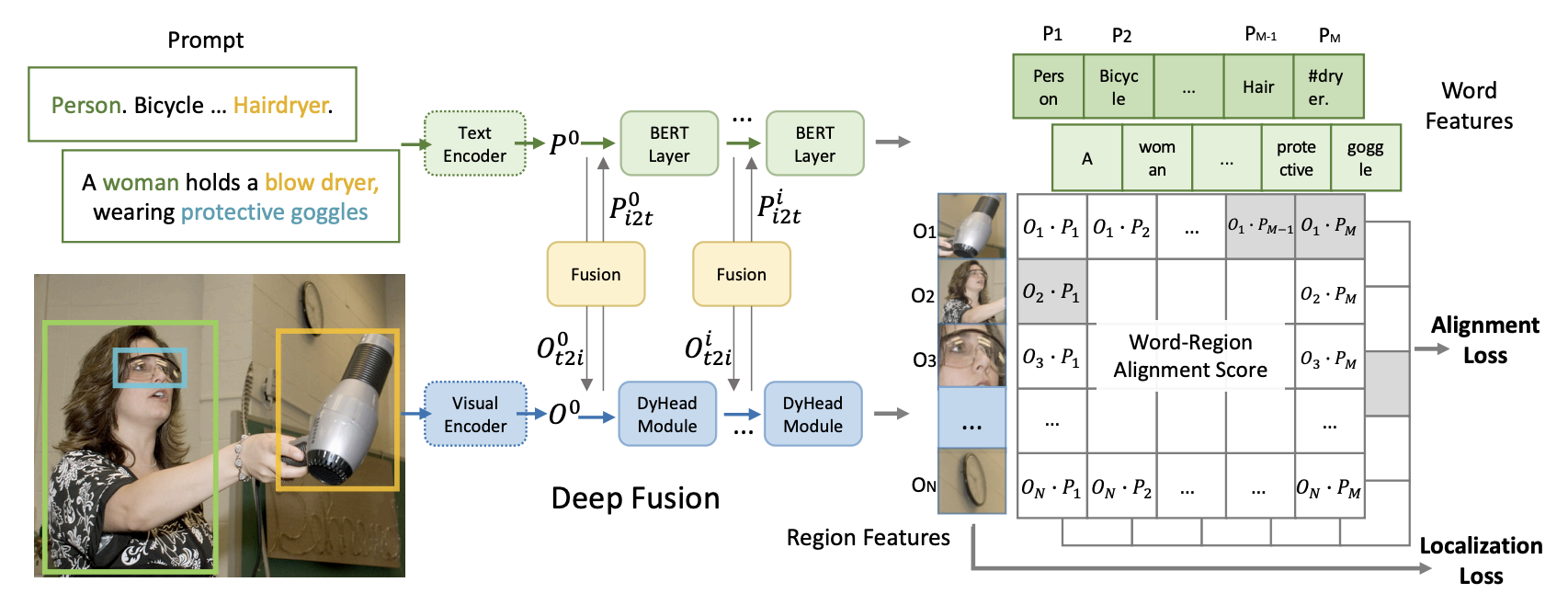}
\end{center}
\vspace{-2mm}
\caption{Overview of GLIP that performs grounded language-image pre-training for open-set object detection. Image credit: \cite{li2022grounded}.}
\label{fig:chp2_glip}
\end{figure}

Specifically, ViLD~\citep{gu2021open} and RegionCLIP~\citep{zhong2021regionclip} distill knowledge from CLIP with a two-stage detector for zero-shot object detection. In MDETR~\citep{kamath2021mdetr} and GLIP~\citep{li2021grounded} (as shown in Figure~\ref{fig:chp2_glip}), the authors propose to reformulate detection as a phrase grounding problem, and perform grounded language-image pre-training. GLIPv2~\citep{zhang2022glipv2} and FIBER~\citep{fiber2022} further perform unified pre-training for both grounding and vision-language understanding tasks. OVR-CNN~\citep{zareian2021open} finetunes an image-text model to detection on a limited vocabulary
and relies on image-text pre-training for generalization to an open vocabulary setting. Detic~\citep{zhou2022detecting} improves long-tail detection performance with weak supervision by training only the
classification head on the examples where only image-level annotations are available. Other works include OV-DETR~\citep{zang2022open}, X-DETR~\citep{cai2022x}, FindIT~\citep{kuo2022findit}, PromptDet~\citep{feng2022promptdet}, OWL-ViT~\citep{minderer2022simple}, GRiT~\citep{wu2022grit}, to name a few. Recently, Grounding DINO~\citep{liu2023grounding} is proposed to marry DINO~\citep{zhang2022dino} with grounded pre-training for open-set object detection. Please refer to Section~\ref{sec:chp4_from_close_to_open} for a detailed review of this topic.

\subsection{Pixel-Level Pre-training}
The Segment Anything Model (SAM)~\citep{kirillov2023segment} is a recent vision foundation model for image segmentation that aims to perform pixel-level pre-training. Since its birth, it has attracted wide attention and spurred tons of follow-up works and applications. Below, we briefly review SAM, as a representative work for pixel-level visual pre-training. 

\begin{figure}[t!]
\begin{center}
\includegraphics[width=\textwidth]{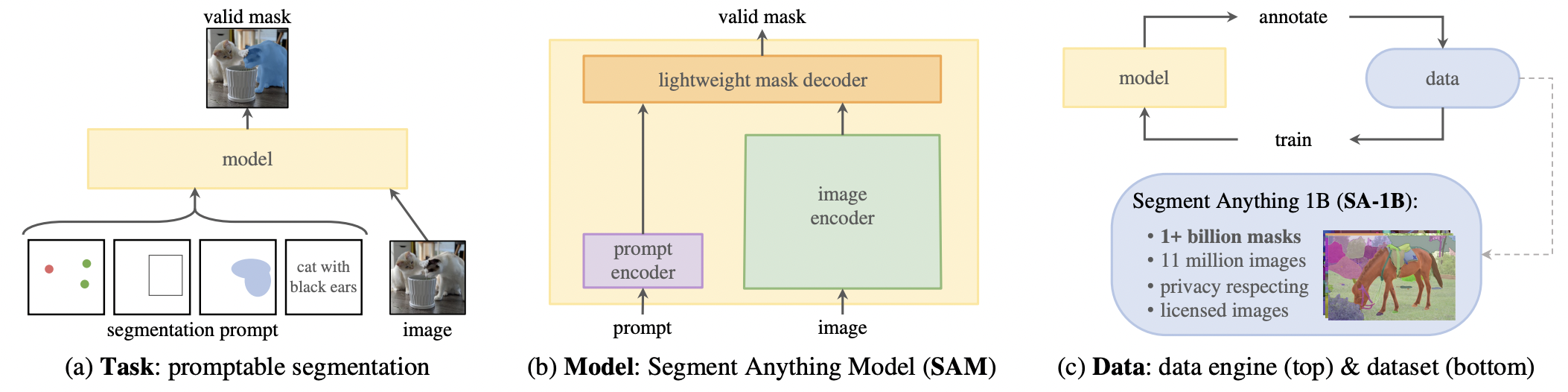}
\end{center}
\vspace{-2mm}
\caption{Overview of the Segment Anything project, which aims to build a vision foundation model for segmentation by introducing three interconnected components: a promptable segmentation task, a segmentation model, and a data engine. Image credit: \cite{kirillov2023segment}.}
\label{fig:chp2_sam1}
\end{figure}

As depicted in Figure~\ref{fig:chp2_sam1}, the objective of the Segment Anything project is to develop a foundational vision model for segmentation. This model is designed to be readily adaptable to a wide range of both existing and novel segmentation tasks, such as edge detection, object proposal generation, instance segmentation, open-vocabulary segmentation, and more. This adaptability is seamlessly accomplished through a highly efficient and user-friendly approach, facilitated by the integration of three interconnected components. Specifically,
\begin{itemize}[leftmargin=*]
    \item \textbf{Task.} The authors propose the promptable segmentation task, where the goal is to return a valid segmentation mask given any segmentation prompt, such as a set of points, a rough box or mask, or free-form text.
    \item \textbf{Model.} The architecture of SAM is conceptually simple. It is composed of three main components: ($i$) a powerful image encoder (MAE~\citep{he2022masked} pre-trained ViT); ($ii$) a prompt encoder (for sparse input such as points, boxes, and free-form text, the CLIP text encoder is used; for dense input such as masks, a convolution operator is used); and ($iii$) a lightweight mask decoder based on transformer.
    \item \textbf{Data.} To acquire large-scale data for pre-training, the authors develop a \emph{data engine} that performs model-in-the-loop dataset annotation.
\end{itemize}

\paragraph{Concurrent to SAM.} Parallel to SAM, many efforts have been made to develop general-purpose segmentation models as well. For example, OneFormer~\citep{jain2023oneformer} develops a universal image segmentation framework; SegGPT~\citep{wang2023seggpt} proposes a generalist in-context learning framework that unifies different segmentation data formats; SEEM~\citep{zou2023segment} further expands the types of supported prompts that a single segmentation model can handle, including points, boxes, scribbles, masks, texts, and referred regions of another image.

\paragraph{Extensions of SAM.} SAM has spurred tons of follow-up works that extend SAM to a wide range of applications, \emph{e.g.}, Inpaint Anything~\citep{yu2023inpaint}, Edit Everything~\citep{xie2023edit}, Any-to-Any Style Transfer~\citep{liu2023any}, Caption Anything~\citep{wang2023caption}, Track Anything~\citep{yang2023track}, Recognize Anything~\citep{zhang2023recognize,li2023semantic}, Count Anything~\citep{ma2023can}, 3D reconstruction~\citep{shen2023anything}, medical image analysis~\citep{ma2023segment,zhou2023can,shi2023generalist,zhang2023segment}, \emph{etc}. Additionally, recent works have attempted to develop models for detecting and segmenting anything in the open-vocabulary scenarios, such as Grounding DINO~\citep{liu2023grounding} and Grounding-SAM\footnote{\url{https://github.com/IDEA-Research/Grounded-Segment-Anything}}. For a comprehensive review, please refer to \cite{zhang2023comprehensive} and some GitHub repos.\footnote{\url{https://github.com/Hedlen/awesome-segment-anything}}

\chapter{Visual Generation}
\label{chp:generation}
\begin{wrapfigure}{r}{4cm}
  \centering
  \vspace{-6cm}
  \includegraphics[width=1.0\linewidth]{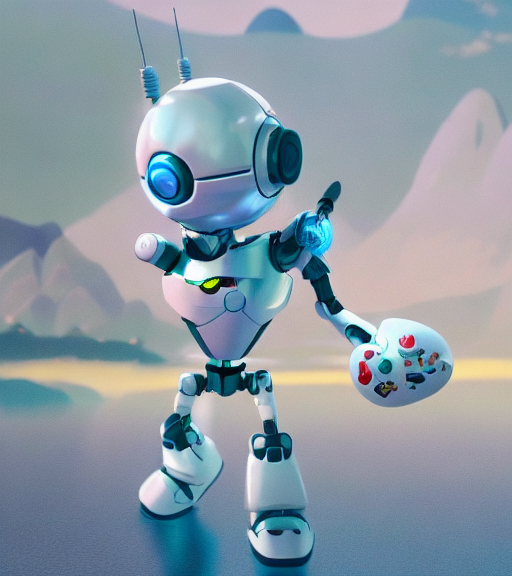}
\end{wrapfigure}

Visual generation aims to generate high-fidelity visual content, including images, videos, neural radiance fields, 3D point clouds, \etc. This topic is at the core of recently popular artificial intelligence generated content (AIGC), and this ability is crucial in supporting creative applications such as design, arts, and multimodal content creation. It is also instrumental in synthesizing training data to help understand models, leading to the closed loop of multimodal content understanding and generation. To make use of visual generation, it is critical to produce visual data that is strictly aligned with human intents. These intentions are fed into the generation model as input conditions, such as class labels, texts, bounding boxes, layout masks, among others. Given the flexibility offered by open-ended text descriptions, text conditions (including text-to-image/video/3D) have emerged as a pivotal theme in conditional visual generation.

In this chapter, we describe how to align with human intents in visual generation, with a focus on image generation. We start with the overview of the current state of text-to-image (T2I) generation in Section~\ref{sec:generation_overview}, highlighting its limitations concerning alignment with human intents. The core of this chapter is dedicated to reviewing the literature on four targeted areas that aim at enhancing alignments in T2I generation, \emph{i.e.}, spatial controllable T2I generation in Section~\ref{sec:generation_spatial}, text-based image editing in Section~\ref{sec:generation_editing}, better following text prompts in Section~\ref{sec:generation_promptfollow}, and concept customization in T2I generation in Section~\ref{sec:generation_conceptcust}. At the end of each subsection, we share our observations on the current research trends and short-term future research directions. These discussions coalesce in Section~\ref{sec:generation_trends}, where we conclude the chapter by considering future trends. Specifically, we envision the development of a generalist T2I generation model, which can better follow human intents, to unify and replace the four separate categories of alignment works.

\section{Overview}
\label{sec:generation_overview}
\subsection{Human Alignments in Visual Generation}
\begin{figure}[h!]
\centering  
\vspace{-4mm}
\includegraphics[width=1.00\textwidth]{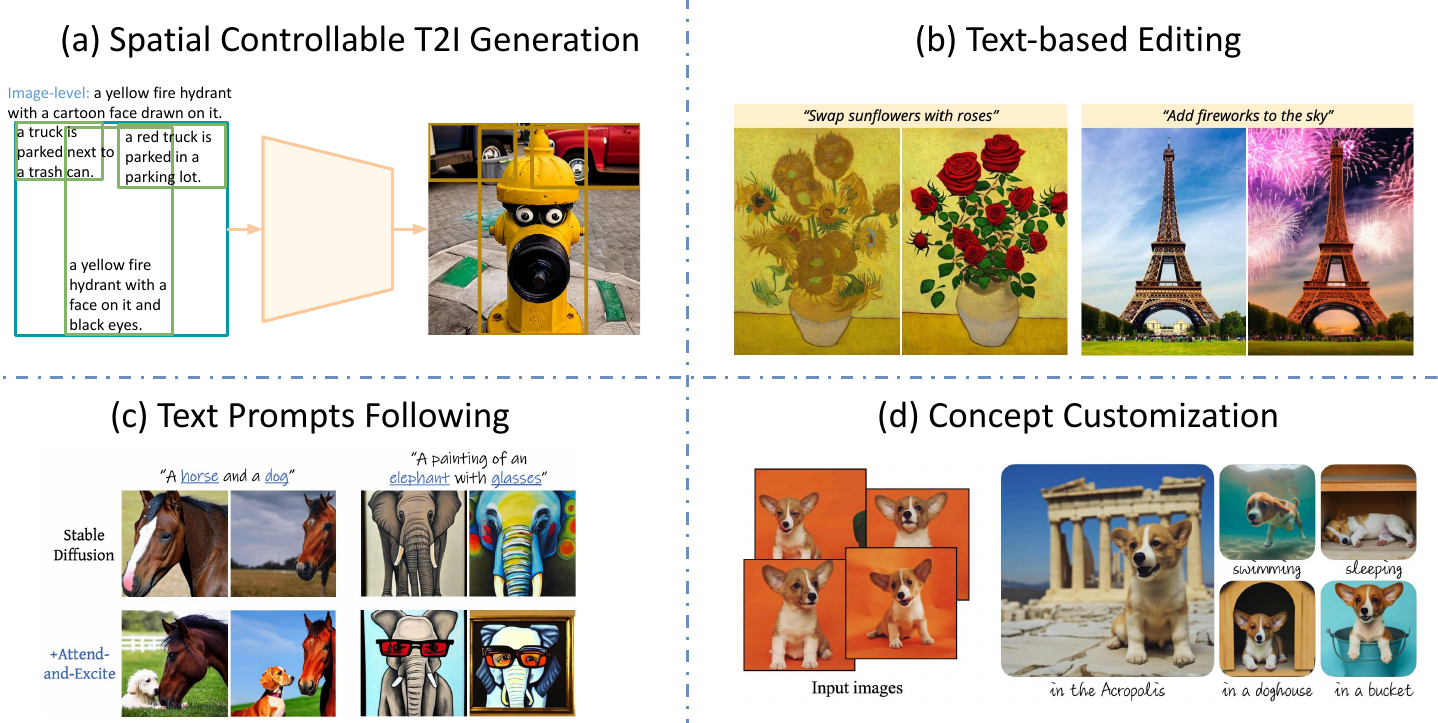} \\
(a) An overview of topics on human alignment for generative foundation models. Image credit:~\cite{yang2023reco,brooks2023instructpix2pix,chefer2023attend,ruiz2023dreambooth}. \\
\vspace{-0mm}
\includegraphics[width=1.00\textwidth]{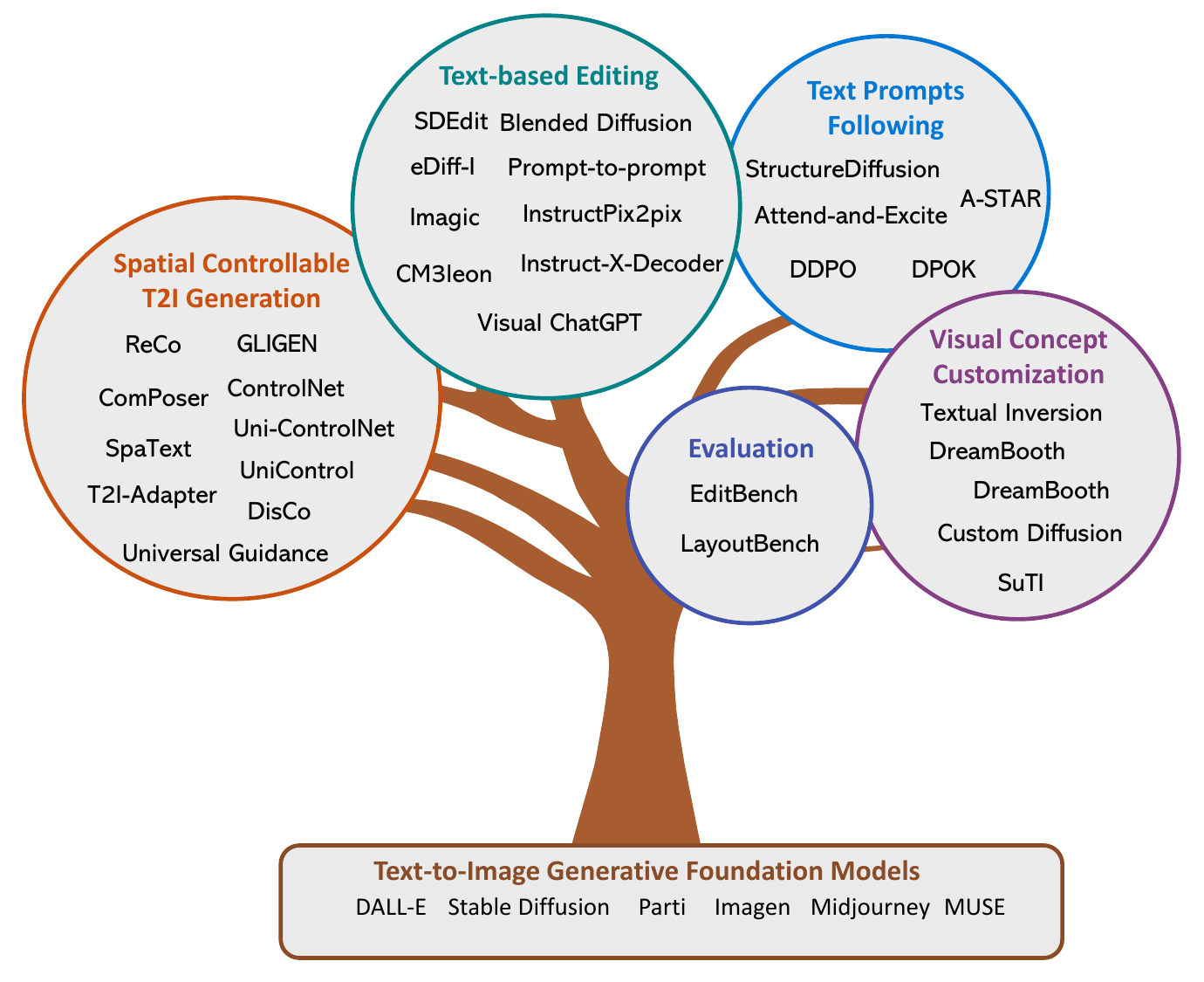} \\
(b) Summary and categorization of papers on ``Human Alignments in Visual Generation.'' \\
\caption{An overview of improving human intent alignments in T2I generation. }
\label{fig:chp3_overview}  
  \vspace{-1mm}
\end{figure}
AI Alignment research in the context of T2I generation is the field of study dedicated to developing image generation models that can easily follow human intents to synthesize the desired generated visual content. Current literature typically focuses on one particular weakness of vanilla T2I models that prevents them from accurately producing images that align with human intents. This chapter delves into four commonly studied issues, as summarized in Figure~\ref{fig:chp3_overview} (a) and follows.
\begin{itemize}[leftmargin=*]
    \item \textbf{Spatial controllable T2I generation.} Text serves as a powerful medium for human-computer interaction, making it a focal point in conditional visual generation. However, text alone falls short in providing precise spatial references, such as specifying open-ended descriptions for arbitrary image regions with precise spatial configurations. Spatial controllable T2I generation~\citep{yang2023reco,li2023gligen,zhang2023adding} aims to combine text inputs with other conditions for better controllability, thereby facilitating users to generate the desired images.
    \item \textbf{Text-based image editing.} Editing is another important means for acquiring human-intended visual content. Users might possess near-perfect images, whether generated by a model or naturally captured by a camera, but these might require specific adjustments to meet their intent. Editing has diverse objectives, ranging from locally modifying an object to globally adjusting the image style. Text-based image editing~\citep{brooks2023instructpix2pix} explores effective ways to create a versatile editing tool.
    \item \textbf{Better following text prompts.} Despite T2I models being trained to reconstruct images conditioned on the paired text input, the training objective does not necessarily ensure or directly optimize for a strict adherence to text prompts during image generation. Studies~\citep{yu2022scaling,rombach2022high} have shown that vanilla T2I models might overlook certain text descriptions and generate images that do not fully correspond to the input text. Research~\citep{feng2022training,black2023training} along this line explores improvements to have T2I models better following text prompts, thereby facilitating the easier use of T2I models.
    \item \textbf{Visual concept customization.} Incorporating visual concepts into textual inputs is crucial for various applications, such as generating images of one's pet dog or family members in diverse settings, or crafting visual narratives featuring a specific character. These visual elements often encompass intricate details that are difficult to articulate in words. Alternatively, studies~\citep{ruiz2023dreambooth,chen2023subject} explore if T2I models can be customized to draw those visual concepts with specialized token embeddings or conditioned images.
\end{itemize}
Before introducing the alignment works in detail, we first review the basics of text-to-image generation in the next section.

\subsection{Text-to-Image Generation}
\begin{figure}[h!]
\centering  
\vspace{-4mm}
\includegraphics[width=1.00\textwidth]{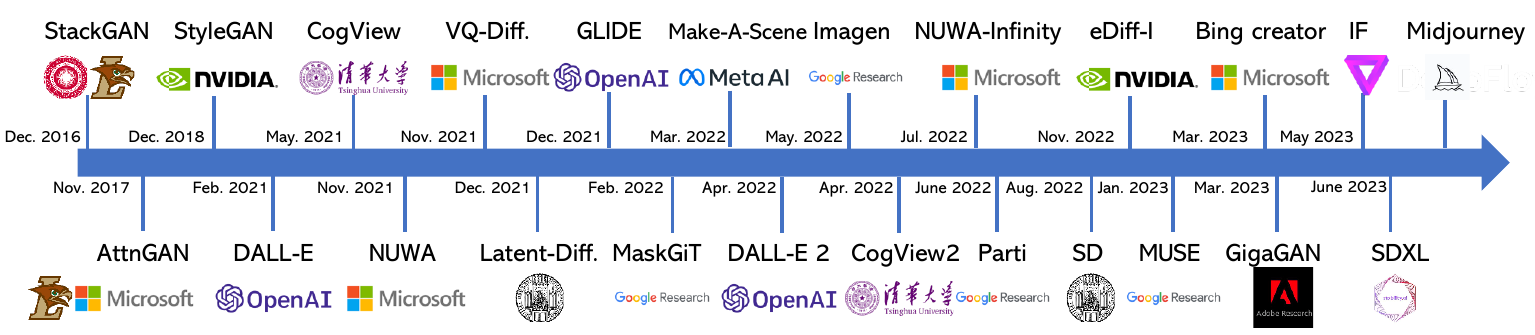} \\
\vspace{-0mm}
\caption{An overview of representative text-to-image generation models until July 2023. }
\label{fig:chp3_t2i_timeline}  
  \vspace{-1mm}
\end{figure}
T2I generation aims to generate images that are not only of high visual quality but also semantically correspond to the input text. T2I models are usually trained with image-text pairs, where text is taken as input conditions, with the paired image being the targeted output. 
Abstracted from the wide range of T2I models shown in Figure~\ref{fig:chp3_t2i_timeline}, we give a high-level overview of the representative image generation techniques.
\begin{itemize}[leftmargin=*]
    \item \textbf{Generative adversarial networks (GAN).} GANs~\citep{goodfellow2020generative,creswell2018generative,kang2023scaling} consist of two key components: a generator and a discriminator. The generator is tasked with creating synthetic images from random noise inputs, and it is trained to adjust these noise inputs based on input text conditions to generate semantically relevant images. In this adversarial process, the discriminator competes with the generator, attempting to differentiate between the synthetically generated images and real ones, thus guiding the generator to improve its image creation capabilities.
    \item \textbf{Variational autoencoder (VAE)} Variational Autoencoder (VAE)~\citep{kingma2013auto,van2017neural,vahdat2020nvae} is a probabilistic model that can generate images by employing paired encoder and decoder network modules. The encoder network optimizes the encoding of an image into a latent representation, while the decoder refines the process of converting the sampled latent representations back into a new image. VAEs are trained by minimizing the reconstruction error between the original and decoded images, whileregularizing the encoded latent space using the Kullback-Leibler (KL) divergence. Vector Quantised-VAE (VQ-VAE)~\citep{van2017neural} further improves VAEs by leveraging the discrete latent space through vector quantization, enabling improved reconstruction quality and generative capabilities.
    \item \textbf{Discrete image token prediction.} At the core of this approach lies a combination of a paired image tokenizer and detokenizer, like Vector Quantized Generative Adversarial Networks (VQGAN)~\citep{esser2021taming}, which efficiently transform continuous visual signals into a finite set of discrete tokens. In this way, the image generation problem is converted to a discrete token prediction task. 
    A widely employed strategy for token prediction is to use an auto-regressive Transformer~\citep{ramesh2021zero,yu2022scaling} to sequentially generates visual tokens, typically starting from the top left corner and moving row-by-row towards the bottom right, conditioned on the text inputs. Alternatively, studies~\citep{chang2022maskgit,chang2023muse} also explore the parallel decoding to speed up the token prediction process. Finally, the predicted visual tokens are detokenized, culminating in the final image prediction.
    \item \textbf{Diffusion model.} Diffusion models~\citep{sohl2015deep,song2020improved,ho2020denoising} employ stochastic differential equations to evolve random noises into images. A diffusion model works by initiating the process with a completely random image, and then gradually refining it over multiple iterations in a denoising process. Each iteration predicts and subsequently removes an element of noise, leading to a continuous evolution of the image, conditioned on the input texts.
\end{itemize}

We use Stable Diffusion (SD)~\citep{rombach2022high} as an example to explain in detail how T2I models work. We choose this model for a variety of reasons. Firstly, SD is one of the most widely used open-source T2I models, which makes it a solid foundation for many alignment techniques we discuss in this chapter. Additionally, as a diffusion-based generation model, it serves as an excellent case study for introducing diffusion models. Finally, its cross-attention-based image-text fusion mechanism is a classic example of various text-conditioned methods, such as auto-regressive T2I generation~\citep{yu2022scaling}, helping us gain an in-depth understand of the image-text interaction in T2I generation.

\begin{figure}[t]
\centering  
\vspace{-4mm}
\includegraphics[width=0.9\textwidth]{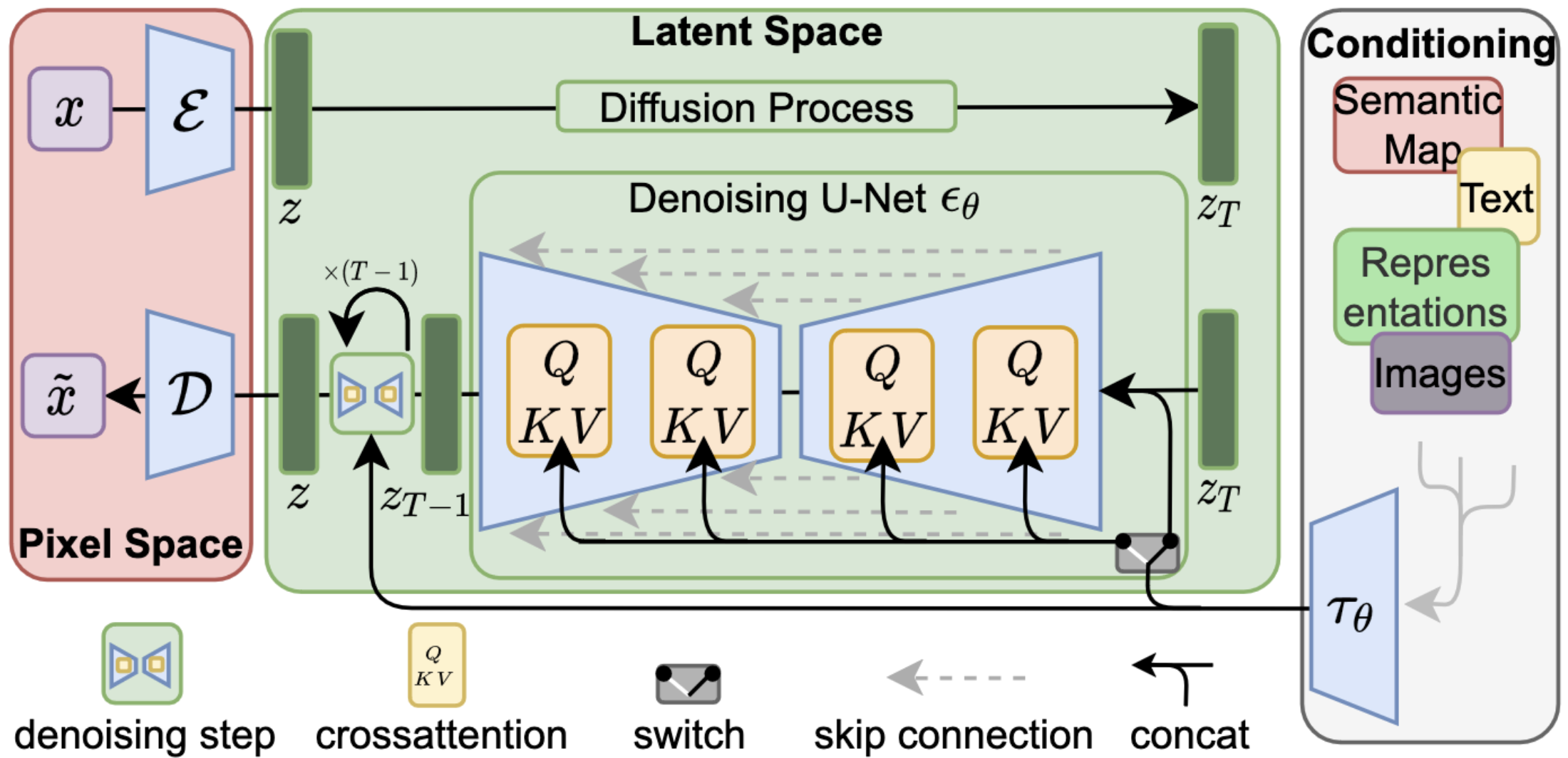} \\
\vspace{-0mm}
\caption{An overview of the latent diffusion model architecture. Image credit:~\cite{rombach2022high}. }
\label{fig:chp3_sd_arch}  
  \vspace{-1mm}
\end{figure}

Stable Diffusion (SD)\footnote{We use Stable Diffusion v1 for the introduction. Later versions such as SD2 and SDXL share the same method but may have different detailed model configurations, such as a larger text encoder, U-Net, and latent dimension.}, and its academic version latent diffusion~\citep{rombach2022high}, contains mainly three modules, \emph{i.e.}, an image VAE, a denoising U-Net, and a condition encoder, as shown in the left, center, and right part of Figure~\ref{fig:chp3_sd_arch}, respectively. We will introduce each module and the inference flow for image generation, following the notations in the original latent diffusion paper~\citep{rombach2022high}.

\begin{itemize}[leftmargin=*]
\item \textbf{VAE.} As introduced in the image generation technique overview, the VAE module contains a paired encoder $\mathcal{E}$ and decoder $\mathcal{D}$, trained to encode RGB image $x$ into a latent random variable $z$ and then decode the latent to best reconstruct the image. Given an RGB image $x \in \mathbb{R}^{H \times W \times 3}$, the encoder $\mathcal{E}$ encodes it into a continuous latent representation $z \in \mathbb{R}^{h \times w \times c}$. With the parameters of $H=W=512$, $h=w=64$, and $c=4$ in SD, latent $z$ is $48$ times smaller than image $x$, thereby significantly improving the computational efficiency by performing the denoising process in this compressed compact latent space.
\item \textbf{Text encoder.} SD is a conditional image generation model, where the input text condition is encoded using a condition encoder $\tau$. Specifically, SD uses the ViT-L/14 CLIP text encoder~\citep{radford2021learning} that encodes the tokenized input text query $y$ into text feature $\tau(y) \in \R^{N\times d_\tau}$, where the maximum length $N$ is $77$ and text feature dimension $d_\tau$ is $768$.
\item \textbf{Denoising U-Net.}
The denoising U-Net is the core module for the diffusion image generation process. The module is trained to predict the noise $\hat{\epsilon}(z_{t},t)$ to subtract in the latent space at each denoising timestep $t$, such that it can step-by-step evolve the initial random noise into a meaningful image latent. The module is trained with the L2 loss between the predicted noise $\hat{\epsilon}(z_{t},t)$ and the target noise $\epsilon$, which is added to the target image latent encoded by VAE encoder $\mathcal{E}$. At inference, the iteratively denoised latent $z$, started from the random noise, is sent through the VAE decoder $\mathcal{D}$ for the final generated image.

In each denoising step, the U-Net takes the text condition as input to generate images that are semantically relevant to the text query. We next detail how the visual stream $z \in \mathbb{R}^{h \times w \times c}$ interacts with the text stream $\tau(y) \in \R^{N\times d_\tau}$. The denoising U-Net, similar to a classic U-Net~\citep{ronneberger2015u,long2015fully}, consists of a series of spatial downsampling and upsampling blocks with skip connections in between. In SD's U-Net, each down/upsampling block has a cross-attention layer and a 2D convolutional down/upsampling layer. Each block takes the visual latent feature, text feature, and denoising step as input and generates the next visual latent feature. The image-text interaction happens in the image-text cross-attention layer.
\begin{equation}
\label{equ:chp3_crossattn}
    \text{Attention}(Q, K, V) = \text{softmax}\left(\frac{QK^T}{\sqrt{d}}\right) \cdot V,
\end{equation}
where $K$, $V$ are projected from the text stream $\tau(y)$ and $Q$ is projected from the visual stream $z$ to share the same hidden dimension $d$. Therefore, the softmax between $Q$ and $K$ produces an attention map $M$ of size $(hw \times d)\cdot (N\times d)^T = hw \times N$. The cross-attention map $M$ indicates the fine-grained image-text interaction among each one of the $N$ text words in all spatial positions $hw$. The attention map $M$ then products $V$ to yield the output of a down/upsampling block.
\end{itemize}

\section{Spatial Controllable Generation}
\label{sec:generation_spatial}
T2I generation takes open-ended text for users to describe their intended images. However, text alone is ineffective in certain descriptions, such as spatial referring. Studies in spatial controllable T2I generation explore extending T2I models to take extra spatial input conditions to guide image generation process. 

We categorize related studies into three topics. 
$(i)$ We start with works~\citep{yang2023reco,li2023gligen,avrahami2023spatext,cho2023diagnostic} that extend the image-level text description in vanilla T2I models to the region-grounded text description, such that open-ended text descriptions can precisely operate on a particularly spatial region. 
$(ii)$ We then extend from boxes to dense spatial conditions represented as 2D arrays, such as segmentation masks, edge maps, depth maps, key points. We review representative works ControlNet~\citep{zhang2023adding} and many others~\citep{mou2023t2i,zeng2023scenecomposer,zhao2023uni,qin2023unicontrol}. 
$(iii)$The previous two threads of work require finetuning T2I models to understand the extended spatial condition. We next review techniques of inference-time guidance~\citep{bansal2023universal,chen2023training} that achieve spatial control without model finetuning.

\begin{figure}[t]
\centering  
\vspace{-4mm}
\includegraphics[width=\textwidth]{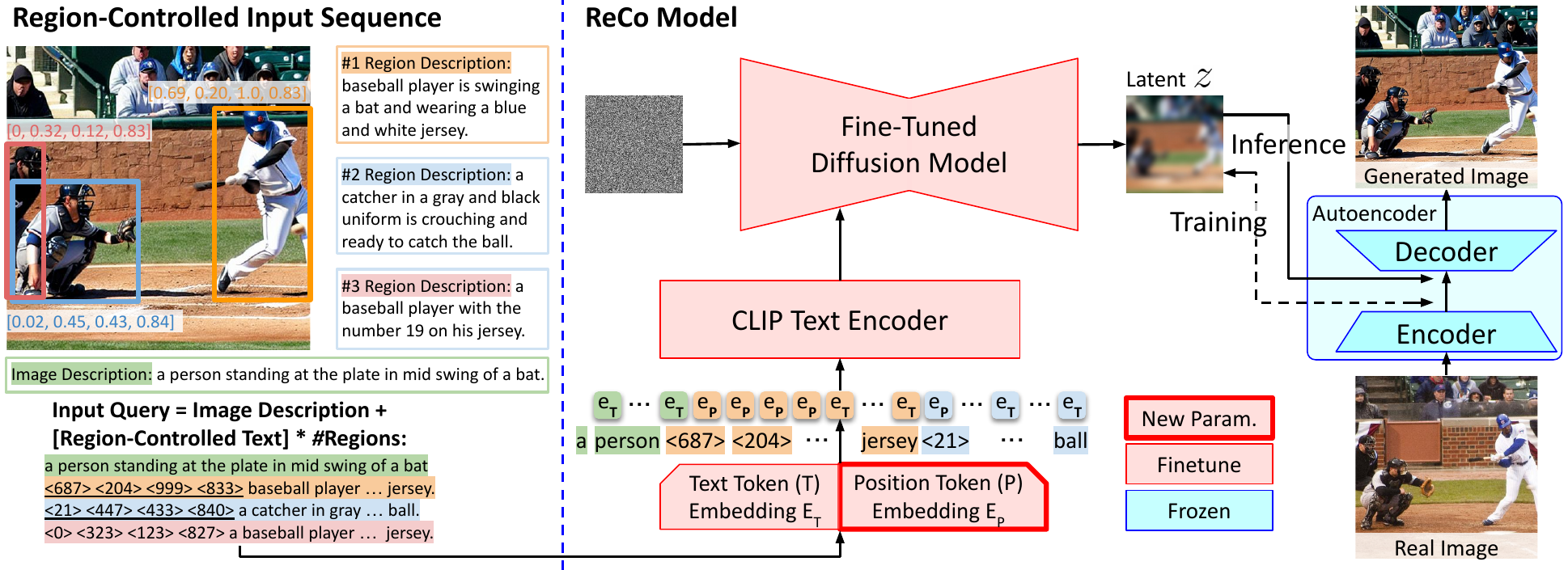} \\
\vspace{-0mm}
\caption{Overview of the ReCo model architecture. Image credit:~\cite{yang2023reco}. }
\label{fig:chp3_reco_arch}  
  \vspace{-1mm}
\end{figure}
\noindent\textbf{Region-controlled T2I generation.}
Large-scale T2I models have demonstrated remarkable efficacy in generating high-resolution images. However, the models lack spatial controllability, \emph{e.g.}, precisely specifying content in a specified area using a free-form text description. This limitation motivates the studies on region-controlled T2I generation. As shown in the left side of Figure~\ref{fig:chp3_reco_arch}, these studies explore the extra input condition of open-ended text descriptions on arbitrary regions (\emph{i.e.}, region-controlled text), augmenting the global image description in T2I models. This new input condition requires T2I models to understand spatial inputs, and associate them with grounded texts.

ReCo~\citep{yang2023reco} is among the most representative works along this direction. The core idea is to extend the text vocabulary of the text encoder $\mathcal{E}$ and arrange different tokens to represent the grounded text inputs. The study augments text tokens understood using pre-trained T2I models with an extra set of position tokens, which represent the quantized spatial coordinates. As shown in Figure~\ref{fig:chp3_reco_arch}, the position tokens (\emph{e.g.}, $\tiny{<}687{>},\tiny{<}204{>},\tiny{<}999{>},\tiny{<}833{>}$) are seamlessly mixed with the text tokens and operate as a spatial modifier, indicating that the text to follow only operates on the specified spatial region, such as the ``baseball player \ldots jersey.'' The pre-trained T2I model is then finetuned to support such a new input interface, thereby facilitating region-controlled T2I generation.

Shared by other approaches along this direction, ReCo discusses several advantages of region-controlled T2I generation in improving the alignment with human intents. 
$(i)$ The grounded texts provide an extra input condition that allows users to specify the desired image easily, \emph{i.e.}, having a free-form regional description precisely at a specific location. The box token and the input sequence design allow users to generate grounded text with the same user interface as query a T2I model  with text, making the extension easy to use. 
$(ii)$ The additional region-level controlled texts help better generate images with correct object count, spatial relationship, and region attributes such as color/size, which may otherwise confuse the vanilla T2I model~\citep{rombach2022high}. 
$(iii)$ Studies also observe a better image generation quality, with the conjecture that the grounded text provides object-level image-text association and therefore simplifies the learning process.

GLIGEN~\citep{li2023gligen} is another representative work. Alternate to generating grounded descriptions through the expansion of input tokens and finetuning the entire T2I model, GLIGEN uses a plug-and-play recipe: freezing the original T2I model and training extra gated self-attention layers to learn the new grounding skills. The grounding tokens carry two types of information: the semantic representation of text words that need to be grounded in and their spatial configurations. These grounding tokens are then added to the pre-trained T2I model via a newly added gated self-attention layer, with all remaining pre-trained parameters frozen. This layer is equipped with a gating parameter, which is initialized to zero, allowing the pre-trained model to incrementally incorporate the grounded text inputs. GLIGEN facilitates various types of grounded controls, including bounding box grounding, keypoint grounding, image prompting, as well as other types of spatially-aligned dense conditions.

\noindent\textbf{T2I generation with dense conditions.}
In addition to spatial coordinates, there exist other spatial conditions often represented as 2D arrays, such as segmentation masks, edge maps and depth maps. ControlNet~\citep{zhang2023adding} is a prominent example of incorporating these dense spatial controls into T2I models. ControlNet is built upon Stable Diffusion, and introduces an additional trainable ControlNet branch that adds an extra input condition to the text prompt. This extra condition can be a canny edge map, hough line, HED boundary, under sketching, human pose maps, segmentation masks, depth images, normal maps, or line drawing, each enabled with its distinct model copy. The added branch is initialized from the pre-trained downsampling blocks in the SD's U-Net. This branch takes the added visual latent and the extra dense condition as input. Before combining input dense conditions with visual latent in the input and merging the ControlNet branch's output back to SD's upsampling blocks, there is a unique zero-initialized $1\times1$ convolutional layer. This layer serves as a gated connector to gradually inject the extra condition into the pre-trained Stable Diffusion model. With the extra dense spatial control, ControlNet provides an effective channel of generation controllability.

\begin{figure}[t]
\centering  
\vspace{-4mm}
\includegraphics[width=1.00\textwidth]{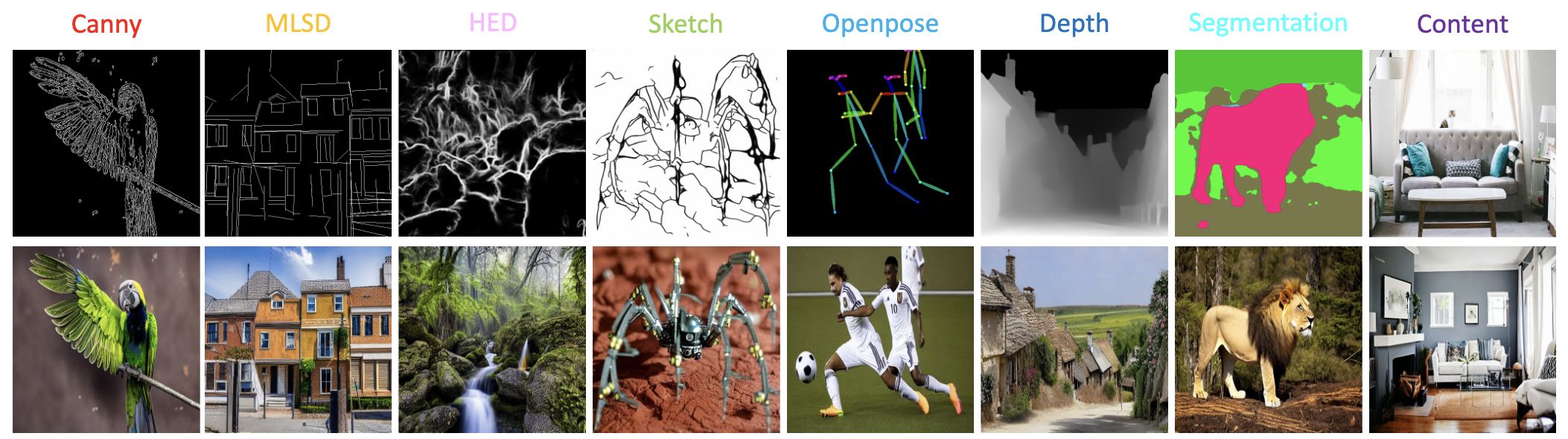} \\
\vspace{-0mm}
\caption{Examples of the dense controls and the corresponding generated images. Image credit:~\cite{zhao2023uni}. }
\label{fig:chp3_controlnet}  
  \vspace{-1mm}
\end{figure}
Follow-up studies such as Uni-ControlNet~\citep{zhao2023uni} and UniControl~\citep{qin2023unicontrol} further improve ControlNet by unifying the input condition, such that a single model can understand multiple input condition types or even take a combination of two conditions. Examples of the dense controls and the corresponding generated images are shown in Figure~\ref{fig:chp3_controlnet}.
Moreover, Disco~\citep{wang2023disco} exemplifies the efficiency of ControlNet in the generation of human dancing videos, which aims to generate videos with controllable elements such as human subjects, video backgrounds, and motion pose sequences. The study successfully separates the background and human pose conditions, which are fed into two distinct branches of ControlNet, which condition on image frames and pose maps, respectively. This disentanglement of control from all three conditions allows Disco to accomplish high fidelity in both the human foregrounds and backgrounds. More importantly, it enables the arbitrary compositionality of human subjects, backgrounds, and dance movements.

\begin{figure}[t]
\centering  
\vspace{-4mm}
\includegraphics[width=1.00\textwidth]{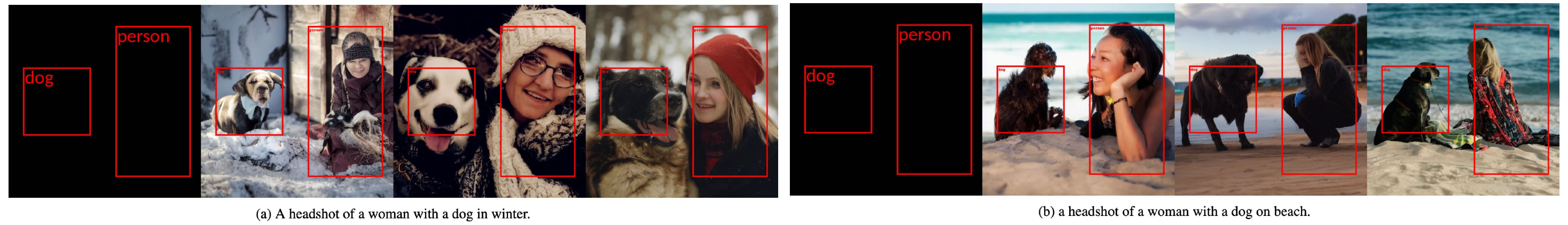} \\
\vspace{-0mm}
\caption{Qualitative results of inference-time spatial guidance. Image credit:~\cite{bansal2023universal}. }
\label{fig:chp3_inference_spatial}  
  \vspace{-1mm}
\end{figure}
\noindent\textbf{Inference-time spatial guidance.}
The aforementioned works require model training, either the T2I models or additional modules to understand the extra spatial conditions. Alternatively, studies~\citep{bansal2023universal,chen2023training} explore providing the inference-time spatial guidance to T2I models without extra model training. The core idea is similar to classifier guidance~\citep{dhariwal2021diffusion}, which takes a discriminator loss to guide the diffusion process as follows:
\begin{equation}
\label{equ:chp3_classguidance}
    \hat{\epsilon}'(z_{t},t) = \hat{\epsilon}(z_{t},t) + s(t)\cdot\triangledown_{z_t}\ell(c,f(\hat{z}_0)).
\end{equation}
Taking spatial control as an example, the discriminator can be a Faster-RCNN object detector~\citep{ren2015faster} indicated by $f$, which operates on the intermediate estimated image $\hat{z}_0$, and compute the object detection loss $\ell$ with the desired layout $c$, to guide the generation $\hat{\epsilon}(z_{t},t)$. $s(t)$ is the guidance strength. This approach enables the spatial control in T2I generation without extra training, with qualitative results shown in Figure~\ref{fig:chp3_inference_spatial}. However, it may not yield results as precise as those from finetuning methods~\citep{yang2023reco,li2023gligen,zhang2023adding}.

\noindent\textbf{Summary and trends.}
Early research on spatial controllable generation, such as layout-to-image and mask-to-image generation, was often treated in parallel with T2I generation. However, with the emergence of advanced large-scale T2I models, recent studies, as discussed in this subsection, are now leaning towards integrating spatial conditions with textual conditions. We identify two primary trends in integrating spatial conditions into T2I models. First, region-controllable T2I generation, such as ReCo, merges spatial coordinate control into text inputs by enlarging the text vocabulary with position tokens. Second, studies extended from ControlNet integrate an additional ``image-like'' condition to T2I frameworks, thereby capturing a broad spectrum of dense conditions. Moving forward, T2I models may have a finetuning stage that allows them to comprehend both image and text inputs. In such a scenario, box coordinates could be incorporated through text, while dense controls could be provided as image inputs. We will explore and elaborate on this idea in Section~\ref{sec:generation_conceptcust}.

\section{Text-based Editing}
\label{sec:generation_editing}
Text-to-image editing synthesizes new images from an given image and input text descriptions. Instead of producing an image entirely from scratch, users might already possess a satisfactory starting point; this could be an image previously generated from a T2I model or a natural image. The objective is to retain the majority of the visual content, only modifying specific components. This could involve altering a local object or the overall image style to precisely match the user's intentions. This text-based editing approach offers users a tool to generate fresh images based on a predecessor, playing a crucial role in creating visual content that accurately follows human intent.

There are various definitions and task setups in text-based editing. We introduce the following representative threads. 
$(i)$ One classic editing scenario is to change a local image region, such as removing or changing an object or adding an object in a certain region. Spatially manipulating the latent in image generation according to the user-generated masks is a simple but effective method~\citep{avrahami2022blended,avrahami2022blendeddiff,meng2021sdedit}. Studies~\citep{balaji2022ediffi,hertz2022prompt} also show that manipulating the image-text cross-attention mask is effective for spatial editing. 
$(ii)$ Extended from spatial editing where the language inputs describe the desired appearance in the spatial region, language can also be used as editing instruction to tell the machine what to do~\citep{kawar2023imagic,brooks2023instructpix2pix}, such as ``change object A in the image to object B.'' 
$(iii)$ Instead of extending a single T2I model for editing, editing systems~\citep{wu2023visual} integrate different specialized modules such as segmentation models~\citep{kirillov2023segment,zou2023segment} and large language models~\citep{brown2020language,gpt4}.

\noindent\textbf{Diffusion process manipulations.}
The multi-step denoising process in diffusion image generation naturally supports a certain extent of image editing. Stochastic Differential Editing (SDEdit)~\citep{meng2021sdedit} shows that first adding noises to the input image to edit and then subsequently denoising the sample, could produce a meaningful edit. Blended Latent Diffusion~\citep{avrahami2022blendeddiff} shows that the diffusion process manipulation can achieve local object editing with a user-generated object mask $m_{latent}$. In each diffusion step, the latent $z$ is a spatial blend of the foreground and background latent: $z=z_{fg}\odot m_{latent} + z_{bg}\odot (1-m_{latent})$, where $z_{fg}$ is the edited object generated from the text description and $z_{bg}$ is the original backgrund image with noises added.

\begin{figure}[t]
\centering  
\vspace{-4mm}
\includegraphics[width=1.00\textwidth]{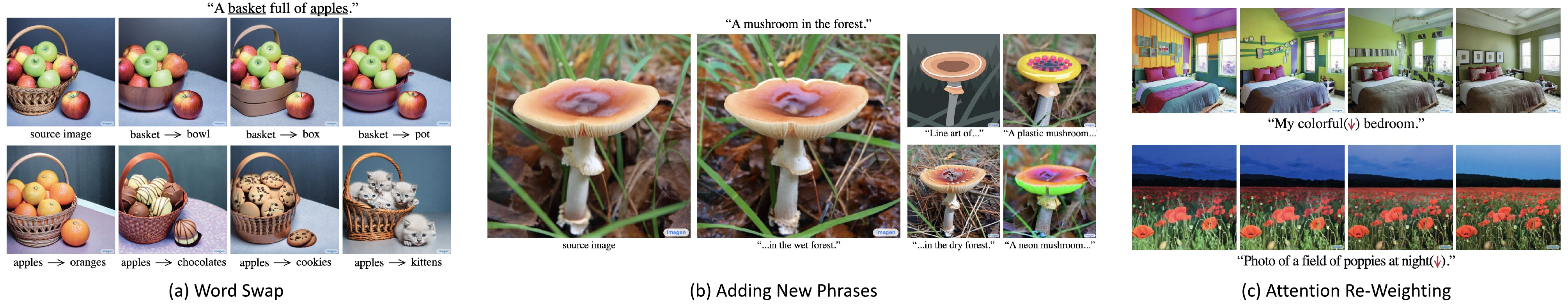} \\
\vspace{-0mm}
\caption{Three types of editing (word swap, adding new phrases, attention re-weighting) on synthetically generated images, enabled by attention map manipulation. Image credit:~\cite{hertz2022prompt}. }
\label{fig:chp3_prompt2prompt}  
  \vspace{-1mm}
\end{figure}
However, there are certain limitations on blending spatial latents. Firstly, it may not always be feasible to require human-generated masks. Secondly, the generation process can sometimes result in artifacts at the edges. Instead of simply blending the latent in a spatial manner, researchers delve into image-text cross-attention maps to unearth clues for object editing. Specifically, Prompt2Prompt~\citep{hertz2022prompt} discovers that cross-attention layers control the interaction among visual regions and text words. Based on this observation, the study enables three types of editing for images generated by a diffusion T2I model, including word swap, adding new phrases, and attention re-weighting, each of which is enabled with corresponding manipulation on the image-text cross-attention map.  Specifically, the Prompt2Prompt tracks both cross-attention maps generated by the original prompt (namely $M_t$) and the edited prompt (namely $M^*_t$), and merges the attention maps with pre-defined rules into the new attention maps $\widehat{M}_t$, which is used for latent computing. For example, while adding a new phrase, attention map $\widehat{M}_t$ remains unaltered for words present in the original prompt. It only incorporates the modified attention maps $M^*_t$ for words that did not exist in the original prompt. Qualitative results of the edits are shown in Figure~\ref{fig:chp3_prompt2prompt}.

Going beyond editing synthetically generated images, Imagic~\citep{kawar2023imagic} explores editing real natural images. The core idea is to represent the image to be edited as text embedding, and blend this embedding with the target text embedding describing the desired image. This blend ensures that the resulting image retains elements from the original while aligning with the aesthetics detailed in the target textual prompt. In practice, test-time finetuning is needed to generate high-quality images.

\begin{figure}[t]
\centering  
\includegraphics[width=1.00\textwidth]{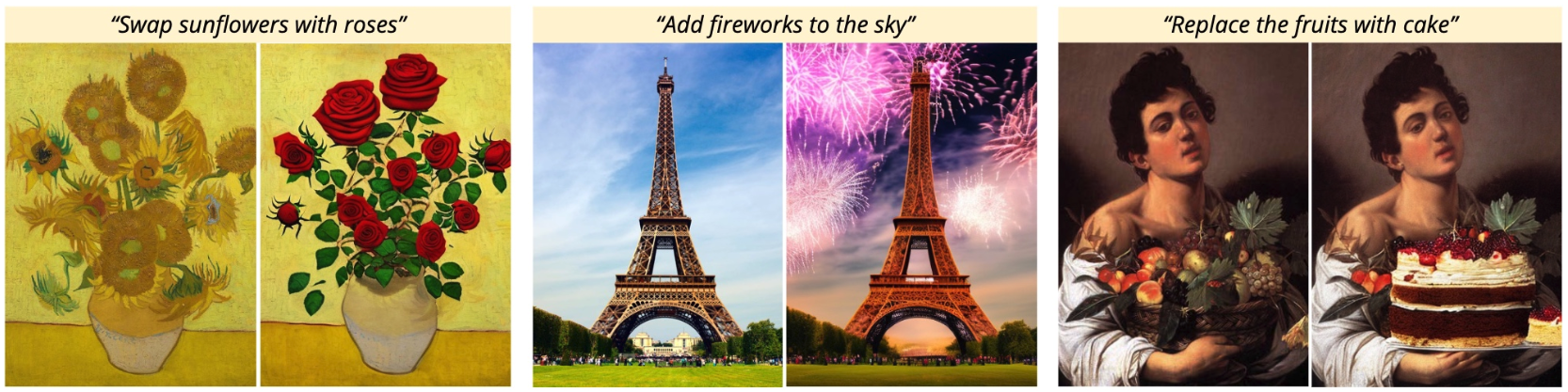} \\
\vspace{-0mm}
\caption{Examples of text instruction editing. Image credit:~\cite{brooks2023instructpix2pix}. }
\label{fig:chp3_instructpix}  
  \vspace{-1mm}
\end{figure}
\noindent\textbf{Text instruction editing.}
Instead of repeating the visual contents of the image to edit in the text prompts, it might be more efficient for users to directly specify editing instructions using language, such as ``swap sunflowers with roses'' in Figure~\ref{fig:chp3_instructpix}. The desired text instruction editing model should work on both model-generated and natural images, and across different types of editing instructions.

\begin{figure}[t]
\centering  
\vspace{-4mm}
\includegraphics[width=1.00\textwidth]{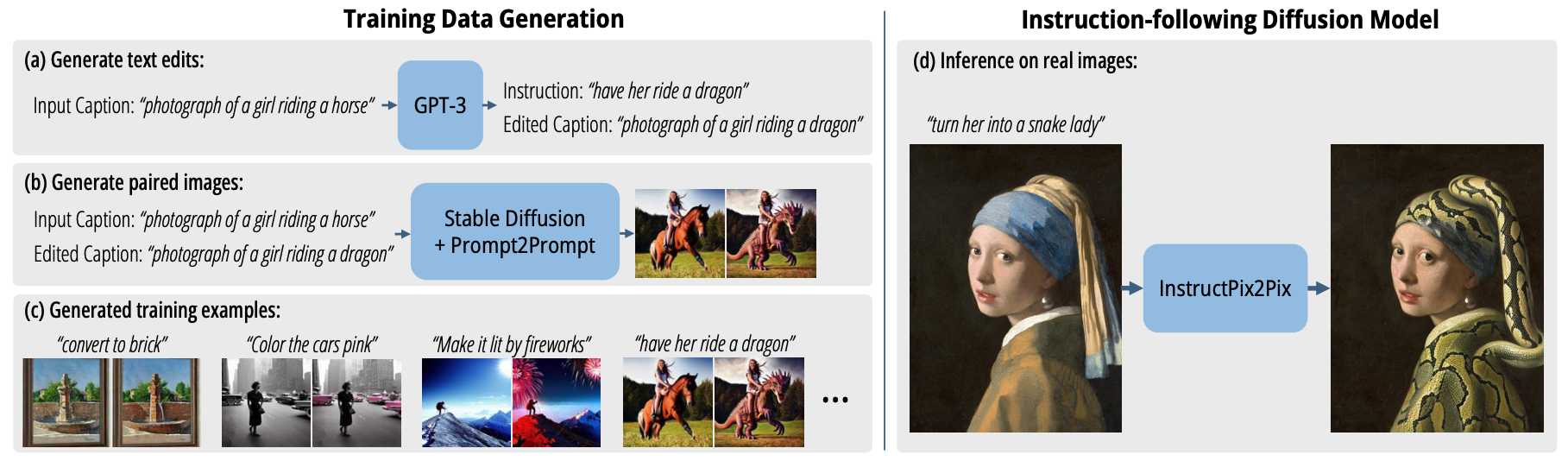} \\
\vspace{-0mm}
\caption{The editing data generation pipeline proposed in InstructPix2Pix. Image credit:~\cite{brooks2023instructpix2pix}. }
\label{fig:chp3_instructpixdata}  
  \vspace{-1mm}
\end{figure}
InstructPix2Pix~\citep{brooks2023instructpix2pix} is designed to accept an image and a text editing instruction to produce an edited version of the input image. The goal is to train an image-to-image model that can understand such editing text instructions. To achieve this, T2I models can be adapted to accept the additional image input by incorporating more input channels into the SD's convolutional layer. The main challenge is how to generate paired editing data. As shown in Figure~\ref{fig:chp3_instructpixdata}, InstructPix2Pix~\citep{brooks2023instructpix2pix} proposes to use a LMM~\citep{brown2020language} to generate a pair of an editing instruction and an edited caption from the original input caption, \emph{e.g.}, ``have her ride a dragon,'' ``photograph of a girl riding a dragon,'' and ``photograph of a girl riding a horse.'' The study then uses Prompt2Prompt~\citep{hertz2022prompt} to convert the original and edited caption pair to a pair of images before and after editing, corresponding to the GPT-generated editing instruction. The study generates over 450K samples to train the editing model. This data generation method has also been adopted in subsequent research, such as CM3Leon~\citep{ge2023planting} for training general-purpose image-text-to-image models.

\noindent\textbf{Editing with external pre-trained models.}
Furthermore, recent studies show the efficacy of incorporating external language and vision models for editing, as opposed to relying solely on a single model. Advancements in generalist segmentation models, such as SAM~\citep{kirillov2023segment} and SEEM~\citep{zou2023segment}, have paved the way for using segmentation models to ground the region for text-based editing. Representative works include Instruct X-Decoder~\citep{zou2023generalized}, Grounded SAM inpainting~\citep{liu2023grounding}, Inpaint anything~\citep{yu2023inpaint}, \etc.
Another emerging trend is the allocation of various generation and editing tools through LMM. Studies such as VisualChatGPT~\citep{wu2023visual} can solve complicated visual editing that requires the collaboration of multiple generation and editing models in multiple steps.

\noindent\textbf{Summary and trends.}
%
Text-based editing models have made significant progress in their capabilities, leading to improved editing quality, expanded domain coverage, and more flexible user interface.
For example, early studies require user-generated masks for object editing, while recent models can work on synthetically generated images without mask inputs, or even directly understand general text editing instructions. As we look to the future, we anticipate an all-encompassing generative foundation model that is capable of processing both image and text inputs. Within this framework, editing instructions would be a specialized form of text input, seamlessly integrated with the image description in T2I generation.
\section{Text Prompts Following}
\label{sec:generation_promptfollow}
Training with image-text pairs encourages T2I models to generate images that semantically correspond to the input text condition. However, the image generation training objective does not directly enforce generated images to exactly follow text prompts. Studies~\citep{feng2022training,chefer2023attend} show that T2I models may fail to follow text prompts, especially when the image description becomes complicated. For example, certain noun phrases may get omitted, attributes may apply to incorrect objects, and generated images may have the wrong object count, relationship, styles, \etc. These limitations motivate work on improving T2I models to better follow text prompts.

The related literature can be broadly categorized into two main groups. 
$(i)$ {\it Inference-time manipulation.}  In the inference stage, the latent and attention adjustment~\citep{liu2022compositional,feng2022training,chefer2023attend,agarwal2023star} design various methods to redistribute the visual latent or image-text cross-attention, such that all noun phrases in the text prompts are represented in the generated image.
$(ii)$  {\it Alignment tuning.} An extra model learning stage is learned~\citep{black2023training,fan2023dpok}, typically with the image-text similarity as rewards, such that the tuned T2I model can better follow text prompts.

\begin{figure}[t]
\centering  
\vspace{-4mm}
\includegraphics[width=1.00\textwidth]{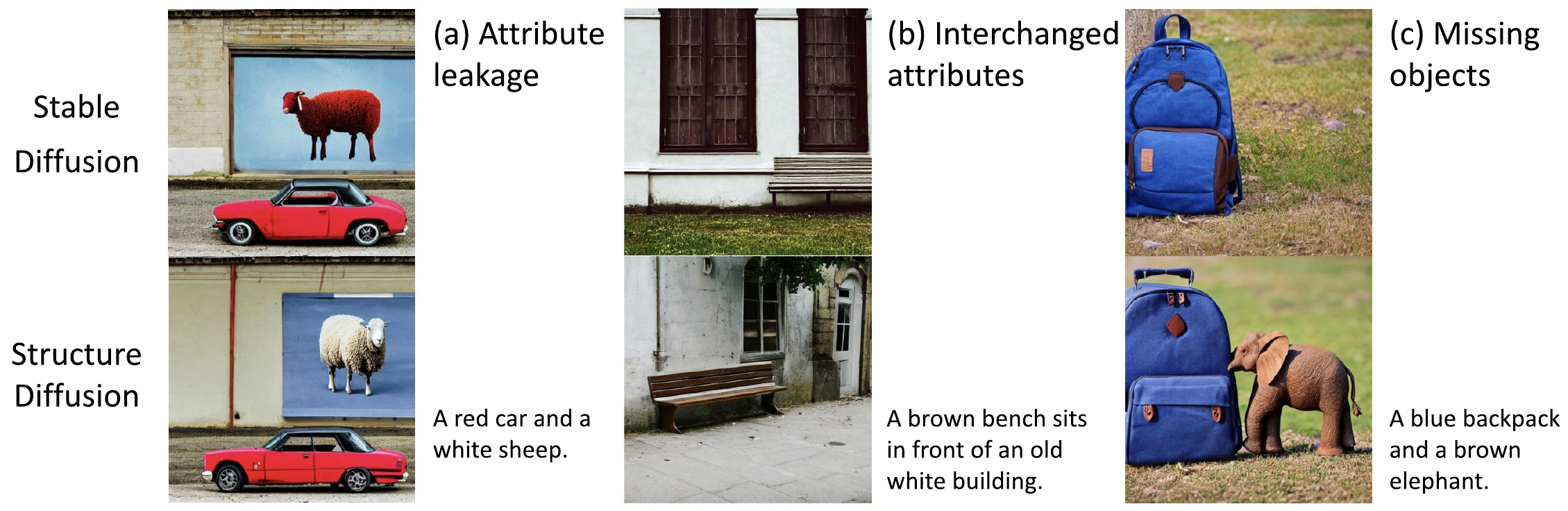} \\
\vspace{-0mm}
\caption{Failure cases of vanilla T2I model in text prompt following. Image credit:~\cite{feng2022training}. }
\label{fig:chp3_structurediff}  
\end{figure}
\noindent\textbf{Inference-time manipulation.}
Training with image-text pairs does not guarantee that T2I models consistently adhere to the text prompts. There can be multiple discrepancies, particularly when the text descriptions are lengthy and intricate. For instance, T2I models may apply attributes to the wrong entity or miss certain objects, as shown in Figure~\ref{fig:chp3_structurediff}. Intuitively, parsing the text query at inference time and explicitly enforcing T2I models to pay closer attention to each noun phrase may generate images that better follow text prompts.

Building upon this intuition, StructureDiffusion~\citep{feng2022training} employs a parsing tree to extract noun phrases and the text prompt's linguistic structure. The study then enforces the model to ``look at'' all extracted noun phrases. This is implemented by modifying SD's cross-attention mechanism introduced in~\eqref{equ:chp3_crossattn}, written as $O=M\cdot V$ where $M$ is the softmax cross-attention map. Instead of producing $M$ with the sentence feature $V$, which may result in words getting overlooked, the study computes the $O=\frac{1}{k+1}\Sigma_{i=0}^{k}(M\cdot V_i)$, where $V_0$ is the sentence feature $V$, and $V_i, i=1,\ldots,k$ is the phrase feature in the parsing tree. This approach ensures that the visual stream maintains a balanced attention across all identified noun phrases, fostering more accurate image generation.

\begin{figure}[t]
\centering  
\vspace{-2mm}
\includegraphics[width=1.00\textwidth]{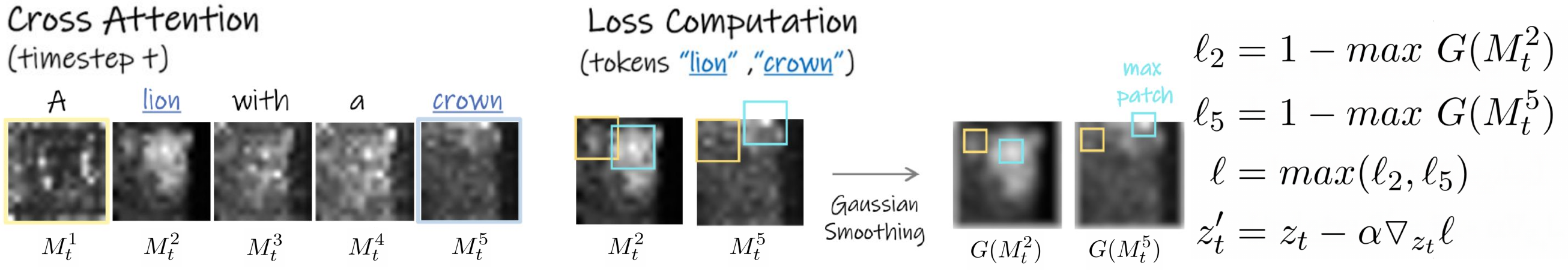} \\
\vspace{-0mm}
\caption{Inference time guidance proposed in Attend-and-Excite. Image credit:~\cite{chefer2023attend}. }
\label{fig:chp3_attend_excite}  
  \vspace{-1mm}
\end{figure}
Motivated by the same objective to ensure that no object is overlooked, Attend-and-Excite~\citep{chefer2023attend} manipulates the attention map. As shown in the right side equations in Figure~\ref{fig:chp3_attend_excite}, a regularization loss $\ell$ is computed to amplify the maximal attention towards the most neglected subject token:
\begin{equation*}
    \ell = \max_{n=1,\ldots,N_\text{sub}}(1-max\ G(M_t^n))
\end{equation*}
where $G$ is a Gaussian kernel to smooth the attention map and $N_\text{sub}$ is the number of subject tokens. The loss is then used to update the latent $z_t$ at inference time:
\begin{equation*}
    z_t' = z_t-\alpha\triangledown_{z_t}\ell,
\end{equation*}
where $\alpha$ is a scalar for the step size. Results show that this inference-time guidance enables T2I models to focus more on objects described in the text prompt, resulting in superior image generation. Follow-up studies~\citep{agarwal2023star} further refine the guidance loss to optimize prompt-following performance.

\noindent\textbf{Model tuning to follow text prompt.}
Instead of inference-time manipulation, one may wonder if we can refine a pre-trained T2I model to better follow text prompts. One promising way to achieve this is via reinforcement learning, using image-text similarity as reward instead of the image generation objective used in the main T2I training. This allows the model to be optimized towards a better image-text alignment.

\begin{figure}[t]
\centering  
\vspace{-3mm}
\includegraphics[width=1.00\textwidth]{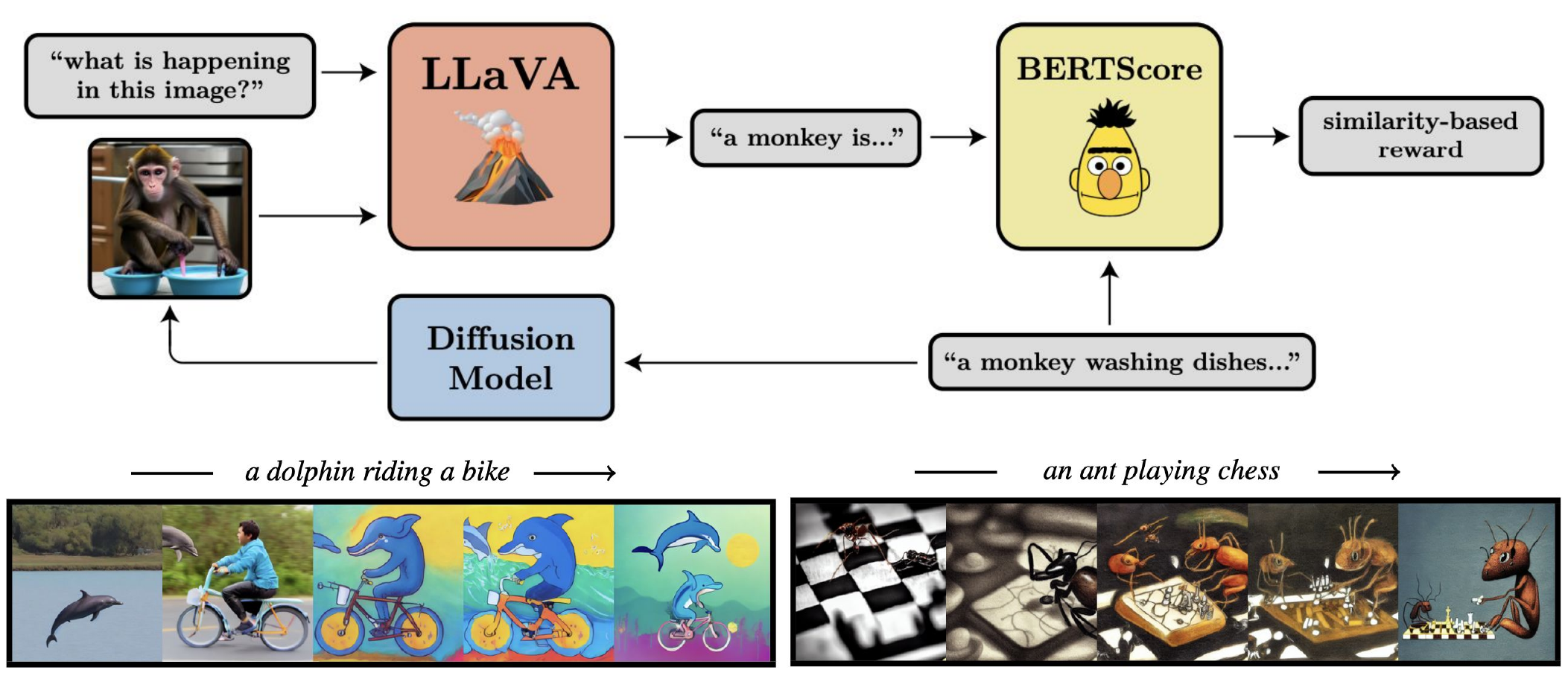} \\
\vspace{-0mm}
\caption{DDPO with vision-language-model-based reward function for image-text alignment tuning. Image credit:~\cite{black2023training}. }
\label{fig:chp3_ddpo}  
  \vspace{-1mm}
\end{figure}

One work along this direction is the denoising diffusion policy optimization (DDPO)~\citep{black2023training}, with the tuning framework shown in Figure~\ref{fig:chp3_ddpo}. Specifically, a vision-language model~\citep{li2023llava} is used to convert the generated image into a text description. This generated caption is compared with the input text prompt, deriving a similarity reward through the use of BERTScore~\citep{zhang2019bertscore}. The similarity reward is then used to finetune the pre-trained T2I model, such that the model can better follow the text prompts. The bottom of Figure~\ref{fig:chp3_ddpo} shows the progression of the generated sample during this similarity-based training. Further, it is worth noting that other human intent may also be formatted as rewards for model tuning, such as compressibility, aesthetic quality, \etc

\noindent\textbf{Summary and trends.}
In this section, we present studies aimed at enhancing the capability of T2I models to better adhere to text prompts. Despite the good performance achieved by the inference-time manipulation, the alignment tuning provides a more intuitive user experience, eliminating the need for extra modifications. In parallel to instruction tuning in LLMs to align human intent for text generation, the TI2 model tuning shares a similar goal, but focuses on image generation. We foresee a similar paradigm emerging in the near future for generative T2I foundational model development. Specifically, the initial training phase still relies on the existing image generation objective on large-scale data, while the subsequent alignment-tuning phase enhances the model's prompt adherence and other aspects of human intent, such as diminishing harmful content. Current RL-based methods show potentials, but they typically zero in on a single optimization goal. Future research could extend these methods for more adaptable alignment tuning, amalgamated with features like accommodating diverse image and text scenarios.

\section{Concept Customization}
\label{sec:generation_conceptcust}
Though language is an powerful medium to express human intent, it is inefficient in comprehensively describing all details of a visual concept for reconstruction. For example, it is challenging to use texts to describe my pet dog or family members with sufficient details, so that they can be generated in different visual scenes. In such applications, directly extending T2I models to understand visual concepts via image inputs is a better option. 

We examine relevant research on visual concept customization, which offers users the ability to generate these personalized concepts.
$(i)$ Pioneer studies~\citep{gal2022image,ruiz2023dreambooth,wei2023elite} start with single-concept customization that involves test-time finetuning to encode multiple images of the visual concept into a new token embedding, such that the learned embedding can be used to refer to the concept during T2I generation.
$(ii)$ Multi-concept customization~\citep{kumari2023multi,avrahami2023break} allows multiple concept tokens to be expanded from the T2I model's token vocabulary, enabling multiple concepts to interact with each other and the remaining visual scene during generation.
$(iii)$ Test-time finetuning requires users to tune T2I models for each new concept to customize. To simplify the usage, studies~\citep{chen2022re,shi2023instantbooth,chen2023subject,yang2023paint} explore customization without test-time finetuning and uses a unified finetuning stage to extend T2I models for accepting image condition inputs. The models take images of the visual concept as an extra input condition, and generate images with the visual concept following the text descriptions.

\begin{figure}[t]
\centering  
\vspace{-3mm}
\includegraphics[width=1.00\textwidth]{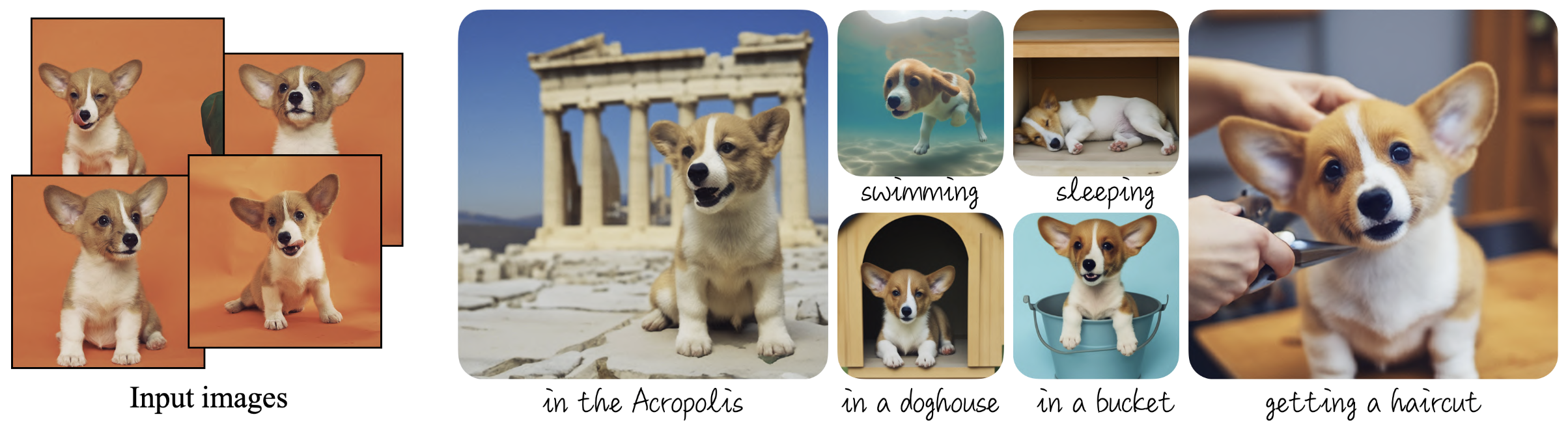} \\
\vspace{-0mm}
\caption{The problem setup and result visualization of the visual concept customization task. Image credit:~\cite{ruiz2023dreambooth}. }
\label{fig:chp3_dreambooth}  
  \vspace{-1mm}
\end{figure}
\noindent\textbf{Single-concept customization.}
The goal of visual concept customization is to enable T2I models to comprehend additional visual concepts tailored to very specific cases.
The problem setup, studied in Textual Inversion~\citep{gal2022image}, involves translating visual concepts from a handful of images into unique token embeddings. As illustrated in the left side of Figure~\ref{fig:chp3_dreambooth}, the T2I model processes four images of a distinct dog breed, subsequently learning the embedding for a new token, denoted as $[\text{V}]$. This $[\text{V}]$ token can be used as a text token to represent this specific dog. The $[\text{V}]$ token can be seamlessly integrated with other textual descriptions to render the specific dog in various contexts, such as swimming, in a bucket, and getting a haircut. 

Textual Inversion~\citep{gal2022image} learns the $[\text{V}]$ token embedding via prefix tuning, \emph{i.e.}, freezing all T2I model's parameters and training the $[\text{V}]$ token embedding to generate the input images. Later studies observe that tuning more model parameters leads to significantly better image generation quality. However, adjusting only the input image may lead to the risk of overfitting the T2I model for a particular concept, and losing the capablity to generate diverse images. For instance, the model might become unable to generate various dog types. To address this, Dreambooth~\citep{ruiz2023dreambooth} proposes the class-specific prior preservation loss. Central to this approach is using the pre-trained T2I model to produce images of the same class as the targeted customization concept. The model is then jointly finetuned on both the input image (with the $[\text{V}]$ token) and the model-generated images (without the $[\text{V}]$ token). This ensures that the model can differentiate between the unique ``$[\text{V}]$ dog'' and other general dogs it was initially trained, thus maintaining its overall T2I capability. Dreambooth then finetunes all T2I model parameters and achieves better image generation quality.

\noindent\textbf{Multi-concept customization.}
Building on studies that focused on learning a single visual concept $[\text{V}]$, recent research has delved into the possibility of integrating multiple visual concepts into a single Text-to-Image (T2I) model, represented as $[\text{V}_\text{1}]$, $[\text{V}_\text{2}]$, and so on. Custom Diffusion~\citep{kumari2023multi} employs a selective subset of model weights, specifically the key and value mappings from text to latent features in the cross-attention layers for concept customization, learned from multiple sets of concept images. The study facilitates the ability to embed multiple customized visual concepts in a single text prompt.
Instead of learning from multiple sets of input images, Break-A-Scene~\citep{avrahami2023break} explores extracting multiple visual concepts in a single image. The study augments input images with segmentation masks to pinpoint the intended target concepts and subsequently transforms them into a series of concept embeddings denoted as $[\text{V}_\text{i}]$.

\noindent\textbf{Customization without test-time finetuning.}
While the concept customization studies, as described above, have achieved good visual quality, the necessity for test-time finetuning hinders its application in real-world settings. Most end users and application platforms lack the compute resources required for finetuning, not to mention the complexities of finetuning process. This naturally leads to the question: can we take concept images as input conditions, and achieve concept customization without finetuning?

\begin{figure}[t]
\centering  
\vspace{-3mm}
\includegraphics[width=1.00\textwidth]{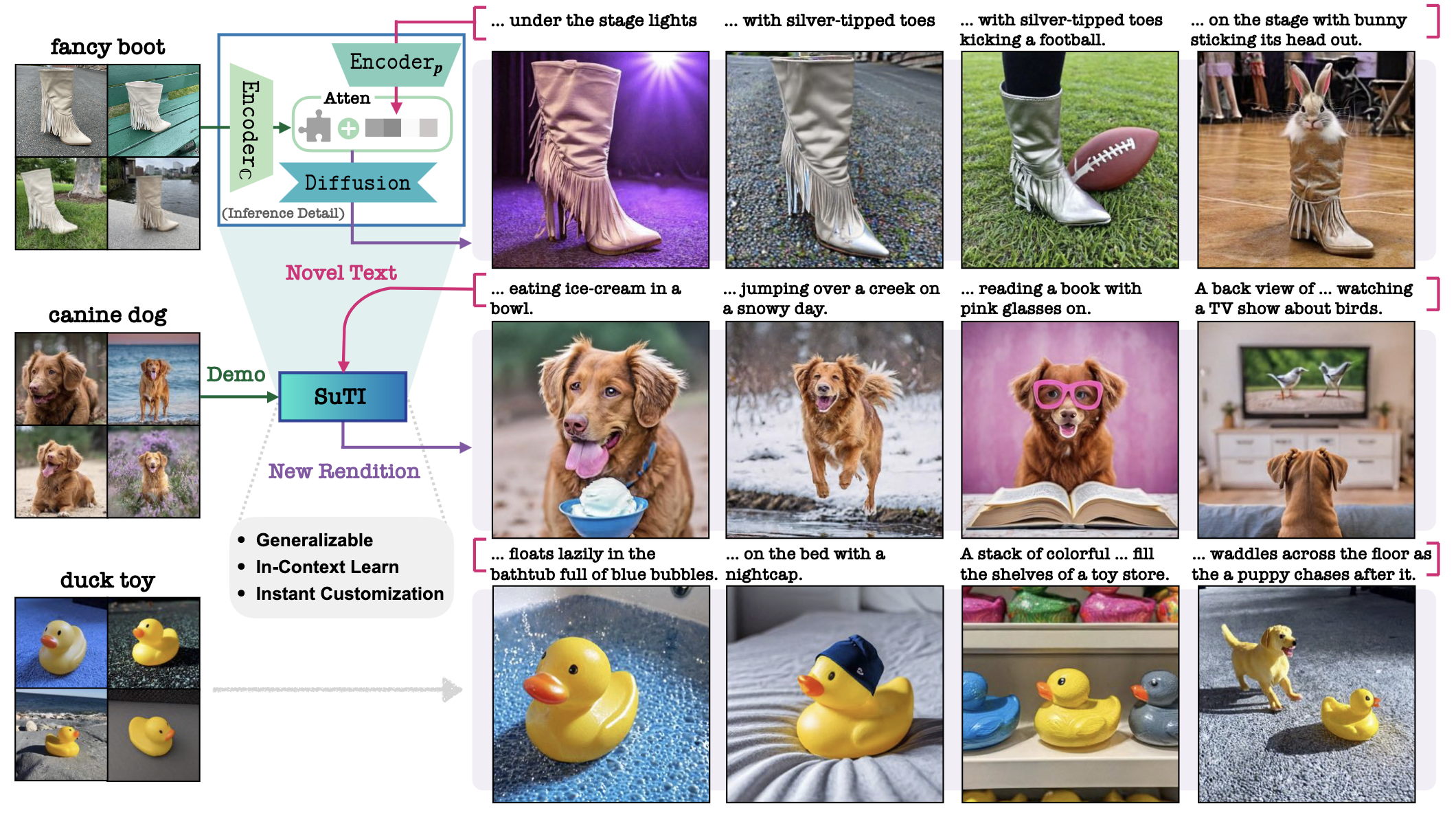} \\
\vspace{-0mm}
\caption{Illustration of in-context concept customization without test-time finetuning. Image credit:~\cite{chen2023subject}. }
\label{fig:chp3_suti}  
  \vspace{-1mm}
\end{figure}

The input/output format of the imagined system is similar to the retrieval-augmented generation~\citep{chen2022re}, which aims to ease the image generation by conditioning on a retrieved similar image. The system supports extra image inputs that contain relevant information for the generation process. By altering the conditioning images during the training phase, the model can potentially achieve a broad in-context learning capability, producing images that align with the given input examples. In line with this framework, SuTI~\citep{chen2023subject} trains a single model to imitate the finetuned subject-specific experts, and generates images conditioning on both text and subject input images, as shown in Figure~\ref{fig:chp3_suti}. As a result, the trained model can perform in-context concept customization without test-time finetuning, and generalize to unseen subjects and descriptions. Another concurrent work, InstantBooth~\citep{shi2023instantbooth}, also shows remarkable results in generating images that are not only aligned with language but also preserve identities, with a single forward pass.

\noindent\textbf{Summary and trends.}
The field of visual concept customization has advanced from finetuning embeddings during the testing stage, to directly performing in-context image generation with a frozen model. The in-context generation pipeline, which incorporates additional image inputs, shows remarkable potentials in real-world applications. In this subsection, we have explored two applications of this approach: facilitating generation through the retrieval of pertinent images~\citep{chen2022re}, and personalizing visual concepts by conditioning them on subject images~\citep{chen2023subject,shi2023instantbooth}. An intriguing direction is to unify the diverse uses of image inputs, directed by descriptive textual instructions. We elaborate on this idea in the following sub-section.

\section{Trends: Unified Tuning for Human Alignments}
\label{sec:generation_trends}
In previous subsections, we presented the literature related to tuning T2I models to more accurately align with human intent. This includes enhancing spatial controllability, editing existing images for improved alignment, more effectively following text prompts, and personalizing T2I models for new visual concepts.
A trend observed across these subtopics is the shift towards integrated alignment solutions that require minimal problem-specific adjustments. Along this direction, we envision a future T2I model having a unified alignment tuning stage, which transforms a pre-trained T2I model into one that resonates more intimately with human intent. Such a model would seamlessly process both text and image inputs, generating the intended visual content without the need for multiple models tailored to different alignment challenges. 

\begin{figure}[t]
\centering  
\vspace{-3mm}
\includegraphics[width=1.00\textwidth]{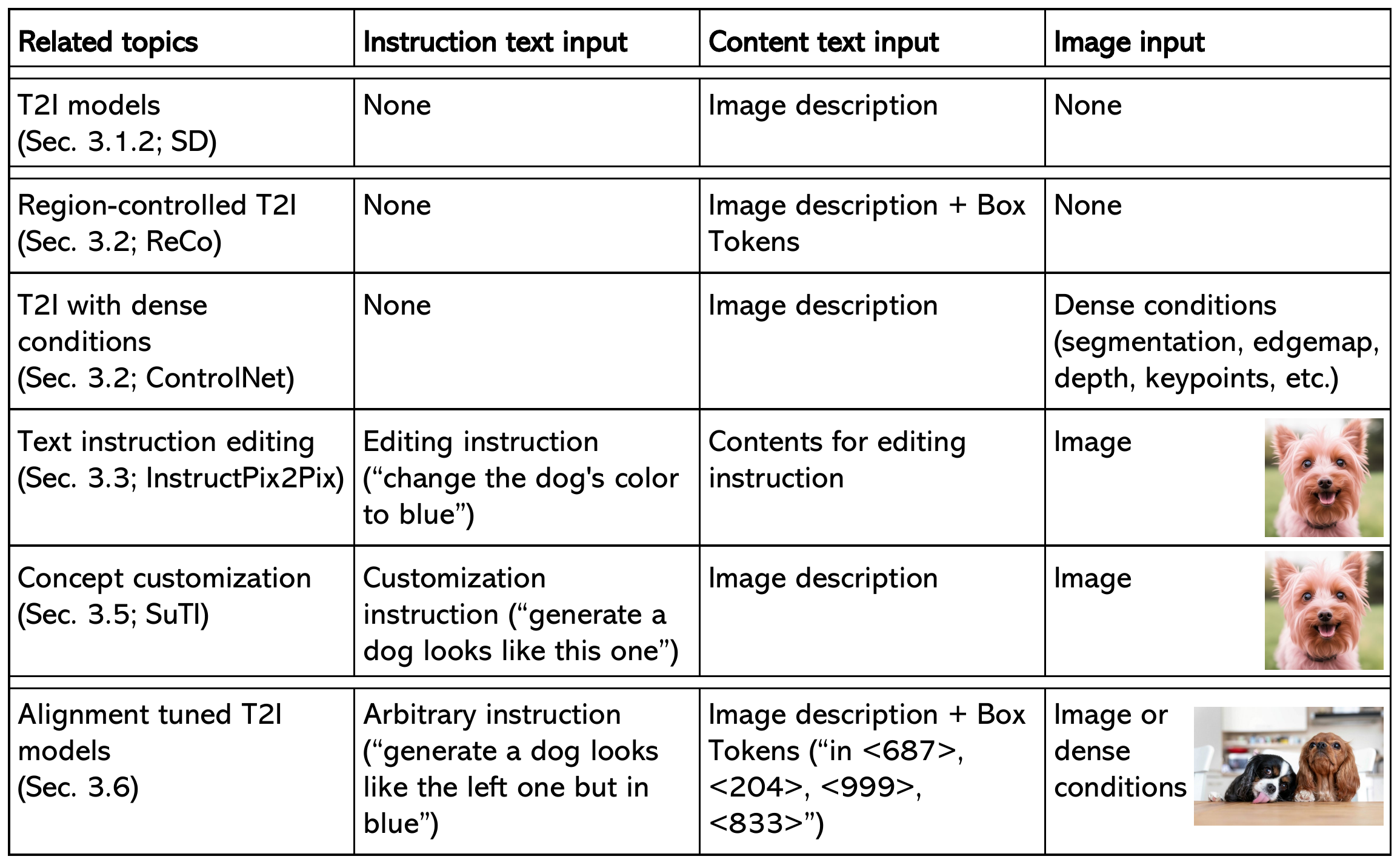} \\
\vspace{-0mm}
\caption{Overview of the unified image and text input interface for human alignments, and its connection to previous sub-sections. }
\label{fig:chp3_unified_t2i}  
  \vspace{-1mm}
\end{figure}

Drawing parallels to the established practice of human-alignment tuning in LLM development, we anticipate that the techniques reviewed in this section will merge into a holistic second-stage tuning for generative foundation model development. This alignment tuning phase serves two primary purposes. First, it extends the T2I's text inputs to incorporate interleaved image-text inputs, as illustrated in Figure~\ref{fig:chp3_unified_t2i}. Second, it finetunes the base T2I model, which has been trained using image generation loss, by the employing data, loss, and rewards that aim to align with human expectations.

\noindent\textbf{Unified image and text inputs.}
We begin with the discussion on interface unification. Specifically, we aim to evolve the textual inputs of T2I models into a multimodal interface that seamlessly integrates both image and text inputs. As shown in Figure~\ref{fig:chp3_unified_t2i}, we consider three types of inputs to begin with: "content text input" characterizes the visual scene to be produced; the "image input" accommodates dense 2D inputs such as images and dense conditions; and the "instruction text input" explains how the input content texts and images should be collectively composed as the condition for generation.

Vanilla T2I models, as shown in the first row of Figure~\ref{fig:chp3_unified_t2i}, take the ``content text input'' of the image description and generate the corresponding image. For the spatial controllable generation in Section~\ref{sec:generation_spatial}, the extra spatial condition can be specified via text inputs by expanding text words with extra box tokens, or via image input by feeding the dense spatial conditions as an image input. For the text-based editing in Section~\ref{sec:generation_editing}, we examine the efficacy of text instruction editing, a task that finetunes the T2I model to comprehend editing instruction texts that manipulate the image input, altering its pixel values accordingly. For visual concept customization in Section~\ref{sec:generation_conceptcust}, the training-free models can now understand customization instructions to extract visual concepts from the image inputs, and combine the concept with context text inputs for image generation.

Incorporating the three elements of the input interface, the envisioned alignment-tuned T2I model can handle all previous tasks described in Section~\ref{sec:generation_spatial}-\ref{sec:generation_conceptcust}. Its behavior is steered by specific text instructions that dictate how the image and text inputs should be jointly processed as the generation condition. 
Given the same image input, different text instructions can invoke different tasks: ``generate a cat image with the same layout'' for spatial control, ``change the dog's color'' for editing, ``generate the same dog sleeping'' for concept customization, and the arbitrary mixture of the existing modes.
Achieving such a unified interface in generative foundational models may be possible through training on a consolidated dataset encompassing data from various tasks, drawing similarities to the success of supervised instruction tuning observed in LLMs. Furthermore, transitioning from processing a single image-text pair to handling interleaved image-text pairs could enable more intriguing capabilities like in-context visual demonstrations~\citep{sun2023imagebrush}.
Another interesting direction is to build a generative model that is capable of generating any combination of output modalities, such as language, image, video, or audio, from any combination of input modalities, as demonstrated in Composable Diffusion (CoDi)~\citep{tang2023anytoany}.

\noindent\textbf{Tuning with alignment-focused loss and rewards.}
In addition to the unified input interface, another noteworthy element deserving consideration is the alignment-focused loss and rewards. As mentioned in Section~\ref{sec:generation_promptfollow}, the image generation loss based on image-text pairs enables models to produce images that match the target data distribution. Yet, it doesn't always perfectly align with human intent. This is reminiscent of the language model loss in LLM training, which necessitates a separate alignment tuning phase~\citep{ouyang2022training}. The recent success in supervised instruction tuning and reinforcement learning from human feedback methods~\citep{black2023training} on image generation provides effective tools for similar alignment tuning in generative foundation models. An intriguing topic left for future exploration is how to balance the different target losses and rewards, such as jointly optimizing for higher aesthetic scores, better image-text alignment, fewer harmful contents, stronger instruction adherence, along with many other desired properties.

\noindent\textbf{Closed-loop of multimodal content understanding and generation.}
As we look ahead, one promising avenue of research is the closed-loop integration of multimodal content understanding and generation. Preliminary studies have shown the benefit of using synthesized data to benefit generation from understanding~\citep{li2023benchmarking,he2022synthetic}, and vice versa. An exciting prospect would be the development of an image-text-input, image-text-output foundational model for both understanding and generation tasks. The ideal balance in combining these two dimensions, and the most efficient approach to achieve it, are left for future explorations.
\chapter{Unified Vision Models}
\label{chp:generalist}
\begin{wrapfigure}{r}{4cm}
  \centering
  \vspace{-6cm}
  \includegraphics[width=1.0\linewidth]{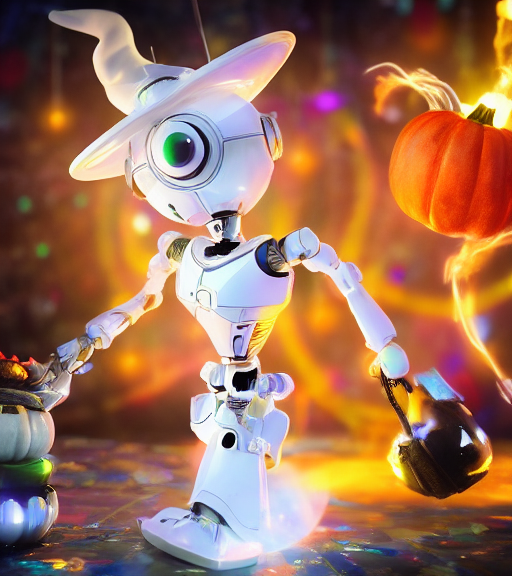}
\end{wrapfigure}

In this chapter, we discuss the unification of vision models. We start with an overview of the challenges in the unification of vision models and the most recent efforts towards this goal in Section \ref{sec:chp4_overview}. What follows are detailed discussions on $(i)$ how to transform closed-set models to open-set ones in Section \ref{sec:chp4_from_close_to_open}; $(ii)$ how to unify different granularities of vision tasks in Section \ref{sec:chp4_from_specific_to_generic}; and $(iii)$ how to build a more promptable interface for vision in Section \ref{sec:chp4_from_static_to_promptable}. Finally, we summarize the chapter and discuss future trends in Section \ref{sec:chp4_discussion}.

\section{Overview}\label{sec:chp4_overview}

Before talking about general-purpose unified vision systems, we revisit how language models and natural language processing (NLP) have evolved in the past years. 
Before 2018, different NLP tasks are addressed with different task-specific models, such as translation~\citep{bahdanau2014neural}, semantic parsing~\citep{berant2013semantic}, summarization~\citep{allahyari2017text}, and so on. With the emergence of the transformer architecture~\citep{vaswani2017attention}, language models for different NLP tasks are unified with a decoder-only architecture, e.g., the GPT models~\citep{brown2020language}. Afterwards, the GPT models learned using the next word prediction task are further finetuned to follow human instructions. This leads to ChatGPT~\footnote{https://chat.openai.com/}, which fundamentally changes our expectations on what AI systems can do. 
The evolution as depicted in Figure~\ref{fig:chp1_stages_nlp_cv} motivates us to wonder whether we can build a general-purpose vision system in a similar manner.

\paragraph{Challenges.}
That computer vision tasks vary greatly presents a great challenge to build a unified vision model. First, vision tasks have different types of inputs, ranging from static images~\citep{russakovsky2015imagenet} to sequential videos~\citep{miech2019howto100m}, from pure vision inputs such as image dehazing~\citep{he2010single} to multi-modality inputs that include e.g., vision and language~\cite{antol2015vqa}. Second, different granularities are required for different tasks, such as image-level tasks like image classification~\citep{he2016deep} and captioning~\citep{vinyals2016show}, region-level tasks like object detection~\citep{girshick2015fast} and grounding~\citep{plummer2015flickr30k}, and pixel-level tasks like image segmentation~\citep{he2017mask}, super-resolution~\citep{wang2020deep}, \emph{etc.} As a result, the outputs of vision systems 
are also of different formats, such as spatial information like edges, boxes, and masks, semantic information like class labels, multi-label tags, or detailed descriptions. In addition to the challenges in modeling, there are also challenges with data. 
First, the cost of annotation varies greatly among different types of labels. As shown in Figure~\ref{fig:cv_task_landscape}, these labels are at different levels of granularity and semantic richness, ranging from whole images, regions (box annotations), to masks (pixel annotations). 
Second, it is in general much more costly to collect image data than text data. So, the scale of vision data is often much smaller than that of text corpora. 

\begin{figure}
    \centering    
    \includegraphics[width=1.00\textwidth]{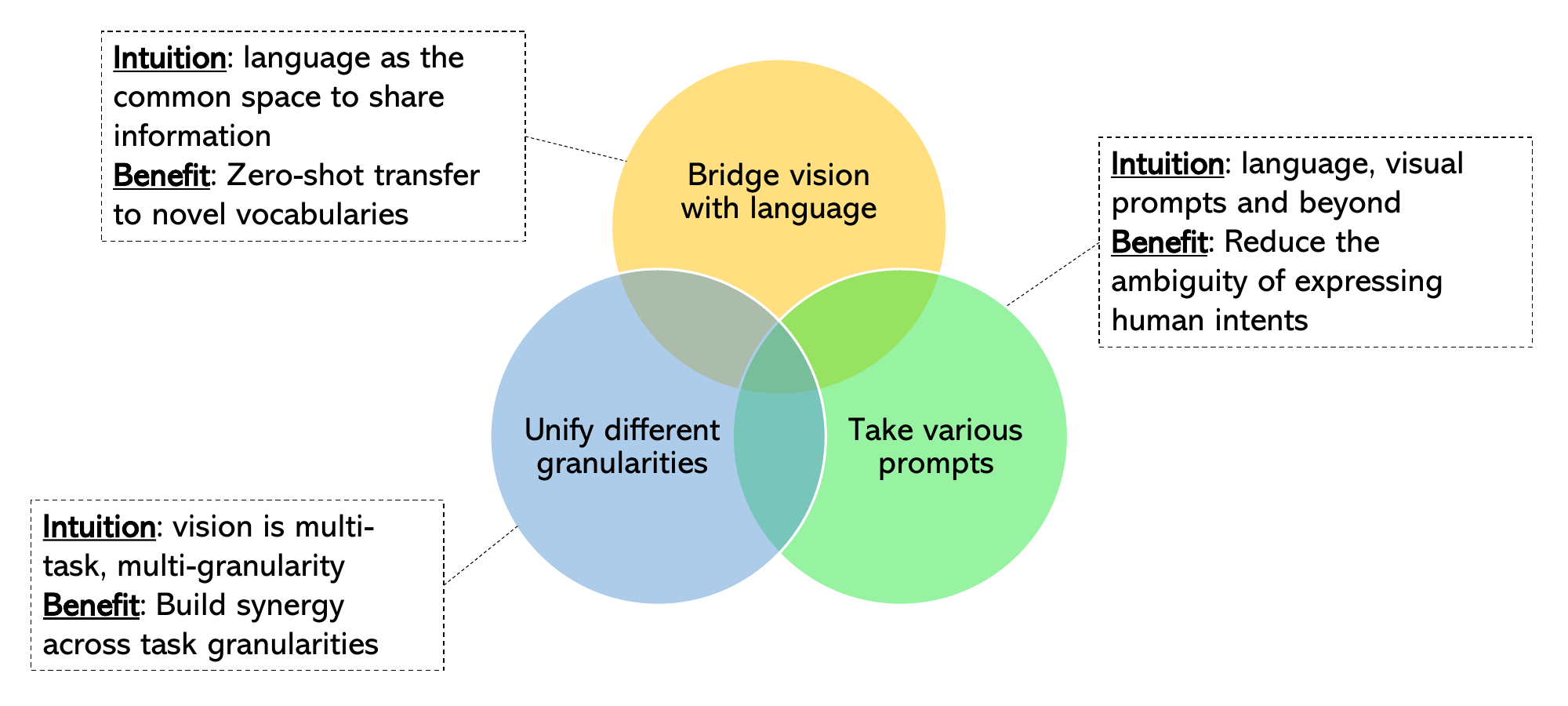}
    \caption{In NLP, we have witnessed a clear trend to build a unified model such as GPT-3~\citep{brown2020language} and then the sophisticated Human-AI interaction system ChatGPT, which has ignited the interests of the whole community and society in AI. A natural question for computer vision (CV) is whether we can unify all different types of vision tasks such as image classification, object detection, segmentation and visual question answering, \textit{etc.}, and likewise build an interaction interface between CV models and humans. Aspired by this, a lot of attempts have been made recently to crack the problem from different angles including but not limited to (a) making vision models open-set; (b) unifying different granularities; and (c) turning the models more promptable.}
    \label{fig:chp4_cv_trend}
\end{figure}

\paragraph{Towards a unified vision model.} Despite these challenges, there is a growing interest in the computer vision community to develop a general-purpose, unified vision system, in particular for visual understanding tasks. As illustrated in Figure~\ref{fig:chp4_cv_trend}, we group these efforts in three categories:

\begin{itemize}[leftmargin=*]
    \item \textbf{Bridging vision and language}. By extending closed-set classification to open-world recognition, the contrastive language-image models like CLIP~\citep{radford2021learning} demonstrate impressive zero-shot transferability for different vision tasks. These models learn the mapping between raw visual signals and rich semantics and can power various open-vocabulary vision recognition tasks~\citep{zhong2022regionclip,gu2021zero,li2022grounded,ghiasi2022scaling}. 
    \item \textbf{Unified multi-task modeling}. Traditional task-specific vision models are trained using task-specific data. 
    It is often prohibitively expensive to develop a model for a new task. Thus, it is desirable to develop a unified vision model that can perform well across many vision tasks~\citep{yang2022unitab,lu2022unified,zou2023generalized,chen2021pix2seq}.
    \item \textbf{LLM-like promptable interface}. LLMs can take different language and in-context prompts as inputs and produce user-desired outputs without finetuning. A general-purpose vision model should have possessed the same in-context learning capability to align the output to various user intents without changing its model parameters~\citep{bar2022visual,kirillov2023segment,zou2023segment,wang2023seggpt,balavzevic2023towards}. 
\end{itemize}

In what follows, we will elaborate the detailed techniques and methods in each category.

\section{From Closed-Set to Open-Set Models}\label{sec:chp4_from_close_to_open}

\begin{figure}[t]
\begin{subfigure}[t]{0.52\linewidth}
    \centering
\includegraphics[width=1.00\textwidth]{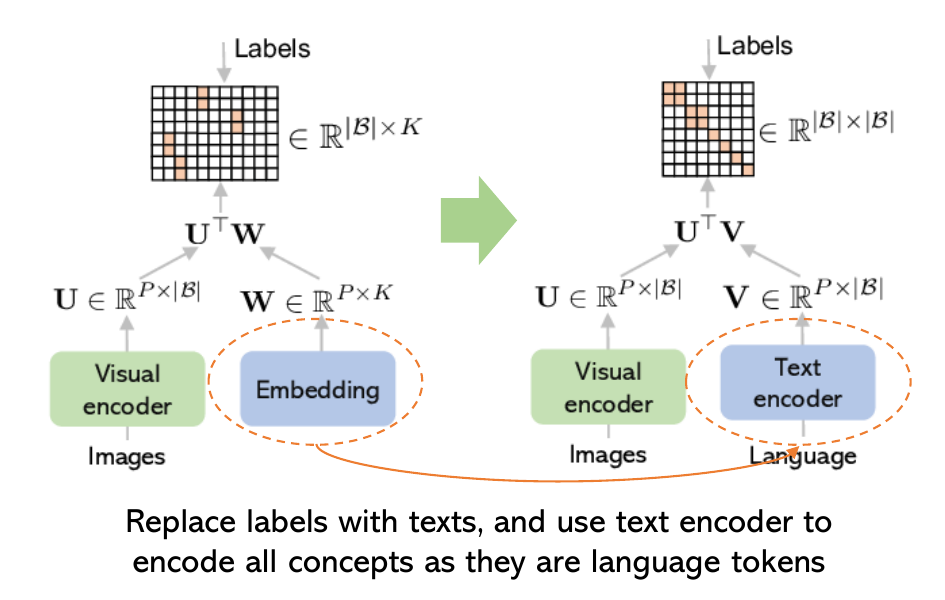} 
\caption{}
\end{subfigure}    
\begin{subfigure}[t]{0.48\linewidth}
\includegraphics[width=1.00\textwidth]{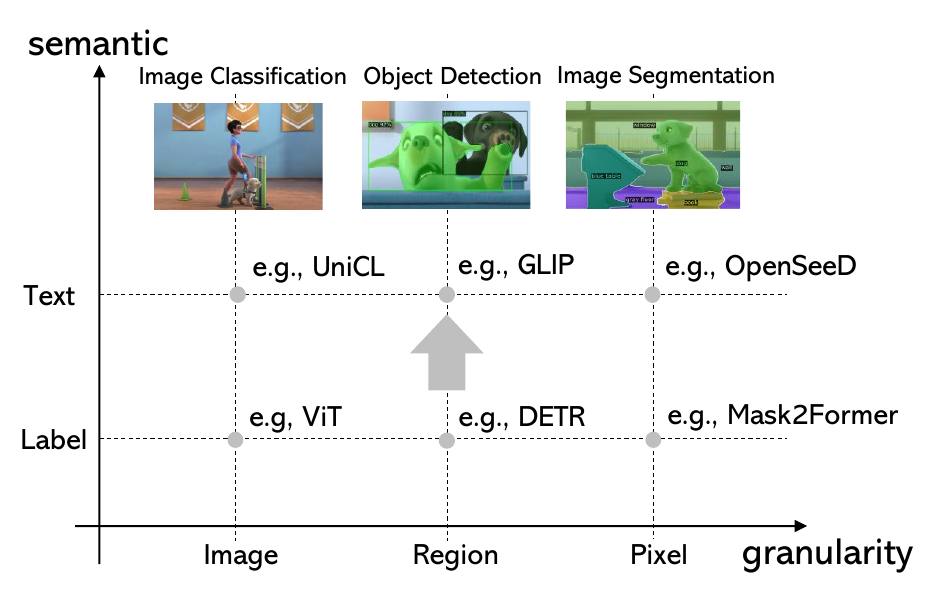}
\caption{}    
\end{subfigure}    
    \caption{(a) As proposed in CLIP~\citep{radford2021learning}, replacing labels with textual descriptions and using a text encoder to encode them can feasibly convert closed-set problems to open-set ones. Image credit from \cite{yang2022unicl}. (b) A number of works have been proposed to transform different computer vision tasks by replacing the label space with language space, such as UniCL~\citep{yang2022unicl}, GLIP~\citep{li2022grounded} and OpenSeeD~\citep{zhang2023simple}.}
    \label{fig:chp4-clip-and-others}
\end{figure}

Traditionally, visual recognition is formulated as a classification problem that maps raw visual data (e.g., images) to discrete text labels. For example, image classification predicts a label from a pre-defined close set for a whole image~\citep{deng2009imagenet}, and object detection identifies the objects, defined in a close set, within an image~\citep{lin2014microsoft}. However, such \emph{closed-set} models can hardly transfer to other tasks where the close set (or vocabulary) is insufficient. For example, it is difficult to apply an object detector trained using the Microsoft COCO object set \footnote{\url{https://cocodataset.org/}} to detect Minecraft objects. Recently, CLIP~\citep{radford2021learning} addresses the limitation of closed-set models by introducing a contrastive language-image pre-training method to train an \emph{open-set} model. As illustrated in Figure~\ref{fig:chp4-clip-and-others}~(a), instead of learning the mapping from input to labels, CLIP learns an aligned visual-semantic space using hundreds of millions of image-text pairs. Mathematically, the traditional vision tasks optimize the log-likelihood of assigning label $y=c$ to an image, often represented as a feature vector $u\in \mathcal{R}^{P}$:
\begin{equation}
    \log \mathcal{P}(y=c|u) = \log \frac{\exp^{w_c\cdot u}}{\sum_{i=1}^{K} \exp^{w_i \cdot u}} , 
    \label{chp4:eq_loglikelihood}
\end{equation}
where $w \in \mathcal{R}^{K \times P}$ is the projection matrix. 
Instead of using a pre-determined project matrix $w$, the CLIP method uses a text encoder ${Enc}_{text}$ to for the projection: 
\begin{equation}
    v_i = {Enc}_{text}(Concept_i) ,
\end{equation}
where $v$ plays the role of $w$ in Eq.~\eqref{chp4:eq_loglikelihood}. The reason why a text encoder can help achieve open-set recognition is that all textual concepts are embedded in the same feature space through large-scale pre-training, and the feature distributions are coherent to the semantic meanings without the need of a pre-defined vocabulary. As such, the aligned visual-semantic space can be easily transferred to a wide range of image recognition tasks in a zero-shot manner. Please refer to Chapter~\ref{chp:understanding} for a detailed discussion. In the following, we focus our discussion on the region-level and pixel-level models.

After the release of the CLIP model~\citep{radford2021learning}, a number of open-set vision models have been developed using large amounts of text-image pairs for visual understanding at different levels of granularity~\citep{yang2022unicl,zhang2023simple,li2022grounded,ghiasi2021open}, ranging from image-level tasks (\emph{e.g.}, image classification~\cite{deng2009imagenet}, image-text retrieval, image captioning~\cite{chen2015microsoftcoco}), region-level localization (\emph{e.g.}, object detection and phrase grounding~\cite{plummer2015flickr30k}), to pixel-level grouping tasks (\textit{e.g.}, image segmentation and referring segmentation~\cite{long2015fully,kirillov2019panoptic,hafiz2020survey}). These models can be categorized along the following three dimensions: model initialization, design and training.

\paragraph{Model initialization.} There are different initialization methods for open-set model training. 
\begin{itemize}[leftmargin=*]
    \item \textbf{CLIP initialized.} Many recent open-set models are trained by using a pre-trained model such as CLIP for initialization since a pre-trained model already provides a well-aligned (but often coarse-grained) visual-semantic feature space. For example, OVR-CNN~\citep{zareian2021open} and RegionCLIP~\citep{zhong2022regionclip} use a CLIP-style pre-trained ResNet~\citep{he2016deep} as the vision encoder and a pre-trained RPN~\citep{ren2015faster} to extract regional features. Likewise, MaskCLIP~\citep{zhou2021maskclip} and FreeSeg~\citep{qin2023freeseg} exploit the CLIP model to extract dense labels for pixels. FC-CLIP~\citep{yu2023convolutions} uses a frozen convolution network ConvNeXt~\citep{liu2022convnet} in CLIP to encode input images of various resolutions.
    \item \textbf{CLIP augmented.} Instead of initializing a model with CLIP parameters, other methods initialize the model parameters as usually (e.g., setting random values to model parameters), but use the pre-trained CLIP to help model training.
    For example, ViLD~\citep{gu2021zero} augments the model with aligned CLIP features via knowledge-distillation. MaskCLIP~\citep{ding2022open} and Mask-Adapted CLIP~\cite{liang2023open} rely on the pre-trained CLIP model to provide features and scores, respectively, during the course of model training.
    \item Other works learn a visual-semantic feature space using supervised pre-trained models or from scratch. For example, GLIP~\citep{li2022grounded} and OpenSeeD~\citep{zhang2023simple} use a pre-trained BERT model~\citep{devlin2018bert} and the CLIP text encoder, respectively, and use a vision backbone pre-trained on ImageNet for image encoding. Though these separately pre-trained image and text encoders do not explicitly learn the alignment between image and language, it turns out that these models still give good representations for images and texts, and are instrumental to efficient model training. Differently, GroupViT~\citep{xu2022groupvit} is trained jointly using an open-set semantic segmentation task and a global image-text alignment task from scratch. ODISE~\citep{xu2023open} exploits pre-trained Stable Diffusion models~\citep{rombach2022high} to extract compact masks. 
\end{itemize}

\paragraph{Model design.} Open-set models can be either multi-stage or end-to-end.

\begin{itemize}[leftmargin=*]
    \item \textbf{Two-stage models.} These models usually follow the design of the pre-DETR based models~\citep{ren2015faster,he2017mask}, which decouples localization and recognition. For object detection, a region proposal network is typically pre-trained for localizing the object of interest~\citep{zhong2022regionclip,gu2021open}, and a mask proposal network for extracting masks~\citep{ghiasi2021open,yao2022detclip}. Given the localization results, a pre-trained CLIP model is used to measure the similarity between visual contents and language concepts. A clear advantage for two-stage models is that they can inherit the open-set semantic understanding capacity without additional training so as to devote modeling training to requiring a well-performed localization network.
    \item \textbf{End-to-end models.} Different from two-stage models, the end-to-end models follow the DETR-based methods~\citep{carion2020end,cheng2022masked} or other one-stage models~\citep{dai2021dynamic}. GLIP~\citep{li2022grounded} is one of the representative works. GLIP formulates object detection as textual grounding and is trained end-to-end on image-text pairs with detection and grounding labels. Follow-up works enhance GLIP by enabling deeper vision-language interactions~\citep{liu2023grounding} or using DETR-like model design~\citep{zang2022open,minderer2022simple}. For segmentation, both ZegFormer~\citep{ding2022decoupling} and OpenSeeD~\citep{zhang2023simple} exploit a DETR-like architecture and predict the masks and categories based on the outputs of their decoders.
\end{itemize}

\paragraph{Model pre-training.} There are mainly three learning methods for pre-training open-set vision models.
\begin{itemize}[leftmargin=*]
    \item \textbf{Supervised learning.} By converting label supervision to language supervision, many works directly leverage the existing supervised annotations for training open-set models. For example, OVR-CNN~\citep{zareian2021open} trains a model with COCO categories and then evaluates its performance on novel categories. Likewise, ViLD~\citep{gu2021open} trains and evaluates two separate models on COCO and LVIS datasets, respectively. Following a similar protocol, many works train the open-set segmentation models on a subset of annotated segmentation data and evaluate the generalization ability on held-out data~\citep{ding2022decoupling,ding2022open,zhang2023simple,xu2023open}.
    \item \textbf{Semi-supervised learning.} One might use both annotated data and unlabeled or weakly-labeled data. For example, both RegionCLIP~\citep{zhong2022regionclip} and GLIP~\citep{li2022grounded} use a teacher model to extract fine-grained region-text alignments from image-text pairs to augment the training data for better open-set detection performance. Differently, OpenSeg~\citep{ghiasi2022scaling} exploits Localized Narrative datasets~\citep{pont2020connecting} as weakly-labeled data, which provides coarse correspondence between language phrases and strokes in images. Empirically, such semi-supervised learning methods often help improve models' generalization ability because they can effectively leverage rich semantics from noisy data.
    \item \textbf{Weakly-supervised learning.} Some works solely use weakly-labeled data for modeling. For example, GroupViT~\citep{xu2022groupvit} 
    uses a contrastive learning method where all supervisions for model training are from positive and negative image-text pairs. Following the same contrastive learning method, SegCLIP~\citep{luo2023segclip} uses a gathering mechanism to learn to merge image patches through the training on image-text pairs. 
\end{itemize}

Below, we review recent models developed for region-level and pixel-level tasks.

\begin{figure}[t]
\begin{subfigure}[t]{0.46\linewidth}
    \centering
\includegraphics[width=1.00\textwidth]{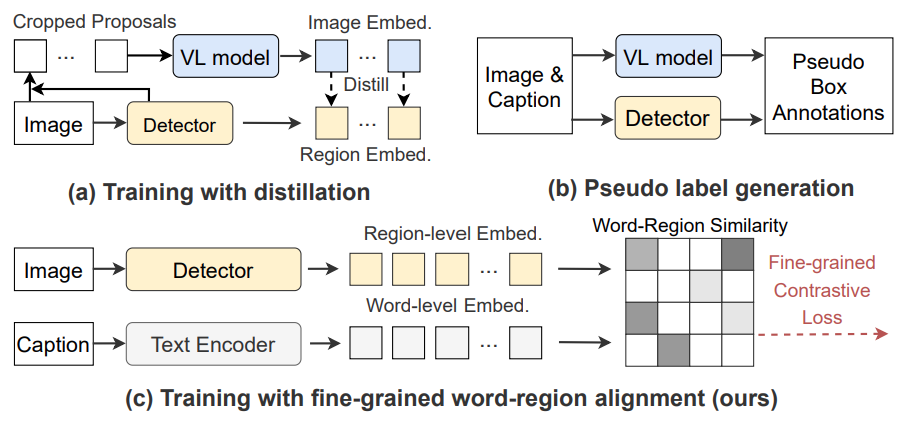} 
\caption{DetCLIPv2}
\end{subfigure}    
\begin{subfigure}[t]{0.54\linewidth}
\includegraphics[width=1.00\textwidth]{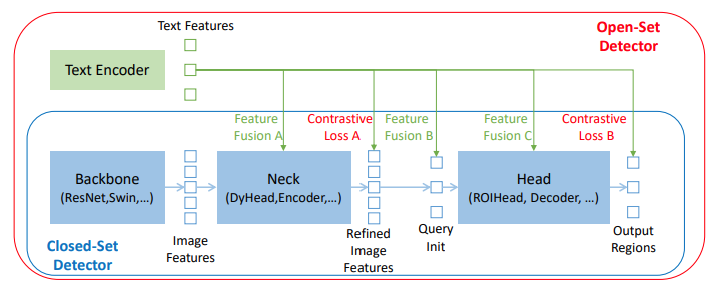}
\caption{Grounding-DINO}    
\end{subfigure}    
    \caption{(a) DetCLIPv2~\citep{yao2023detclipv2} learns fine-grained word-region alignment from object detection and grounding data and large-scale image-text pairs. (b) Grounding-DINO~\citep{liu2023grounding} injects text conditions into different stages of the Transformer encoder-decoder, which significantly improves the grounding performance. Image credit:~\cite{yao2023detclipv2} and~\cite{liu2023grounding}.}
    \label{fig:chp4-detclip-and-others}
\end{figure}

\subsection{Object Detection and Grounding}

Object detection is a fundamental task in computer vision that involves identifying and localizing objects of interest within an image or a video sequence~\citep{viola2001rapid}. Over the years, various techniques and algorithms have been developed to improve the accuracy and efficiency of object detection. In the past, region-based approaches such as R-CNN~\cite{girshick2015region}, Fast R-CNN~\citep{girshick2015fast} and Faster R-CNN~\citep{ren2015faster} have been fostering the development of advanced techniques for object detection. To improve real-time performance, YOLO~\citep{redmon2016you} proposes a single neural network that simultaneously predicts object classes and bounding box coordinates. Some improvements are made by either using multiple feature maps at different scales~\citep{liu2016ssd} or introducing a focal loss to address the class imbalance problem in dense object detection scenarios~\citep{lin2017focal}. After the emergence of Transformer~\citep{vaswani2017attention}, DETR~\citep{carion2020end} applies the transformer architecture to object detection, treating it as a set prediction problem. Since DETR, a number of methods have been proposed to improve transformer-based detection models from various aspects, such as DINO~\citep{zhang2022dino}, Group DETR~\citep{chen2022group}, and Co-DETR~\citep{zong2023detrs}. 

Open-set object detection models aim to detect arbitrary concepts beyond the vocabulary provided in training data. Three main evaluation settings have been developed in the literature:
\begin{itemize}[leftmargin=*]
    \item  \textbf{Zero-shot object detection.} Similar to zero-shot image classification~\citep{xian2018zero}, zero-shot object detection restricts the object classes used for training, and evaluates models' transferrability to novel classes. Methods falling in this category mainly focus on evaluating how a model leverages pre-trained concept embeddings (\textit{e.g.}, word2vec~\citep{mikolov2013efficient}) and learns good visual-semantic alignments~\citep{bansal2018zero,rahman2020improved,zhu2019zero,zhu2020don}. 
    
    \item \textbf{Strict open-vocabulary object detection.} First introduced in OV-RCNN~\citep{zareian2021open}, this setting differs from zero-shot object detection in that there is no limit on the training vocabulary as long as it does not cover any target classes. Under this protocol, some representative works are ViLD~\citep{gu2021open}, RegionCLIP~\citep{zhong2021regionclip} which leverage large-scale language-image models~\citep{radford2021learning,jia2021scaling}, and Detic~\citep{zhou2022detecting} that learns from image-label data.
    
    \item \textbf{Generalized open-vocabulary object detection.} Some recent works like GLIP~\citep{li2022grounded}, and OWL-VIT~\citep{minderer2022simple} advocate a more flexible setting to evaluate the dataset or task transferrability for object detection models. This setting allows vocabulary overlap between training and test sets, \textit{e.g.}, Objects365 for training while COCO for evaluation. This is arguably a more practical setting than the two settings described above in that models can be trained using any arbitrary set of training data and their detection performance evaluated in the wild~\citep{li2022elevater}.  
\end{itemize}

\begin{figure}
    \centering
\includegraphics[width=1.0\linewidth]{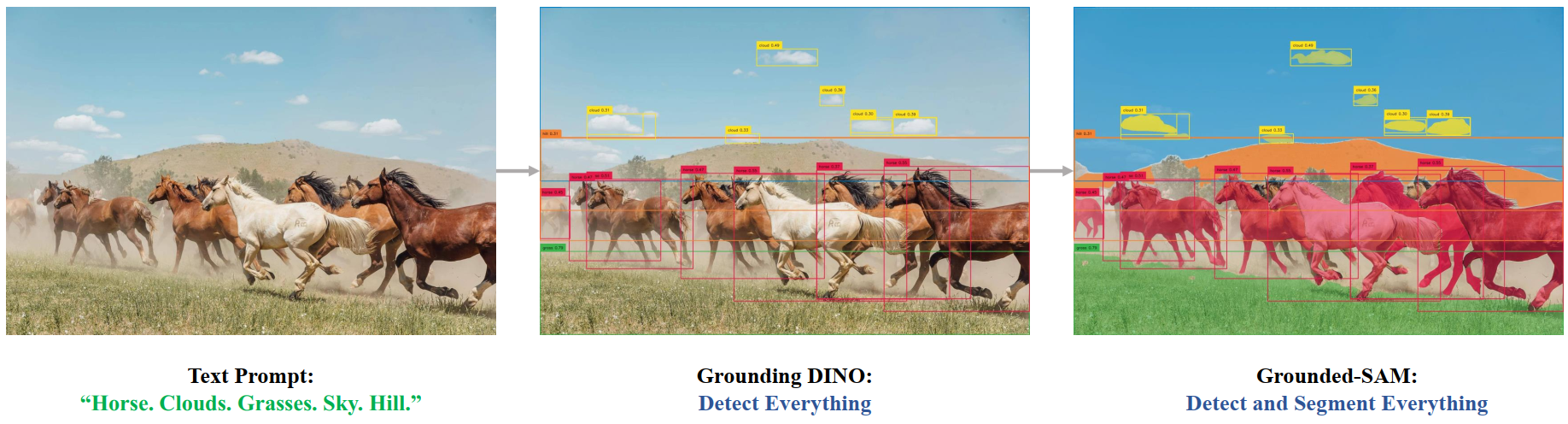}
    \caption{Grounding-SAM consisting of Grounding-DINO~\citep{liu2023grounding} and SAM~\citep{kirillov2023segment}. Image credit:~\cite{liu2023grounding}.}
    \label{fig:chp4-grounding-sam}
\end{figure}

Object grounding can be considered as a generalized open-set object detection task~\citep{plummer2015flickr30k,kazemzadeh2014referitgame,chen2019object,deng2018visual}. In this task, models take a sentence and an image as input and localize objects that are associated with the noun phrases. Recently, M-DETR~\citep{kamath2021mdetr} employs a transformer-based architecture to build an end-to-end modulated detector to detect objects in an image given a raw text query.
Unlike previous works where models are trained on specific datasets, the network is pre-trained with 1.3M pairs of text and images, sourced from multi-modal datasets where the connections between text phrases and corresponding image objects are labeled. Inspired by M-DETR, GLIP~\citep{li2022grounded} casts object detection as a grounding problem, and jointly learns a model using object detection and grounding data for open-set scenarios. 
Following this line of research, DetCLIPv2~\citep{yao2023detclipv2} proposes a simple joint learning method where multiple tasks are converted into a word-region alignment task, and then a model is trained end-to-end on a corpus consisting of object detection data, grounding data and image-text pairs. Grounding-DINO~\citep{liu2023grounding} is a state-of-the-art grounded object detection method, where the object detector is composed of components: a backbone, a neck, and a head, and inject language conditions at every stage. A combined text and image backbone is employed to extract features at multiple scales, which are then passed on to the neck. The text and image characteristics generated by the neck are subsequently used for language-driven query selection. Grounding-SAM is developed by combining Grounding-DINO with SAM~\citep{kirillov2023segment}. As shown in Figure~\ref{fig:chp4-grounding-sam}, an image and a group of concepts are first fed into Grounding-DINO to produce the boxes, and then the boxes are used as prompts for SAM to predict masks for each box.

\subsection{Image Segmentation and Referring}

Image segmentation is a long-standing and challenging vision problem. There are mainly three subtasks, including semantic~\citep{long2015fully}, instance~\citep{hafiz2020survey}, and panoptic~\citep{kirillov2019panoptic} segmentation. Semantic segmentation cares about the per-pixel semantic within an image~\citep{long2015fully, chen2017rethinking, chen2022vision}, whereas instance segmentation groups pixels of the same semantic meaning into objects. Models for both tasks have evolved from CNN-based architectures~\citep{long2015fully} to transformer-based ones~\citep{chen2022vision}, and from two-stage models~\citep{he2017mask} and one-stage models~\citep{bolya2019yolact,tian2020conditional} to the recent query-based approaches~\citep{dong2021solq,zou2022end}. With the capability of per-pixel and instance-level understanding, a natural step was taken to formulate panoptic segmentation~\citep{kirillov2019panoptic,wang2021max,cheng2022masked}. Most recently, Mask2Former~\citep{cheng2022masked} proposed to address all three tasks with a unified encoder-decoder architecture. Nevertheless, all these works cope with a limited number of categories. In the following, we will review the most recent works on open-set image segmentation and referring segmentation.

\begin{figure}
    \centering
\includegraphics[width=1.0\linewidth]{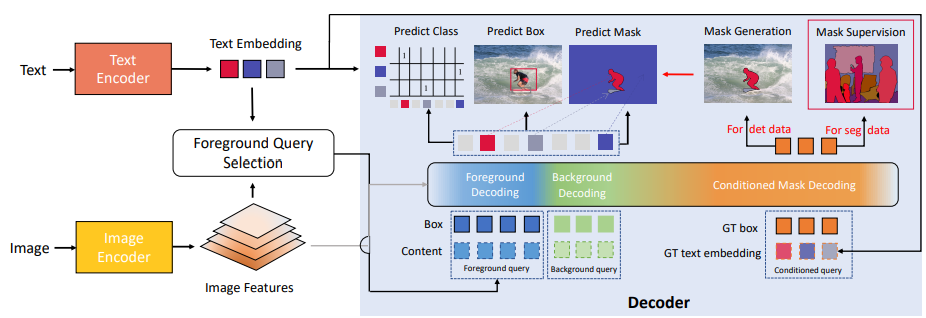}
    \caption{OpenSeeD~\citep{zhang2023simple} leverages both mask and box supervision for learning a universal open-vocabulary image segmentation model. Image credit:~\cite{zhang2023simple}.}
    \vspace{-5pt}
    \label{fig:chp4-openseed}
\end{figure}

\paragraph{Open-Vocabulary Segmentation.} Recently, a number of methods have been proposed to transfer or distill the rich visual-semantic knowledge from foundation models~\citep{radford2021learning,jia2021scaling} to specific segmentation tasks. Prominent examples include LSeg~\citep{li2022language}, OpenSeg~\citep{ghiasi2021open}, and \cite{huynh2022open}. Instead of using existing models, GroupViT~\citet{xu2022groupvit} performs language-image pre-training from scratch with a bottom-up grouping ViT~\citep{dosovitskiy2020image}, while DenseCLIP~\citep{rao2022denseclip} demonstrates the superiority of foundation models in finetuning settings compared with supervised models. Recently, MaskCLIP~\citep{ding2022open} is proposed to tackle open-vocabulary panoptic and semantic segmentation simultaneously by leveraging CLIP, and achieves impressive performance on ADE20K~\citep{zhou2017scene} and PASCAL~\citep{mottaghi2014role,everingham2011pascal}. Instead of using the ViT backbone, a recent work called FC-CLIP~\citep{yu2023convolutions} exploits a convolutional CLIP backbone (\textit{i.e.}, ConvNeXt trained by OpenCLIP~\citep{ilharco_gabriel_2021_5143773}) as both a feature extractor and a vision encoder. Based on a simplified pipeline, FC-CLIP shows plausible efficiency and lefts the state of the art on various open-vocabulary segmentation benchmarks. Rather than only using CLIP, a recent work ODISE~\citep{xu2023open} leverages text-to-image diffusion models, and shows that the latent features in the pre-trained UNet can provide useful compact segmentation information for open-vocabulary segmentation.

A big challenge in open-vocabulary segmentation is the lack of segmentation data annotated with semantic labels. Thus far, most of the works are still using COCO segmentation annotations. A few recent works attempt to leverage object detection data as the extra supervision to augment the training of segmentation models, such as OpenSeeD~\citep{zhang2023simple} (shown in Figure~\ref{fig:chp4-openseed}) and DataSeg~\citep{gu2023dataseg}. In addition to these new modeling techniques, new datasets have been developed to mitigate this problem, including curating multi-domain segmentation datasets~\citep{lambert2020mseg}, collecting high-quality annotations~\citep{qilu2023high} or scaling up to billions of masks~\citep{kirillov2023segment}.

\textbf{Referring Segmentation} by design is open-vocabulary. Models are usually designed specifically to learn from target datasets using various multimodal fusion strategies~\citep{hu2016segmentation,liu2017recurrent,margffoy2018dynamic,ye2019cross,yu2016modeling,wu2022language}. CLIPSeg~\citep{luddecke2022image} extends a textual query to a visual query and shows superior performance not only on referring segmentation but also on semantic segmentation. Since the emergence of vision transformers, works like LAVT~\citep{yang2022lavt} enhance the cross-modal interactions from the very beginning, which leads to a decent performance on RefCOCO~\citep{yu2016modeling}, RefCOCO+~\citep{yu2016modeling} and G-Ref~\citep{mao2016generation,nagaraja2016modeling}. Differently, PolyFormer~\citep{liu2023polyformer} converts masks into polygons and asks the transformer decoder to decode a sequence of polygon coordinates. Inspired by Pix2Seq~\citep{chen2021pix2seq}, a similar method in object detection, PolyFormer presents an alternative way to represent masks for state-of-the-art referring segmentation. As we discussed earlier, one can also compose Grounding DINO~\citep{liu2023grounding} with SAM~\citep{kirillov2023segment} for referring segmentation.

\paragraph{Unified Segmentation.} Given the above methods for open-vocabulary and referring segmentation, an open question is how to unify all segmentation tasks in a single framework. Recently, X-Decoder~\citep{zou2023generalized} uses a generalized encoder-decoder architecture to unify all these segmentation tasks. The referring segmentation task is reformulated as a conditioned panoptic segmentation that takes some textual phrases as input to the decoder. UNINEXT~\citep{yan2023universal} is another work that attempts to unify all instance-level segmentation in images and videos. Different from X-Decoder, UNINEXT uses early fusion to fuse the various prompts and vision features, which are then fed to the transformer encoder-decoder.



\section{From Task-Specific Models to Generic Models}\label{sec:chp4_from_specific_to_generic}

\begin{figure}
\begin{subfigure}[t]{0.52\linewidth}
    \centering
\includegraphics[width=1.00\textwidth]{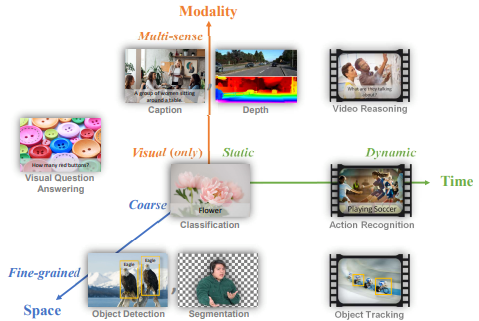} 
\caption{}
\end{subfigure}    
\begin{subfigure}[t]{0.48\linewidth}
\includegraphics[width=1.00\textwidth]{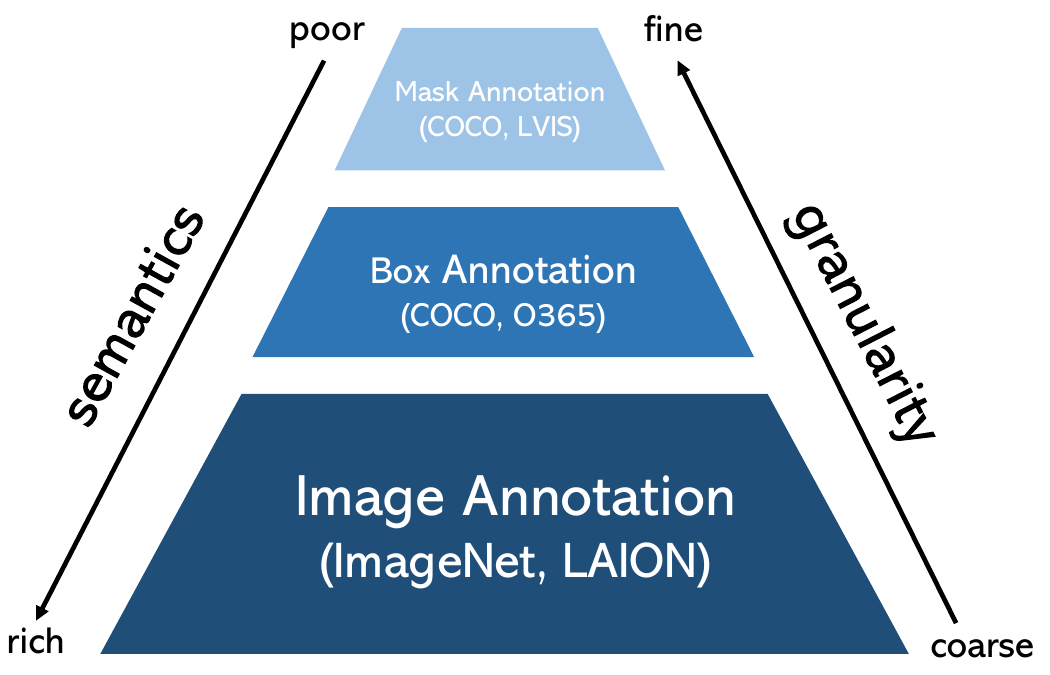}
\caption{}    
\end{subfigure}    
\caption{(a) CV task landscape: CV tasks can span different axes, including modality, space and time, which renders significant challenges to unify all of them in a single model. Image credit: \cite{yuan2021florence}. (b) The data scale pyramid: In particular, datasets in different tasks usually contain different types of supervision. Image-level datasets like ImageNet~\citep{deng2009imagenet} and LAION~\cite{schuhmann2021laion} have annotations that have rich semantics coverage but are coarse-grained, while pixel-level datasets like COCO panoptic segmentation~\citep{chen2015microsoftcoco} provides fine-grained annotations but with limited concepts.}
\label{fig:cv_task_landscape}
\end{figure}

Above we have discussed the recent efforts of transforming closed-set models to open-set ones for detection and segmentation. Until recently, however, most vision tasks have been separately tackled with specialized model designs, preventing the synergy of tasks across different granularities or domains from being exploited. This is arguably due to two reasons: 
\begin{itemize}[leftmargin=*]
    \item \textbf{Vision tasks are fragmented}. As shown in Figure~\ref{fig:cv_task_landscape}~(a), computer vision tasks span across different axes including space, time, and modality. From the space aspect, it can be image-level, region-level and pixel-level tasks as we discussed before. Along the time axis, we need to tackle not only static images but also temporal video sequences. Regarding the modality, the inputs and outputs can be images, texts, or other types (\textit{e.g.}, human pose, depth map). Such diverse task formats significantly impede the development of a unified model for all tasks. 
    \item \textbf{Data scales are different}. In addition to the complicated task landscape, the scarcity of human annotations and their different scales for different tasks also make building a unified model challenging. In Figure~\ref{fig:cv_task_landscape}~(b), we can see a clear pyramid of data scale, where different layers of human annotations have different semantics. More specifically, image-text datasets like LAION~\cite{schuhmann2021laion} contain up to 2B samples, while object detection datasets like Objects365~\citep{shao2019objects365} have 1.7M images in total. More significant gap is observed in segmentation datasets due to the high cost of annotating masks.
\end{itemize}

Despite the aforementioned challenges, 
we are now witnessing a growing interest in building unified, general-purpose models that can learn from and be applied to a diverse set of vision and vision-language tasks, thanks to the versatility of transformers~\citep{vaswani2017attention}. These attempts can be grouped into two main categories: 

\begin{itemize}[leftmargin=*]
    \item \textbf{I/O Unification.} Following the development of unified LLMs, a number of recent works reformulate many vision tasks as a sequence-to-sequence problem~\citep{wang2022ofa,yang2022unitab,chen2022unified, lu2022unified}. They typically use a tokenizer to tokenize the original inputs and outputs (I/O) in different modalities used in various tasks into a coherent sequence (visual or text) tokens and then exploit a unified, sequence-to-sequence model. 
    \item \textbf{Functionality Unification.} In addition to I/O unification, one might built a generic model via functionality unification. Extending multi-task learning methods~\citep{lu202012,gupta2022towards,hu2021unit}, many recent use a coherent encoder-decoder architectures~\citep{yu2022coca, zhang2022glipv2, zou2023generalized}. This line of work usually does not need task-specific or modality-specific tokenizers but requires a sophisticated model design to accommodate various tasks.
\end{itemize}

\begin{figure}
    \centering
\includegraphics[width=1.0\linewidth]{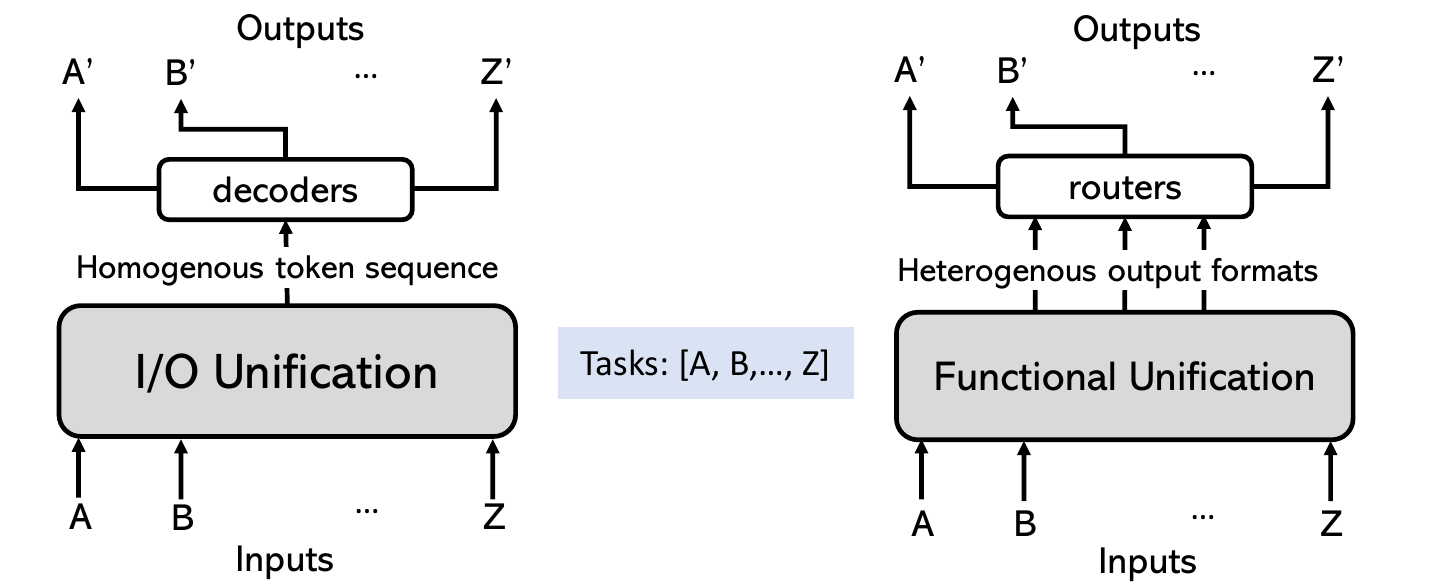}
    \caption{Side-by-side comparison between I/O unification and functionality unification. I/O unification is aimed at utilizing a decoder to decode homogeneous token sequences, which are then decoded by task-specific decoders. In contrast, functionality unification predicts heterogeneous outputs and then uses different routers or headers to produce the final outputs for various tasks.}
    \label{fig:unification}
\end{figure}

Figure~\ref{fig:unification} illustrates the difference between the two categories of unification methods. For I/O unification, the I/O unification module always generates a sequence of tokens, and exploits a separate decoder to decode the final outputs for different tasks. For functionality unification, the functional unification module generates 
heterogeneous outputs for different task, \emph{e.g.}, semantic outputs and spatial outputs. Then, these different types of outputs are combined to produce the final task-specific outputs. 
Both unification methods strive to make use of synergy across tasks with different levels of granularity. For example, coarse-grained data is expected to contribute to rich semantic understanding required by fine-grained tasks, while fine-trained data to enhance the grounding ability for coarse-grained tasks. In the following, we review some recent works of these two categories.

\subsection{I/O Unification}

This line of work is mainly inspired by LLMs that unify many NLP tasks as sequential modeling. In the vision domain. the methods of building generic models via I/O unification can be grouped into two categories depending on the tasks of interest and output formats.

\subsubsection{Sparse and discrete outputs} 
For vision tasks that produce sparse or discrete token outputs, we can easily exploit a language tokenizer, such as byte-pair encoding (BPE)~\citep{sennrich2016neural}, for I/O unification. 
In contrast, spatial outputs like boxes, masks, or human skeletons can be formulated as a sequence of numeric coordinates which are then tokenized into discrete tokens~\citep{cho2021unifying,yang2022unitab,liu2023polyformer}. As a result, the decoded output tokens are interleaved with organic textual tokens and numeric textual tokens to support a wide range of tasks. Without the loss of generality, the decoding process is formulated as auto-regressive generation and the model trained with the objective function defined as:
\begin{equation}
    L(\theta) = -\sum_{t=1}^{T} \log p(s_t | s_{<t}, v; \theta),
\end{equation}
where $\{s\}_{t=1}^{T}$ is the discrete token sequence of length $T$, and $v$ is the visual feature. Below, we review some representative works.

\begin{figure}[t]
\begin{subfigure}[t]{.56\linewidth}  
    \centering
    \includegraphics[width=1.0\textwidth]{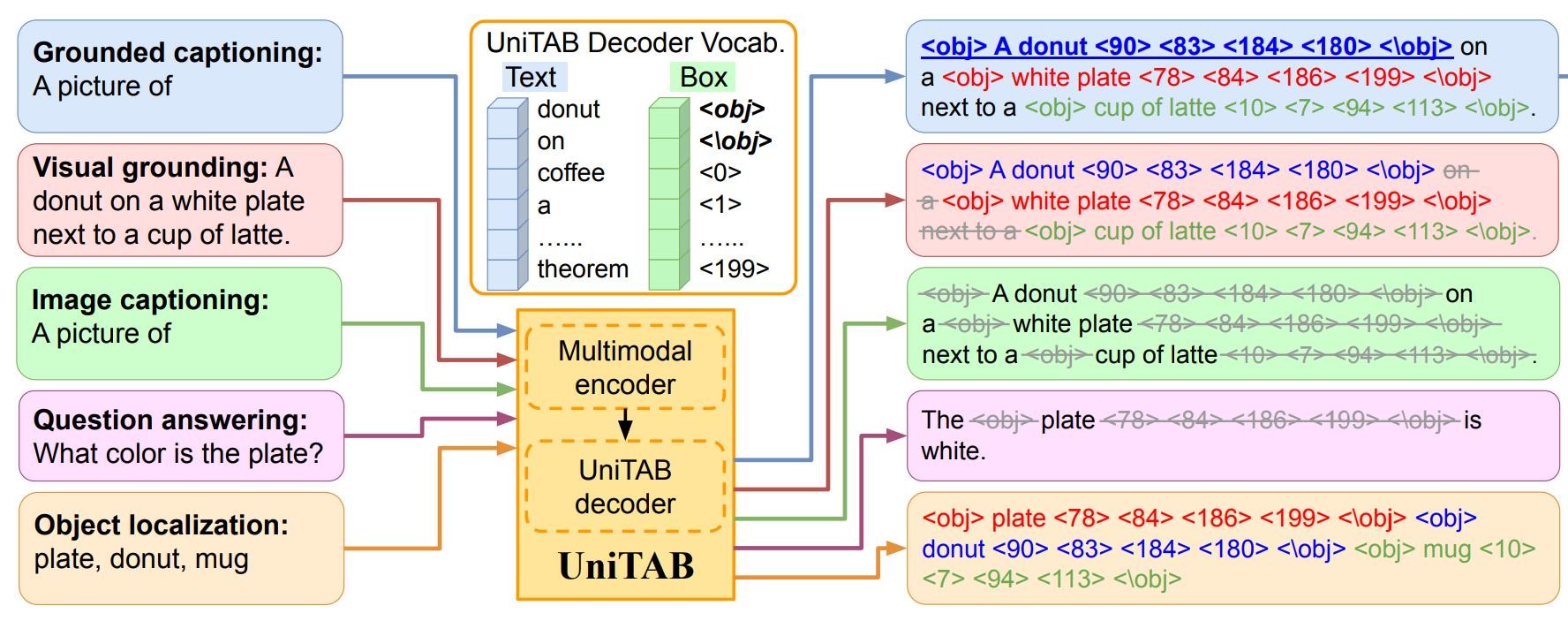}
    \caption{UniTab}    
\end{subfigure}
\begin{subfigure}[t]{.46\linewidth}  
    \centering
    \includegraphics[width=.95\textwidth]{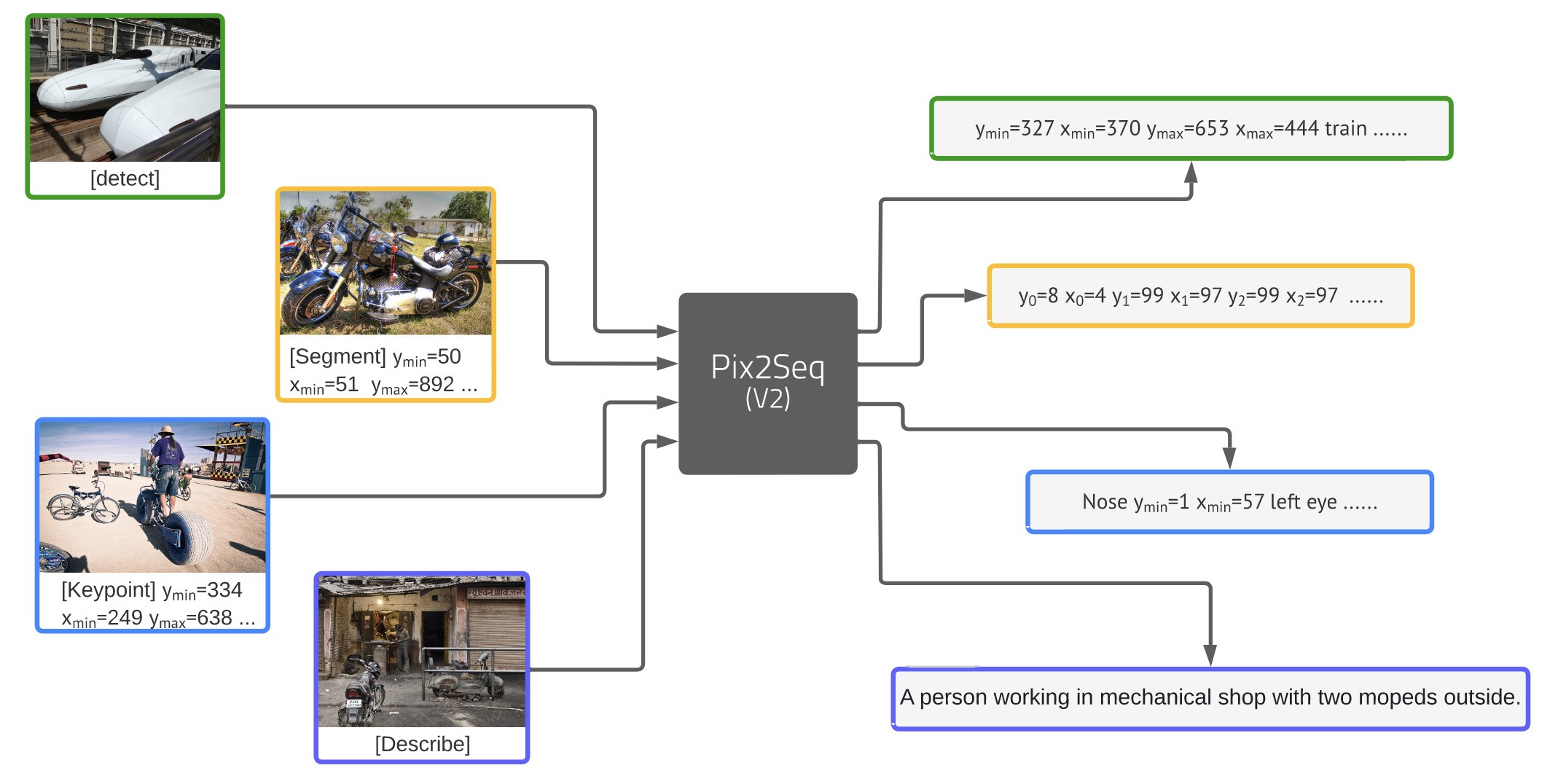}
    \caption{Pix2Seqv2}        
\end{subfigure}
    \caption{(a) UniTab~\citep{yang2022unitab} is proposed to unify grounded captioning, visual grounding, image captioning, VQA, and object localization. (b) Pix2Seqv2~\citep{chen2022unified} is proposed to unify object detection, referring segmentation, keypoint detection, and image captioning. Image credit:~\cite{yang2022unitab} and~\cite{chen2022unified}.}
    \label{fig:chp4-unitab-pix2seqv2}
\end{figure}

UniTab~\citep{yang2022unitab} unifies text and box output in a sequence decoding manner. As shown in Figure~\ref{fig:chp4-unitab-pix2seqv2}~(a), the box coordinates are represented by numerical numbers with $<>$ and then a special token $<$obj$>$ is used to encompass the location information. In this way, the model can unify a variety of tasks that require textual and location outputs, including image captioning~\citep{chen2015microsoftcoco}, grounded captioning~\citep{plummer2015flickr30k}, visual grounding, object localization and visual question answering~\citep{antol2015vqa}. The model is trained in three stages: pre-training, multi-task finetuning, and task-specific finetuning. 

Pix2SeqV2~\citep{chen2022unified} slightly differs from UniTab in that it unifies two different vision tasks: referring segmentation and keypoint detection. Following Pix2Seq~\citep{chen2021pix2seq}, Pix2SeqV2 represents objects in an image as $[y_{min}, x_{min}, y_{max}, x_{max}, \textit{text}]$. Then, it introduces a unique task prompt for each task, which contains task type information or a combination of task types and specific locations. For mask decoding, a mask contour is converted into a polygon and then its coordinates extracted from the polygon~\citep{castrejon2017annotating}. A similar strategy is also used for referring segmentation, as in Polyformer~\citep{liu2023polyformer}.


\begin{figure}
    \centering
    \includegraphics[width=1.0\linewidth]{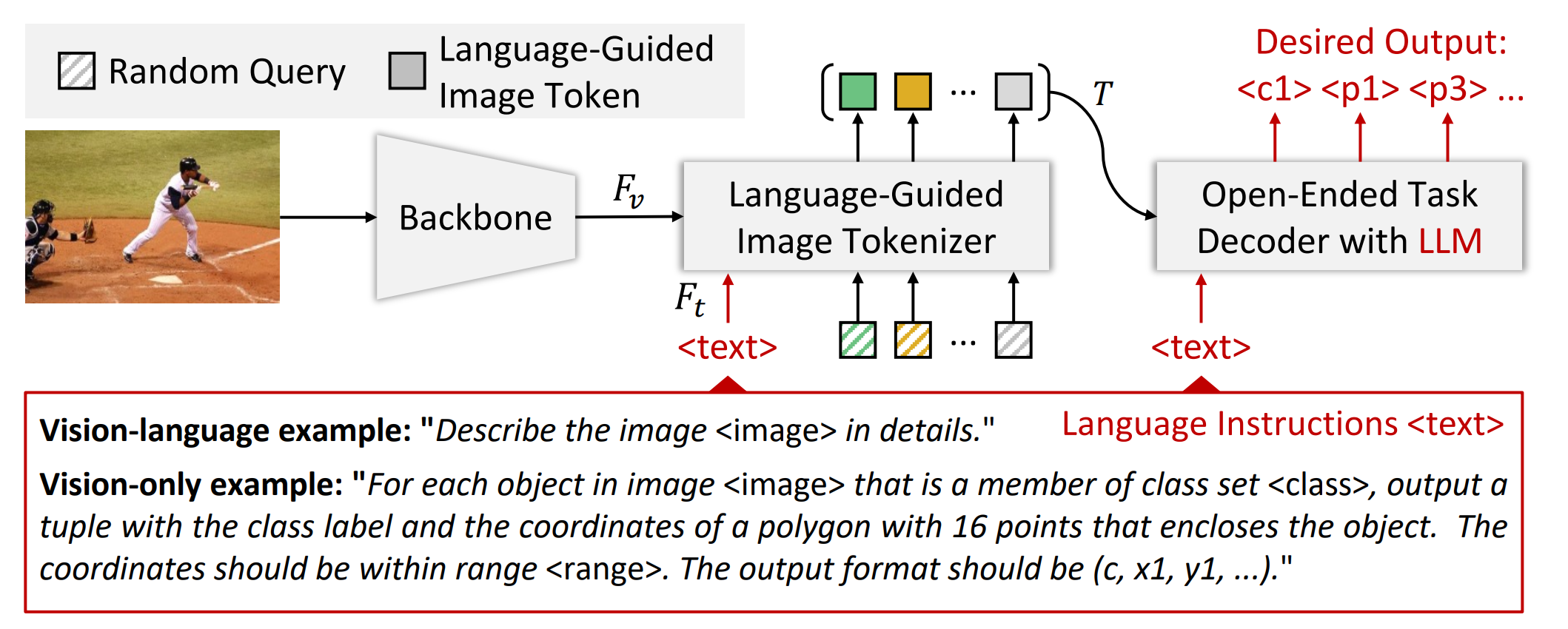}
    \caption{VisionLLM~\citep{wang2023visionllm} is proposed to bridge vision systems with LLMs in a sequential decoding manner. Image credit:~\cite{wang2023visionllm}.}
    \label{fig:chp4-visionllm}
\end{figure}

\paragraph{LLM-augmented.} Recent works have also explored building a generic decoding interface based on LLMs, which are pre-trained on large amounts of text data and human instructions. Kosmos-2~\citep{peng2023kosmos} exploits the pretrained LLMs of Kosmos-1~\citep{huang2023language} and augments the grounded multi-modal data by collecting a web-scale grounded image-text pair dataset (GRIT) consisting of 91M images. VisionLLM~\citep{wang2023visionllm} appends an even larger LLM (\textit{e.g.}, LLaMa~\citep{touvron2023llama}) on top of an image tokenizer, as shown in Figure~\ref{fig:chp4-visionllm}. The resultant model exhibits a very strong vision-language reasoning capacity and decent localization ability for object detection, segmentation, \textit{etc}. Some other works that combine LLMs with grounding are DetGPT~\citep{pi2023detgpt} and GPT4ROI~\citep{zhang2023gpt4roi}. To further equip the model with the segmentation capability, both BubaGPT~\citep{zhao2023bubogpt} and LISA~\citep{lai2023lisa} use an extra referring segmentation model to segment images by taking texts or embeddings as input, respectively. PaLI-X~\citep{chen2023pali} is by far the largest unified model that can cope with multilingual vision and vision-language tasks.

\subsubsection{Dense and continuous outputs}

\begin{figure}[t]
\centering
\begin{subfigure}[t]{0.85\linewidth}  
    \includegraphics[width=1.0\textwidth]{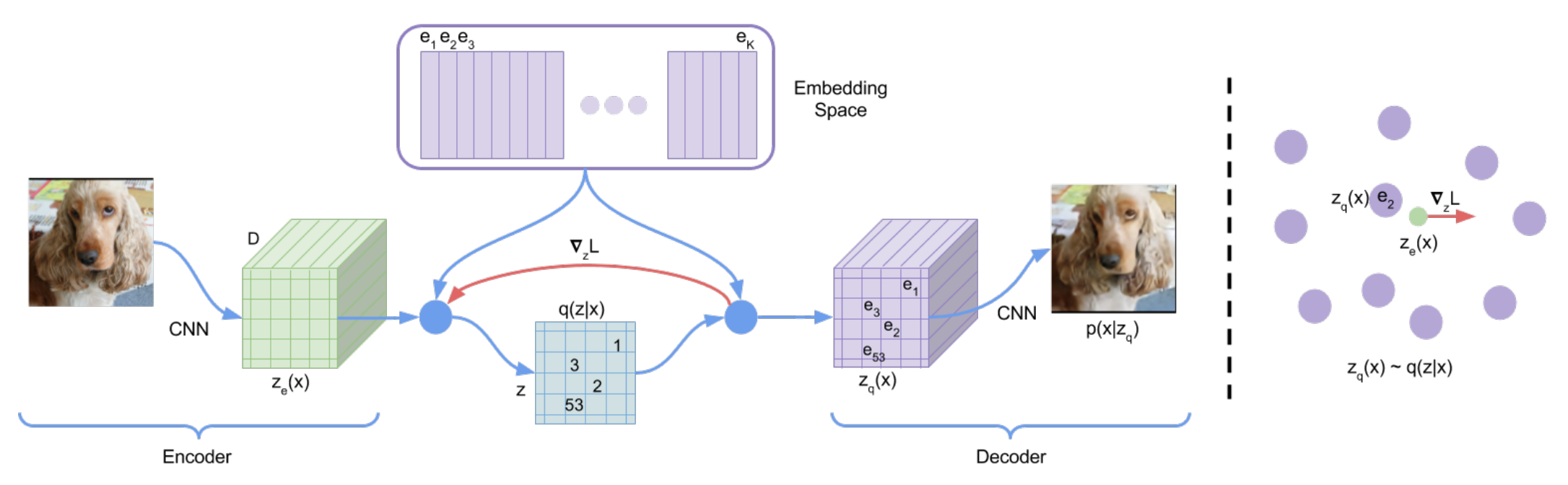}    
    \caption{VQ-VAE}    
\end{subfigure}
\begin{subfigure}[t]{0.8\linewidth}  
    \includegraphics[width=1.0\textwidth]{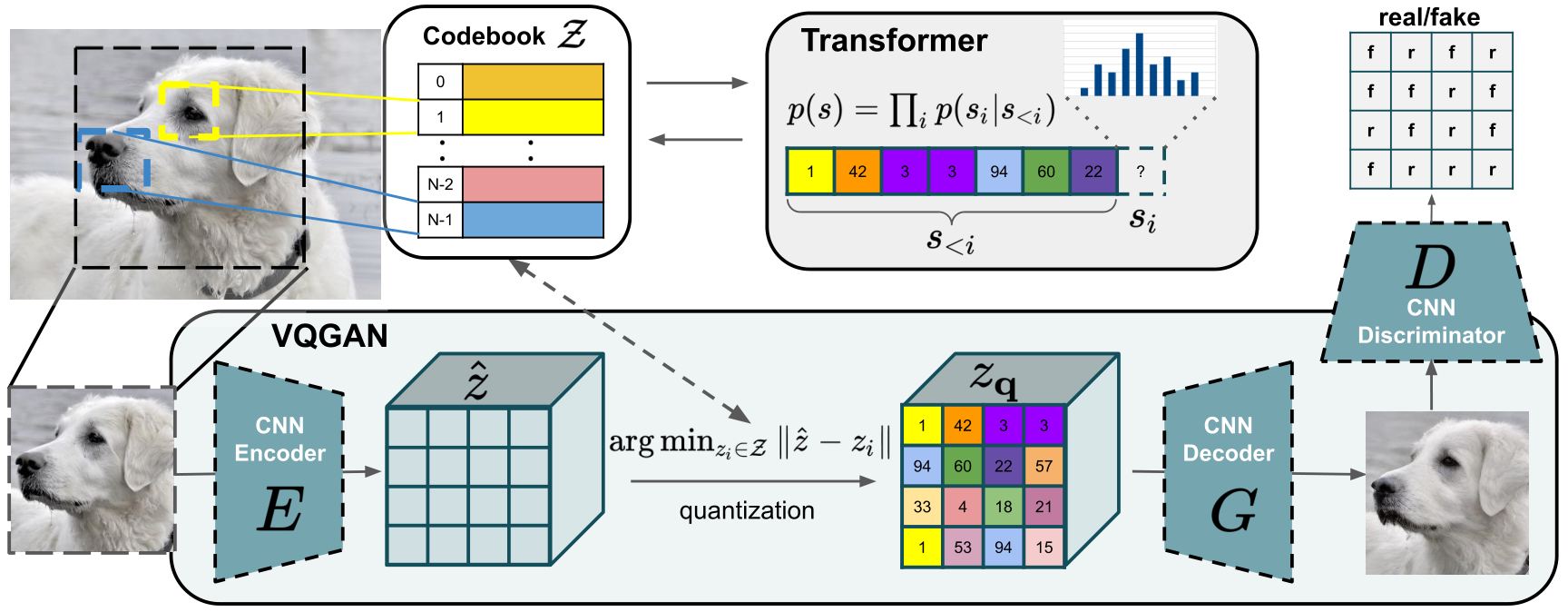}  
    \caption{VQ-GAN}    
\end{subfigure}
\caption{Illustration of VQ-VAE~\citep{oord2017neural} and VQ-GAN~\citep{esser2021taming}.}
\label{fig:chp4-vqvaegan}
\end{figure}

There are also some tasks that require dense and continuous outputs, such as image segmentation~\citep{he2017mask}, depth estimation~\citep{mertan2022single}, image inpainting and editing~\citep{elharrouss2020image,brooks2023instructpix2pix}. Except for segmentation masks which can be approximated by polygons~\citep{liu2023polyformer,chen2022unified}, most dense and continuous outputs cannot be easily converted into discrete tokens due to the high-dimensional space. Thus, we have to resort to an image-oriented tokenizer. Akin to the language tokenizer, an image tokenizer encodes raw images and extracts discrete tokens spanning the visual feature space. The most representative work is VQ-VAE~\citep{oord2017neural,razavi2019generating}. As shown in Figure~\ref{fig:chp4-vqvaegan}~(a), VQ-VAE learns an encoder $z_e$, a decoder $z_q$ and a discrete codebook $\textbf{e}=\{e_1,...,e_K\}$ consisting of $K$ embeddings. Given the input $x$, the posterior categorical probability $q(z|x)$ is defined as:
\begin{equation}
  q(z=k|x)=\begin{cases}
    1, & \text{for $k=\argmin_i \| z_e(x) - e_i \|$}.\\
    0, & \text{otherwise}.
  \end{cases}    
\end{equation}
where the decoder $z_q$ takes $x$ (or its representation $e_k$) as input to predict class label. As a variant of VQ-VAE, VQ-GAN uses a discriminator and the perceptual loss~\citep{larsen2016autoencoding,lamb2016discriminative} to maintain a good balance between output quality and model efficiency (via high compression rate). In Figure~\ref{fig:chp4-vqvaegan} (b), we see that the discriminator is applied at the patch level to regularize the decoding of images at high resolution. Below, we discuss some most recent works that attempt to unify different vision and multi-modal tasks that involve dense outputs.

UViM~\citep{kolesnikov2022uvim} is one of the first works that employ a dense decoding process to unify various core vision tasks, including panoptic segmentation, depth estimation and colorization. The learning process consists of two 
stages: $(i)$ 
Base encoder-decoder $f$ and restricted oracle $\Omega$ are learned to predict outputs given input images, where $f$ takes raw image as input and $\Omega$ takes the desired output as input to decode the oracle code; $(ii)$ Instead of using the desired output as input to the oracle $\Omega$, the model learns a language model to produce the oracle code for the input raw image. Notably, the encoder-decoder model used here is trained with VQ-VAE objectives. As the first step to unify vision tasks with a single model, UViM shows promising results on three vision tasks.

\begin{figure}
    \centering
    \includegraphics[width=1.0\linewidth]{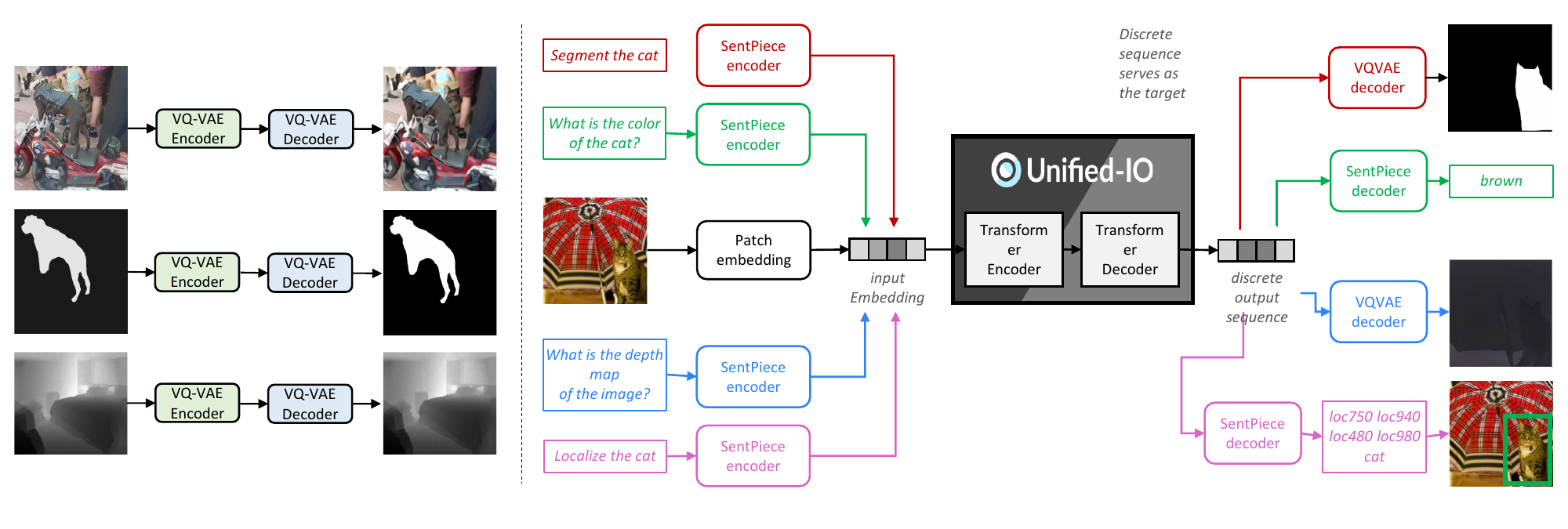}
    \caption{Unified-IO~\citep{lu2022unified} unifies different vision tasks by first pre-training VQ-VAE for each task and then an encoder-decoder for tasks jointly. Image credit:~\cite{lu2022unified}.}
    \label{fig:chp4-unified-io}
\end{figure}

Unified-IO~\citep{lu2022unified} is another representative work. Compared to UVIM, it scales to many more vision tasks and datasets. Unlike the training procedure of UViM, Unified-IO first trains different VQ-VAE models for different tasks, as depicted in Figure~\ref{fig:chp4-unified-io} left. After obtaining all VQ-VAE encoder-decoders, 90 datasets are combined to train another transformer encoder-decoder end-to-end, as shown on the right side. Similar to previous works, it also uses a language decoder to obtain the organic and numeric texts to generate coordinate outputs. After the second-stage pre-training, the model achieves state of the art on the GRIT benchmark~\citep{gupta2022grit} and exhibits compelling compositionality, although the performance still lags behind the strongest models on common tasks. As a follow-up, a soft-token strategy is proposed in~\cite{ning2023all} to improve the accuracy for next token decoding. In addition, a masked modeling strategy is proposed to learn robust representations. Evaluated on instance segmentation and depth estimation, the model achieves state-of-the-art performance on NYUv2~\citep{silberman2012indoor} and competitive performance on segmentation. A recent work uses image inpainting as the general task to unify different pixel-level vision tasks~\citep{bar2022visual}. Given the target discrete tokens produced by VQ-GAN, the method exploits a masked autoencoder to decode the missed image regions, using the task input-output examples as prompts. Painter~\citep{wang2023images} extends this pipeline to facilitate more vision tasks and obtains competitive performance on various standard benchmarks.

\paragraph{Diffusion-augmented.} Unlike the above works that learn their own decoding models, some recent works utilize the off-the-shelf stable diffusion model to build generalist vision models. For example, Prompt Diffusion~\citep{wang2023incontext} initializes a model using Stable Diffusion and ControlNet~\citep{zhang2023adding}, and trains the in-context image-to-image model jointly on six different vision-language tasks, including segmentation, depth estimation, \textit{etc.} InstructDiffusion~\cite{geng2023instructdiffusion} also uses the diffusion model but explicitly introduces task-specific instructions to the diffusion process. Moreover, it uses task-specific training and human alignment training to enable a generalist interface for vision tasks.

\subsection{Functionality Unification}

Unlike I/O unification, functionality unification attempts to unify different tasks based on the task characteristics, with the awareness that they are neither fully isolated nor fully aligned. At a high level, vision tasks produce three types of outputs: $(i)$ location outputs, $(ii)$ semantic outputs, and $(iii)$ pixel-level outputs. For example, both object detection and phrase grounding need to localize objects in the image, while both generic segmentation and referring segmentation produce masks. On the other hand, many tasks require semantic (or text) outputs to represent either concept names or textual descriptions.

\subsubsection{Multi-task learning}

\begin{figure}
\begin{subfigure}[t]{0.53\linewidth}
    \centering
    \includegraphics[width=1.0\textwidth]{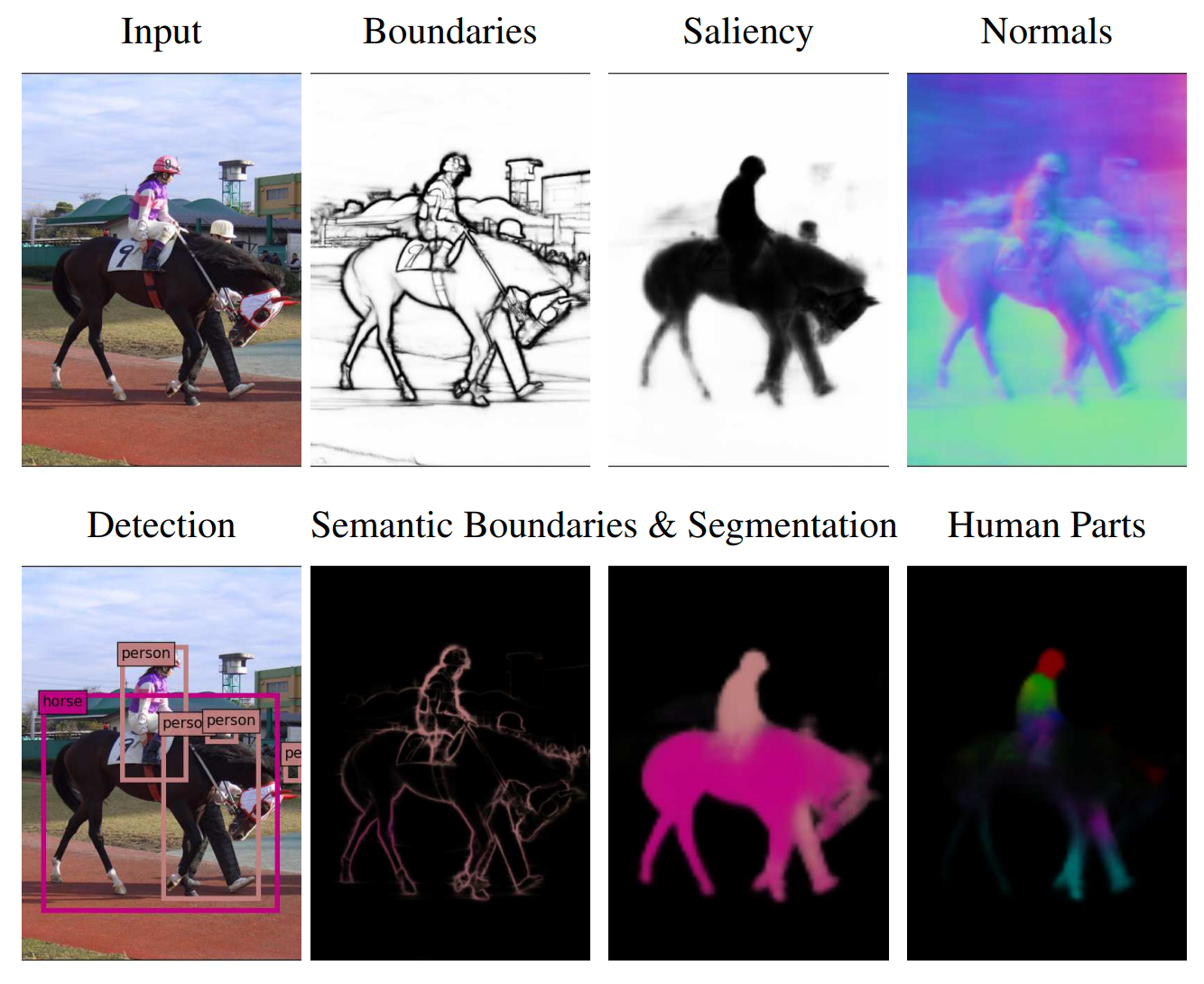}   
    \caption{UberNet}
\end{subfigure}
\begin{subfigure}[t]{0.47\linewidth}
    \centering
    \includegraphics[width=1.0\textwidth]{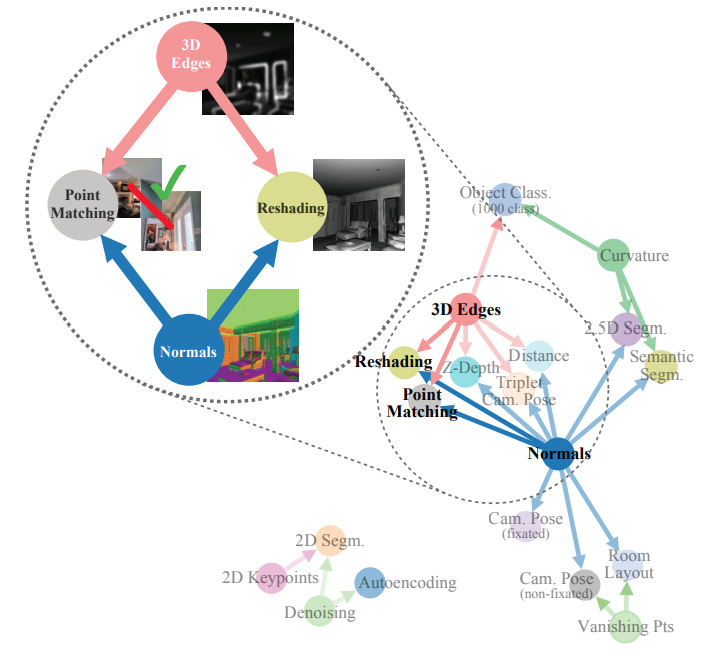}    
    \caption{Taskonomy}
\end{subfigure}
\caption{(a) UberNet~\citep{kokkinos2017ubernet} can be applied to 7 vision tasks using a unified and budget-controllable CNN architecture. (b) Taskonomy~\citep{zamir2018taskonomy} further studies the relationship across vision tasks by exploiting a multi-task transfer modeling. Image credit:~\cite{kokkinos2017ubernet} and~\cite{zamir2018taskonomy}.}
\vspace{-5pt}
\label{fig:chp4-ubernet}
\end{figure}

Some early works explore multi-task learning methods for unifying different vision or vision-language tasks. 

\paragraph{Vision models.} A few works explore using CNNs for learning with different vision tasks at different levels. For example, Cross-stitch Networks~\citep{misra2016cross} develops a strategy to split different numbers of layers from the top in CNNs so as to adapt to different vision tasks. Results show that the best-performing multi-task architecture depends on the tasks of interest and can hardly generalize to new tasks. UberNet~\citep{kokkinos2017ubernet} takes one step further to use a single universal CNN architecture and sophisticatedly design a routing mechanism to save the memory and computing cost, as shown in Figure~\ref{fig:chp4-ubernet}~(a). Both works require some tweaking to the CNN architecture so that they can adapt to different levels of tasks and loss types. But they unfortunately fail to build the synergy across tasks to improve model performance. Taskonomy~\citep{zamir2018taskonomy} specifically studies the relationship among vision tasks. It first trains task-specific models for each individual task and then performs transfer modeling across tasks in the latent space. The task affinity is then calculated in the latent space, providing us with the taskonomy. The result shows that vision tasks have different affinities for different groups, as shown in Figure~\ref{fig:chp4-ubernet}~(b). For example, surface normal estimation is heavily related to reshaping and point matching. Curvature extraction is related to image segmentation tasks. This study provides deep insights for multi-task vision modeling~\citep{xu2018pad,crawshaw2020multi}.

\paragraph{Multi-modal models.} The emergence of Transformers significantly facilitates the advancement of multi-task multi-modal learning. Among them, 12in1~\citep{lu202012} is one of the pioneering works that combine 12 vision-language tasks in a single BERT-based architecture. It uses task-specific heads for individual tasks and a commonly shared trunk ViLBERT~\citep{lu2019vilbert}. Results show that multi-task learning can achieve substantial improvements over single-task learning while reducing the model parameters significantly. Later on, UniT~\citep{Hu_2021_ICCV} exploits an encoder-decoder architecture and expands to vision-only tasks like object detection. Additionally, it allows end-to-end training on the task pool without relying on pre-trained detectors. Similar to 12in1, it also uses a task-specific head for each task, motivated by the empirical result  that sharing the same head usually hurts performance. Likewise, E2E-VLP~\citep{xu2021e2evlp} proposes an end-to-end pipeline for both localization tasks and text generation. Both UniT and E2E-VLP demonstrate the versatility of the encoder-decoder architecture of DETR~\citep{carion2020end}. Following the same spirit, GPV~\citep{Gupta_2022_CVPR} proposes an end-to-end task-agnostic architecture for different vision and vision-language tasks. It uses DETR to extract boxes and region features and then exploits a cross-attention module for fusion, followed by a vision decoder and a language decoder for decoding different outputs.

The above vision and multi-modal models unify different tasks by incorporating different modules or heads designed to cope with different tasks, and can hardly achieve synergy across tasks. In the following, we discuss recent model unification research that aims to make the best use of synergy among various vision and multi-modal tasks.

\begin{figure}
    \centering
    \includegraphics[width=1.0\linewidth]{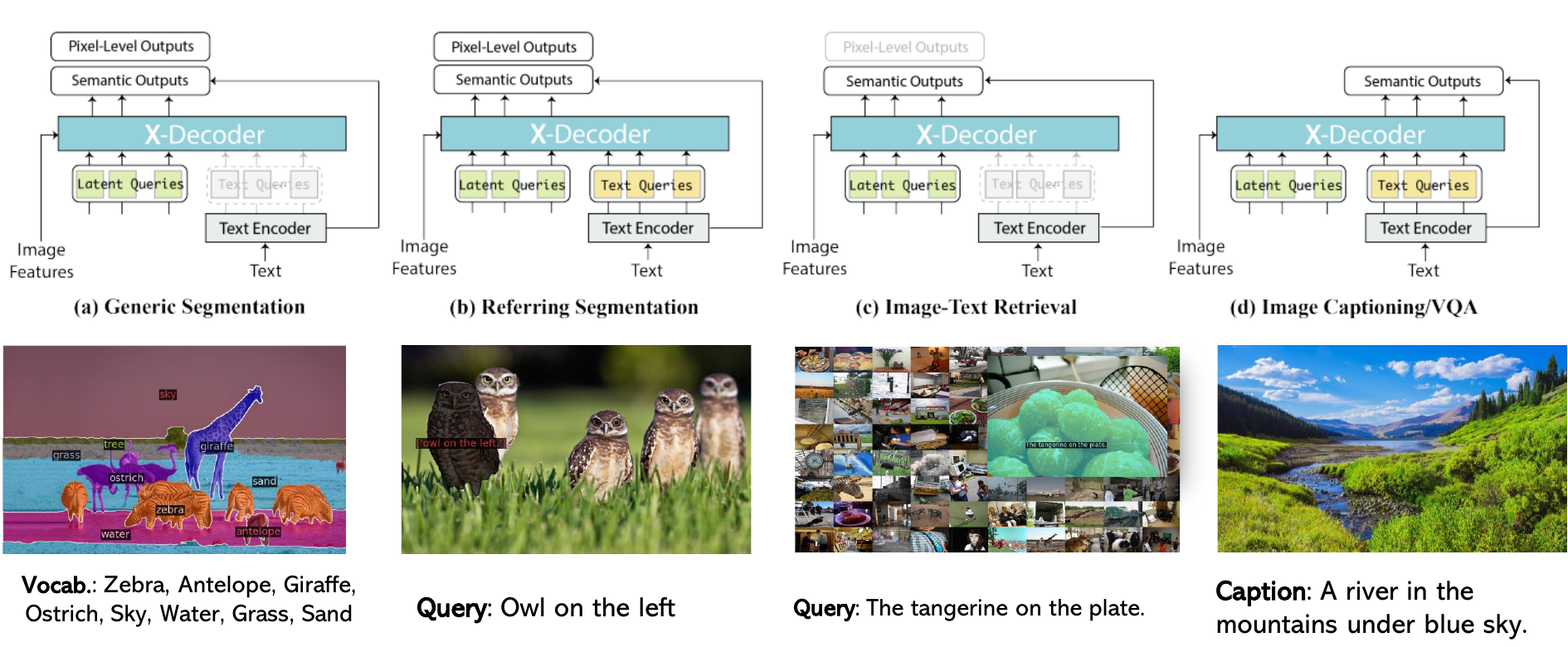}
    \caption{A generalist model X-Decoder~\citep{zou2023generalized} that unifies different vision and vision-language tasks in a functional manner. It uses a single decoder with the same suite of parameters, but different routing mechanisms to tackle different tasks. Image credit:~\cite{zou2023generalized}.}
    \label{fig:chp4-x-decoder}
    \vspace{-3mm}
\end{figure}

\subsubsection{Unified learning}

The barrier across tasks is gradually blurred thanks to the use of Transformers~\citep{vaswani2017attention} and the development of open-set models as we discussed earlier. It is now possible to bind inputs from different modalities to learn a shared semantic space. 
A number of works~\citep{zhang2022glipv2,zou2023generalized,li2023uni} have recently been proposed to unify vision and vision-language tasks by \emph{using one model for all}. After pre-training, the single model can be applied to tackle all tasks in a zero-shot manner and the performance can be further improved via task-specific finetuning. Note that unified learning in this context differs from previous works of large-scale pre-training. 
Like GPT which serves as a universal language interface after pre-training, a unified vision model is not only a representation learning engine but also an interface that supports as many tasks as possible in a zero-shot manner. Below, we review a few representative works.

\textbf{GLIPv2}~\citep{zhang2022glipv2} is proposed by extending GLIP~\citep{li2022grounded} to support a wide range of vision and vision-language tasks, including grounded captioning, visual question asnwering, \textit{etc}. GLIPv2 seamlessly integrates localization pre-training and Vision-Language Pre-training (VLP) through three distinct pre-training tasks: $(i)$ phrase grounding, which serves as a vision-language adaptation of detection tasks; $(ii)$ region-word contrastive learning, introducing a novel contrastive task at the region-word level; and $(iii)$ masked language modeling. In a zero-shot manner, this pre-trained model can be applied to different tasks and attain plausible performance across the board. Unlike previous works (\emph{e.g.}, GPV~\citep{Gupta_2022_CVPR}), it merges the localization module and vision-language matching module in a coherent manner, which makes model training from fused data much more efficient and effective.

\textbf{X-Decoder}~\citep{zou2023generalized} follows the generic design of encoder-decoder architecture. Given an input image, it first uses an image encoder to extract features at multiple scales. Afterward, a text encoder is used to encode a textual query into a sequence of embeddings. The visual features, textual queries and the non-semantic or latent queries are fed to a decoder to predict the outputs. Three critical designs are proposed to empower the generalization ability of X-Decoder to a variety of vision and vision-language tasks: $(i)$ It defines two types of queries and outputs. Specifically, the queries for the decoder are categorized into latent queries and text queries, which undertake generic vision and vision-language tasks, respectively. Likewise, the output is categorized into pixel-level masks and semantic embeddings; $(ii)$ A single text encoder is exploited to encode the textual corpus from all tasks. The common text encoder is used to encode referring phrases, text descriptions, and image captions in the task of referring segmentation, image-text retrieval and image captioning, respectively; $(iii)$ It fully decouples the image and text encoder, and use all the outputs as queries. As such, it can learn from both intra-image supervisions and inter-image ones, which is essential to learn stronger pixel-level representations and support different granularity of tasks. As shown in Figure~\ref{fig:chp4-x-decoder},  the pre-trained model can support different tasks by taking different routing while sharing the same suite of parameters. 

\textbf{Uni-Perceiver-v2}~\citep{li2023uni} is another generalist model that unifies vision and vision-language tasks. Similar to X-Decoder, the model exploits a vision encoder, a text encoder and a general decoder. Differently, it introduces a region proposal network on top of the vision backbone to explicitly predict the boxes and masks, which are then encoded as ``queries'' for the general decoder. To jointly train on datasets at different levels, it introduces a unified max-likelihood estimation strategy for tasks with localization and without localization.


\section{From Static to Promptable Models}\label{sec:chp4_from_static_to_promptable}

The success of Large Language Models (LLMs) such as ChatGPT~\citep{openai2023gpt4} have shown the importance of modern AI models in interacting with humans, and have provided a glimpse of AGI~\citep{bubeck2023sparks}. The ability to interact with humans requires a user-friendly interface that can take as many types of human inputs as possible and generate responses that humans can easily understand. In NLP, such a universal interaction interface has emerged and evolved for a while from early models like GPT~\citep{brown2020language} and T5~\citep{raffel2020exploring}, to more advanced techniques like prompting~\citep{shin2020autoprompt,zhao2021calibrate,li2021prefix} and chain-of-thought~\citep{wei2022chain,kojima2022large, schick2023toolformer}. However, most vision models are still static in that they are less flexible than LLMs to various prompts. Most recently, a number of works have proposed to enhance the static vision models with the capabilities to support: $(i)$ multi-modal prompting; $(ii)$ in-context prompting.

\subsection{Multi-modal Prompting}

Vision is different from language by nature. To enable a smooth interaction between humans and AI, a model requires not only language prompts but also other types of prompts to complement the missing information or resolve the ambiguity in language. Recently, a number of works have explored how to combine or augment language prompts with other types of prompts, such as spatial prompts~\citep{kirillov2023segment}, visual prompts~\citep{zou2023segment} and other modalities~\citep{girdhar2023imagebind,liu2023prismer}. In the following, we review some representative works.

\paragraph{Spatial prompting.} Vision is rooted in the physical world, and as such it is not only semantic but also spatial by nature. Spatial prompting can be considered as a way to modulate the vision models through the inputs of location information, which could be a point, a box, or an arbitrary stroke, \textit{etc}. Such clues have been heavily used in UI designs of computers (\textit{e.g.}, mouse) and mobile devices (\textit{e.g.}, touch screen). In computer vision, interactive segmentation~\citep{mortensen1998interactive,mcguinness2010comparative,chen2021conditional,chen2022focalclick} naturally requires such capability so that the model can take multiple clicks from users and gradually refine the segmentation mask. However, most of these works are still designed task-specifically and lack enough flexibility to support different types of spatial prompts.

\begin{figure}
    \centering
    \includegraphics[width=1.0\linewidth]{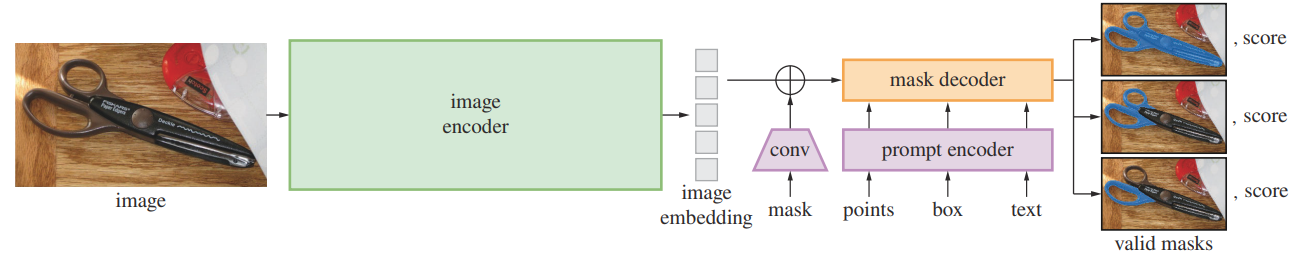}    \caption{SAM~\citep{kirillov2023segment} proposes a promptable segmentation model which can take different spatial prompts in addition to text prompts. It further develops a data annotation engine to scale up the mask-annotated data. Image credit:~\cite{kirillov2023segment}.
    }
    \label{fig:chp4-sam}
\end{figure}

SAM~\citep{kirillov2023segment} is one of the pioneering works that propose a convenient spatial prompting interface and learn a foundation model for image segmentation. As shown in Figure~\ref{fig:chp4-sam}, the model can take points or boxes as the prompts, and segment images in arbitrary granularity. The ability to segment images following the user instructions from humans makes it readily a foundation to build many more models and applications~\citep{zhang2023survey}. To name a few, a number of works~\citep{ma2023segment,roy2023sam} start with SAM and train a promptable segmentation model for the medical domain. Spatial prompting is particularly beneficial in that the textual annotations for medical images are usually limited and hard to interpret. Similar cases also happen in other industry domains~\citep{tang2023can}. To further improve point prompting, SAMAug~\citep{dai2023samaug} proposes to refine the points using the max entropy criterion and saliency map, which can help to determine the most informative locations the model should look at.

\begin{figure}
    \centering
    \includegraphics[width=.95\linewidth]{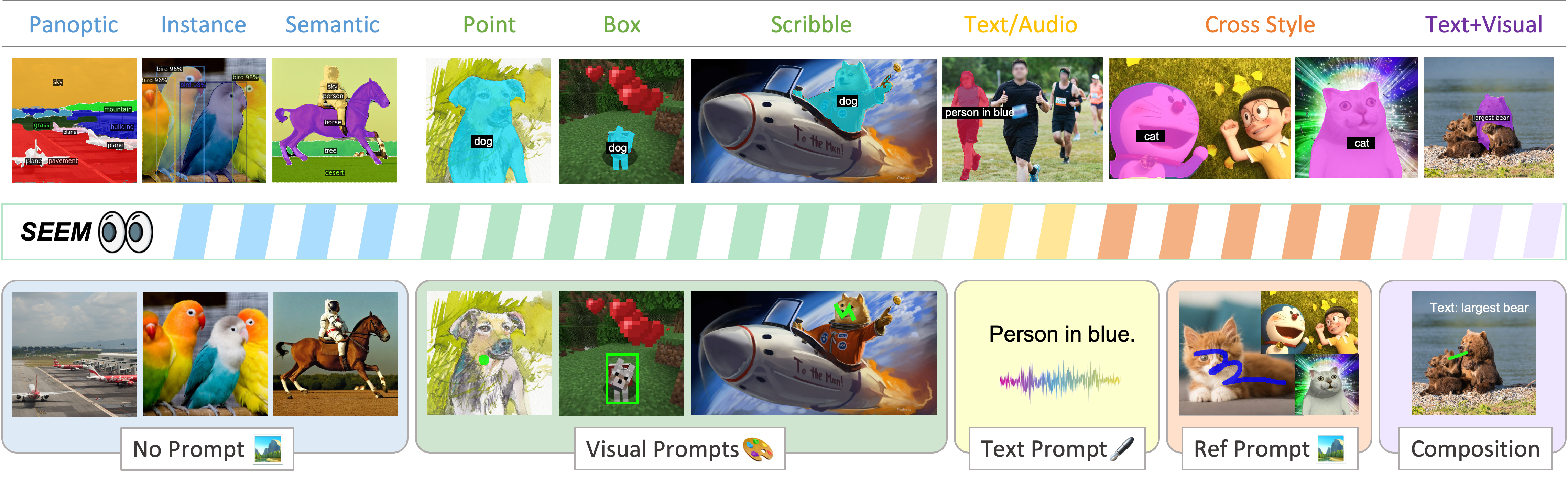}
    \caption{ SEEM~\citep{zou2023segment} can take different types of prompts as inputs for various image segmentation tasks. Image credit:~\cite{zou2023segment}.}
    \label{fig:chp4-seem}
    \vspace{-6pt}
\end{figure}

\paragraph{Visual prompting.} In many cases, textual descriptions of objects are not necessarily clear to convey the information. For example, given an unrecognizable or indescribable object, people may fail to express themselves clearly about the object. In this case, showing one or a few examples would be more informative and straightforward. With this idea, a lineup of works have studied exemplar-based visual modeling, such as image-to-image retrieval~\citep{yoon2021image,datta2008image,zhang2018generative}, image co-segmentation~\citep{joulin2010discriminative,jerripothula2016image} and visual object tracking~\citep{yilmaz2006object,luo2021multiple,wu2013online}. Most recently, this strategy has been formulated as visual prompting in that different types of visual inputs are usually encoded to some unified format and then fed into a Transformer architecture, as shown in LLMs. 

SEEM~\citep{zou2023segment} is one of the representative works that enable visual prompting to a vision model for image segmentation. As shown in Figure~\ref{fig:chp4-seem}, SEEM differs from the aforementioned SAM and can take visual prompts by drawing points, boxes, and strokes on an image that can be the target image or another reference image. It develops a new module called a visual sampler that can extract visual features from an image according to the locations specified by users. Based on the visual sampler, the model can even take another reference image as input without any training like that. As a result, it shows impressive performance not only for various image segmentation tasks but also for video object segmentation in a zero-shot manner.

PerSAM~\citep{zhang2023personalize} develops a personalized segmentation model on top of SAM and takes one shot as the input. It learns a specific model that takes a source image plus a mask as input and then predicts the mask for a target image. To extract the visual prompts, mask pooling is taken and used as the input tokens to the decoder of PerSAM. It also proposes a way to extract the positive and negative priors based on feature matching to facilitate pre-trained SAM models with comprehensive clues. Like most prompt learning methods in LLMs, a plausible feature for PerSAM is that it can be easily attained by some off-the-shelf models like SAM. SAM-PT~\citep{rajivc2023segment} further applies this strategy to video object segmentation. Inspired by the spatial prompting in SAM, it exploits a point-tracking system~\citep{harley2022particle} to track different points (both positive and negative ones) and then ask SAM to segment the image given the points. It exhibits strong point tracking performance as well as segmentation. 

\paragraph{Others.} Some other works combine a wide range of visual prompting types. For example, Painter~\citep{wang2023images} reformulates different vision tasks (\emph{e.g.}, depth estimation, image segmentation) all as prompting and learns a decoder in an in-context learning manner. The prompts are combinations of raw images and the corresponding dense annotations (\emph{e.g.}, depth or segmentation maps). In contrast, Prismer~\citep{liu2023prismer} makes use of many off-the-shelf vision models to extract different information from the raw images and then feed the information to a vision-language model. To facilitate the interplay across multiple modalities, ImageBind~\citep{girdhar2023imagebind} learns a universal alignment among image/video, language, audio and depth. Once the embedding space is learned, it can be used to compose different types of prompts by simply doing the summations.

\subsection{In-context Prompting}

The in-context learning capability has been observed in many LLMs such as GPT-3~\citep{radford2019language}, which makes the model more configurable via prompting without any model parameter updates. In contrast, till now, the in-context learning capability for vision models is still less studied. Flamingo~\citep{alayrac2022flamingo} is one of the pioneering works that demonstrate in-context language generation for multi-modal inputs, which is acquired by learning from interleaved image-text pair data. Likewise, Kosmos-1~\citep{huang2023language} is another work that takes visual inputs as a foreign language so that the in-context learning ability in LLMs can be naturally translated to multi-modal inputs. However, both methods take multi-modal data as inputs but merely generate texts as outputs. As we discussed earlier, vision tasks require outputs of different types beyond texts. How to endow the in-context learning ability for vision systems is still an open question. Below, we review some recent attempts toward the goal.

\begin{figure}[t]
    \centering
\includegraphics[width=0.98\linewidth]{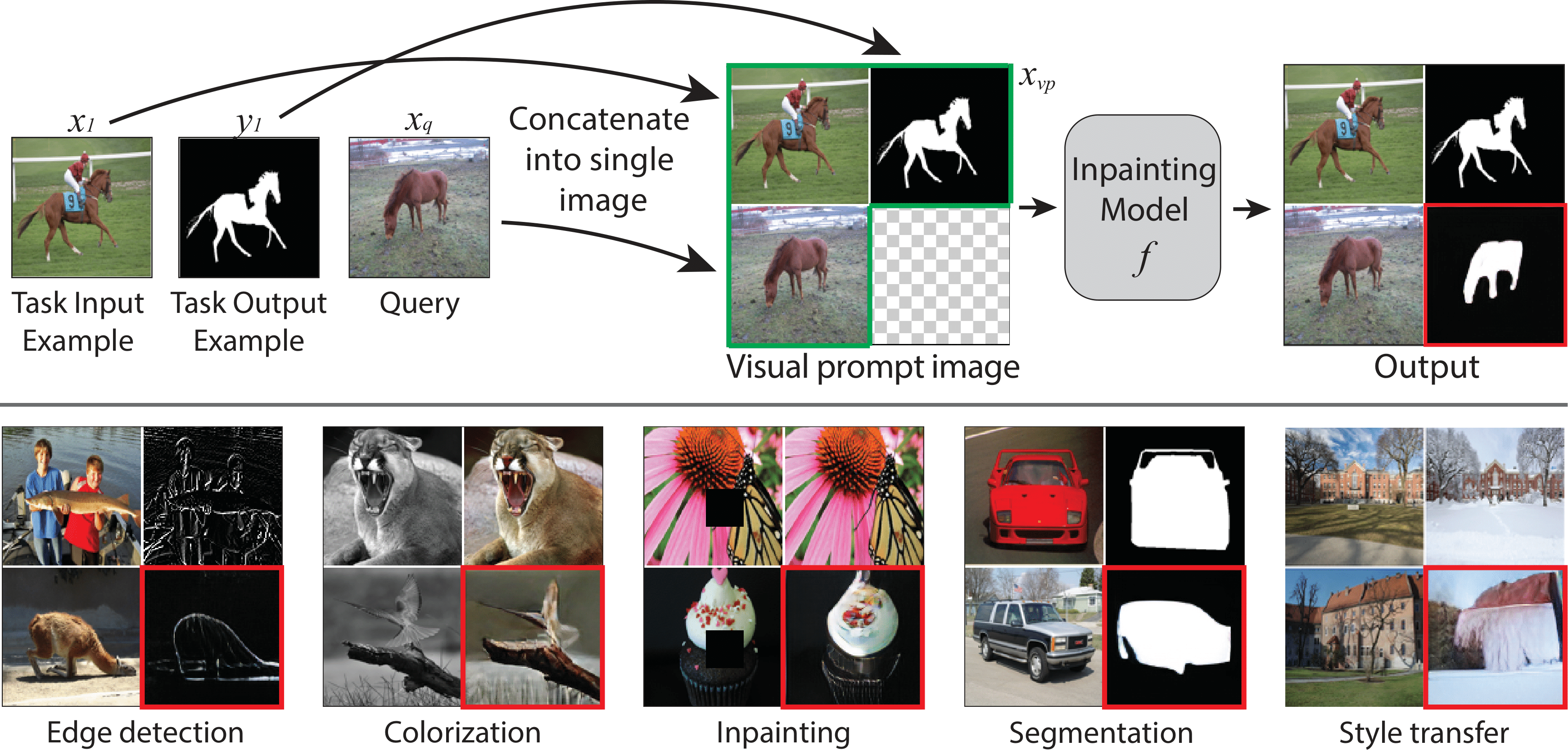}
    \caption{Visual prompting via visual inpainting proposed in~\cite{bar2022visual}. Image credit:~\cite{bar2022visual}.}
    \label{fig:chp4-visual-inpainting}
    \vspace{-5pt}
\end{figure}

Visual prompting via inpainting is proposed in~\cite{bar2022visual} to teach the model to predict dense outputs, such as edges, masks, depths, \textit{etc.} as shown in Figure~\ref{fig:chp4-visual-inpainting}. Given an input image $\textbf{x} \in \mathcal{R}^{H \times W \times 3}$ and a binary mask $\textbf{m} \in \{0,1\}^{H \times W}$, an inpainting model is to predict the missing region $\textbf{y} = f(\textbf{x}, \textbf{m})$. The authors exploit a pre-trained VQ-GAN to encode the original image into discrete tokens and ask another ViT encoder to predict the masked regions. To make sure the model understands the visual ``context'' in the images, the authors collected a new dataset called \textit{Computer Vision Figures} dataset which consists of 88k images from Arxiv papers. After pre-training, the model is used to predict the content at the bottom-right corner.

\begin{figure}
    \centering
    \includegraphics[width=1.0\linewidth]{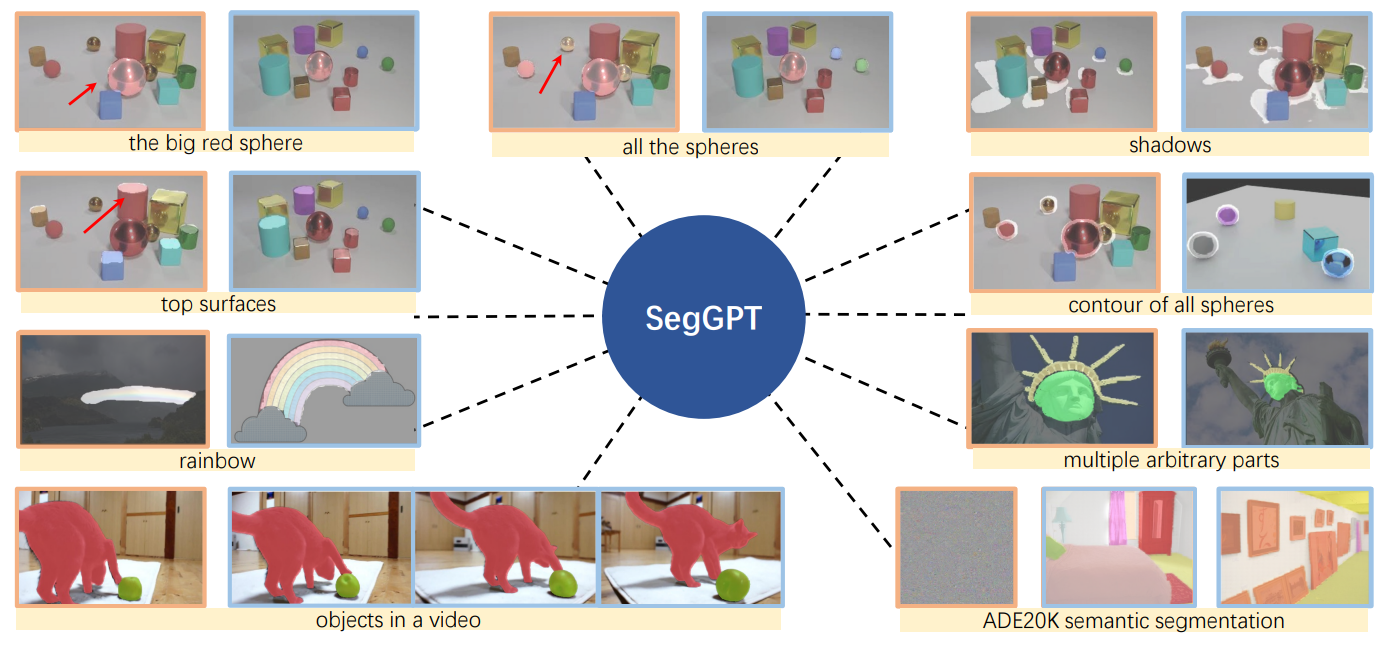}
    \caption{SegGPT~\citep{wang2023seggpt} proposes to perform in-context learning for image segmentation. Image credit:~\cite{wang2023seggpt}.}
    \label{fig:chp4-seggpt}
    \vspace{-3pt}
\end{figure}

\begin{figure}
    \centering
    \includegraphics[width=1.0\linewidth]{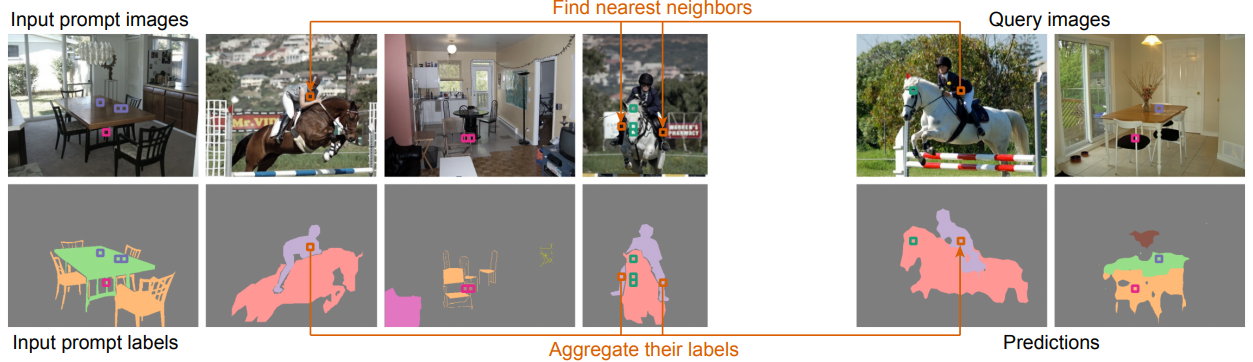}
    \caption{Hummingbird~\citep{balavzevic2023towards} is proposed for in-context visual scene understanding through nearest-neighbor retrieval. Image credit:~\cite{balavzevic2023towards}.}
    \label{fig:chp4-hammingbird}
\end{figure}

Concurrently, Painter~\citep{wang2023images} extends a similar idea of visual in-context learning to more diverse datasets and benchmarks. Unlike~\cite{bar2022visual}, it predicts the output in the continuous pixel space instead of discrete tokens. For different tasks, the authors define rules to convert the output spaces into image spaces. For example, it uses different colors to represent different individual instances in the image for the segmentation task. After unifying the input and output format, the authors use vanilla ViT as the encoder and masked image modeling~\citep{he2022masked}. A follow-up work called SegGPT~\citep{wang2023seggpt} is built on top of Painter and designed specifically for image segmentation tasks. The pre-trained model can be easily extended for exemplar-based image segmentation tasks.

Hummingbird~\citep{balavzevic2023towards} resorts to a different method for in-context visual learning. Instead of using masked modeling, the authors propose to leverage attention across target and source images to aggregate the information. As shown in Figure~\ref{fig:chp4-hammingbird}, the models take multiple input images (first row) and corresponding semantic label maps (second row). Given a query image, it first finds the nearest neighbor feature locations in the prompt images for the query points and then projects the same matches to the semantic label maps so as to aggregate the label for the target query. This strategy is akin to earlier works that build classification models based on K-nearest-neighbor but differently applied to dense prediction tasks. 

\paragraph{Discussion.} In-context learning is arguably an appealing feature. On one hand, there are a number of works that attempt to bridge vision with LLM so as to inherit the in-context learning capability such as Flamingo~\citep{alayrac2022flamingo} and Kosmos-1~\citep{huang2023language}. On the other hand, researchers resort to pure vision-based in-context learning to address vision-specific tasks such as image segmentation, depth estimation, \textit{etc}. Thus far, there is no single model that can take multi-modal inputs and predict different types of outputs as well in an in-context learning manner, which may render a promising future direction along this line. 

\section{Summary and Discussion}\label{sec:chp4_discussion}

\begin{figure}
    \centering
    \includegraphics[width=1.0\linewidth]{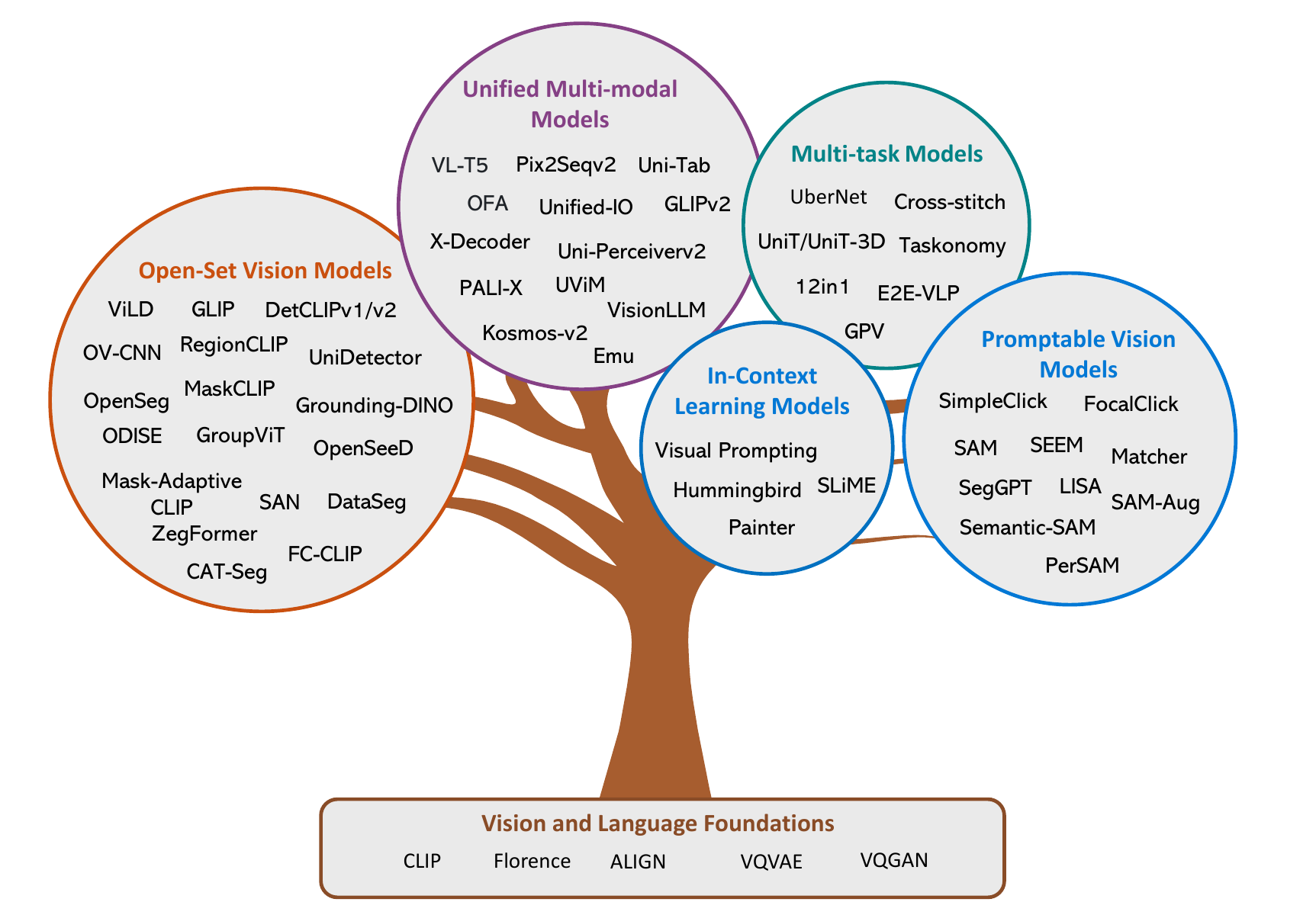}
    \caption{A summary of topics covered in this chapter. A lot of effort has been spent to unify vision models from different aspects to enable more intelligent vision systems.}
    \label{fig:chp4-tree}
    \vspace{-3pt}
\end{figure}

To the end, an illustrative summary of the works that have been covered in this chapter is shown in Figure~\ref{fig:chp4-tree}. There is a clear trend in the vision community to build \emph{open-world, unified and interactive} vision models. Nevertheless, there are still some intrinsic differences between vision and language. First, vision differs from language in that it captures the physical world with raw signals. We need to develop some sophisticated tokenization methods to compress the raw data into compact ``tokens''. In the language domain, this can be easily done by using some well-established heuristic tokenizers~\citep{sennrich2016neural}. Second, unlike language, vision data itself is not labeled and thus difficult to convey information or knowledge. It always requires human labors to annotate the visual contents in either a semantic or spatial manner. Third, language data is homogeneous while vision data and tasks are heterogeneous. Last but not least, storing vision data is much more costly than language data. For example, GPT-3 consumes 45 TB of training data, while the ImageNet dataset which contains 1.3M images costs more than hundreds of gigabytes. When it comes to video data like Howto100M~\citep{miech2019howto100m}, the storage cost already exceeds that of training corpus for GPT-3. All these differences cast some open questions that need to be addressed in the vision community, detailed below.
\begin{itemize}[leftmargin=*]
    \item \textbf{Computer vision in the wild}. Due to the heterogeneous nature, the current vision data we use for training models can hardly cover the full picture of the physical world. Despite the effort in building open-set vision models, we are still facing significant challenges in coping with novel or long-tail scenarios. 
    \item \textbf{Scaling law in vision}. As discussed in~\cite{kaplan2020scaling,hoffmann2022training}, the performance of large language models improves smoothly with the increase of model size, data scale, and amount of computes. As the scale increases, some intriguing emerging properties are further observed in LLMs. In contrast, it is still not clear what 
    is the right path to scale vision models, not to mention the emerging properties in such models.
    \item \textbf{Vision-centric or language-centric models}. Currently, the boundary between vision and language is gradually dismissed. However, due to intrinsic differences between vision and language, it is still not clear whether we should further scale up the vision models and integrate language models or the combination of moderate vision models and LLMs is sufficient to address most (if not all) of the problems. 
\end{itemize}

With that being said, we are close yet still far away from an intelligent vision system that can perceive the world like humans. We hope the literature review in this chapter could provide an overall picture of the existing efforts, and inspire the pursuit of next-generation vision models.
\chapter{Large Multimodal Models:\\
Training with LLM}
\label{chp:training_with_llm}
\begin{wrapfigure}{r}{3.6cm}
  \centering
  \vspace{-6cm}
  \includegraphics[width=0.97\linewidth]{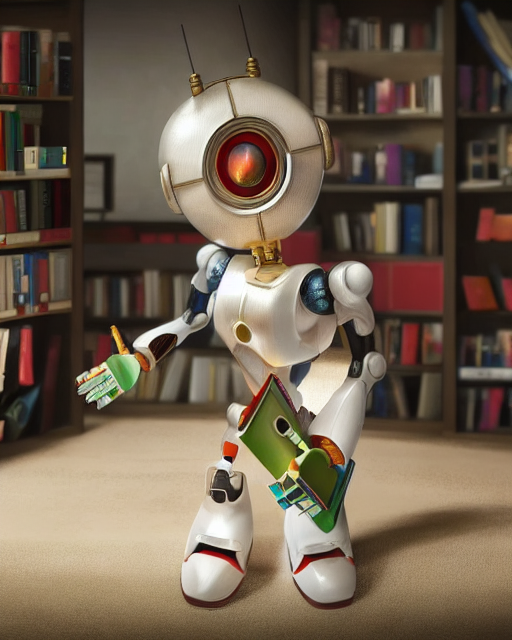}
\end{wrapfigure}

In this chapter, we comprehensively explore large multimodal models~\citep{alayrac2022flamingo,gpt4}. We begin with Section~\ref{sec:background_lmm} to delve into the background of such models, with the focus on the basics of image-to-text generative models and their representative model instances in various case studies. We also discuss the state-of-the-art OpenAI Multimodal GPT-4~\citep{gpt4} and identify the existing research gaps in the field. To better understand the process of instruction tuning in large language models, Section~\ref{sec:instruct_tuning_llm} examines its importance and its role in self-instruct and open-source LLMs. Moving forward, we explore instruction-tuned large multimodal models in Section~\ref{sec:instruct_tuning_lmm}, shedding light on their basics, significance and applications. Additionally, Section~\ref{sec:emerging_topics}  touches upon advanced topics in the realm of multimodal models to provide a deeper understanding of the subject. Finally, we assess the current progress in the field by evaluating how close we are to achieving the OpenAI Multimodal GPT-4 in Section~\ref{conclusions_lmm}, a major milestone in AI research.

\section{Background}
\label{sec:background_lmm}

\subsection{Image-to-Text Generative Models}

\begin{figure}[h!]
\centering  
\vspace{-4mm}
\hspace{-2mm}
\begin{tabular}{p{1.0\textwidth}}
\includegraphics[width=1.00\textwidth]{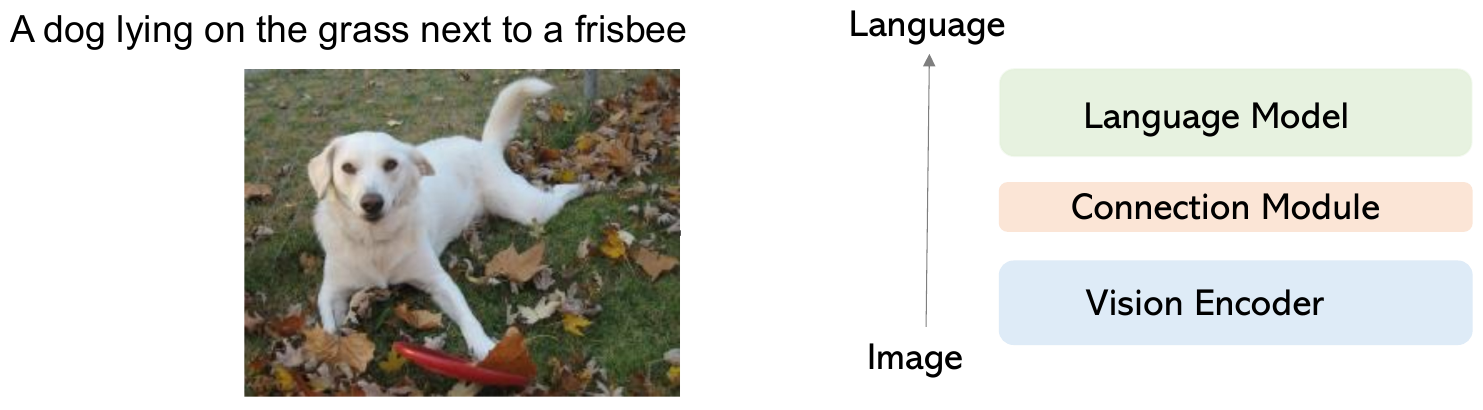} \\
(a) Left: An example of image-to-text generation task; Right: Model architecture. \\
\includegraphics[width=1.00\textwidth]{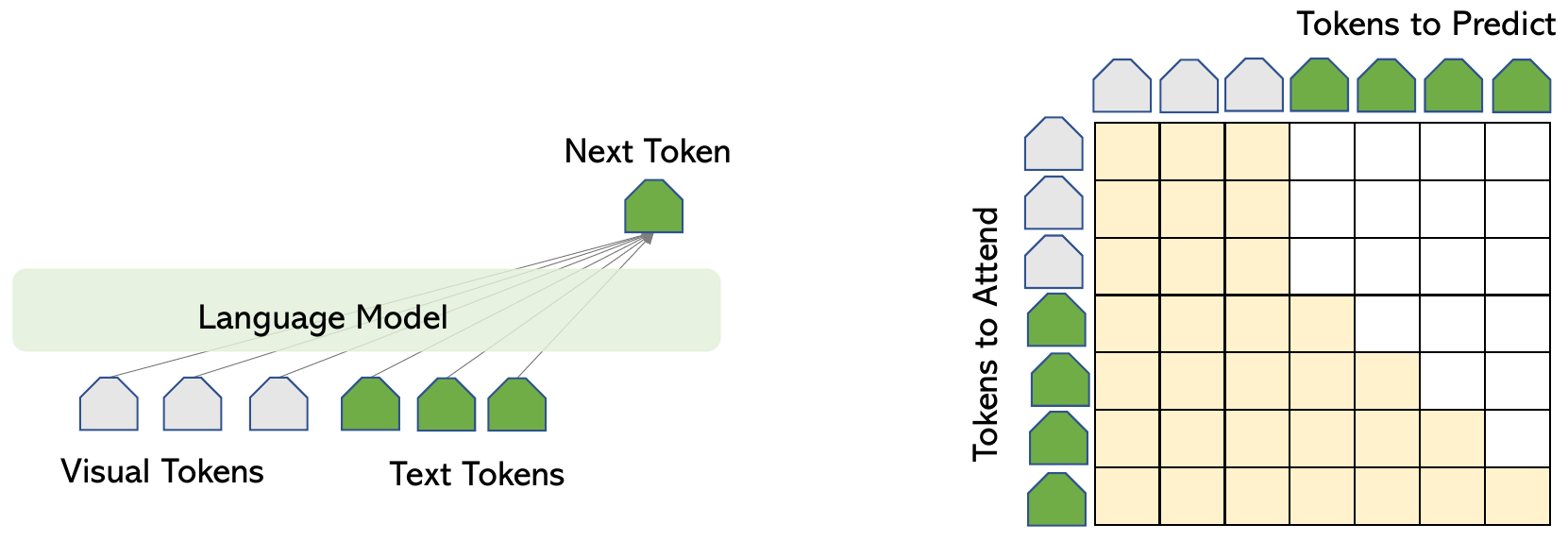} \\
(b) Training objective and attention mask. For each row, the yellow elements indicate that the prediction token attends the tokens on the left. \\
\end{tabular}
\vspace{-0mm}
\caption{Illustration of image-to-text generation task, architecture, and training objective. }
\label{fig:image2text}  
  \vspace{-1mm}
\end{figure}

LMMs in their current form is primarily an image-to-text generative model, which takes images as input, and outputs a text sequence. One example is illustrated in Figure~\ref{fig:image2text} (a) Left. All of the model variants share a very similar model architecture and training objective.

\begin{itemize}[leftmargin=7.5mm]
\setlength{\itemsep}{2pt}
\item 
{\bf \it Model Architecture}. As illustrated in Figure~\ref{fig:image2text} (a) Right, the model typically consists of an image encoder to extract visual features, and a language model to decode the text sequence. The vision and language modalities can be optionally connected by trainable connection module. The image encoder and language model can be either trained from scratch or initialized from pre-trained models.

\item
{\bf \it Training Objective}. As illustrated in Figure~\ref{fig:image2text} (b), it typically employs an auto-regressive loss on the output text tokens.
For the attention map in the Transformers~\citep{vaswani2017attention},  image tokens can attend to each other, and the current text token attends to all image tokens and the previous text tokens.
\end{itemize}

\subsection{Case Studies}
We use some prominent LMMs as examples to illustrate how the aforementioned network architecture can be instantiated in different models, while maintaining the same auto-regressive training objective.

\begin{figure}[t!]
\centering  
\vspace{-4mm}
\hspace{-2mm}
\begin{tabular}{l}
\includegraphics[width=1.00\textwidth]{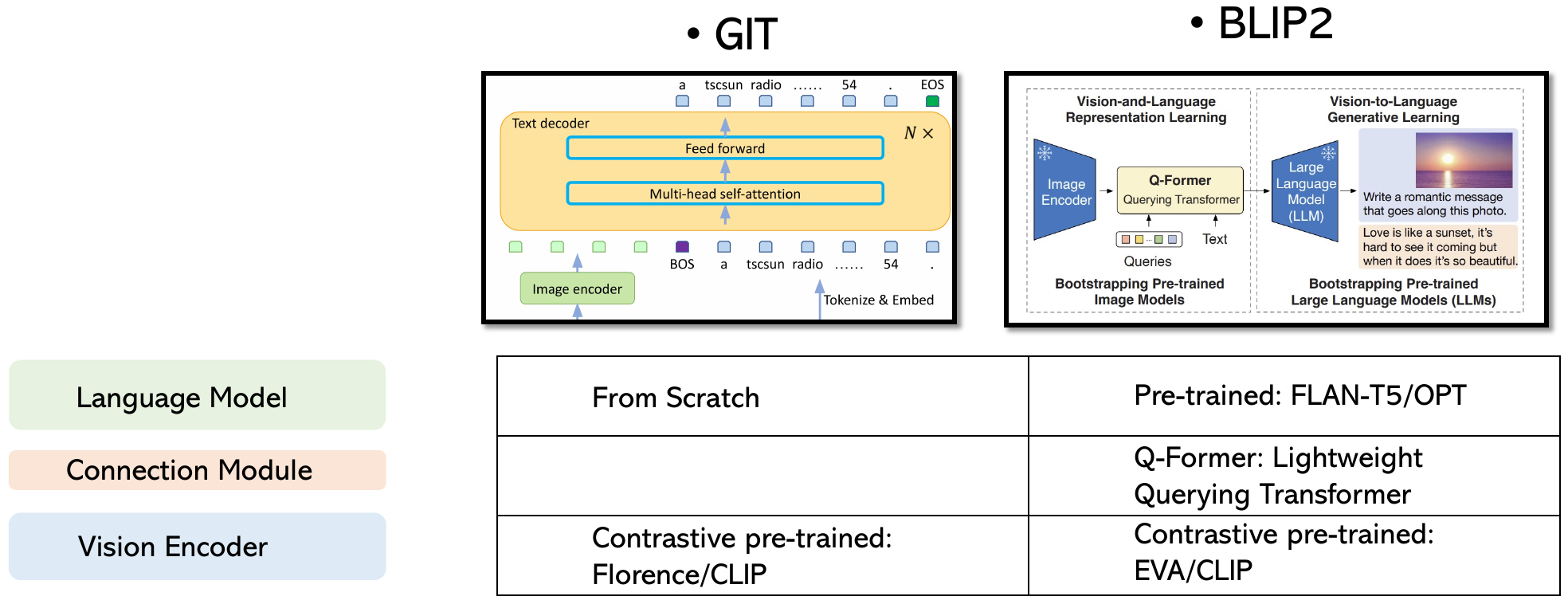} \\
(a) Example 1: LMM trained with image-text pairs. \vspace{2mm}\\
\includegraphics[width=1.00\textwidth]{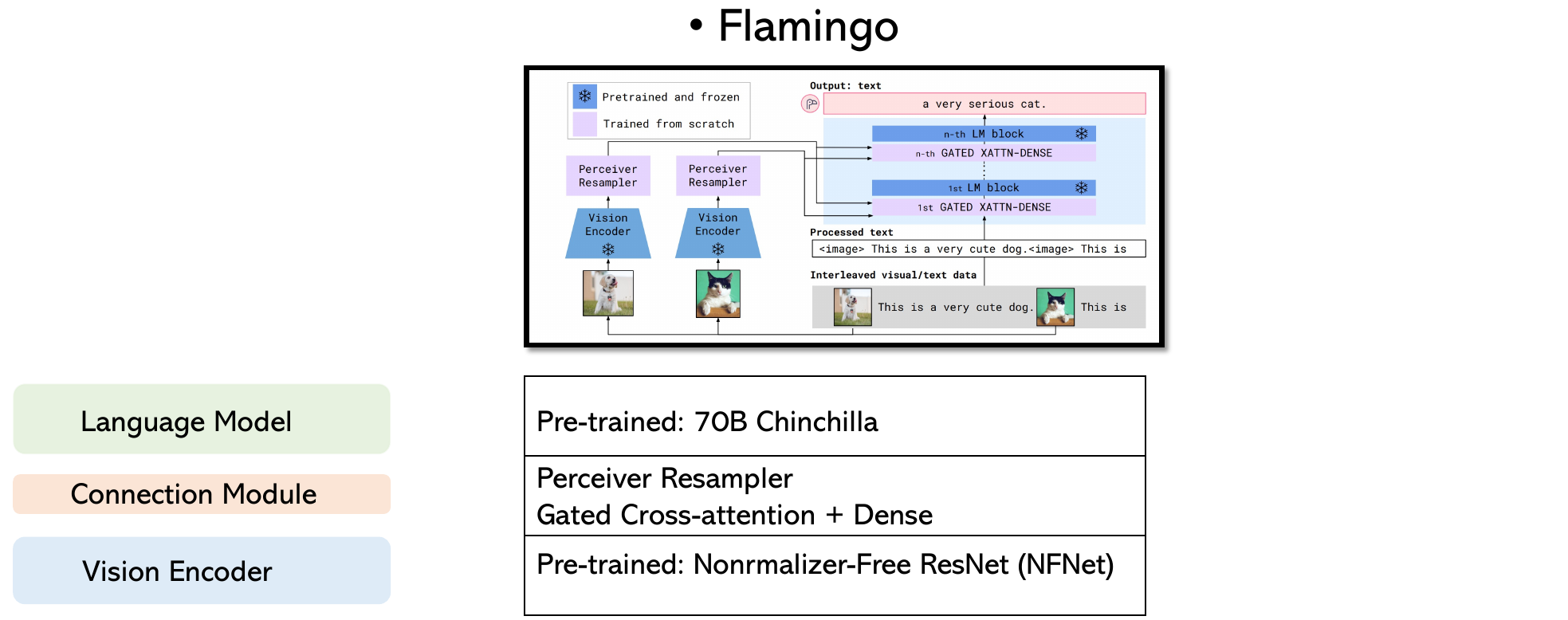} \\
(b) Example 2: LMM trained with image-text pairs and interleaved image-text data. \\
\end{tabular}
\vspace{-0mm}
\caption{Examples of image-to-text generation models. Image credit:~\cite{wang2022git,li2023blip,alayrac2022flamingo}.}
\label{fig:image2text_examples}  
  \vspace{-1mm}
\end{figure}

\paragraph{Case study I: LMM trained with image-text pairwise instances.}
Most LMMs are trained on a large number of image-text pairs, where each training sample is a pair. GIT~\citep{wang2022git} and BLIP2~\citep{li2023blip} are two large models that achieve state-of-the-art (SoTA) performance on many datasets. The comparisons are shown in Figure~\ref{fig:image2text_examples}(a). GIT initializes the image encoder with contrastively pre-trained Florence model~\citep{yuan2021florence}, and trains the language model from scratch. On the other hand, BLIP2 freezes the weights of a pre-trained image encoder and a pre-trained language model, while  trains a lightweight Q-former module to connect the image encoder and the language model. 

\paragraph{Case study II: LMM trained with interleaved image-text sequence instances.} We use Flamingo~\citep{alayrac2022flamingo} as an example, shown in Figure~\ref{fig:image2text_examples}(b).
It connects the frozen pre-trained image encoder and language model – by adding novel architectural components in between. Specifically, Perceiver Sampler module helps reduce computational complexity, and Gated Transformer module helps to stabilize training during the initial stage.
Flamingo is trained on a mixture of complementary large-scale multimodal data coming only from the web, without using any data annotated for machine learning purposes. After this training is done, Flamingo can be directly adapted to vision tasks via simple few-shot learning without any additional task-specific tuning.

\begin{figure}[t!]
\centering  
\vspace{-0mm}
\includegraphics[width=1.00\textwidth]{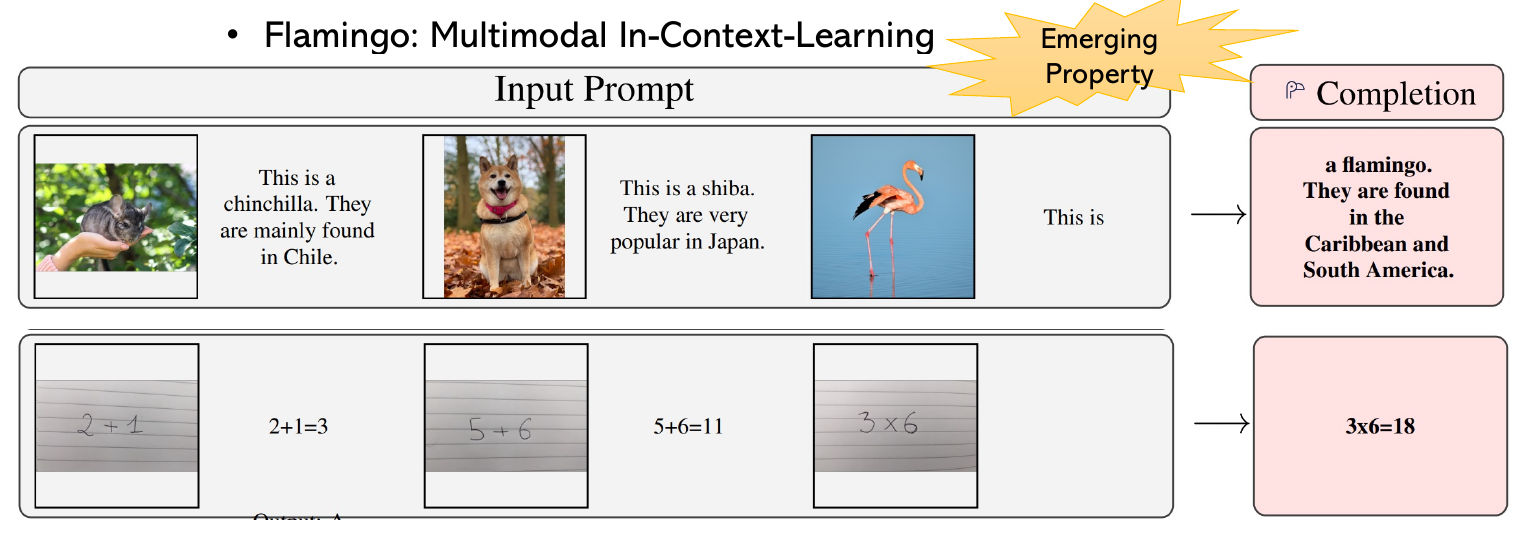} \\
\vspace{-0mm}
\caption{The emerging property of pre-training on web-scale interleaved image-text data: multimodal in-context-learning. Examples are adopted from~\cite{alayrac2022flamingo}.}
\label{fig:image2text_mic_emgerging_property}  
  \vspace{-1mm}
\end{figure}

\paragraph{Multimodal in-context-learning.}
Beside the SoTA performance on dozens of academic benchmarks, probably the most appealing aspect of Flamingo is the emerging property: Multimodal In-Context-Learning. Specifically, given a couple of image-text pairs as examples, Flamingo can zero-shot task transfer to unseen problems, such as solving visual math problems.
This means Flamingo can tackle a number of difficult problems with just a handful of task-specific examples, without any additional training required. 
For example in Figure~\ref{fig:image2text_mic_emgerging_property}, two new tasks are presented to Flamingo. The top row provides two image-text pairs as the context in the prompt, where the text describes the name of the animal in the image, followed by the geographical information of the animal. Flamingo is able to understand the patterns presented in the examples, and output the corresponding information for a new image. In the bottom row, the text first shows the optical character recognition (OCR) result of the image, followed by the answer to the math problem. Flamingo follows the task instruction illustrated in the multimodal context, and outputs the correct answer for a new math problem in the third image.
This intriguing in-context learning capability makes Flamingo 
the GPT-3 moment~\citep{brown2020language} in the multimodal domain.

\begin{figure}[h!]
\centering  
\vspace{-4mm}
\includegraphics[width=1.00\textwidth]{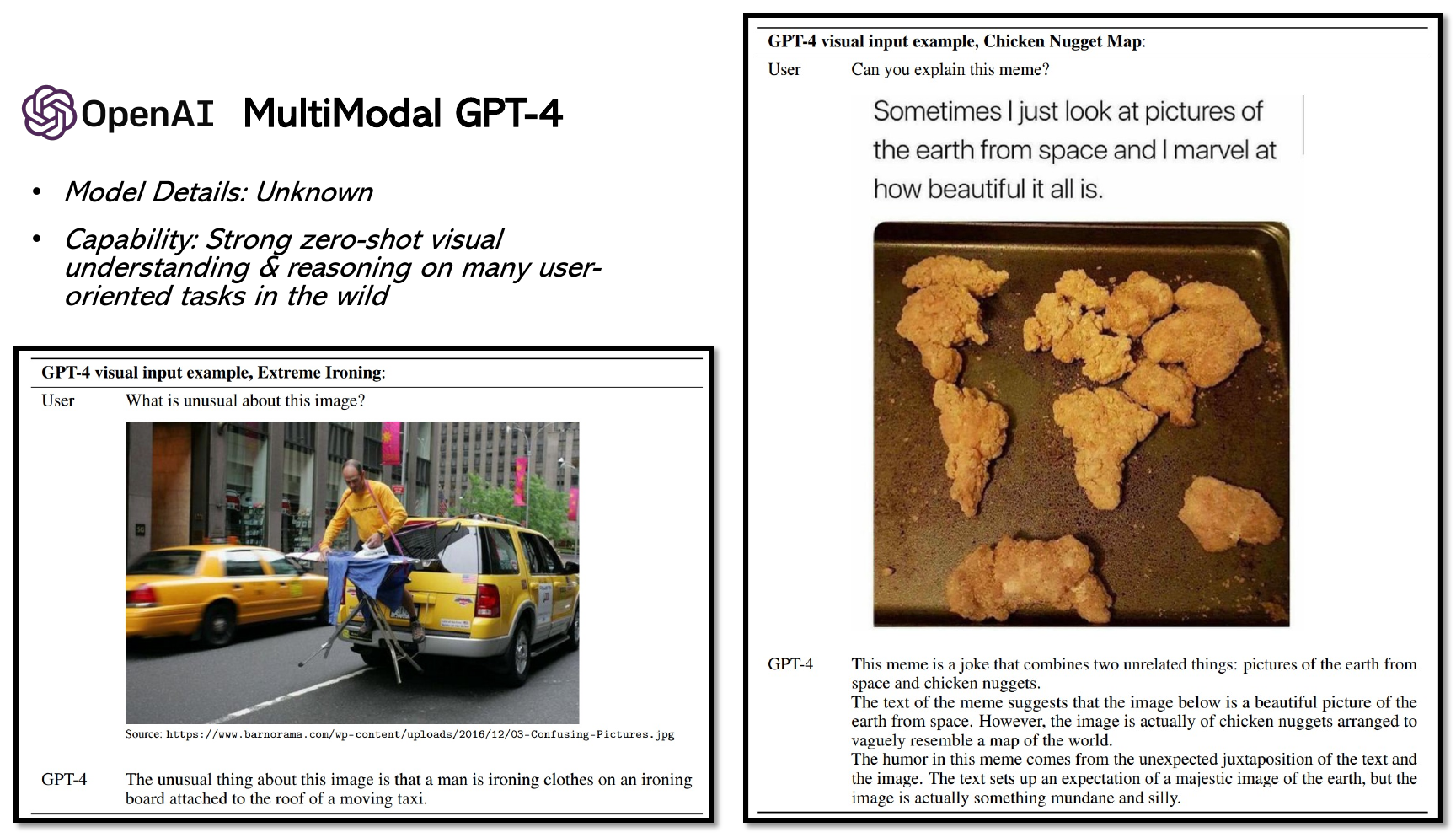} \\
\vspace{-0mm}
\caption{OpenAI Multimodal GPT-4. Visual examples are from~\cite{gpt4}.}
\label{fig:gpt4_examples}  
  \vspace{-1mm}
\end{figure}

\subsection{OpenAI Multimodal GPT-4 and Research Gaps}
In March 2023, OpenAI released GPT-4~\citep{gpt4}, with impressive capability in visual understanding and reasoning. Though the model details are not revealed, there is no doubt that GPT-4 enables many new scenarios, based on the examples highlighted in the technique report. For instance, two popular visual examples are illustrated in Figure~\ref{fig:gpt4_examples}. The first one identifies  the uncommon visual region and exhibits strong complex reasoning performance. The second one recognizes text in the image and captures the mere across image-text. For a while, the research community had no clue how this new ability is achieved (probably because they are not tightened to any established academic tasks/datasets), but all are determined that these are exciting results.
It naturally raises a question: how can we build Multimodal GPT-4 like models?

\begin{figure}[t!]
\centering  
\vspace{-4mm}
\includegraphics[width=1.00\textwidth]{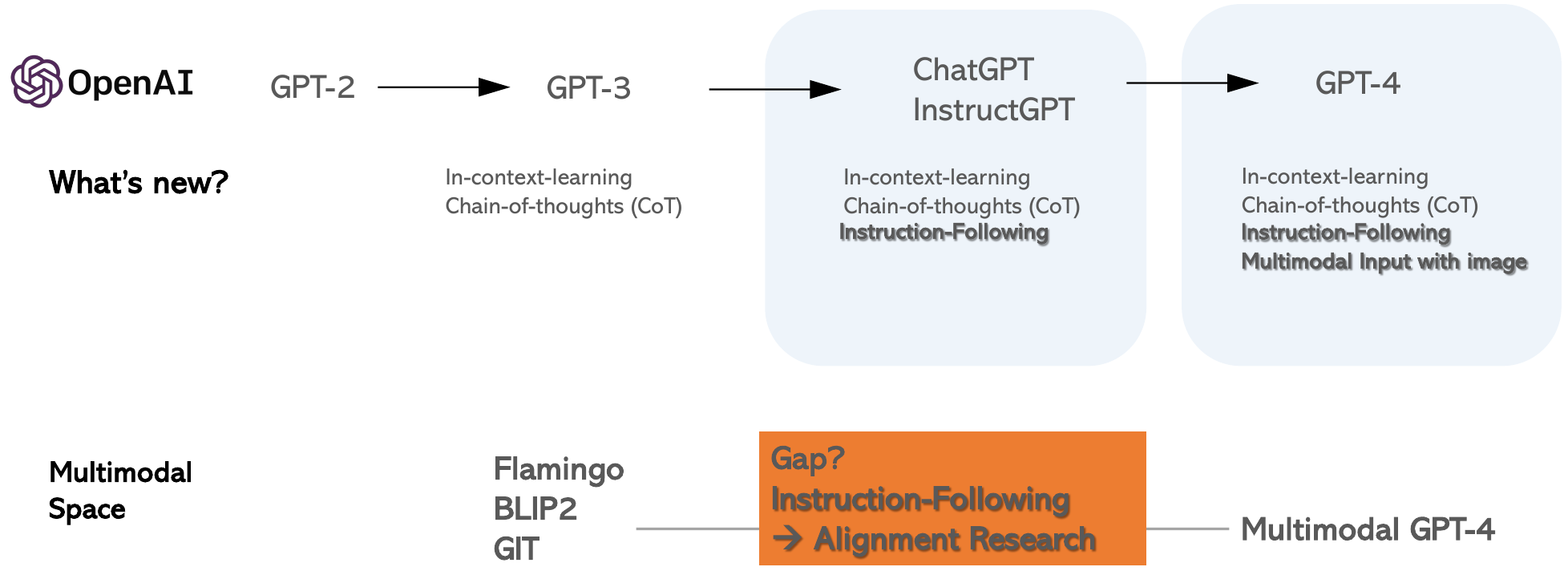} \\
\vspace{-0mm}
\caption{Recap on Language Modeling: OpenAI LLM development history. The unique properties for each generation model are highlighted, from which the research gap is revealed for LMM.}
\label{fig:lmm_research_gap}  
  \vspace{-1mm}
\end{figure}

To answer it, we start to review the big models from OpenAI, by highlighting the most appealing properties for each model in Figure~\ref{fig:lmm_research_gap}. There are several key observations:
$(i)$
GPT-2~\citep{radford2019language} is the auto-regressive counterpart in the BERT era~\citep{devlin2018bert} for the pre-train-then-finetune paradigm. Compared with GPT-2, GPT-3~\citep{brown2020language} is a 175B model trained on web-scale text corpus, which exhibits two emerging properties with a frozen model: in-context-learning~\citep{brown2020language} and chain-of-thoughts (CoT) reasoning~\citep{wei2022chain}. 
This means, without any additional training,  the model can tackle a wide range of new problems with just a few task-specific examples and by properly prompting it step-by-step, respectively. 
It further leads to the modeling paradigm from task-specific finetuning to prompting frozen models, where the latter shows higher generalizability and lower adaptation cost in task transfer.
$(ii)$ ChatGPT and InstructGPT~\citep{ouyang2022training} show the importance of instruction-following and alignment with human intents for LLMs, by finetuning the base language model GPT-3/GPT-3.5 on high quality instruction-following data, and improving them with a reward model via reinforcement learning with human feedback.
$(iii)$ GPT-4 not only improves the language ability of previous models, but also allows visual signals as additional input for visual understanding and reasoning. We see that the newer generation model  maintains/improves the existing properties of the previous ones, and enable new properties. 

In other words, from GPT-3 to GPT-4, we see two new properties: instruction-following and multimodal input.
This reveals the gap between existing LMMs (\textit{e.g.}, Flamingo) and multimodal GPT-4: how to perform instruction-following and alignment research in the multimodal space, which is the focus of this chapter.

\section{Pre-requisite: Instruction Tuning in Large Language Models}
\label{sec:instruct_tuning_llm}
Note that instruction-following is a notion originated in NLP. To study the intuition behind it and have a full picture of its history, we first revisit instruction tuning with LLMs.

\begin{figure}[h!]
\centering  
\vspace{-4mm}
\hspace{-2mm}
\begin{tabular}{p{1.0\textwidth}}
\includegraphics[width=1.00\textwidth]{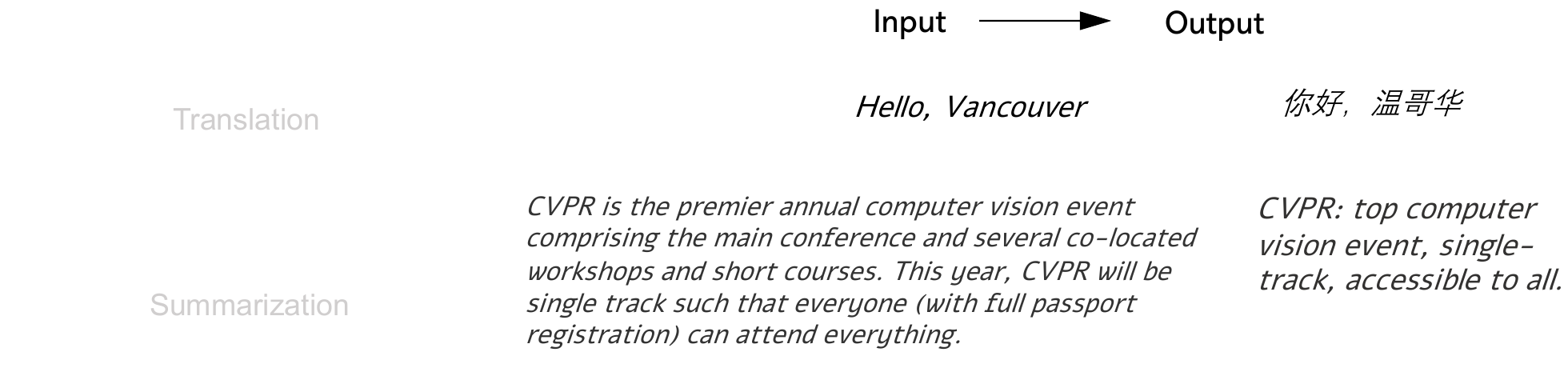} \\
(a) Training: Implicit task instructions in traditional language data. \vspace{2mm}\\
\includegraphics[width=1.00\textwidth]{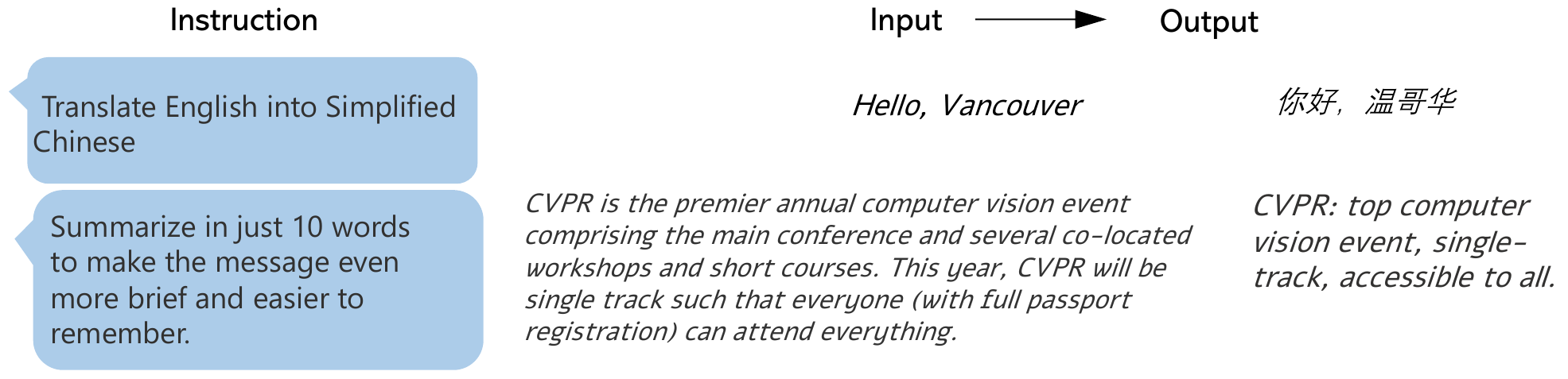} \\
(b) Training: Explicit task instructions in instructional language data. \\
\includegraphics[width=1.00\textwidth]{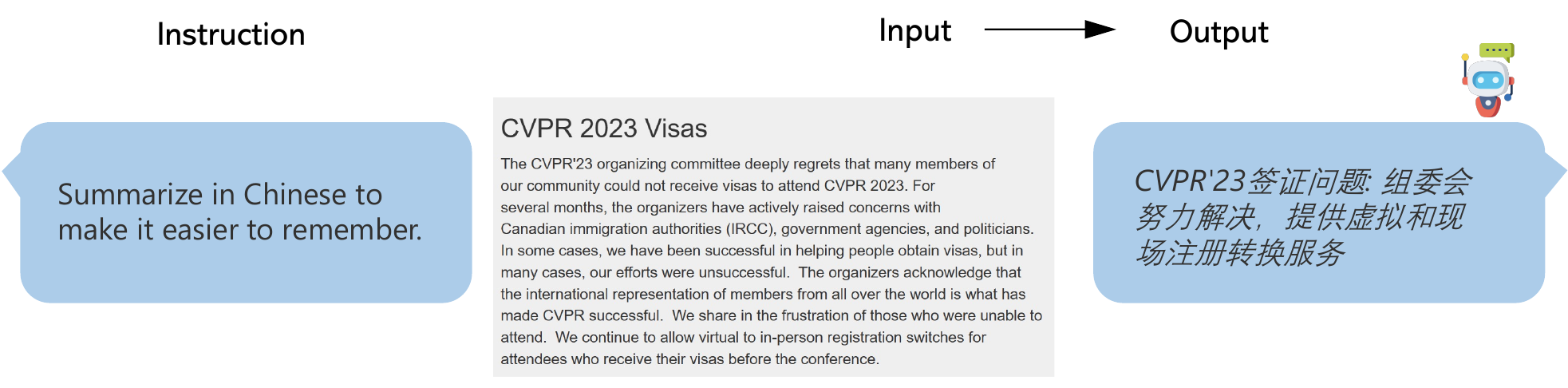} \\
(c) Inference: Explicit task instructions in instructional language data. The English meaning of the output is {\em ``CVPR'23 visa issue: organizing committee works to solve and provide virtual and in-person registration switch services."
} \\
\end{tabular}
\vspace{-0mm}
\caption{Examples of task instructions in traditional and instructional language data, respectively.}
\label{fig:task_instruction_lm}  
  \vspace{-1mm}
\end{figure}

\paragraph{Traditional language data.}
As a typical data instance in NLP, sequence-to-sequence (seq2seq) representation is widely adopted for many language tasks: each data instance consists of two parts: one sequence as the input and another sequence as the output. We provide two examples in Figure~\ref{fig:task_instruction_lm} (a). Without any task instruction specified, we know they are translation and summarization tasks, respectively.

This seq2seq representation is also the conventional data format in NLP research, where task instructions are implicit. Based on each data domain, individual models are trained. Or sometimes one model is trained with multi-task objectives over multiple data domain without specifying the task instructions. For both cases, the models are hard to generalize to new tasks in a zero-shot fashion, as they are not trained to understand task instructions, thus cannot distinguish and generalize what task to perform during testing time.

\paragraph{Instructional language data.}
Recently, researchers have started to explicitly add task instructions into the model training, as shown in Figure~\ref{fig:task_instruction_lm} (b). Interestingly, the task instruction of most NLP tasks can be expressed in natural language as well. It leads a new data format: instruction-input-output triplets. Based on the new format, one single model can be trained to perform multiple tasks, each with its specific instructions. Since models have observed many task instructions and many instances for each task during training, it is more natural and easier for them to generalize to new tasks by task composition in the inference stage.

For example, in the evaluation stage, a new task that requires both summarization and translation is provided in Figure~\ref{fig:task_instruction_lm} (c). Though the model has never seen this new task during training, it observes individual task basis, and learns to perform on new tasks. Note that we humans are always creating new tasks in our daily life, and presumably these new tasks would never been observed by models. It is thus appealing if a model is able to solve thousands of new tasks in the wild without training. This is partially why ChatGPT is becoming popular and prevalent so quickly.

\subsection{Instruction Tuning}
\label{sec:what_is_instruct_tuning_llm}

How can we collect a diverse set of high-quality instruction-following data? There are two general schemes. One is through human-human interaction, where humans (task providers) provide the annotation statement and requirements, based on which another group of humans complete the annotation tasks. Such a scheme is typically costly and time consuming. The other scheme is via human-machine interaction, where similarly humans provide the annotation statement and requirements, but it is now the machines/models that complete the annotation tasks. 

\begin{figure}[h!]
\centering  
\vspace{-0mm}
\includegraphics[width=1.00\textwidth]{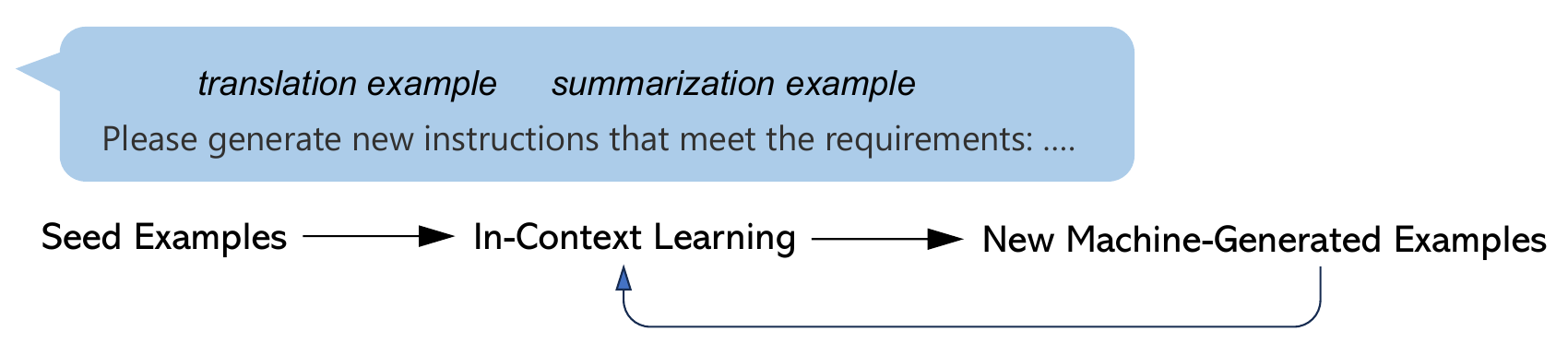} \\
\vspace{-0mm}
\caption{Illustration of the self-instruct pipeline~\citep{wang2022self}.}
\label{fig:self_instruct}  
  \vspace{-1mm}
\end{figure}

To enable LLMs to follow natural language instructions 
and complete real-world tasks, researchers have been exploring methods to instruction-tune LLMs. This is implemented by either finetuning the model on a wide range of tasks using human-annotated prompts and feedback~\citep{ouyang2022training}, or supervised finetuning using public benchmarks and datasets augmented with manually or automatically generated instructions~\citep{wang2022benchmarking}. Among these methods, Self-instruct tuning~\citep{wang2022self} is a simple and effective method of aligning LLMs to human intent, by learning from instruction-following data generated by SoTA LLMs. It turns out that the line of instruction-tuning research has produced effective means to improve zero-shot and few-shot generalization abilities of LLMs. Self-instruct leverages the in-context-learning ability of LLM. The pipeline is illustrated in Figure~\ref{fig:self_instruct}. Humans create a few examples (\ie seed examples) as the context, and ask LLM such as GPT-3 or GPT-4 to create more instructions and responses that follow the requirements stated in the prompt. The machine-generated instruction-following data can be further selected to construct with the prompt for in-context-learning in the next data generation iteration. The procedure iterates until a given number of samples are collected.
Due to the relatively lower cost and higher response speed of API calls (compared with human annotations), self-instruct is becoming more favorable in the research community.

\subsection{Self-Instruct and Open-Source LLMs}
\label{sec:what_is_instruct_tuning_llm_open}

\begin{figure}[t!]
\centering  
\vspace{-0mm}
\includegraphics[width=1.00\textwidth]{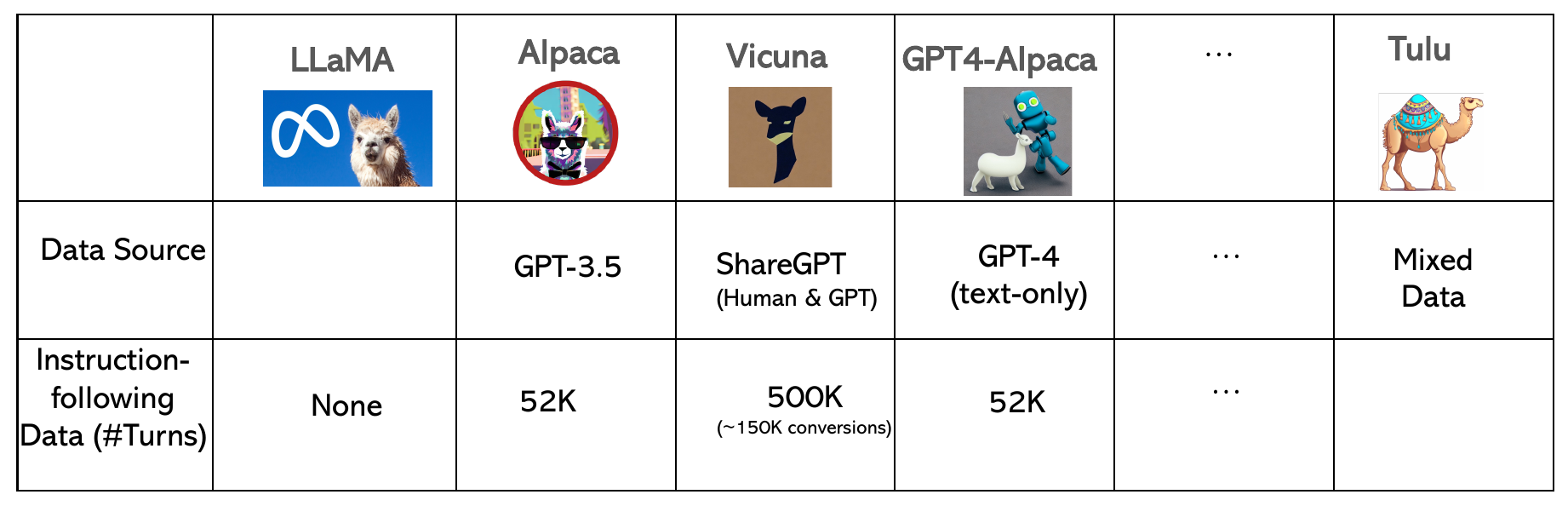} \\
\vspace{-0mm}
\caption{Model examples of the LLaMA family.}
\label{fig:llama_family}  
  \vspace{-1mm}
\end{figure}

The open-source community has witnessed a surge of open LLMs.
The success of ChatGPT~\citep{chatgpt} and GPT-4~\citep{gpt4} offers tremendous opportunities to improve open-source LLMs using instruction-tuning. 
Figure~\ref{fig:llama_family} compares several open-source instruction-tuned LLMs. 
LLaMA~\citep{touvron2023llama} is a series of open-sourced LLMs, which match the performance of proprietary LLMs such as GPT-3. To teach LLaMA to follow instructions, Self-instruct tuning has been quickly adopted given its superior performance and low cost. For example, to name a few early attempts in this line of research, Stanford Alpaca~\citep{alpaca} uses 52K instruction-following samples generated by GPT-3.5, while Vicuna~\citep{vicuna} uses around 500K high-quality instruction-following samples (150K conversions) between user and GPT~\citep{sharegpt}. To advance the SoTA of instruction-tuning for LLMs, \cite{peng2023instruction} uses GPT-4 as the teacher to generate the responses to the Alpaca instructions. Many follow-up works~\citep{zhang2023instruction}
improve the instruction-following data to enable the open LLMs  with better alignment quality in chat. For a comprehensive review, we refer the readers to a recent paper~\citep{wang2023far}, where a LLM Tulu is trained on a mix of several high-quality instruction data, and comprehensive comparisons are conducted across multiple benchmarks.

\paragraph{Quick assessment of LLM chatbots.} 
To study the quality of LLM Chatbots, we consider {\it Vicuna-Instructions-80}\footnote{\footnotesize \url{https://github.com/lm-sys/FastChat/blob/main/fastchat/eval/table/question.jsonl}}~\citep{vicuna}, a dataset with 80 questions that baseline models~\citep{touvron2023llama} 
find challenging. Besides generic instructions, the instructions fall into 8 categories, including knowledge, math, Fermi, counterfactual, roleplay, generic, coding, writing and common-sense.
To quantitatively compare the performance,  GPT-4 is used to rate the response from score 1 to 10 for any two given chatbots, then compute the relative score. 
Surprisingly, it turns out this evaluation metric is quite consistent across different settings. The open-source LLaMA family seems to perform closely to SoTA proprietary chatbots.

\paragraph{Further discussions.} There are several important topics on LLMs that we have not covered in this chapter, but are worthwhile future exploring.

\begin{itemize}[leftmargin=7.5mm]
\setlength{\itemsep}{2pt}
\item 
{\bf \it Data-centric AI}. We emphasize that the development of these open-source LLMs is data-centric~\citep{mazumder2022dataperf}, rather than model-centric, so that we hope the readers could align with this perspective when discussing the topic. As the training objectives and network architectures are becoming similar or even identical to GPT-like models, the key differential factor is data. For example, behaviors of the aforementioned LLMs are determined by the instruction tuning data. 

\item
{\bf \it False promise?} There is a debate on that the open LLMs could catch up with the proprietary LLMs is a false promise~\citep{gudibande2023false}. 
To align the discussions, we argue that there are two distinctive abilities for LLMs: the instruction-following ability to know which task to perform, and massive knowledge storage to complete the task with high quality. Imitation models are good at the former, by mimicking ChatGPT's style but perform poorly in terms of  factuality in their responses. In~\cite{gudibande2023false}, the authors conclude that there exists a substantial capabilities gap between open and closed LLMs that, with current methods, can only be bridged using an unwieldy amount of imitation data or by using more capable base LLMs. They also advocate that the highest leverage action for improving open-source models is to tackle the difficult challenge of developing better base LLMs. However, unfortunately, the resources to train such base LLMs are only available in a few industry labs. It seems more promising for most academic research labs to explore the opportunities in alignment research with affordable resources, or explore the techniques to reduce the compute barriers.

\item
{\bf \it Base LLMs}. Developing more capable or commercial usable LLMs is of great value. Besides LLaMA, the open-source community has developed variants of base LLMs such as LLaMA-2, OpenLLaMA~\citep{openlm2023openllama}, MPT~\citep{MosaicML2023Introducing} and Falcon~\citep{refinedweb}, or released the training recipe~\citep{together2023redpajama}.
\end{itemize}

\section{Instruction-Tuned Large Multimodal Models}
\label{sec:instruct_tuning_lmm}

In this section, we illustrate how to build the minimum prototype of multimodal GPT-4 with open-source resources. Specially, we use LLaVA~\citep{liu2023visual} as the running example, a similar idea is also proposed in its con-current work MiniGPT-4~\citep{zhu2023minigpt4}.


The research in the multimodal space has often been inspired by the latest advances in NLP in recent years. One successful recipe is to explore what would happen if the most intriguing and successful NLP ideas are borrowed for the vision-and-language community, for example, self-instruct. 
However, the unique challenge with self-instruct in multimodal research is that there is no strong multimodal teacher publicly available. Therefore, the research question becomes: how can we use language models such as language-only GPT-4 to create multimodal instruction following data.

\begin{figure}[h!]
\centering  
\vspace{-4mm}
\hspace{-2mm}
\begin{tabular}{l}
\includegraphics[width=1.00\textwidth]{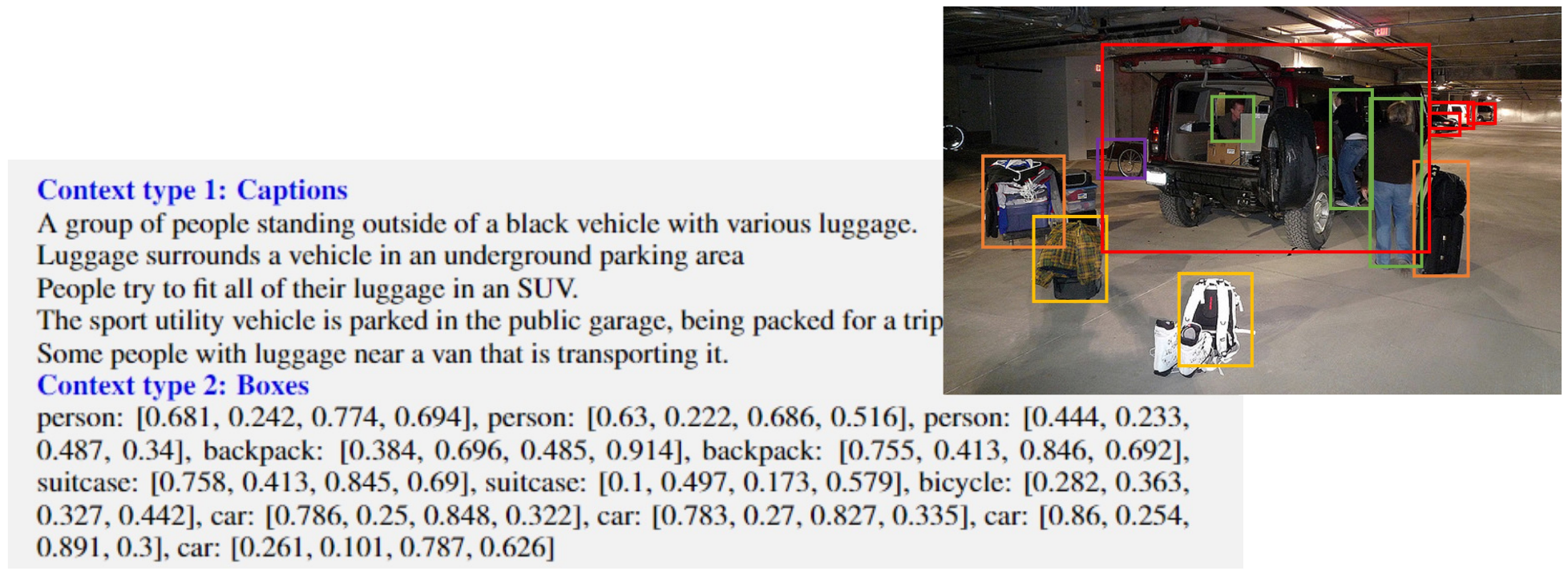} \\
(a) The sequence representation of the image data. \vspace{2mm}\\
\includegraphics[width=0.8\textwidth]{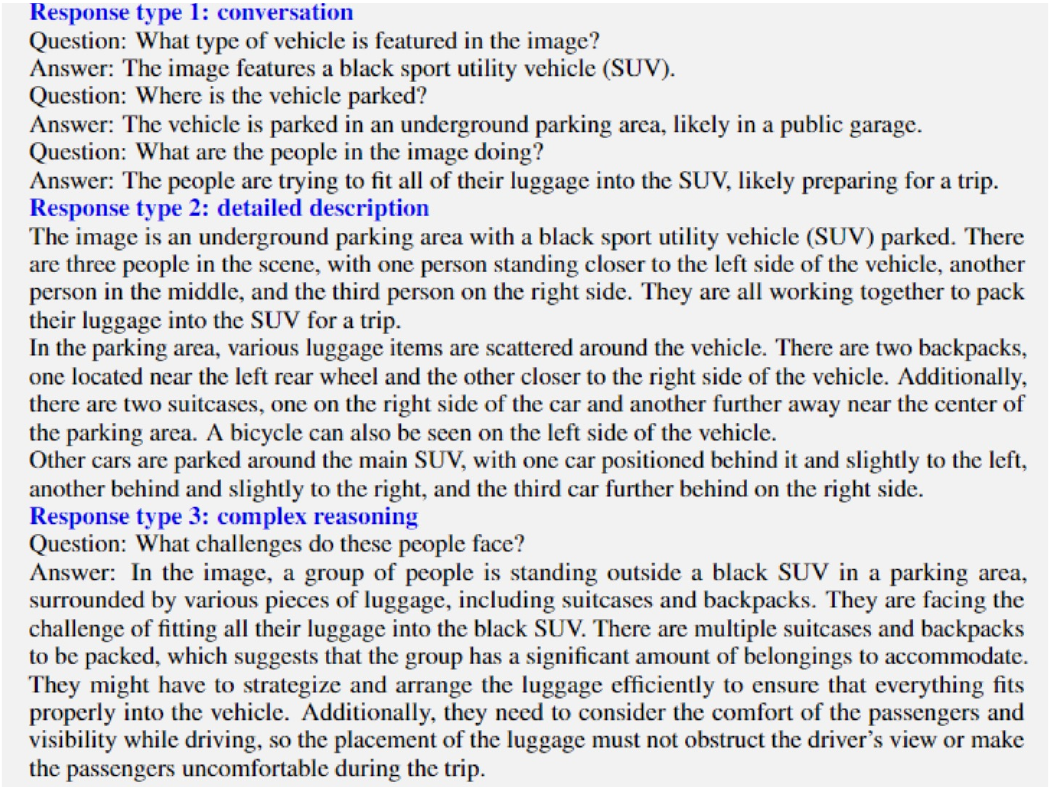} \\
(b) Three types of instruction-following data for the given image. \\
\end{tabular}
\vspace{-0mm}
\caption{Examples of multimodal instruction-following data. Image credit: ~\cite{liu2023visual}.}
\label{fig:multimodal_instruction_data}  
  \vspace{-1mm}
\end{figure}

\subsubsection{Data Creation}
Instead of directly feeding images into OpenAI GPT-4, we use their symbolic sequence representations shown in Figure~\ref{fig:multimodal_instruction_data} (a). In LLaVA, both captions and bounding boxes are considered, due to the following reasons: $(i)$ it is empirically found that GPT-4 can understand both well, in contrast to the poor performance of ChatGPT in understanding bounding box coordinates. $(ii)$  They are often complementary to each other and hence can represent the image as informative as possible.

As shown in Figure~\ref{fig:multimodal_instruction_data} (b), three types of instruction-following data are considered: $(i)$ multi-turn conversations so that users can chat with the model; $(ii)$ detailed description so that long-form responses can be generated from the model; and $(iii) $ complex reasoning, which is more about the implication of the image, rather than the image content. For example, \textit{``what challenge do these people face?''}, which requires to first recognize that the image is about a SUV in the parking area, and there are quite a few luggage placed on the ground, and then to infer that the challenge is how the luggage can be packed into the SUV due to the tight space of the trunk. In total, 158K samples are collected over three types.
To summarize, the spirit is that whatever tasks one wants the model to perform in the serving stage, it is important to create the corresponding instruction-following data for training.

\begin{figure}[t!]
\centering  
\vspace{-0mm}
\includegraphics[width=0.34\textwidth]{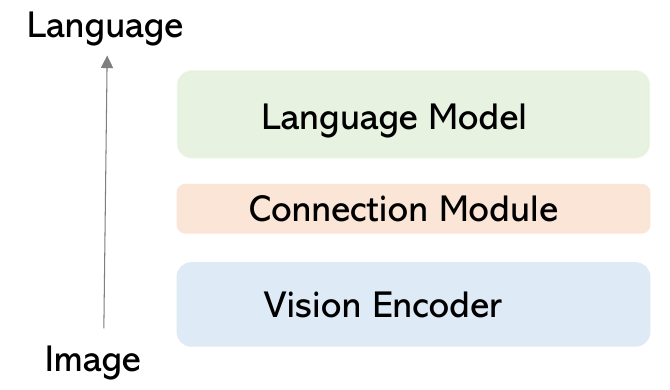} ~
\includegraphics[width=0.6\textwidth]{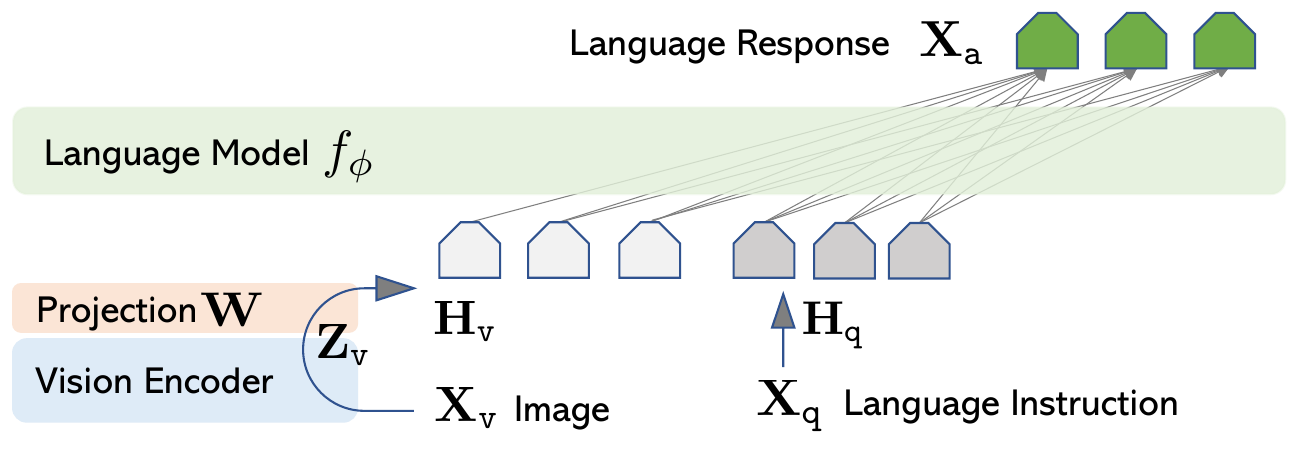} \\
\vspace{-0mm}
\caption{Network architecture: Left: General LMM; Right: LLaVA. Image credit: ~\cite{liu2023visual}.}
\label{fig:llava_arch}  
  \vspace{-1mm}
\end{figure}

\subsubsection{Network Architecture and Training}
As illustrated in Figure~\ref{fig:llava_arch}, the network architecture of LLaVA is an instantiation of the general image-to-text generative model framework introduced in Figure~\ref{fig:image2text} of Section~\ref{sec:background_lmm}. Specifically, LLaVa connects the pre-trained CLIP ViT-L/14 visual encoder~\citep{radford2021learning} and large language model Vicuna~\citep{vicuna}, via a simple projection matrix (\textit{i.e.}, the linear projection layer). A two-stage instruction-tuning procedure is adopted to train the model.
$(i)$ {\it Stage 1: pre-training for feature alignment}. Only the projection matrix is updated, based on a subset of CC3M~\citep{changpinyo2021conceptual}.
$(ii)$ {\it Stage 2: finetuning end-to-end}. Both the projection matrix and LLM are updated on the proposed multimodal instruction-following data for daily user-oriented applications.

\subsubsection{Performance}
\paragraph{Visual chat: towards building multimodal GPT-4 level chatbot.} LLaVA is finetuned on the generated multimodal instruction-following data, which contains a diverse set of task instructions and responses for daily user-oriented applications. It is empirically observed that finetuning the linear projection layer only is sufficient for the chat demo/scenarios, though it requires longer training time.
To evaluate the model performance,  an evaluation dataset named LLaVA-Bench is constructed, with two subsets: $(i)$ LLaVA-Bench (COCO): 30 unseen COCO images with 90 new language-image instructions, $(ii)$ LLaVA-Bench (In-the-Wild): 24 images with 60 questions. Each image can be associated with three types of instructions: conversation, detailed description and complex reasoning. The ground-truth answers are collected by manually re-writing GPT-4 output.
We test LLaVA and use language-only GPT-4 to rate their responses from score 1 to 10. Overall, LLaVA achieves 85.1\% relative score compared with ground-truth on LLaVA-Bench (COCO), and 73.5\% on LLaVA-Bench (In-the-Wild). On the latter dataset, Google Bard (July 19, 2023) and Microsoft BingChat (June 29, 2023) achieves 77.8\% and 71.5\%, respectively.
It indicates the effectiveness of the proposed self-instruct method in multimodal settings. One examples is shown in Table~\ref{tab:visual_example_chichken}.

\paragraph{Science QA: New SoTA with the synergy of LLaVA with GPT-4.}
LLaVA is finetuned on a multimodal reasoning dataset in the science domain~\citep{lu2022learn}.  LLaVA alone achieves 90.92\% in accuracy. We further explores with  language-only GPT-4 as the judge, to predict the final answer based on its own previous answers and the LLaVA answers. This ``GPT-4 as judge'' scheme yields a new SoTA of 92.53\%.

\paragraph{OCR in the wild: An emerging property.} LLaVA has never been explicitly trained on OCR data, \ie images that contains scene text that is described in the corresponding caption. Surprisingly, the model shows strong zero-shot OCR task transfer ability in the wild.

\begin{table}[t!]
  \begin{minipage}{1.0\linewidth}
\centering
\scalebox{0.70}{
\begin{tabular}{l p{1.1\linewidth} }
\toprule
 \multicolumn{2}{l}{\bf Visual input example, Chicken Nugget Map:}  \\
\midrule
&  \includegraphics[height=4.5cm]{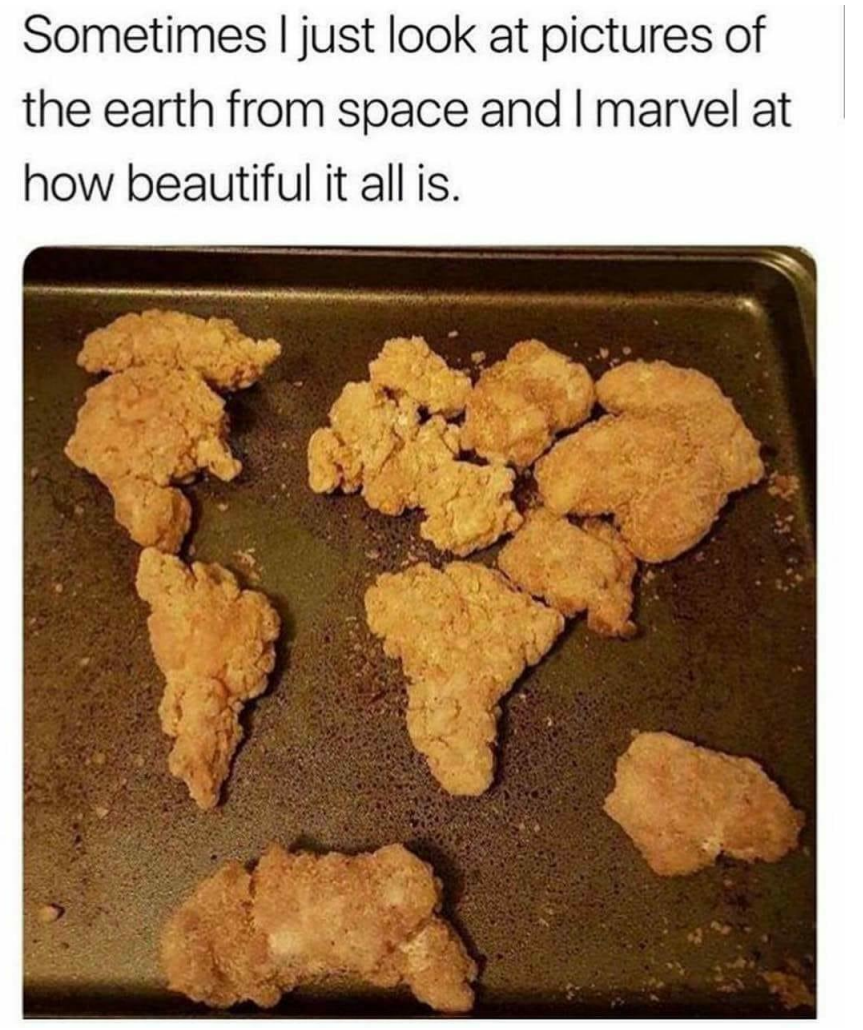} \\

User & Can you explain this meme in detail? \\
LLaVA & The meme in the image is a creative and humorous take on food, with a focus on chicken nuggets, as the center of the universe. The meme begins with the phrase "Sometimes I just look at pictures of the Earth from space and I marvel at how beautiful it all is..." The main image features a tray of cooked chicken nuggets, with the chicken pieces positioned to resemble the continents and islands on a world map.
The meme playfully suggests that the chicken nuggets represent the Earth, and the various locations depicted in the photo are actually chicken nugget versions of different places. The punchline of the meme appears in smaller text at the bottom, reading "I mean, it's not the real Earth, but how beautiful it is all is." This lighthearted and entertaining meme combines food and humor, using the image of chicken nuggets to create a funny and imaginative representation of the world.
\\
\midrule
User & Can you explain this meme? \\
GPT-4 &  This meme is a joke that combines two unrelated things: pictures of the earth from
space and chicken nuggets.
The text of the meme suggests that the image below is a beautiful picture of the
earth from space. However, the image is actually of chicken nuggets arranged to
vaguely resemble a map of the world.
The humor in this meme comes from the unexpected juxtaposition of the text and
the image. The text sets up an expectation of a majestic image of the earth, but the
image is actually something mundane and silly.
\\ \midrule
User & Can you explain this meme in detail? \\
Bing Chat & Sure, I can try to explain this meme in detail. This is a **chicken nugget meme**, which is a popular online meme that typically features a picture of a chicken nugget with a caption that is funny or relatable. The meme has been popular since the early 2000s, and continues to be one of the most popular memes today. This particular meme is based on a contrast between the text and the image. The text above the image reads "Sometimes I just look at pictures of the earth from space and I marvel at how beautiful it all is." This is a sentiment that many people can relate to, as looking at the earth from space can inspire awe and wonder. However, the image below the text shows a baking tray with chicken nuggets arranged in the shape of the world map. This is a humorous twist, as chicken nuggets are not usually associated with beauty or marvel, but rather with fast food or junk food. The image also implies that the person who made this meme loves chicken nuggets so much that they see them everywhere, even when looking at pictures of the earth from space. The meme is meant to be funny and absurd, and to poke fun at the person's obsession with chicken nuggets.I hope this explanation helps you understand this meme better. \\ \midrule
User & Can you explain this meme in detail? \\
Bard & Sure. The meme you sent me is a humorous take on the idea of looking at pictures of the earth from space and marveling at its beauty. The text in the meme says ``Sometimes I just look at pictures of the earth from space and I marvel at how beautiful it all is.'' However, the image shows a map of the world made out of chicken nuggets. This juxtaposition of the text and the image is what creates the humor in the meme. The text suggests that the speaker is looking at something beautiful and awe-inspiring, but the image shows something that is more mundane and humorous. The meme also plays on the fact that chicken nuggets are often seen as a junk food. By depicting a map of the world made out of chicken nuggets, the meme suggests that even something as mundane as chicken nuggets can be beautiful, or at least, more beautiful than we might think. Overall, the meme is a humorous way to poke fun at the idea of finding beauty in the mundane. It also shows how our perception of beauty can be influenced by our expectations. Here are some other interpretations of the meme: (1) The meme could be seen as a commentary on the way we consume images of the earth from space. We often see these images as representations of beauty and perfection, but the meme reminds us that the earth is also a messy and imperfect place. (2) The meme could also be seen as a critique of the fast food industry. The use of chicken nuggets to represent the world suggests that we are increasingly seeing the world through the lens of consumerism. Ultimately, the meaning of the meme is up to the interpretation of the viewer. However, it is clear that the meme is a humorous and thought-provoking way to explore the idea of beauty. \\   \midrule
User & Can you explain this meme in detail? \\
MiniGPT-4 & This meme is a picture of a pan full of fried chicken. The chicken is cooked and appears to be ready to be served. The caption reads, “Sometimes it’s just like looking at pictures from space and marveling at how beautiful it all is.” This meme is likely meant to be humorous and playful, as it compares the beauty of space to the simple pleasure of eating fried chicken. It is likely meant to be a lighthearted and funny way to express appreciation for both the wonders of space and the simple pleasures of food.
\\ \midrule
User & Can you explain this meme in detail? \\
BLIP-2 & sometimes i just look at pictures of the earth from space and marvel how beautiful it is
\\ \midrule
User & Can you explain this meme in detail? \\
OpenFlamingo & It's a picture of a chicken nugget on the International Space Station. 
\\
\bottomrule
\end{tabular}
}
\vspace{2mm}
\captionof{table}{Example prompt comparing LLaVA, GPT-4, BingChat, Bard, MiniGPT-4, BLIP-2, and OpenFlamingo's visual reasoning capabilities in understanding the humor. LLaVA and GPT-4 both explain the meme and its humor, while GPT-4 produces a more concise answer. Table credit: \cite{liu2023visual}.}
\label{tab:visual_example_chichken}  
  \end{minipage}
  \vspace{-0mm}
\end{table}

\section{Advanced Topics}
\label{sec:emerging_topics}

\begin{figure}[h!]
\centering  
\vspace{-3mm}
\hspace{-2mm}
\begin{tabular}{p{1.0\textwidth}}
\includegraphics[width=1.0\textwidth]{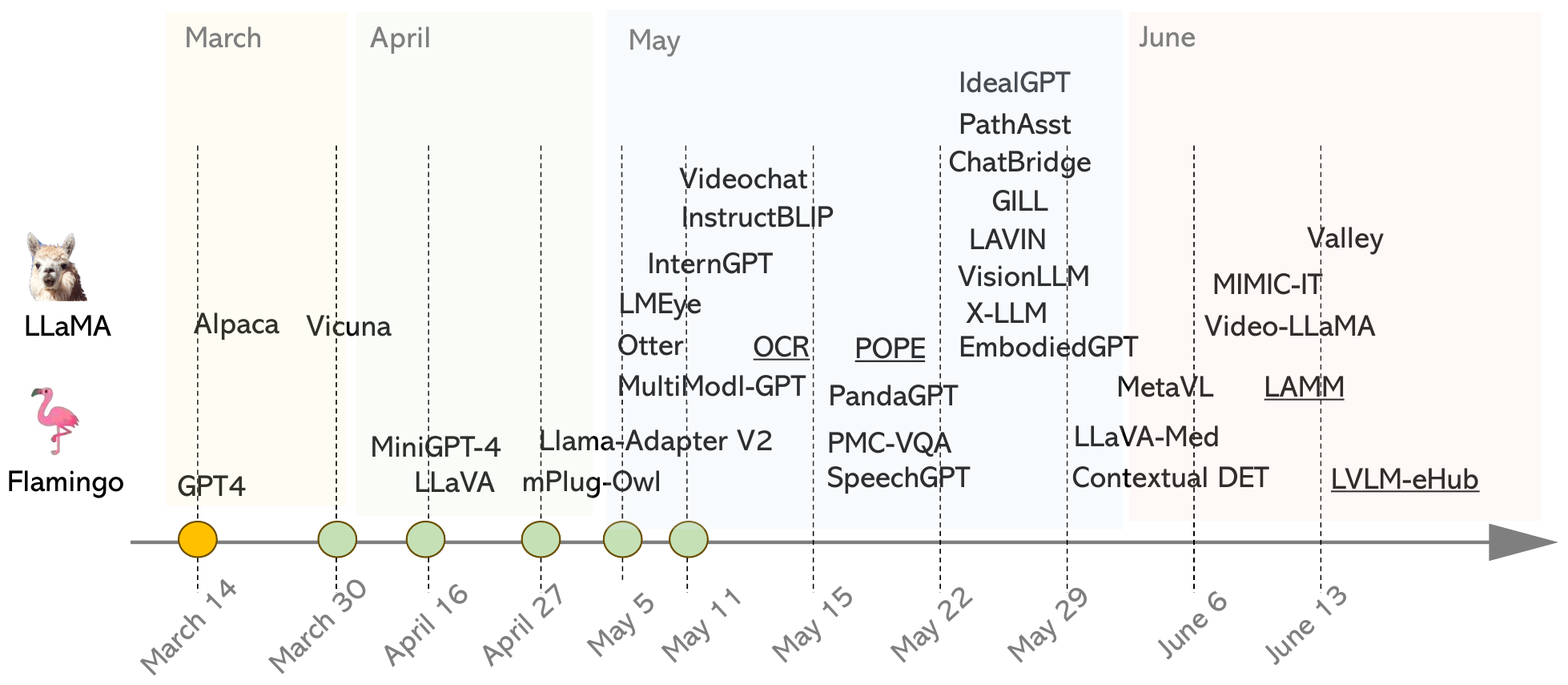} \\
(a) The surge of papers on LMMs 
from March 14, 2023 to June 19, 2023. Those with an underline indicate benchmarks, otherwise indicate models. \vspace{2mm}\\
\includegraphics[width=1.0\textwidth]{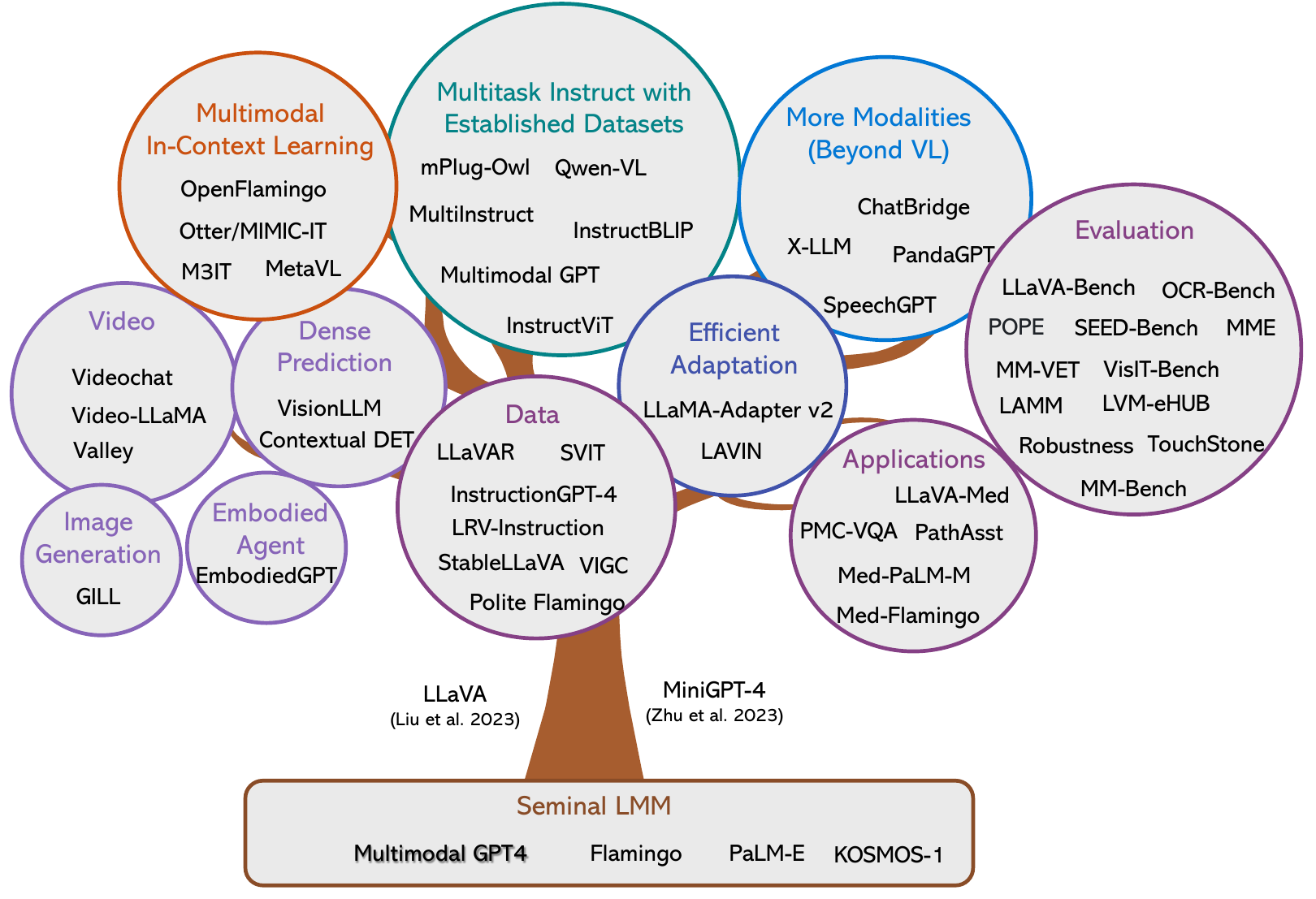} \\
(b) Summary and categorization of papers on LMMs. \\
\end{tabular}
\vspace{-0mm}
\caption{Review and summary for the emerged LMM literature. Due to space constraints, some methods are not displayed visually, but we aim to address them in the accompanying text.}
\label{fig:emgerging_topics}  
  \vspace{-1mm}
\end{figure}

The history of recent instruction-tuned LMMs are illustrated in Figure~\ref{fig:emgerging_topics} (a). 
Due to the popularity of ChatGPT and GPT-4,  instruction-tuned LMM appears as an emerging line of research in the past three months after GPT-4 was proposed. Alpaca~\citep{alpaca} and Vicuna~\citep{vicuna} were proposed to make LLaMA more instruction-following in the language domain in March.
In two weeks, MiniGPT-4~\citep{zhu2023minigpt4} and LLaVA~\citep{liu2023visual} were proposed to make Vicuna to see and chat about the visual world. In ten days, LLaMA-Adapter v2~\citep{gao2023llama} and mPlug-OWL~\citep{ye2023mplug} started to compare performance with MiniGPT-4/LLaVA, indicating the beginning of model evolution. The data points in April are relatively sparse.
In May, a large number of LMM papers appeared on arXiv, which improve this line of research from many different aspects. The momentum is till going in June.

It is easy to lose track of all the recent papers for the readers, so as well in our literature review. To better organize the literature, we group them based on specific research topics, shown in Figure~\ref{fig:emgerging_topics} (b). The early LMMs with billions of parameters include GPT-4~\citep{gpt4}, Flamingo~\citep{alayrac2022flamingo}, 
PaLM-E~\citep{driess2023palm} and KOSMOS-1~\citep{huang2023language}. In contrast to these proprietary LMMs, LLaVA and MiniGPT-4 open the opportunities to build LMMs with open-source resource.
We will discuss several topics as below, in addition to the extensions of RLHF~\citep{gunjal2023detecting}, dense prediction~\citep{wang2023visionllm,zang2023contextual,chen2023shikra}, video~\citep{zhang2023video,luo2023valley,li2023videochat}, image generation~\citep{koh2023generating} and embodied agent~\citep{mu2023embodiedgpt}.

\subsubsection{More Modalities (Beyond VL)}

While LMM extends LLM by adding the vision modality, it is natural to further extend the framework to include more modalities beyond vision and language. Following this spirit, several attempts have been made, including ChatBridge~\citep{zhao2023chatbridge}, PandaGPT~\citep{su2023pandagpt}, SpeechGPT~\citep{zhang2023speechgpt} and X-LLM~\citep{chen2023x}. PandaGPT leverages ImageBind to add more modalities into LMMs. The ImageBind model~\citep{girdhar2023imagebind} learns a single, shared representation space for text, image/video, audio and sensors that record depth (3D), thermal (infrared radiation), or inertial measurement units (IMU), which calculate motion and position. ImageBind provides a holistic understanding of the visual world that connects objects in a photo with how they will sound, their 3D shape, how warm or cold they are, and how they move. By training a projection layer for one modality in LMM, the model can zero-shot transfer to infer over other modalities, thanks to the shared multimodal embedding space. 
Another representative model is SpeechGPT, where language and speech modalities are enabled for both inputs and outputs. Despite of rich model variations, the idea to connect diverse modalities is similar to LMM that adds images into LLMs. 
NExT-GPT~\citep{wu2023nextgpt} connects an LLM with multimodal adaptors and different diffusion decoders, enabling NExT-GPT to perceive inputs and generate outputs in arbitrary combinations of text, images, videos, and audio.
The LMM framework has also been successfully extended to speech~\citep{zhao2023bubogpt}, 3D~\citep{wang2023chat,hong20233d}, and point cloud~\citep{xu2023pointllm}.

\subsubsection{Improving Visual Instruction Data Quantity and Quality}

Given the convergence of model architectures to GPT-like network, the performance of LMM is primarily determined by its training data. Therefore, it is cricial to improve the quantity and quality of visual instruction tuning data.  SVIT~\citep{zhao2023svit} follows the same data generation pipeline as in LLaVA, but further includes region description to prompt GPT-4, in addition to the caption and box data as shown in Figure~\ref{fig:multimodal_instruction_data} (a). The data is scaled up to 3.2 million, which is 20 times larger than the data used in LLaVA.

Unlike existing studies that primarily focus on positive instruction samples, LRV-Instruction~\citep{liu2023aligning} includes both positive and negative instructions for more robust instruction-tuning.
Other examples along this line include LLaVAR~\citep{zhang2023llavar} that adds OCR-related instruction-tuning data for text-rich image understanding, and StableLLaVA~\citep{li2023stablellava} that considers model-synthesized images for image-dialogue data. Polite Flamingo~\citep{chen2023visual} trains LLM to re-write the instruct data. Instead of leveraging GPT-4 for data generation, VIGC~\citep{wang2023vigc} considers to utilize LMM to generate instruction-tuning data and progressively enhance its quality on-the-fly.
Similar to the ``less is more'' observation in LIMA~\citep{zhou2023lima} from the NLP domain, InstructionGPT-4 
 shows that the quality of the instruction-tuning data is more important than its quantity, where they finetune a better version of MiniGPT-4 with 200 high-quality samples (6\%), selected from the 3500 samples used in the original MiniGPT-4.

\subsubsection{Multitask Instruct with Established Academic Datasets/Tasks}

As discussed earlier in Section~\ref{sec:instruct_tuning_llm}, instruction tuning in the language domains is implemented in two different ways: finetuning the model on a wide range of tasks using human-annotated prompts and feedback~\citep{ouyang2022training}, or supervised finetuning using public benchmarks and datasets augmented with manually or automatically generated instructions~\citep{wang2022benchmarking}. The former is good at user-oriented daily life tasks, and the latter is good at achieving decent performance on established benchmarks. LLaVA and MiniGPT-4 fall into the former class. Several other works either target for the latter class or combine both classes, including MultiInstruct~\citep{xu2022multiinstruct}, mPlug-OWL~\citep{ye2023mplug}, InstructBLIP~\citep{dai2023instructblip}, Multimodal-GPT~\citep{gong2023multimodal}, Instruction-ViT~\citep{xiao2023instruction} and Qwen-VL~\citep{bai2023qwen}.

\begin{figure}[t!]
\centering  
\vspace{-0mm}
\includegraphics[width=1.00\textwidth]{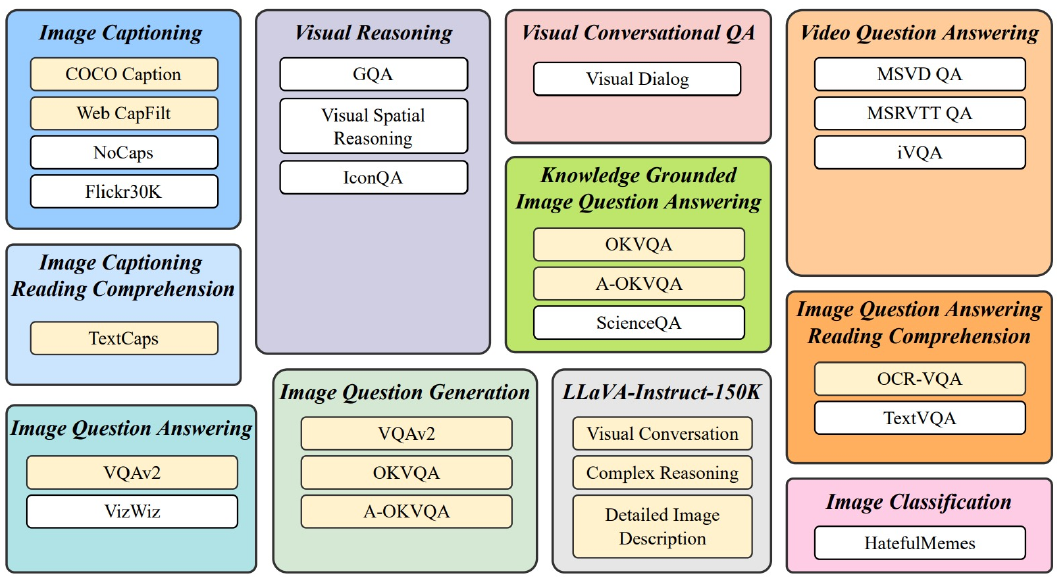} \\
\vspace{-0mm}
\caption{The vision-language tasks covered in InstructBLIP. Image credit:~\cite{dai2023instructblip}.}
\label{fig:instructblip_tasks}  
  \vspace{-1mm}
\end{figure}

For example, MultiInstruct is an early attempt before open-source LLaMA for instruction tuning with multimodal datasets. 
InstructBLIP is a recent work that combines chat and benchmark instruction-following data.
As shown in Figure~\ref{fig:instructblip_tasks}, InstructBLIP transforms 26 publicly available datasets, covering a wide variety of tasks and capabilities, into instruction tuning format. Trained on 13 held-in datasets, InstructBLIP attains SoTA zero-shot performance across all 13 held-out datasets, substantially outperforming BLIP-2 and larger Flamingo models. Qwen-VL scales up both image-text pair data for pre-traning and academic datasets for multi-task pre-traning, and achieve excellent performance on many tasks.

\subsubsection{Multimodal In-Context-Learning}

Similar to the behavior of LLMs, which can address a language task by processing examples of the task in their text prompt, multimodal in-context-learning refers to an visual and text interface that can steer the model towards solving a multimodal task. Given a few example pairs of visual inputs and expected text responses composed in the multimodal prompt, the model can be queried with a question about a new image or video, and then generate an answer. The direction to extend in-context-learning from language to multi-modalities has been explored, including  OpenFlamingo~\citep{anas_awadalla_2023_7733589}, Otter~\citep{li2023otter}, M$^3$IT~\citep{li2023m3it},  MetaVL~\citep{monajatipoor2023metavl} and Sparkles~\citep{huang2023sparkles}.

OpenFlamingo~\citep{anas_awadalla_2023_7733589} is an open source version of DeepMind's Flamingo model, trained on Multimodal C4 dataset~\citep{zhu2023multimodal}, which is a billions-scale corpus of interleaved image-text data. To explicitly enhance the multimodal in-context-learning ability of LMMs,  MIMIC-IT~\citep{li2023mimic} dataset is constructed, which is 2.4M multimodal instruction instances with in-context examples. By tuning OpenFlamingo on MIMIC-IT, a new model Otter is obtained with a stronger instruction-following ability. Using two image-text pairs as the context, Otter learns the concise answer style demonstrated by the examples, otherwise a tedious response is generated.

\subsubsection{Parameter-Efficient Training}

While finetuning very large models often leads to high performance, it is prohibitively expensive; For example, regular 16-bit finetuning of a LLaMA-65B model~\citep{touvron2023llama}  requires more than 780 GB of GPU memory~\citep{dettmers2023qlora}. Therefore, it is critical to reduce the memory footprint of LLMs/LMMs, especially when it comes to improve the accessibility of large models to a wider community. 

Parameter-efficient training is an effective approach for LMM adaptation. It freezes most of the model parameters, and only allows a fraction of trainable parameters to update with domain-specific data. For example, LLaMA Adapter v2~\citep{gao2023llama} and LAVIN~\citep{luo2023cheap} only have 14M and 3.8M trainable parameters, compared with 7B/13B LLM parameters, respectively.
Another efficient training method is quantization. The recent QLoRA~\citep{dettmers2023qlora} finetunes 65B LLaMA for 24 hours on a single GPU, achieving 99.3\% of the performance level of ChatGPT. Since instruction tuning typically involves a small amount of data, it makes parameter-efficient training or model quantization the practical approach, especially when with limited GPU resources. Both LoRA~\citep{hu2021lora} and QLoRA are supported in LLaVA codebase to allow LMM training with less GPUs. It is empirically shown in~\cite{lu2023empirical} that LoRA/QLoRA can achieve similar performance with full-modal tuning when scaling LLaVA to 33B and 65B, when training with around 150K instruct data and evaluating with LLaVA-Bench.

\subsubsection{Benchmarks}

While LMMs have shown excellent visual recognition and reasoning in an open-set manner with free-form text across many scenarios,
the evaluation of LMMs is becoming an urgent and challenging problem. Several related benchmarks have been developed to evaluate various aspects of LMMs, ranging from their specific  abilities including OCR~\citep{liu2023hidden}, hallucination (POPE~\citep{li2023evaluating} and HaELM~\citep{wang2023evaluation}) and adversarial robustness~\citep{zhao2023evaluating}, to comprehensive evaluation such as LAMM~\citep{yin2023lamm}, LVLM-eHub~\citep{xu2023lvlm}. 
We summarize the LMM evaluation benchmarks in Table~\ref{tab:lmm_benchmarks}. Among them, LLaVA-Bench is the first attempt to designed open-world visual chat benchmark specifically  for LMM.
Recently, early multimodal experiments have been conducted to compare open-source LMM with commercial ones such as BingChat and Bard and LLaVA-Bench~\citep{liu2023visual} and LVLM-eHub~\citep{shao2023tiny}.

\begin{table}[h!]
\centering
\scalebox{0.75}{
\begin{tabular}{ p{3.1cm} p{5.9cm} p{3.8cm}p{2.8cm}}
\toprule
Benchmark& Capability to Evaluate & Statistics & Metric \\
\midrule
LLaVA-Bench ~~~~~~\citep{liu2023visual} & Multi-turn QA, detailed description, reasoning  & Two subsets: 90 samples on COCO and 60 samples on In-the-Wid & Relative score via GPT-4 evaluation\\ \midrule
OCR-Bench ~~~~~~~~~~ \citep{liu2023hidden} &  Zero-shot OCR & A suite of 23 OCR-related academic tasks & Accuracy \\ \midrule

MMBench ~~~~~~~~~~~~~~~ \citep{liu2023mmbench} &  Perception (coarse, fine-grained 
single-instance and cross-instance) and Reasoning (attribute, relation, logic) & 2974 multiple-choice samples in 20 ability
dimensions &   Circular evaluation via ChatGPT answer extraction \\ \midrule

M3Exam ~~~~~~~~~~~~ \citep{zhang2023m3exam} & Multilingual, multimodal, and multi-level assessment & 12,317 questions in 9 languages, with 2,816
questions involving images &  Accuracy on multiple-choice questions \\ \midrule

MME ~~~~~~~~~~~~~~~~~~~~~~~~~~~~~~ \citep{fu2023mme} & Perception and Cognition &  
14 tasks & Accuracy on “yes” or “no”\\ \midrule

LAMM ~~~~~~~~~~~~~~~~~~~~~~ \citep{yin2023lamm} & Various 2D/3D vision tasks & 9 image task with 62K
samples, and 3 point cloud tasks with 12K samples & Traditional CV task metrics \\ \midrule
LVLM-eHub ~~~~~~~~~ \citep{xu2023lvlm} &  six  multimodal capabilities such as VQA and embodied AI &   47 standard text-related visual benchmarks & CIDEr and accuracy; Arena with human judgment\\ \midrule

SEED-Bench ~~~~~~~~~~~~~ \citep{li2023seed} & Comprehension of both the image and video modality & 19K multiple choice questions in 12 dimensions & Accuracy on multiple-choice questions  \\ \midrule

VisIT-Bench ~~~~~~~~~~ \citep{bitton2023visitbench} & Real-life vision-language instructions & 592 samples in 70 tasks & Elo, matches\\ \midrule

MM-VET ~~~~~~~~~~~~~~~~~ \citep{yu2023mm}  & Integrated capabilities in recognition, OCR, spatial, knowledge, math, language & 200 samples & GPT-4 evaluation \\ \midrule

TouchStone ~~~~~~~~~~~~~~ \citep{bai2023touchstone} & Five  abilities: basic description, visual recognition, visual comprehension, visual storytelling, and multi-image analysis & 908 dialogues in 27 tasks & GPT-4 evaluation \\ \midrule

SciGraphQA ~~ \citep{li2023scigraphqa}  & Scientific graph question-answering & 3K test samples & CIDEr, BLEU-4, and ROUGE\\ 

\bottomrule
\end{tabular}
}
\vspace{1mm}
\caption{Comparisons of recently proposed LMM evaluation benchmarks.}
\vspace{-6mm}
\label{tab:lmm_benchmarks}
\end{table}

\begin{figure}[h!]
\centering  
\vspace{-0mm}
\includegraphics[width=1.00\textwidth]{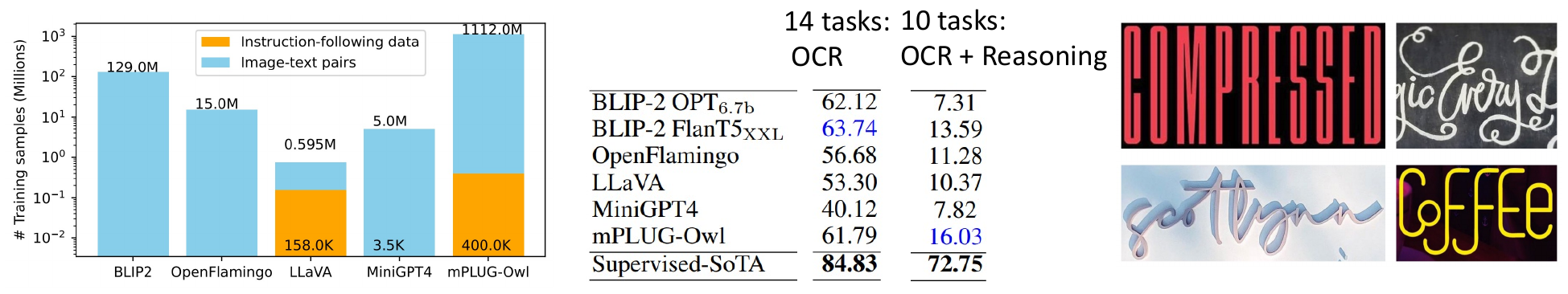} \\
\vspace{-0mm}
\caption{Zero-shot OCR performance of LMMs on 24 datasets. Image credit:~\cite{liu2023hidden}.}
\label{fig:ocr_bench}  
  \vspace{-1mm}
\end{figure}

It is surprising that LMMs shows strong zero-shot OCR performance in the wild, without explicitly training on text recognition data. To shed light on the hidden mystery of OCR in LMMs, a comprehensive empirical study is conducted in~\cite{liu2023hidden} to compare open-source LMMs on 24 academic text recognition datasets, shown in Figure~\ref{fig:ocr_bench}. Three observations are highlighted: 
$(i)$ LLaVA consistently outperforms MiniGPT-4 on 21 out of 24 datasets, despite that the training data in LLaVA is an order of magnitude smaller. $(ii)$ Training with significantly more training data leads to higher OCR performance, as demonstrated by BLIP2~\citep{li2023blip} and mPLUG-Owl. $(iii)$ In most cases, supervised SoTA results significantly outperform zero-shot LMM. However, it is worth noting that in the WordArt dataset~\citep{xie2022understanding}, which primarily features challenging artistic text, BLIP2 surpasses supervised SoTA. This reveals the potential of LMM in recognizing more complex text types.

\subsubsection{Applications}

The success of ChatGPT/GPT-4 in the general domain has inspired the interests in building assistants in the vertical domains such as medicine, gaming and education. Such domain-specific assistants can have the several advantages over the general domain counterpart: $(i)$  training with high-quality domain-speicifc data makes the assistants more helpful; $(ii)$  the model size can be smaller, with lower severing cost; and $(iii)$  the sensitive user prompt data can be maintained internally by serving the model locally, to avoid privacy issue.

\begin{figure}[t!]
\centering  
\vspace{-0mm}
\includegraphics[width=1.00\textwidth]{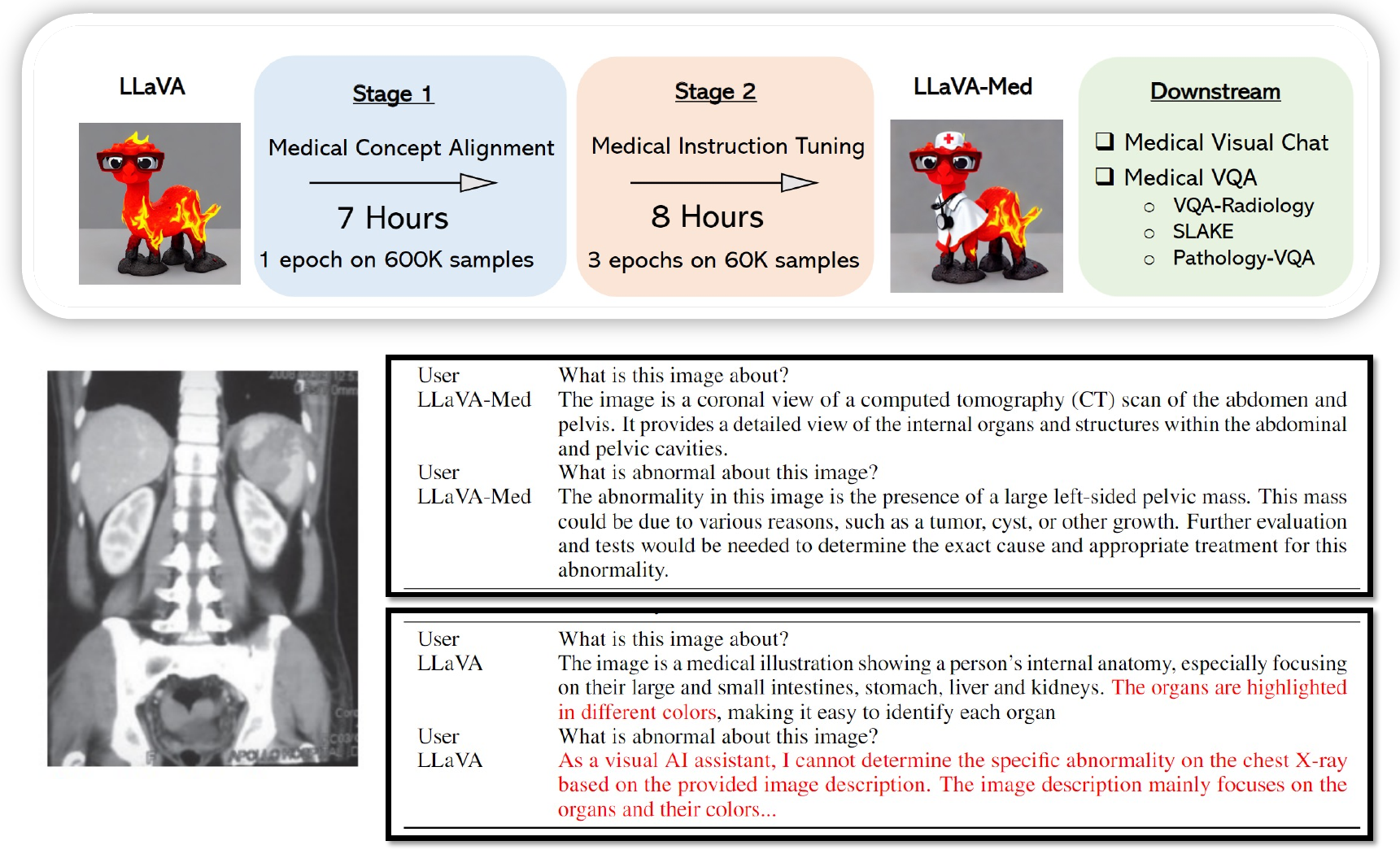} \\
\vspace{-0mm}
\caption{Application of LMMs to biomedical images. Top: Domain adaptation from LLaVA to LLaVA-Med. Bottom: The chat behaviors of two chatbots. Image credit: ~\cite{li2023llava}.}
\label{fig:llava_med}  
  \vspace{-1mm}
\end{figure}

To improve text recognition ability of LMM, OCR-specific models have been developed, including BLIVA~\citep{hu2023bliva}, LLaVAR~\citep{zhang2023llavar}, mPlug-DocWL~\citep{ye2023mplugdocowl}.
LMMs have been recently explored in the biomedical domain~\citep{sun2023pathasst,zhang2023pmc,li2023llava}, where conversational generative AI has demonstrated remarkable promise for empowering biomedical practitioners.
LLaVA-Med~\citep{li2023llava} is a cost-efficient approach for training a vision-language conversational assistant that can answer open-ended research questions about biomedical images. The key idea is to leverage a large-scale, broad-coverage biomedical figure-caption dataset extracted from PubMed Central, use GPT-4 to self-instruct open-ended instruction-following data from the captions, and then finetune a large general-domain vision-language model LLaVA using a novel curriculum learning method. Specifically, the model first learns to align biomedical vocabulary using the image-caption pairs as is, then learns open-ended conversational semantics using GPT-4 generated instruction-following data, broadly mimicking how a layperson gradually acquires biomedical knowledge. In Figure~\ref{fig:llava_med}, we provide examples of the biomed visual conversations with different chatbots. LLaVA-Med precisely answers the questions requiring biomedical knowledge, while LLaVA behaves like a layperson, that hallucinates based on commonsense.
LLaVA-Med has inspired several generalist biomedical AI models, including Google Med-PaLM-M~\citep{tu2023towards}, Stanford Med-Flamingo~\citep{moor2023med} and radiology generalist \citep{wu2023towards}.

\section{How Close We Are To OpenAI Multimodal GPT-4?}
\label{conclusions_lmm}

With all the works mentioned above, are we close to (or, even surpassing) OpenAI Multimodal GPT-4? It is encouraging to see that the open-source community has quickly developed a variety of models and prototypes for various new capabilities. For example, LLaVA/Mini-GPT4 paves the way towards building multimodal chatbots, with some examples that reproduce the results in OpenAI GPT-4 technique report; CM3leon~\citep{yu2023CM3Leon}, Emu~\citep{sun2023generative}, GILL~\citep{koh2023generating} extends LMMs for end-to-end image generation, to the best of our knowledge, this is a capability that the current GPT-4 does not exhibit. From the perspective of enabling new capabilities with the minimum prototypes, the open-source community seems close to OpenAI Multimodal GPT-4, by exploring the baby steps towards building the general-purpose multimodal assistant.


However, there is still a clear large gap in terms of scaling a given capability, \emph{e.g.}, for the visual reasoning capability that we have observed in LLaVA. There are two more visual examples from OpenAI technical report, to correctly answer the questions, it requires models to understand multiple high-resolution images and long sequence text depicted in the image, as well as responding with domain knowledge. It requires much more compute and more powerful language models, which are not available to most people.

In summary,  we have presented the background and strong capabilities of LMM, reviewed instruction tuning in LLMs, and showed how to build a prototype such as LLaVA and MiniGPT-4 using open-source resources. We also summarized the most recent papers emerged on this line of research to help those who are interested to gain the momentum to start the journey of LMM research.
To discuss the next steps to work on as a community, one sustainable suggestion can be that those with resources can continue focusing on the scaling success and study new emerging properties, while others focus on prototypes for new functionalities and evaluation, as well as developing techniques to reduce the computational barriers and thus allow easier accessibility to large models.
\chapter{Multimodal Agents:\\ Chaining Tools with LLM}
\label{chp:chaining_with_llm}
\begin{wrapfigure}{r}{4.0cm}
  \centering
  \vspace{-6cm}
  \includegraphics[width=0.97\linewidth]{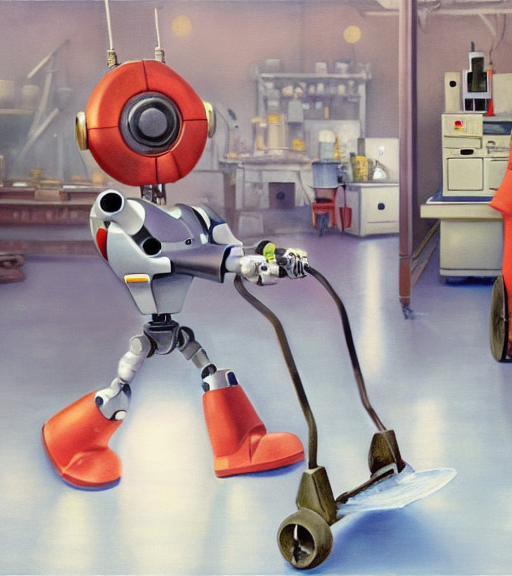}
\end{wrapfigure}

Large Language Models (LLMs)~\citep{chowdhery2022palm,gpt4} have shown intriguing properties 
generalizing to user prompts in various domains,
and rapidly adapting to new scenarios, using in-context  learning with a few examples.
Inspired by such strong capabilities, researchers are now exploring a new modeling paradigm that shifts from standalone models for solving finite, pre-defined problems, into synergistically chaining multiple tools or experts with LLMs to solve complicated, open problems. Unlike what has been introduced in Chapter~\ref{chp:training_with_llm}, such a system can be built without any training involved, just by using a few demonstration examples to teach the LLM to generate proper calling to existing tools. 

In this chapter, we review the fast-evolving literature on chaining different multimodal experts with LLMs to solve complicated multimodal understanding problems, referred to as {\it multimodal agents}. We start with an overview on the evolution of this modeling paradigm in Section~\ref{chp6_sec:overview}, highlighting the differences between traditional approaches and the new modeling paradigm of chaining tools with LLM. Section~\ref{chp6_sec:model} gives a general overview of multimodal agents. Pivoting on an exemplary multimodal agent \textsc{MM-ReAct}~\citep{yang2023mmreact}, Section~\ref{chp6_sec:mmreact} comprehensively reviews how to build a multimodal agent, its emerging capabilities in multimodal understanding, and how it can be easily extended to incorporate the latest and strongest LLM and potentially millions of tools.  Finally, in Section~\ref{chp6_sec:advanced_topics}, we end the chapter with discussions on advanced topics, such as how to improve/evaluate multimodal agents, the diverse applications powered by multimodal agents.

\section{Overview}\label{chp6_sec:overview}
We first revisit the evolution of modeling paradigms, from task-specific models to the most recent large multimodal models, which all require data curation and model training. We then introduce the new modeling paradigm of chaining tools with LLM, which may not require any training, but instead directly takes advantage of a pre-trained LLM and existing tools that are widely available through open-source platforms or APIs. 

\begin{figure}[htb!]
\centering  
\vspace{-0mm}
\includegraphics[width=1.00\textwidth]{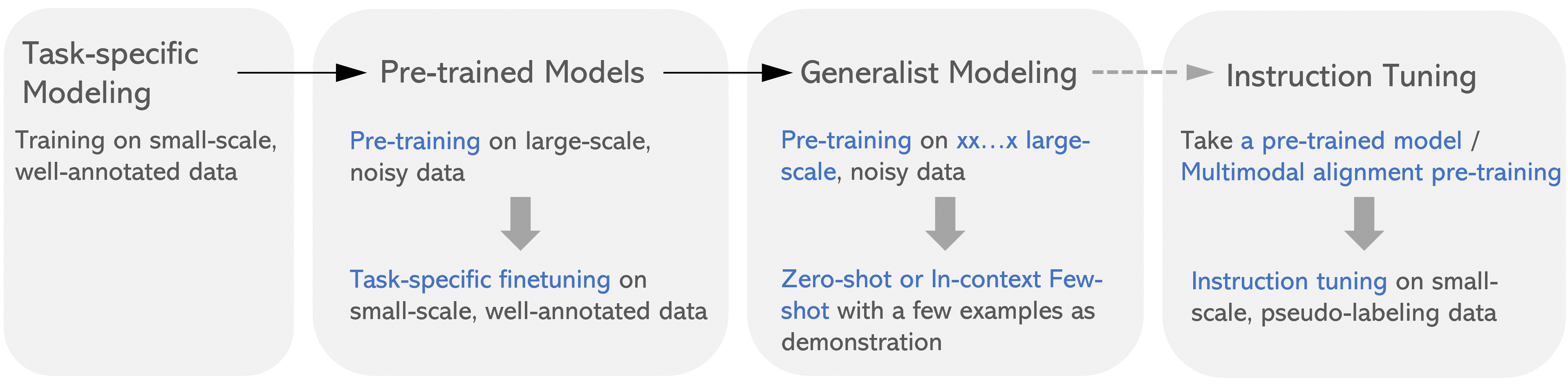} \\
\vspace{-0mm}
\caption{Evolution of modeling paradigm.}
\label{fig:chp6_model_evolution}  
  \vspace{-1mm}
\end{figure}

\paragraph{Evolution of modeling paradigm.}  As summarized in Figure~\ref{fig:chp6_model_evolution}, we are witnessing the transition from task-specific models towards general-purpose assistants across language, vision, and multimodal research. 

We started with \textbf{task-specific models} that are trained on small-scale well-annotated data. This results in dedicated models~\citep{anderson2018bottom,li2019relation,yu2019deep} for each task or even each dataset. 

We then transitioned to the phase of \textbf{pre-trained models}, 
with the pretrain-then-finetune paradigm widely adopted across both NLP and vision-language (VL) research. During pre-training, the model can take advantages of large-scale, web-crawled noisy data, for example, millions to billions of image-text pairs~\citep{chen2020uniter,wang2022git}, or billions of text tokens~\citep{devlin2018bert,liu2019roberta}. However, it is still mostly task-specific finetuned, requiring similarly small-scale, well-annotated data as the ones used in training task-specific models. This paradigm has led to many well-known models, such as BERT~\citep{devlin2018bert}, RoBERTa~\citep{liu2019roberta} in NLP, and UNITER~\citep{chen2020uniter}, OSCAR~\citep{li2020oscar} in VL. These early VL foundation models were considered to be large-scale (trained with 10M image-text pairs), but may be of intermediate or even small size in today’s view (billions of pairs). 

Nowadays, we are entering a new era of \textbf{generalist modeling}, 
where the pre-training has been further scaled up to trillions of text tokens~\citep{gao2023llama}. For downstream adaptation, these generalist models have shown strong performance with in-context few-shot learning on a few demonstration examples, or even zero-shot evaluation. These models are what we now refer as large language/multimodal models, including the GPT family~\citep{chatgpt,gpt4}, PaLM family~\citep{chowdhery2022palm,driess2023palm}, LLaMa~\citep{touvron2023llama}, Flamingo~\citep{alayrac2022flamingo}.

Based on the generalist models, the pipeline of building \textbf{instruction-following models} 
covered in Chapter~\ref{chp:training_with_llm}, similarly follows the pretrain-then-finetune paradigm. For example, Alpaca~\citep{alpaca}, is built on top of the pre-trained LLaMa~\citep{touvron2023llama}, then finetuned on a smaller-scale instruction tuning dataset. Similarly, for instruction-following VL models (\eg~LLaVA~\citep{li2023llava}), an additional stage of image-text alignment pre-training is introduced to align the visual representations to the frozen LLM first, followed by visual instruction tuning. 

\paragraph{New modeling paradigm: chaining tools with LLM.}
LLMs~\citep{brown2020language,chowdhery2022palm,gpt4} have demonstrated exceptional
abilities to tackle new tasks with only a few
examples or textual instructions, showing the promise of serving as general-purpose foundations for many applications. Despite being versatile and impressive, they  encounter challenges with the basic functionalities, such as mathematical reasoning and information retrieval. Furthermore, a fundamental limitation of not only LLMs but also other large-scale models nowadays, is that they only represent the world described by their training data, which will inevitably become outdated over time. Regularly re-training the model with the latest information is simply not feasible.

Meanwhile, many tasks with real-world impact cannot be readily tackled by by LLMs alone. For example, accessing up-to-date information and performing computations, can be done via existing tools (\textit{\eg}, search engine or calculator). Hence, recent research in language modeling has explored a new modeling paradigm by supplementing LLMs with external NLP tools~\citep{nakano2021webgpt, huang2022language, ahn2022can}, including calculators, search engines, translation systems, calendars, or even API calls on other models. 

The above studies mainly focus on a single modality, \textit{i.e.}, language, in which the output of the tools are in text format, thereby can naturally be fed into LLMs as additional knowledge. However, we live in a multimodal world and a truly intelligent agent should be able to perform advanced multimodal reasoning and actions.
How to enable LLMs with perception of multimodal signals via tool using, is the focus of the remaining part of this chapter.

\section{Multimodal Agent}\label{chp6_sec:model}

\begin{figure}[t!]
\centering  
\vspace{-0mm}
\includegraphics[width=\textwidth]{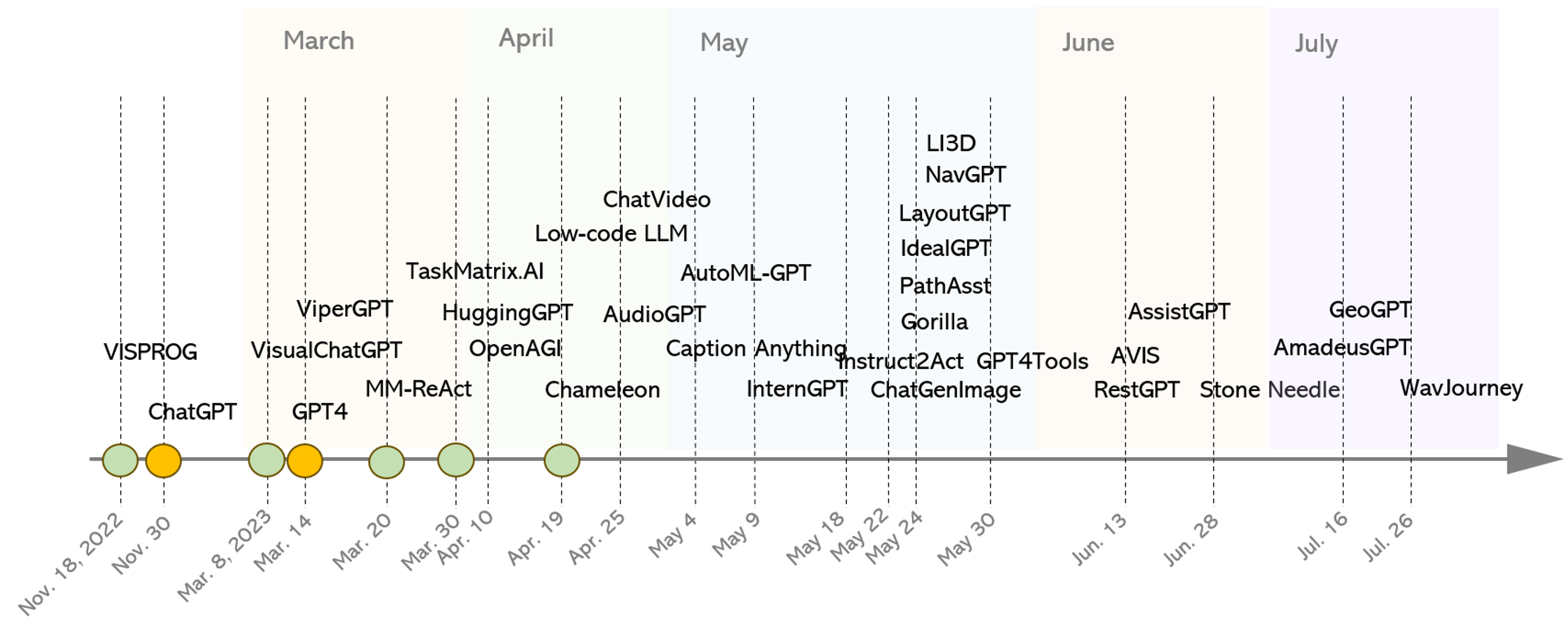} \\
\vspace{-0mm}
\caption{The surge of multimodal agents from November 18, 2022 to July 26th, 2023.}
\label{fig:chp6_timeline}  
  \vspace{-1mm}
\end{figure}

There are several representative works on building multimodal agent with tool use of vision experts, including VISPROG~\citep{Gupta2022VisProg},  Visual ChatGPT~\citep{wu2023visual} and MM-ReAct~\citep{yang2023mmreact}. VISPROG is the very first work on using programming language to chain different vision tools with a LLM. Visual ChatGPT enables dialogue-based image editing by complementing ChatGPT~\citep{chatgpt} with various image generation tools. MM-ReAct shows that when collaborating various advanced vision experts, ChatGPT can perform complex multimodal actions and reasoning.  
Figure~\ref{fig:chp6_timeline} presents the fast-evolving literature in multimodal agents from November 18, 2022 to July 26th, 2023. Among them, we include a few more exemplary multimodal agents in Table~\ref{tab:chp6_agent_glossary}, along with two representative works in the NLP domain.

 \begin{table*}[!t]
\resizebox{1.0\textwidth}{!}
{
  \begin{tabular}{l|ccccc}
     \multirow{2}{*}{\bf Model}  & \multirow{2}{*}{\bf LLM} & \multirow{2}{*}{\bf Tools} & \bf Tool & \bf Tool & \multirow{2}{*}{\bf Multimodal}\\
     & & & \bf Size & \bf Execution &   \\
     \hline
     ART~\citep{paranjape2023art} & GPT-3 &
     \includegraphics[width=12pt,valign=c]{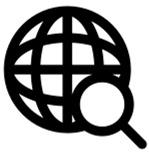}
     \includegraphics[width=12pt,valign=c]{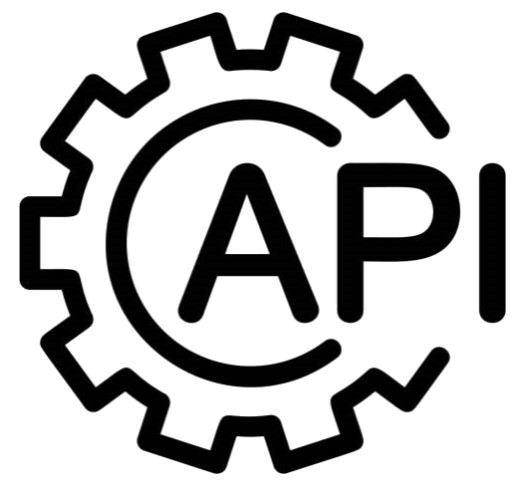}
      \includegraphics[width=12pt,valign=c]{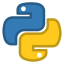}
      & 3 & Program & \ding{55}\\
     Toolformer~\citep{schick2023toolformer} & GPT-J & 
      \includegraphics[width=12pt,valign=c]{figs/chp6_icons/search_engine.png}
     \includegraphics[width=12pt,valign=c]{figs/chp6_icons/api.jpg}
     & 5 & Natural language & \ding{55} \\
     \hline
     VISPROG~\citep{Gupta2022VisProg} & GPT-3 & 
     \includegraphics[width=12pt,valign=c]{figs/chp6_icons/python.png} 
     \includegraphics[width=12pt,valign=c]{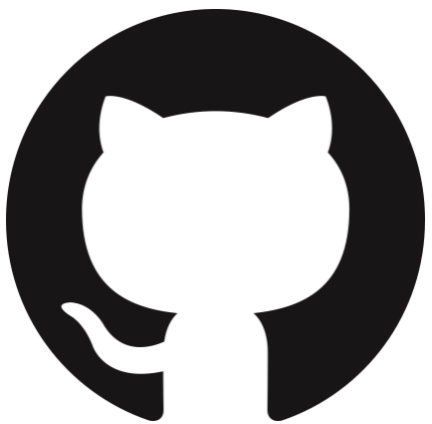}
     & $>$ 10 & Program & \ding{51} \\
     Visual ChatGPT~\citep{wu2023visual} & ChatGPT & 
     \includegraphics[width=12pt,valign=c]{figs/chp6_icons/GitHub-Mark.png}
     & $>$ 10 & Natural Language & \ding{51} \\
     ViperGPT~\citep{suris2023vipergpt} & GPT-3 Codex & 
     \includegraphics[width=12pt,valign=c]{figs/chp6_icons/api.jpg}
     \includegraphics[width=12pt,valign=c]{figs/chp6_icons/GitHub-Mark.png}
     \includegraphics[width=12pt,valign=c]{figs/chp6_icons/python.png}
     & $>$ 10 & Program & \ding{51}\\
     MM-ReAct~\citep{yang2023mmreact} & ChatGPT/GPT-4 & 
     \includegraphics[width=12pt,valign=c]{figs/chp6_icons/search_engine.png}
     \includegraphics[width=12pt,valign=c]{figs/chp6_icons/api.jpg}
     & $>$ 10 & RegExp Match & \ding{51} \\
     HuggingGPT~\citep{shen2023hugginggpt} & ChatGPT & 
     \includegraphics[width=12pt,valign=c]{figs/chp6_icons/GitHub-Mark.png}
     & $>$ 10 & Natural Language & \ding{51} \\
     Chameleon~\citep{lu2023chameleon} & GPT-4 & 
     \includegraphics[width=12pt,valign=c]{figs/chp6_icons/search_engine.png} 
     \includegraphics[width=12pt,valign=c]{figs/chp6_icons/GitHub-Mark.png}
     \includegraphics[width=12pt,valign=c]{figs/chp6_icons/python.png}
     & $>$ 10 & Natural Language & \ding{51} \\
  \end{tabular}
  }
  \caption{Glossary of representative works on chaining tools with LLMs. The LLMs used in these works include GPT-3~\citep{brown2020language}, GPT-J~\citep{gpt-j}, ChatGPT~\citep{chatgpt}, GPT-3 Codex~\citep{chen2021evaluating} and GPT-4~\cite{gpt4}. \includegraphics[width=12pt,valign=c]{figs/chp6_icons/search_engine.png}: search engine. \includegraphics[width=12pt,valign=c]{figs/chp6_icons/python.png}: code. \includegraphics[width=12pt,,valign=c]{figs/chp6_icons/api.jpg}: APIs other than search engine. \includegraphics[width=12pt,valign=c]{figs/chp6_icons/GitHub-Mark.png}: open-source models.
  }
  \label{tab:chp6_agent_glossary}
\end{table*}

\begin{figure}[tb!]
\centering  
\vspace{-0mm}
\includegraphics[width=.98\textwidth]{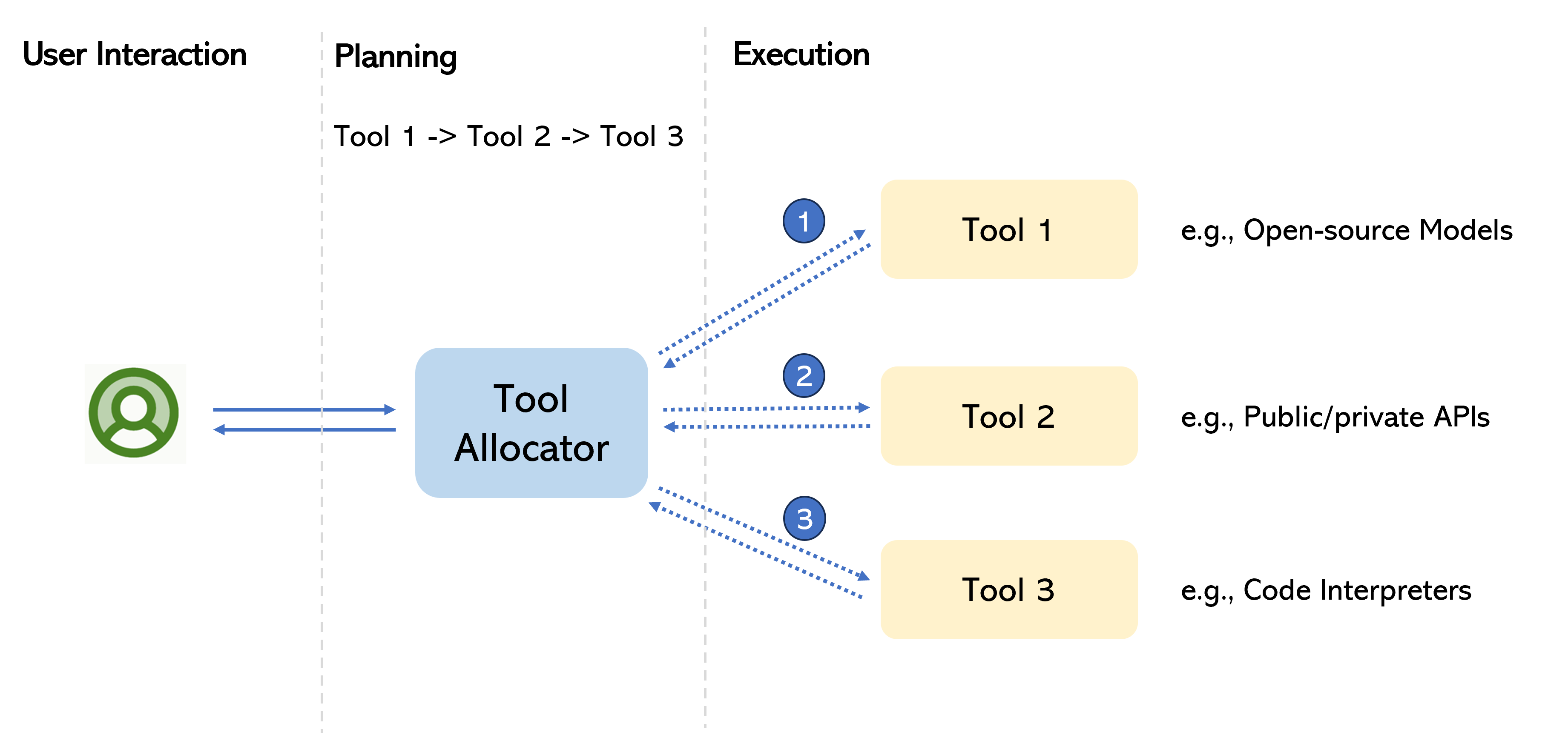} \\
\vspace{-0mm}
\caption{An overview of multimodal agent framework.}
\label{fig:chp6_new_modeling}  
  \vspace{-1mm}
\end{figure}

An overview of a typical multimodal agent framework is illustrated in Figure~\ref{fig:chp6_new_modeling}. The user directly interacts with the \textbf{Tool Allocator}, which functions as the brain of the agent. In current literature, the tool allocator is usually a LLM. To achieve the user's goal, the LLM will outline all the steps necessary with either a single tool or collaborating multiple tools together.  
Subsequently, it will retrieve from all the candidate tools for the needed tools, and execute possibly multiple
rounds of tools to fulfill the human requirement. Finally, the execution results from the tools are gathered as inputs of the LLM to generate a response to the user. Next, we cover the three key components of multimodal agents. 

\paragraph{Tools.} Tools are external modules that are callable by the LLM to obtain extra information that is missing from the model weights, including open-source models, public/private APIs, or code interpreters. As LLMs only accept language inputs, one must include tools that can process multimodal inputs to build a multimodal agent.

\paragraph{Planning.}
During planning, the LLM decomposes the user requests into smaller, manageable sub-problems, and outlines a step-by-step solution, each of which involves calling an external tool.  There are two ways to teach LLMs for planning. One is to prompt the LLM with in-context few-shot examples of all candidate tools. This approach can extend the general model directly but is limited by the context length. The other approach relies on large amounts of annotated data to fine-tune the LLM, which most likely will damage the robustness and generalizability of the model.

\paragraph{Execution.} The generated plan is further translated into executable calls to the required tools, which can be done via regular expression matching~\citep{yang2023mmreact}; directly prompting LLMs to generate executable programs~\citep{suris2023vipergpt}; or leveraging in-context few-shot learning capability of LLMs by providing natural language instructions that describe the roles of each module together with a few calling examples~\citep{lu2023chameleon}. The execution results are fed back to the LLM to generate a response to the user.


\begin{figure}[htb!]
\centering  
\vspace{-0mm}
\includegraphics[width=.95\textwidth]{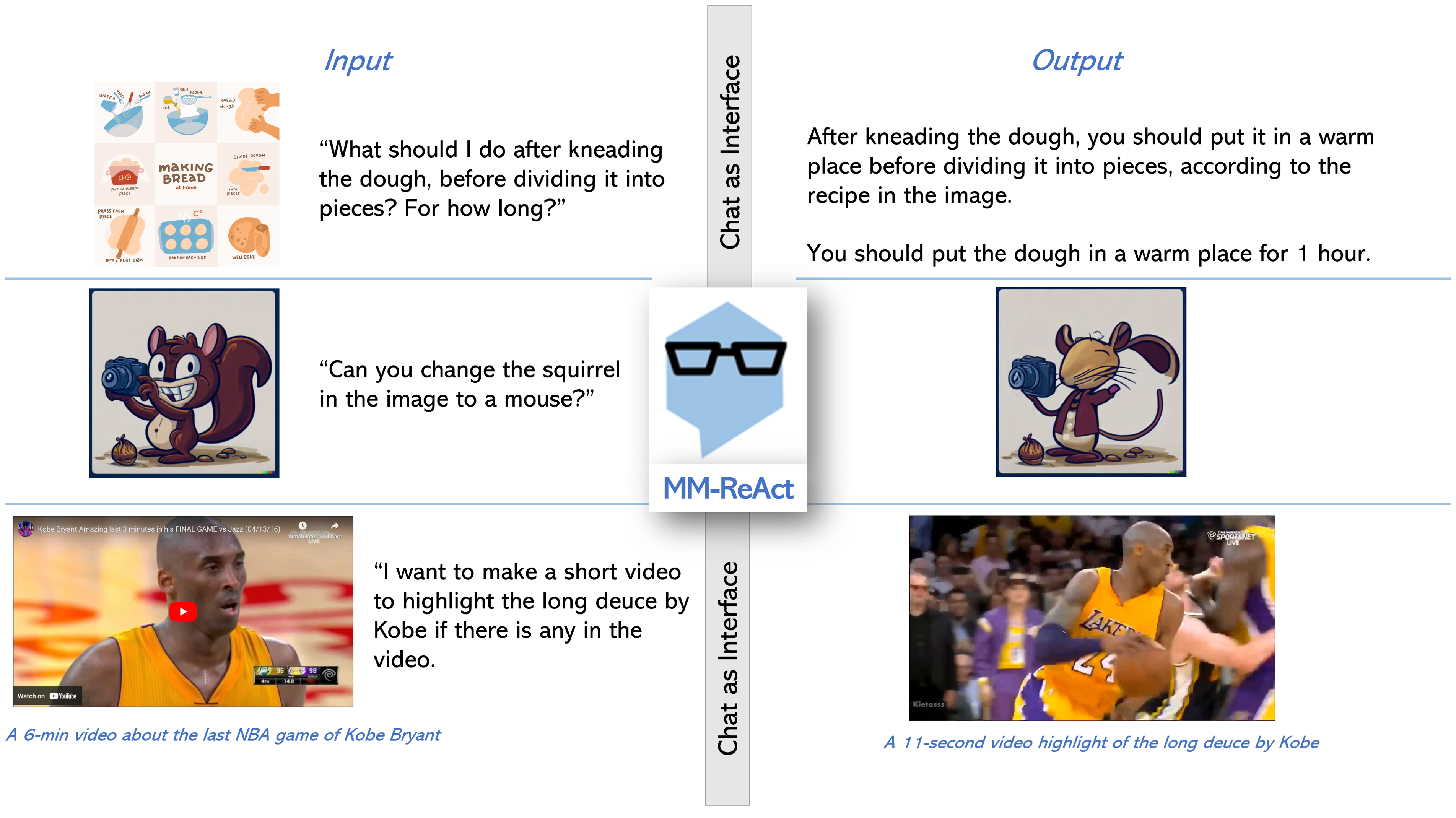} \\
\vspace{-0mm}
\caption{Input/Output modalities of \textsc{MM-ReAct}~\citep{yang2023mmreact}.}
\label{fig:chp6_mmreact_intro}  
  \vspace{-1mm}
\end{figure}

\section{Case Study: \textsc{MM-ReAct}}\label{chp6_sec:mmreact}
We use \textsc{MM-ReAct}~\citep{yang2023mmreact} as a case study to show how to build a multimodal agent, its emerging capabilities in multimodal understanding, and how it can be easily extended to incorporate the latest and strongest LLM and potentially millions of tools.

\subsection{System Design}
MM-ReAct designs the system paradigm that composes numerous multimodal tools\footnote{In \cite{yang2023mmreact}, these tools are referred as experts. We unify the terminology as tools throughout this chapter.} with ChatGPT~\citep{chatgpt} for multimodal reasoning and action. 
By augmenting the language-only ChatGPT with various multimodal tools,  \textsc{MM-ReAct} supports both inputs and outputs in multi-modalities, including text, image and video, as shown in Figure~\ref{fig:chp6_mmreact_intro}.

\begin{figure}[htb!]
\centering  
\vspace{-0mm}
\includegraphics[width=1.00\textwidth]{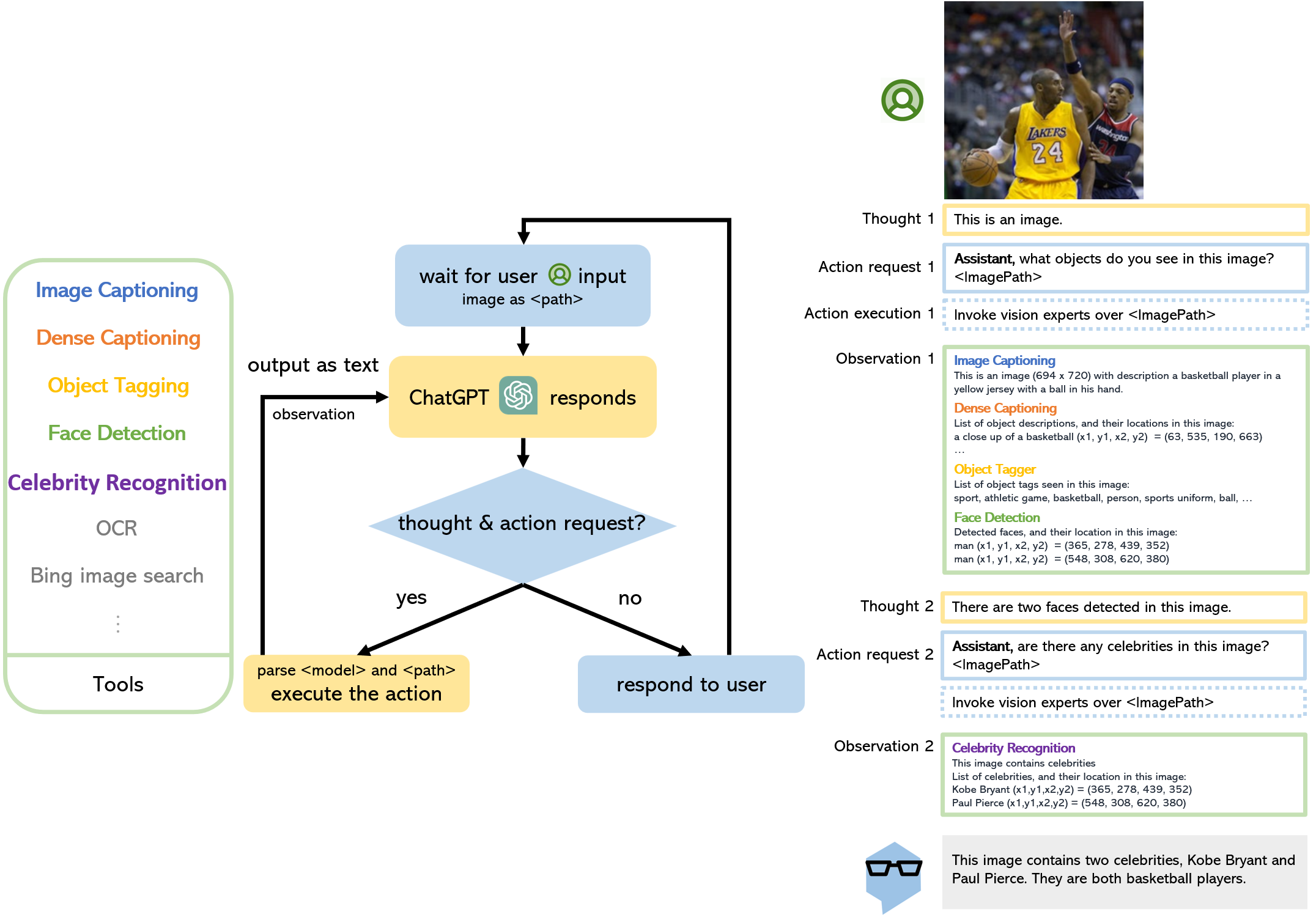} \\
\vspace{-0mm}
\caption{System design of \textsc{MM-ReAct} \citep{yang2023mmreact}. 
}
\label{fig:chp6_mmreact_flowchart}  
  \vspace{-1mm}
\end{figure}

Figure~\ref{fig:chp6_mmreact_flowchart} shows the system design of \textsc{MM-ReAct}. The \textbf{multimodal tools} explored in \textsc{MM-ReAct} are mainly computer vision models that take an image as input and 
interpret the image content from different perspectives. For instance, the image captioning model generates a natural description, the OCR model extracts the scene text in the image, the celebrity recognition model identifies the celebrity names, and the object detection model extracts the salient object with bounding box locations. 
LLMs such as ChatGPT serves as the brain of the agent, which plans on which tools to use, and in what order, based on the input image and the user intent. 
Next, with the example in 
Figure~\ref{fig:chp6_mmreact_flowchart}, we unfold the planning and execution of \textsc{MM-ReAct} behind the scene.

\paragraph{User prompt.} As ChatGPT only accepts language inputs, to enable image as inputs, we simply use the file path as the input to ChatGPT. The file path functions as a placeholder, allowing ChatGPT to treat it as a black box and later seek help from different tools during the planning stage. Besides the input image, the user can also provide the intent in text format (\textit{\eg}, a question about the input image). When there is no text input from the user, the goal is to get a general understanding about the image.


\paragraph{Planning.} Upon receiving the input image and user prompt, ChatGPT plans for what tools to use.  Inspired by \textsc{ReAct}~\citep{yao2022react}, \textsc{MM-ReAct} instructs ChatGPT to respond with certain watchwords, such as \textit{``Assistant, what objects are there in the image? $<$file path$>$''}, if a specific tool is required (i.e., \textit{action request} in Figure~\ref{fig:chp6_mmreact_flowchart}). In practice, one can tell whether a multimodal tool is needed by simply string-matching the keyword  ``Assistant,'' in ChatGPT's response. 

MM-ReAct encourages ChatGPT to show the \textit{thought} (reasoning)
process to highlight why an external tool is needed, which has been proven beneficial  in NLP studies~\citep{yao2022react}.
In addition, for generating proper calling to each tool, both instructions and in-context examples are added as the prefix when prompting ChatGPT. Each tool is described with the model name, a general description of its capability, the input data format, and the output information. After describing each tool, a few in-context dialogue examples are also included for enhanced performance. 

\paragraph{Execution.} Given the action request from ChatGPT, the tool name and the file path can be parsed via regular expression matching, which are used to invoke the tool (\textit{action execution}).

Take the example shown in Figure~\ref{fig:chp6_mmreact_flowchart}, upon receiving the input image, ChatGPT first invokes a series of tools for a general understanding about the image. The invoked tools include image captioning for an overall description of the image; dense captioning to get the region-level, more detailed description about the objects in the image; object tagging to get the tags of the objects in the image; face detection to get the box coordinates of the two faces mentioned in the object tags.  The outputs from the tools (\textit{i.e.} \textit{observations}) are serialized as text, and fed back to ChatGPT. 

Combining observations with the chat history, ChatGPT can further invoke additional experts or return the final answer to the user. In this specific example, ChatGPT invokes a second round of \textit{thought-action-observation} over the two faces detected in the image and calls celebrity recognition to get the names of these two persons. 


\paragraph{Response generation.} When ChatGPT decides no external tools are needed, it takes consideration of all observations gathered and summarize them as the response to the user, which is ``This image contains two celebrities,
Kobe Bryant and Paul Pierce. They are both
basketball players.'' for the example shown in Figure~\ref{fig:chp6_mmreact_flowchart}.

If the user continues to interact with \textsc{MM-ReAct}, it repeats the process described above, but with all observations and chat history available when planning for the tools needed.  For instance,  if the user then asks ``how many championship rings did the player on the left win in his career'', it is not available in the existing observations nor chat history, but ChatGPT has the bounding boxes to decide who is on the left, and also the names of the players. It plans to invoke Bing Search to find the right answer, which should be 5.

\begin{figure}[tb!]
\centering  
\vspace{-0mm}
\includegraphics[width=1.00\textwidth]{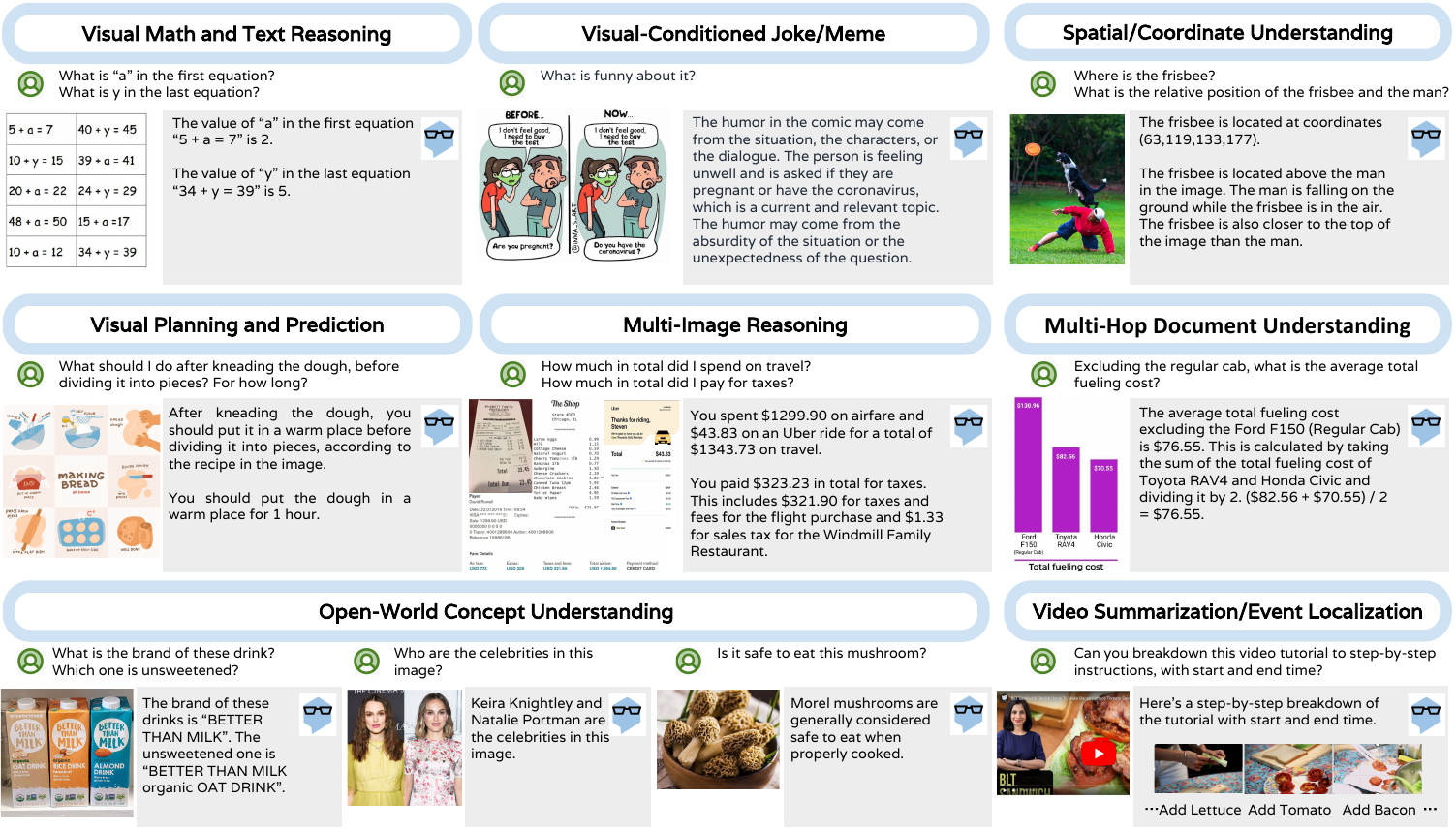} \\
\vspace{-0mm}
\caption{Emerging capabilities of \textsc{MM-ReAct} for multimodal reasoning and action. Image credit:~\cite{yang2023mmreact}.}
\label{fig:chp6_mmreact_capabilities}  
  \vspace{-1mm}
\end{figure}

\subsection{Capabilities}
Figure~\ref{fig:chp6_mmreact_capabilities} shows the representative capabilities and application scenarios that \textsc{MM-ReAct} demonstrates, including visual
math and text reasoning, understanding visual-conditioned jokes/memes, spatial/coordinate understanding, visual planning and prediction, multi-image reasoning, multi-hop document understanding, open-world concept understanding, video analysis and summarization. 

\begin{figure*}[htb!]
\centering
\includegraphics[width=1.00\textwidth]{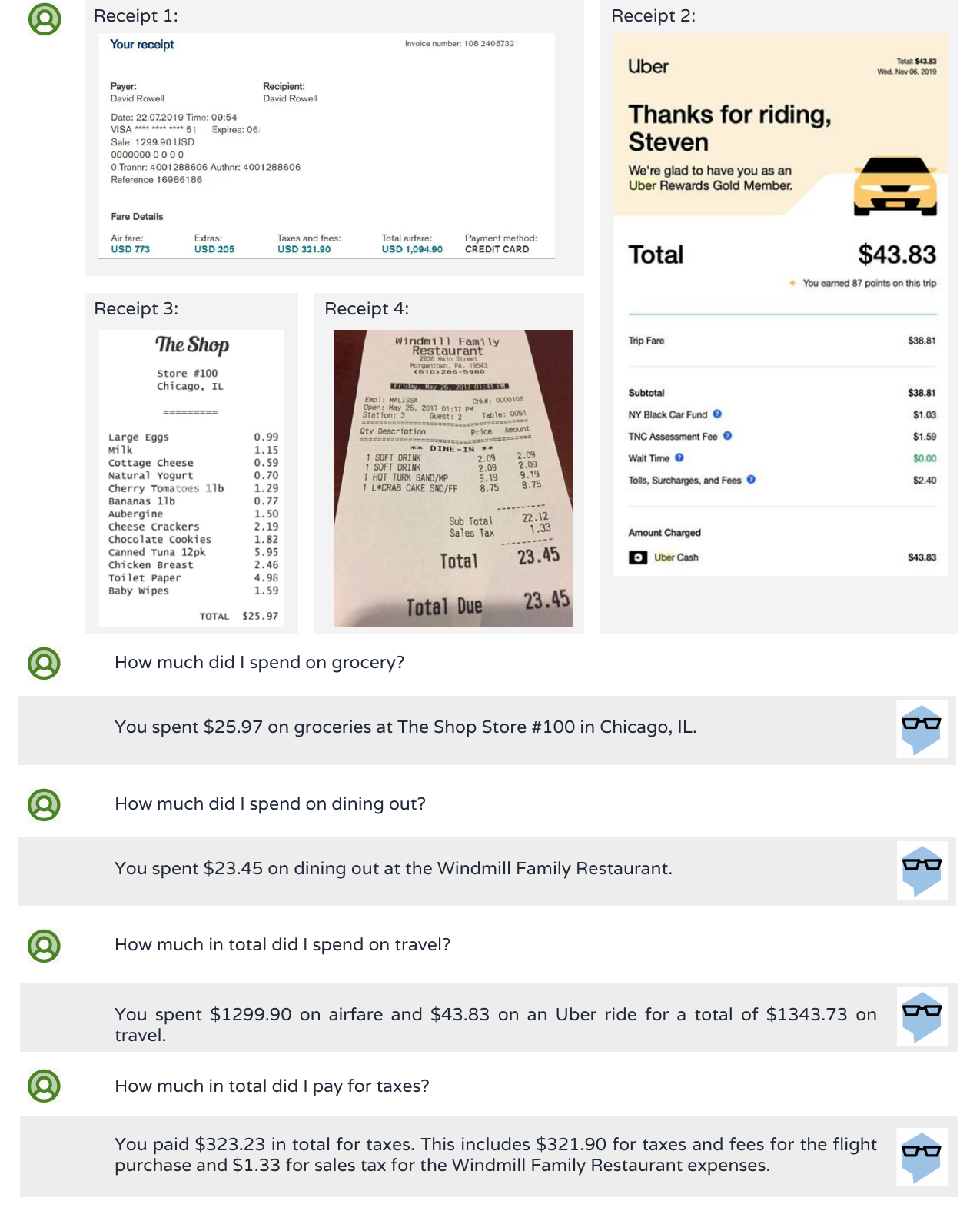}
\caption{Examples of \textsc{MM-ReAct}'s capabilities and application scenarios: \textbf{multi-image reasoning}. Image credit:~\cite{yang2023mmreact}.}
\label{fig:chp6_capability_multi_image}
\end{figure*}

In addition, we show an example of the full response from \textsc{MM-ReAct} on multi-image reasoning in Figure~\ref{fig:chp6_capability_multi_image}, which may not be easily achievable by visual instruction tuning in Chapter~\ref{chp:training_with_llm}.
For more comprehensive examples of all emerging capabilities of \textsc{MM-REACT}, we refer the reader to the original paper.

\begin{figure*}[htb!]
\centering
\vspace{-2mm}
\centering
\includegraphics[width=.95\textwidth]{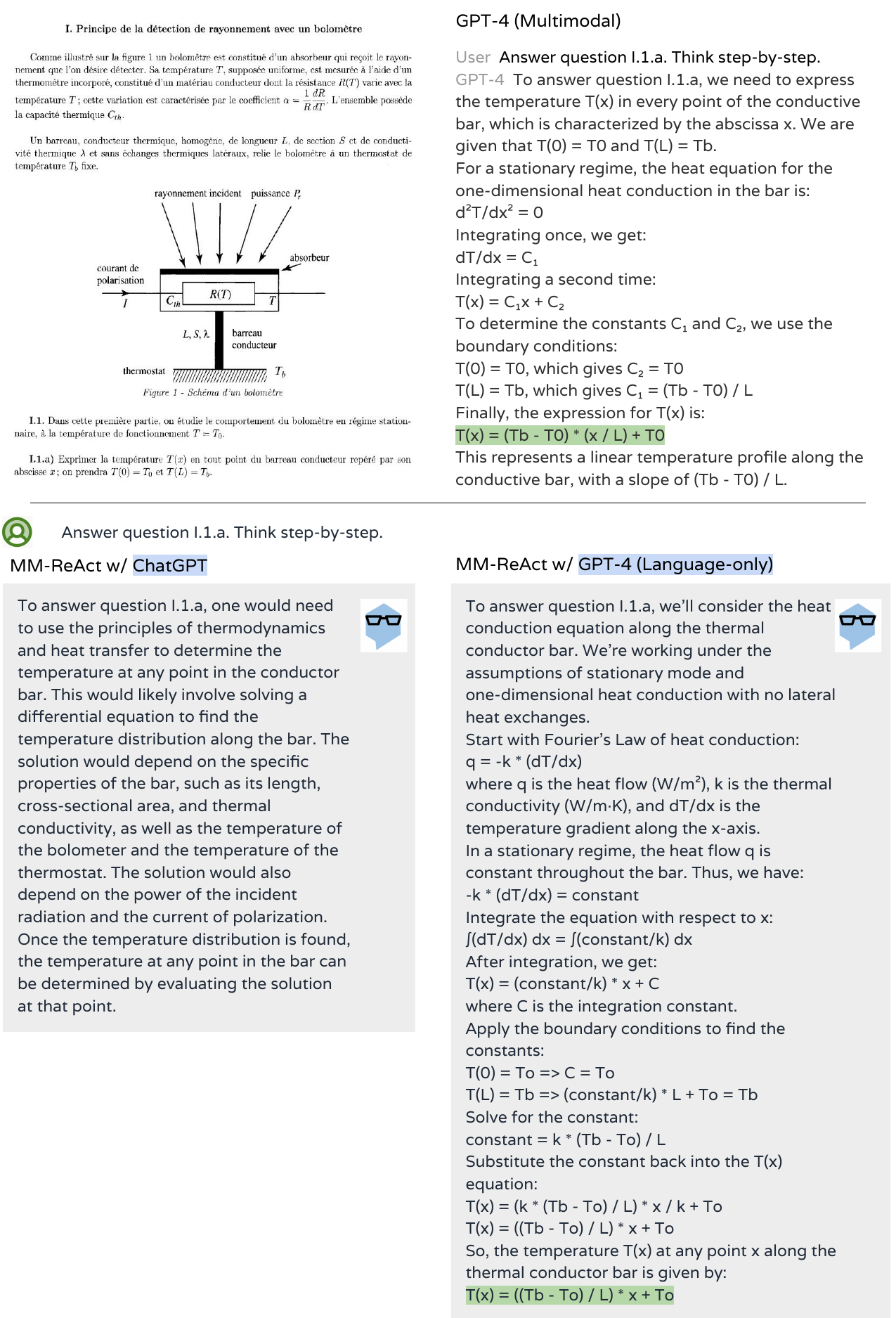}
\caption{Extensibility of multimodal agents: upgrading LLMs. Image credit: \cite{yang2023mmreact}. 
}
\label{fig:chp6_upgrading_llm}
\end{figure*}

\subsection{Extensibility}
One favorable property of tool chaining to build multimodal agents is that the system can be easily extended and enhanced, from two perspectives. One is to upgrade the core part of the system, the LLM, and the other is to expand the number of external tools. 

\paragraph{Upgrading LLM.} The system design of \textsc{MM-ReAct} allows for upgrading the core part of the system, the LLM, to newer and more powerful models as they come out, without the need of re-training. We show an example in Figure
~\ref{fig:chp6_upgrading_llm} on upgrading ChatGPT to language-only GPT-4, which improves \textsc{MM-ReAct} to potentially match the performance of multimodal GPT-4.




\paragraph{Plug-and-play (adding more tools).}  Existing multimodal agents incorporates tools via a plug-and-play mechanism, allowing adding more tools without training. One prominent work along this direction is HuggingGPT~\citep{shen2023hugginggpt}, which proposes to leverage all open-source models hosted on huggingface. Chameleon~\citep{lu2023chameleon}, incorporates not only huggingface models, but also open-source models from GitHub, Bing search API, and python compiler. RestGPT~\citep{song2023restgpt} proposes a multi-level online planning framework that effectively handles the practical challenges associated with integrating LLMs with more than 100 RESTful APIs. However, it remains challenging in scaling this framework to thousands to millions of tools, which is the potential future demonstrated in TaskMatrix.AI~\citep{liang2023taskmatrix}. 


Moreover, one can leverage SAM~\citep{kirillov2023segment} as a tool to allow for more types of human interaction with the multimodal agent other than text. Recall in \textsc{MM-ReAct}, the user intent is all captured by the natural language query from the user. In InternGPT~\citep{liu2023internchat}, by connecting the tool SAM with GPT, it allows for more ways to interact with the system, for example, via clicks, scribbles, and drawing bounding boxes.
These additional interactions, to some extent, are mimicking the action of finger-pointing when we humans are having a conversation. 




\section{Advanced Topics}\label{chp6_sec:advanced_topics}
In this section, we discuss more advanced topics and shed light on potential future directions. 


\subsection{Comparison to Training with LLM in Chapter~\ref{chp:training_with_llm}}
We have covered two directions on building advanced multimodal systems based on LLMs. As the key distinction, the multimodal agents in this chapter leverages LLMs' high-level planning abilities to allocate various multimodal tools, while training multimodal models with LLMs in Chapter~\ref{chp:training_with_llm} solely leverages LLMs for text generation conditioned on multimodal inputs.

Nonetheless, both of these methods exhibit their respective advantages and disadvantages.
On one hand, instruction tuning enables an end-to-end model that directly interprets rich semantics in multimodal inputs, but requires data curation and training, hence more computationally expensive. However, limited instruction tuning data may limit its capabilities in certain scenarios, such as OCR. On the other hand, one can build a multimodal agent without any training by chaining LLMs with abundant off-the-shelf models/APIs/code interpreters as tools, and leveraging in-context few-shot examples to teach LLMs on planning. However, as there is no training, the system may fail to invoke the right tool. Moreover, weak domain experts may produce noisy outputs, that can confuse LLM on planning or reasoning, leading to weak performance. 

Though these two approaches exhibit distinct variations,, we envision the possibility of an intermediate domain that amalgamates the strengths of both paradigms, and raise the following questions. Now that we have open-source LMM such as LLaVA~\citep{liu2023visual}, can we replace the LLM with LLaVA as the tool allocator? If so, what capabilities would require a tool to be enabled? And what problems can be solved by instruction tuning. These are interesting directions that may worth exploring in the near future.

\subsection{Improving Multimodal Agents}
Existing multimodal agents mainly rely on in-context few-shot examples to teach LLM on planning, which can be unreliable, leading to inaccurate tool using. To improve the accuracy in planning, several works have been proposed and we group them into three categories below.

\paragraph{Composing tools via code generation.} Most existing multimodal agents uses natural language to prompt LLM for planning which tool to use. Researchers~\citep{gupta2023visual,suris2023vipergpt} have also been exploring using programming language for more accurate execution. Visual programming~\citep{gupta2023visual} is a prominent work along this direction, which uses the in-context learning ability of GPT-3~\citep{brown2020language} to generate python-like modular programs from natural language instructions for compositional visual tasks  ViperGPT~\cite{suris2023vipergpt} instructs GPT-3 Codex~\citep{chen2021evaluating} to generate Python code to compose multimodal tools for a one-round query answering.  However, as the codes are still generated by a LLM, the problem of inaccurate tool using still remains. 

\paragraph{Improving accuracy in tool using: self-assessment.}  A recent work AssistGPT~\citep{gao2023assistgpt} tries to improve the accuracy in tool using via self-assessment. It adds a stage of inspection and learning loop into the system. When the round of plan and execution is finished, the system inspects the outcome, and determines whether the reasoning path of calling the tool is a success or not, if so, save it as an in-context example, to teach LLM for a more accurate tool calling in the future rounds.

\paragraph{Improving accuracy in tool using: instruction tuning.} Another thread on improving accuracy in tool using is to combine the system with instruction tuning~\citep{patil2023gorilla,yang2023gpt4tools}.  One can generate a dataset of instruction-API pairs via self-instruct to tune a smaller LLM (\textit{\eg}, Vicuna-7B~\citep{vicuna}).

\paragraph{LMM as the tool allocator?} In addition, as LMMs evolve, we envision that the LLM  can be replaced by a LMM as the tool allocator in the system, to enable even more advanced application scenarios. 
If the tool allocator can take multimodal inputs, there is no need to unify the outputs of tools into text sequence, allowing more natural interactions between the tool allocator and multimodal tools, particularly those producing multimodal outputs. For instance, one can imagine using multimodal GPT-4~\citep{gpt4} to coordinate various image or video generation tools to make a short movie by providing it with a sketch of the storyline and visual examples of the main characters.

\subsection{Diverse Applications of Multimodal Agents}
By composing tools from a specific domain, this new system paradigm can also support diverse domain-specific applications.

\cite{yu2023interactive} composes LLMs with image synthesis tools and object-level/pixel-level image understanding tools to build a data synthesis pipeline to provide diverse annotations on synthesized image. Instruct2Act~\citep{huang2023instruct2act} complements the LLM with robotic executors, to enable robotic actions based on multi-modal instructions. When chaining a pool of audio models with LLM, AudioGPT~\citep{huang2023audiogpt} can understand and generate speech, music, sound and talking head. Similarly, WavJourney~\citep{liu2023wavjourney} further supports compositional audio creation  with storylines encompassing speech, music, and sound effects. With tracking, captioning, audio understanding models, ChatVideo~\citep{wang2023chatvideo} enables ChatGPT to understand multi-channel videos. Other application scenarios include 3D scene generation~\citep{lin2023towards,feng2023layoutgpt}, medical image understanding~\citep{liu2023stone,sun2023pathasst} and vision-language navigation~\citep{zhou2023navgpt}.

\begin{figure*}[htb!]
\centering
\includegraphics[width=.95\textwidth]{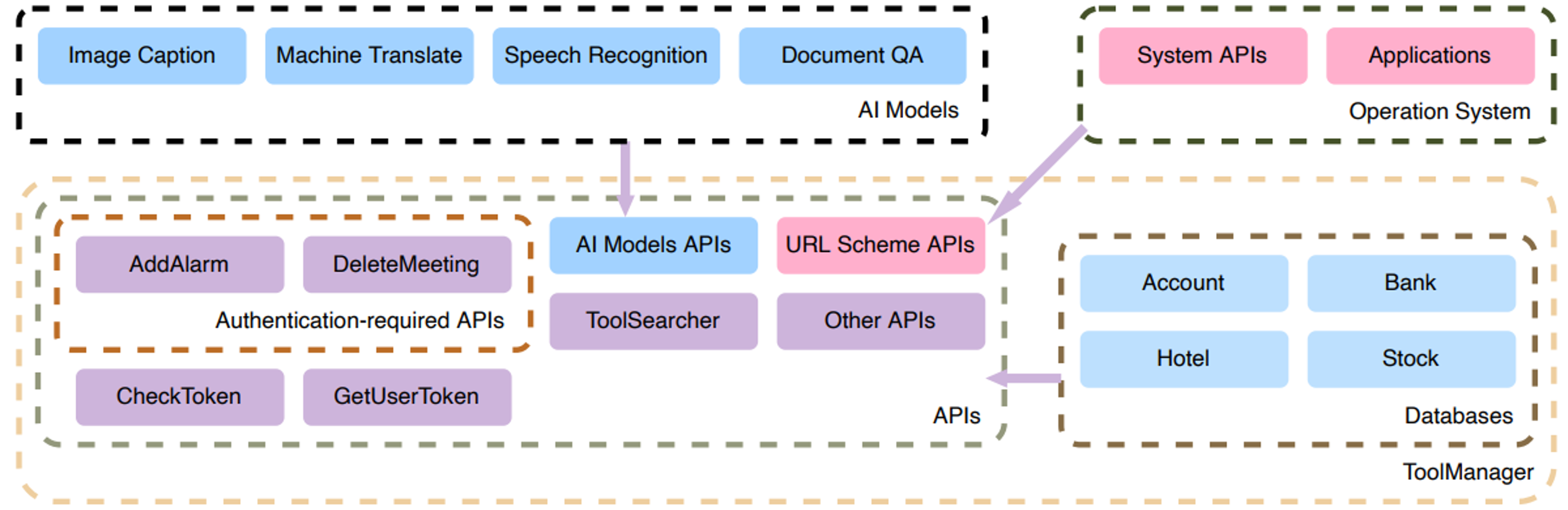}
\caption{Example of evaluation benchmark focusing on toll using accuracy.
}
\label{fig:chp6_api-bank}
\end{figure*}

\begin{figure*}[htb!]
\centering
\includegraphics[width=.95\textwidth]{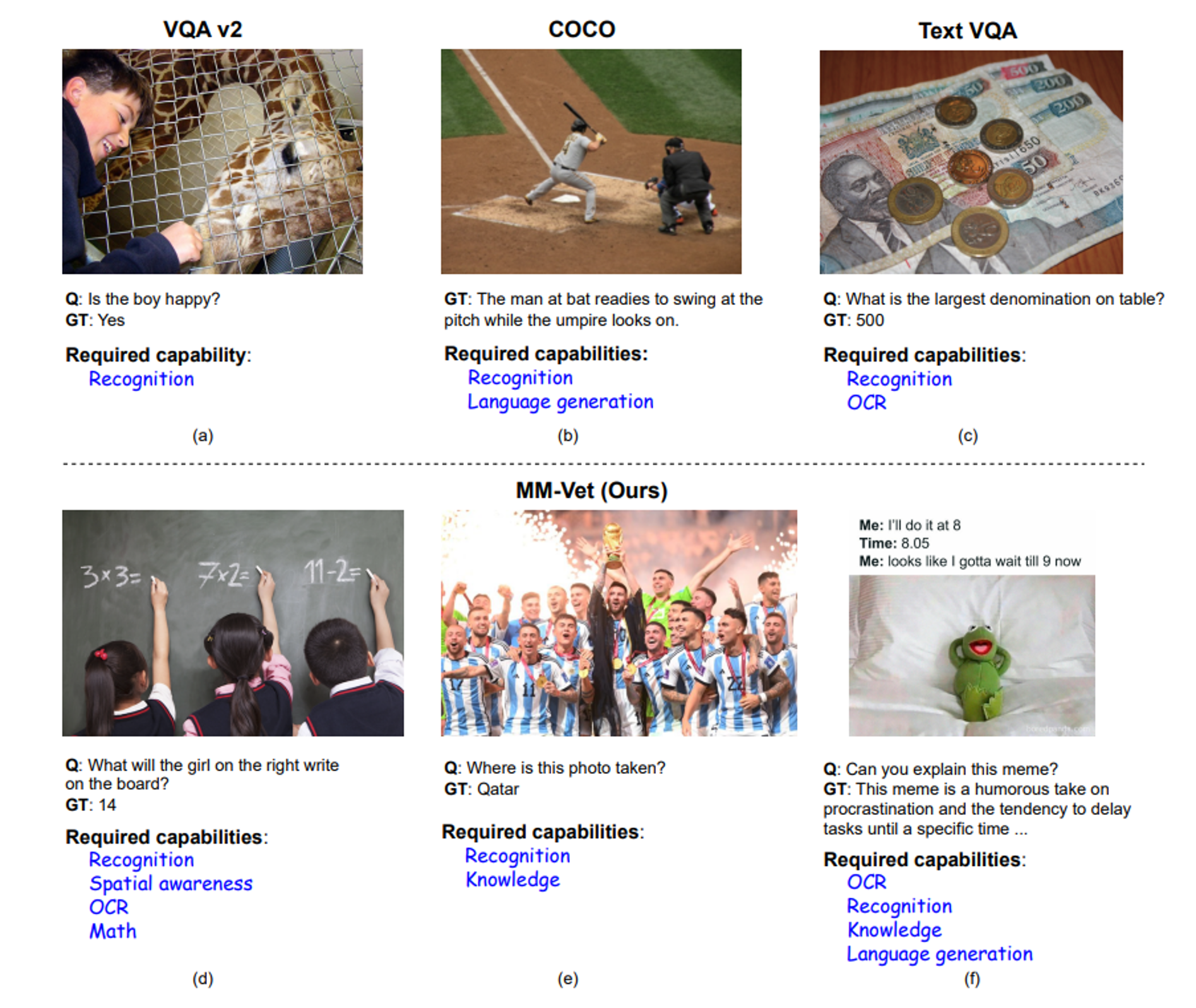}
\caption{Example of evaluation benchmark focusing on emergent capabilities. MM-Vet focuses on the integration of different core VL capabilities, including recognition, OCR, knowledge, language generation, spatial awareness, and math. Image credit: \cite{yu2023mm}. 
}
\label{fig:chp6_mm-vet}
\end{figure*}

\subsection{Evaluation of Multimodal Agents}
\paragraph{Multimodal tool using.} Although we have seen qualitative examples of new scenarios enabled by multimodal agents, it remains unclear how these agents perform in terms of the accuracy in tool using. API-Bank~\citep{li2023api} is a starting point on building pipeline in systematically evaluating tool-augmented LLMs.

\paragraph{Emergent capabilities.}  Existing VL benchmarks focus on specific capabilities of interest, such as visual
recognition~\citep{antol2015vqa}, image description~\citep{chen2015microsoftcoco,agrawal2019nocaps}, as well as other benchmarks for specialized capabilities
such as scene text understanding~\citep{sidorov2020textcaps,gurari2018vizwiz}, commonsense reasoning~\citep{zellers2019recognition}, outside knowledge~\citep{schwenk2022okvqa}. The intriguing abilities shown in large multimodal models and multimodal agents are not examined by existing benchmarks, such as solving math problems written on the blackboard, reasoning about events and celebrities in news images, or explaining visual jokes. Furthermore, the long, chatty outputs from these systems poses challenges to today’s evaluation metrics. Researchers~\citep{fu2023mme,liu2023mmbench} have started to design comprehensive evaluation samples to facilitate the LMM evaluation. As an attempt to test multimodal systems on integrated capabilities,  MM-Vet~\citep{yu2023mm} defines 6 core VL capabilities and examines the 16 integrations of interest derived from the capability combination (Figure~\ref{fig:chp6_mm-vet}). In addition, to accommodate for the open-ended free-form text outputs, MM-Vet proposes an LLM-based evaluator to enable evaluation across different question types and answer styles.

\subsection{Tool Creation} Imagine if we have a completely new scenario without a robust tool to use. Can we create a tool based on the user need on-the-fly? In NLP, CREATOR~\citep{qian2023creator} proposes to create tools by writing python code for math reasoning, as opposed to calling math solver API such as Wolfram Alpha. \cite{cai2023large} further explores the capabilities of LLMs to make tools, and experiment with two LLMs, one as the tool maker and the other as the tool user to collaboratively solve complicated tasks, such as scheduling a meeting. In terms of multimodal agents, the challenge is how to create a tool that can process multimodal inputs. One may follow ViperGPT~\citep{suris2023vipergpt} to instruct LLMs to generate python programs leveraging pre-existent python packages such as Open-CV. AutoML GPT~\citep{zhang2023automl} envisions that one can utilize LLMs to automate the model training pipeline. There may be potential to develop novel multimodal deep learning tools tailored to more effectively address the requirements of users.

\subsection{Retrieval-Augmented Multimodal Agents} In real-life applications, a substantial portion of information resides within databases, and user needs may require accurate retrieval of such information. Meanwhile, it is infeasible to encode all the world knowledge into the weights of pre-trained models, particularly when it comes to the long-tail concepts and fast-evolving data. 

In NLP, several works augment LLMs with external data encoded with structured language and relation representations~\citep{peters2019knowledge,guu2020realm,lewis2020retrieval}. Given input texts, such retrieved-augmented models utilize a retriever that retrieves relevant documents from an external memory, and uses a generator to generate predictions given the retrieved documents.

Motivated by retrieval-augmented models in NLP, several recent works leverage visual and / or textual knowledge to improve vision tasks, such as image classification~\citep{long2022retrieval}, captioning~\citep{yang2023re}, question answering~\citep{wu2021multi,marino2021krisp,yang2021empirical,chen2022murag}, image generation~\citep{blattmann2022retrieval,sheynin2022knn,chen2022re,zhou2022lafite2}, and multi-modal tasks simultaneously~\citep{yasunaga2022retrieval}.  RAC~\citep{long2022retrieval} improves long-tail classification by retrieving from a non-parametric memory consisting of pre-encoded images and text. K-LITE~\citep{shen2022k} enhances the text prompts with the retrieved external knowledge that is encoded in natural language.
REACT~\citep{liu2023learning} retrieve from billions of the paired knowledge of image-text and aims to improve task transfer performance for core vision problems.
Among them, RA-CM3~\citep{yasunaga2022retrieval} builds the first retrieval-augmented LMM with a multimodal retriever to retrieve multimodal documents, and a retrieval-augmented generator that can generate both text and image. 
Chaining tools with LLM shares a strong connection with the retrieval-augmented methods in that both leverage external knowledge to provide additional information for the core model to utilize. In the multimodal regime, the image itself can be used as the query to gain external knowledge, either retrieved from a knowledge base, or extracted from another pre-trained vision expert models.


\chapter{Conclusions and Research Trends}
\label{chp:conclusion}

Multimodal foundation models have garnered significant interest among scholars in the fields of computer vision and multimodal vision-language research. 
Although prevailing research topics, approaches and methodologies have been evolving – encompassing image self-supervised learning, language-image contrastive learning, text-to-image generation, unified vision modeling, and large language-and-vision assistants – they converge on a common overarching objective:  the creation of general-purpose models and systems capable of following human intents and effortlessly executing a diverse array of vision and vision-language tasks in the wild.   In this chapter, we provide a concise summary of what has been reviewed, and delve into the prevailing research tendencies in the field.

\section{Summary and Conclusions}
This paper surveys the most recent advances at the frontier of multimodal foundation model research, categorized into two classes discussed below.
\begin{itemize}[leftmargin=*]
    \item \textbf{Specific-purpose multimodal foundation models.} There is a diverse set of problems to tackle in the computer vision community. To lay a comprehensive foundation for the introduction of general-purpose visual assistants, we have discussed many seminar papers in the era of pre-training. The major paradigm during this period is pre-training on a large amount of problem-related data, and then transferring to a number of real-world scenarios of the same problem type in a zero- or few-shot fashion.
    More specifically, we have presented two general topics: ($i$) \emph{Visual Understanding} in Chapter~\ref{chp:understanding}: individual multimodal foundation models have developed to analyze the content of visual data in the image, region, pixel levels, prospectively. The language-augmented vision models are a popular family, contributing to the recent success of visual understanding tasks in the wild.
    ($ii$) \emph{Visual Generation}  in Chapter~\ref{chp:generation}: text-to-image generation models have served the foundation for image synthesis, which has been successfully extended to allow user controllability and customization at more fine-grained manners. The availability and creation of large amount of problem-related data has played a key role in making these multimodal foundation models possible.
    
    \item \textbf{General-purpose assistants.} We have reviewed recently emerged literature on building general-purpose assistants, which often possess an unified network architecture, an unified input-output data format, and a general interface that facilitates easy interaction with humans. Inspired by the success in NLP that LLM such as ChatGPT/GPT-4 is a general assistant for a wide range of language tasks, researchers in computer vision have explored various solutions to their counterpart for vision tasks. Based on how LLM is leveraged in the methodology, existing works can be categorized into three topics:  
    ($i$) \emph{Unified Vision Models} in Chapter~\ref{chp:generalist}: The spirit of unifying modeling in LLM is borrowed to build unified vision models at different levels and across different tasks. 
    ($ii$) \emph{Training with LLM} in Chapter~\ref{chp:training_with_llm}: Starting with a pre-trained LLM, visual data is connected to LLM for end-to-end training. 
    ($iii$) \emph{Chaining with LLM} in Chapter~\ref{chp:chaining_with_llm}: By freezing LLM, existing vision experts can be triggered by prompt engineering LLM to complete specific vision tasks.  
\end{itemize}
The comparisons among these models are summarized in Table~\ref{tab:pros_and_cons}.

\begin{table}[t!]
\centering
\scalebox{0.87}{
\begin{tabular}{@{\hspace{3pt}}l@{\hspace{3pt}}l p{5.5cm}p{5.5cm}}
\toprule
& Models & Advantages & Disadvantages \\
\midrule
\multirow{2}{*}{\rotatebox{90}{\footnotesize \it Specific-Purpose Models}}
&
Visual Understanding & Well studied and scalable solutions on image-level understanding; Emerging interests and success on region-level and pixel-level visual understanding &
High training cost; No successful scalable recipe beyond the billion-image level \\ \cmidrule{2-4}
& 
Visual Generation & Well studied and scalable solutions on image-level generation; Emerging interests and success in controllable/customized image generation & 
High training and inference cost; Debate between diffusion and auto-regressive solutions for the best recipe; 
More studies are needed for video generation\\ \midrule 
\multirow{3}{*}{\rotatebox{90}{\footnotesize \it General-Purpose Assistants}} 
& 
Unified Vision Models & Promises to unlock new emerging capabilities and scenarios & High risks in modeling and high training cost \\ \cmidrule{2-4}
& 
Training with LLM &  
Some new emerging capabilities and scenarios are enabled with light model training & 
The performance is bounded by LLM \\ \cmidrule{2-4}
& 
Chaining with LLM & 
Fast system development cycles with low cost as no training is involved & 
Low flexibility in improving system performance; No new emerging capabilities  \\
\bottomrule
\end{tabular}
}
\vspace{1mm}
\caption{Comparisons of different multimodal foundation model families covered in this paper.}
\label{tab:pros_and_cons}
\end{table}

\section{Towards Building General-Purpose AI Agents}

At the end of each chapter, we have discussed future trends for individual topics. The paper itself is organized to demonstrate the transition from specialist multimodal foundation models to general-purpose visual assistants. Though powerful, existing visual assistants such as Flamingo~\citep{alayrac2022flamingo} and multimodal GPT-4~\citep{openai2023gpt4} are in the preliminary form, compared with grand vision on building a general-purpose multimodal AI agent via foundation models. 
In what follows, we highlight a number of research trends towards this goal.

\paragraph{Generalist agents with multi-modality.} This aligns with the grand goal of building a single generalist agent that interacts with world like humans through fusing multiple channels such as language, vision, speech and actions. From this perspective, the notion of multimodal foundation models becomes somewhat indistinct on its own. Instead, it serves as a crucial component of the agent for perceiving and synthesizing visual signals. For example, Gato~\citep{reed2022generalist} and PaLM-E~\citep{driess2023palm} perform a wide range of language, multimodal and control tasks with a single set of model weights, where visual perception is a crucial component in understanding the environment. It also raises challenges in the effective and scalable pre-training objectives for unified vision and multimodal modeling. 

\paragraph{Alignment with human intents.} AI alignment research focuses on steering AI systems towards humans' intended goals, values, or ethical guidelines. An AI system is deemed aligned when it effectively promotes the desired goals. Though language has exhibited its generality in expressing human intents, it is not always the best option. As demonstrated in SAM~\citep{kirillov2023segment} and ControlNet/GLIGEN~\citep{zhang2023adding,li2023gligen}, human intents can be more precisely and conveniently represented in visual prompts such as key points, bounding boxes and sketch drawing, for visual understanding and generation tasks, respectively. Building foundation models that are well equipped with such a multimodal human-machine interaction interface is a key step to unlock new use scenarios, where human intents are best represented visually. For example, the spatial arrangement of elements within a scene, as well as the artistic style and visual appeal of a piece of visual art.

\paragraph{Planning, memory, and tool use.} 
It is highlighted in~\cite{weng2023agent} that a LLM-powered autonomous agent system can be built, where LLM functions as the agent’s brain, complemented by several key components: planning, memory and tool use. Following the framework, we could foresee the role of multimodal foundation models in this AI agent system.
($i$) Planning. To complete complex tasks in real-world scenarios, the agent should be able to decompose large tasks into smaller, manageable subgoals, enabling efficient handling of complex tasks. In the ideal case, the AI agent possesses the self-improvement ability, engaging in self-assessment and introspection regarding previous actions, enabling it to learn from errors and enhance its approach for subsequent endeavors, ultimately leading to better outcomes. Visual modality is a common channel to represent state of the environment. To facilitate planning, it raises challenges in improving the capability of the current visual understanding models in perceiving more fine-grained visual details and longer sequence videos.  
($ii$) Memory. 
For short-term memory, in-context learning (or prompt engineering) is utilized as short-term memory for the model to learn. Interleaved multimodal prompts can enable new scenarios to clarify the human intents.  For long-term memory, it provides the agent with the capability to  recall external knowledge over extended sessions, which can be implemented by fast retrieving from a multi-modal vector space~\citep{liu2023learning}. In term of modeling, foundation models are required to learn the new skills to effectively leverage both types of memory.
($iii$) Tool use. 
The agent learns to utilize external APIs for knowledge that is missing from the foundation model weights. New capabilities are required to deal with the vision modality in several scenarios. For example, based on both the input visual signal and instructions, the model decides and plans whether certain external APIs are needed to complete the goal, such as code execution of detection/segmentation/OCR/generator experts.

The field of multimodal foundation models is evolving at a rapid speed, with new directions/methods emerging frequently. There are many important research topics that are not discussed in this paper, mostly due to the daily-updated research innovation. 
We are optimistic about the future of multimodal foundation models, not only because we are convinced that foreseeable exciting research innovations/ideas in individual areas are becoming reality by following the path of LLM in the near future, but also because connecting computer vision with the broader AI community, and building general-purpose AI agents is going to significantly advance the daily life of human being.

\chapter* {Acknowledgments} 
 
This book is largely based on our CVPR 2023 tutorial on vision foundation models.
Many people have supported us and provided valuable feedback to the writing of this book.
We thank all the authors who have contributed to the related papers, which makes the tutorial and book possible.
We are also grateful to Mark de Jongh, the editor from the journal of {\it Foundations and Trends® in Computer Graphics and Vision}, for inspiring and encouraging us to write the book on multimodal foundation models.

\bibliographystyle{apalike}
\bibliography{ref}

\end{document}